\documentclass[journal]{IEEEtran}
\ifCLASSINFOpdf
\else
\fi

\usepackage{color}
\usepackage{epsfig}
\usepackage{graphicx}
\usepackage{amsmath}
\usepackage{amssymb}
\usepackage{bm}
\usepackage{algorithm}
\usepackage{algorithmic}
\usepackage{multirow}
\usepackage{subfigure}
\usepackage{amsthm}
\usepackage[colorlinks,linkcolor=green]{hyperref}
\usepackage{url}

\newtheorem{thm}{Theorem}[section]
\newtheorem{prop}{Proposition}[section]
\newtheorem{corollary}{Corollary}[section]

\newtheorem{lemma}{Lemma}[section]
\newtheorem{defi}{Definition}[section]

\newtheorem{condition}{Condition}[section]

\begin{document}
%
\title{Learning Converged Propagations with \\Deep Prior Ensemble for Image Enhancement}
%
%
%

\author{Risheng~Liu,~\IEEEmembership{Member,~IEEE,}
	Long~Ma,
	Yiyang~Wang,
	and Lei~Zhang,~\IEEEmembership{Fellow,~IEEE}
	\thanks{This work was supported by the National Natural Science Foundation of China (Nos. 61672125, 61733002, 61572096, 61632019, and 61806057), the Fundamental Research Funds for the Central Universities, and the Project funded by the China Postdoctoral Science Foundation  (No. 2018M632018).}
	\thanks{R. Liu is with DUT-RU International School of Information Science \& Engineering and the Key Laboratory for Ubiquitous Network and Service Software of Liaoning Province, Dalian University of Technology, Dalian, China. He is also with the State Key Laboratory of Integrated Services Networks, Xidian University, Xi'an, China. (Corresponding author, e-mail: rsliu@dlut.edu.cn).}
	\thanks{L. Ma is with DUT-RU International School of Information Science \& Engineering, and Key Laboratory for Ubiquitous Network and Service Software of Liaoning Province, Dalian University of Technology, Dalian, China. (e-mail: malone94319@gmail.com).}
	\thanks{Y. Wang is with Institute of Atmospheric Sciences, Fudan University and Shanghai Institute of Meteorological Science, Shanghai, China (e-mail: yywerica@fudan.edu.cn).}
	\thanks{L. Zhang is with the Department of Computing, The Hong Kong Polytechnic University, Hong Kong, China (e-mail: cslzhang@comp.polyu.edu.hk).}
	\thanks{Manuscript received October 21, 2017; revised May 05, 2018 and July 25, 2018; accepted September 30, 2018. }}

\maketitle

\begin{abstract}
	Enhancing visual qualities of images plays very important roles in various vision and learning applications. In the past few years, both knowledge-driven maximum a posterior (MAP) with prior modelings and fully data-dependent convolutional neural network (CNN) techniques have been investigated to address specific enhancement tasks. In this paper, by exploiting the advantages of these two types of mechanisms within a complementary propagation perspective, we propose a unified framework, named deep prior ensemble (DPE), for solving various image enhancement tasks. Specifically, we first establish the basic propagation scheme based on the fundamental image modeling cues and then introduce residual CNNs to help predicting the propagation direction at each stage. By designing prior projections to perform feedback control, we theoretically prove that even with experience-inspired CNNs, DPE is definitely converged and the output will always satisfy our fundamental task constraints.
	The main advantage against conventional optimization-based MAP approaches is that our descent directions are learned from collected training data, thus are much more robust to unwanted local minimums. While, compared with existing CNN type networks, which are often designed in heuristic manners without theoretical guarantees, DPE is able to gain advantages from rich task cues investigated on the bases of domain knowledges. Therefore, DPE actually provides a generic ensemble methodology to integrate both knowledge and data-based cues for different image enhancement tasks.
	More importantly, our theoretical investigations verify that the feed-forward propagations of DPE are properly controlled toward our desired solution.
	Experimental results demonstrate that the proposed DPE outperforms state-of-the-arts on a variety of image enhancement tasks in terms of both quantitative measure and visual perception quality.
\end{abstract}


%
\IEEEpeerreviewmaketitle

\section{Introduction}
%
%
%
%

The basic purpose of image enhancement task is processing given images to make the restored results be more
suitable than the original observations for specific applications. There are plenty of ways that bring about degradations on visual data, corresponding to specific image enhancement tasks. For example, image deblurring (also named as deconvolution) is a classical but high-profile application in the image enhancement society~\cite{Pan2014Deblurring}.
The degraded blurry images are produced by the movements of image sensors during exposures, through accumulating incoming lights for amount of times.
For another instance, super-resolution is a class of techniques that aim at enhancing the resolutions of low-quality visual devices, such as hand-held computers and mobile phones, in an accurately and quickly manner~\cite{Wang2017Single}.
While in foggy and hazy weather or underwater scenarios, the atmospheric particles or waters will absorb and scatter not only atmospheric lights, but also reflected lights to cameras.
Thus the image acquired under such scenarios are seriously degraded and thus usually have poor visibilities, in which the targets and obstacles are quite difficult to be recognized.
Therefore, it is necessary to recover the authentic images from these corrupted observations~\cite{Fattal2008Single,Cai2016DehazeNet}. Actually all these image enhancement tasks play active roles not only in improving the visualization of data but also in supporting many subsequent applications in various computer vision, pattern recognition and image analysis problems.

\subsection{Related Work}

In general, existing knowledge-based approaches for settling image enhancement problems can be roughly grouped into two categories, by distinguishing their mechanisms with the prior information, i.e, \emph{optimizing designed-priors} and \emph{learning parameterized-priors}. Besides, recently developed deep models
often aim to train \emph{fully-dependent neural networks} to address
particular tasks. So in this part, we would like to briefly review related work following the above categories.

\subsubsection{Optimizing Designed-Priors}

Due to the ill-posed nature of most image enhancement tasks, it is necessary to design prior regularizations for getting desired solutions, under the employment of the maximum a posteriori (MAP) approach~\cite{Deng2016Guided}. Following this way, it is tending to put forward more and more complex priors to better characterize the generative modes of specific tasks and the structures of desired solutions.

Many image enhancement methods exploit the sparsity prior of natural images. Based on the fact that natural
image gradients exhibit heavy-tailed distributions~\cite{weiss2007makes}, image gradient histograms or total variations~\cite{rudin1992nonlinear} are widely employed as sparsity priors to regularize image enhancement~\cite{osher2005iterative}. By using the $\ell_1$-optimization techniques, the sparse coding based methods~\cite{daubechies2004iterative} encode an image over an over-complete dictionary, and have been demonstrated better image enhancement results than linear transform methods. Though it is non-convex and discontinuous, the $\ell_0$ penalty attracts wide attentions since it globally controls the number of non-zero elements. This sparse prior is also a major participant in the fields of image filtering~\cite{Xu2011Image}, deconvolution~\cite{Xu2013Unnatural} and image layer separation~\cite{Guo2014Robust}, etc.

Some of the previously mentioned methods also exploit the local self-similarity prior of images, i.e., restraining a pixel being similar to its neighboring pixels. The image nonlocal self-similarity prior, which is based on the fact that similar patches to a local patch can be quite far from it, which has been demonstrated to achieve better image enhancement
performances. The representative works along this line include the nonlocal means~\cite{buades2005non}, nonlocal regularization~\cite{zhang2010bregmanized}, BM3D~\cite{dabov2007image}, etc. In particular, the BM3D method, which stacks the nonlocal similar patches as a 3D cube and applies the 3D wavelet transform to it, has become a benchmark for image denoising; it has also been extended to many other image enhancement tasks~\cite{danielyan2012bm3d}. The nonlocal self-similarity can be coupled with low-rank modeling~\cite{gu2014weighted} to further improve the image enhancement results.

Though designing complicated priors probably helps narrowing feasible regions of variables for image enhancement tasks, it also brings difficulties in optimization and obtaining desired solutions.
In recent decades, there are some numerical algorithms that have been proposed for solving non-convex optimization problems~\cite{Attouch2010Proximal,Bolte2014Proximal,Zeng2016Sparse}. However, the theoretical convergence results of those algorithms are relatively weaker compared to the convex problems. Thus it is usually time-consuming for standard numerical algorithms to globally optimize the MAP formulations with sophisticated priors. Moreover, to avoid trapping to the unwanted local minimum on these challenging tasks, the initial points and algorithm parameters are very sensitive to both the final results and the convergence performances of iterative directions.

\subsubsection{Learning Parameterized-Priors}

Instead of using pre-designed priors, one can learn parameterized prior models from natural
images directly for image enhancement. It is popular to model image priors with statistical distributions. In~\cite{zoran2011learning}, a multi-dimensional Gaussian mixture model (GMM) is learned to model image patches, and the so-called EPLL method shows promising denoising results. Moreover, GMM is also used in~\cite{yu2012solving} to learn piecewise linear estimators for image restoration. In the pioneering work called Fields of Experts (FoE)~\cite{Roth2009Fields}, the filtering responses to images are modeled with the Student’s t-distribution, and the filters are learned with Markov random field. Despite the successes of prior learning based methods, they usually learn a generic image prior model from
high quality images, without exploiting the statistics of degraded images and the characteristics of specific
enhancement tasks. Therefore, discriminative learning methods have been proposed to model the relationship
between the degraded images and their corresponding latent ones, as we briefly introduce below.

Discriminative prior learning methods aim to learn a model to map the degraded image to its corresponding latent image, meanwhile, maximizing the posterior probability of latent data. For example, with conditional random fields~\cite{schmidt2016cascades}, the cascade of shrinkage fields (CSF)~\cite{Schmidt2014Shrinkage} learns a set of filters from clean images and their degraded counterparts from denoising and deblurring. An adaptive version of shrinkage fields~\cite{xiao2016learning} was proposed for blind image deconvolution, and a conditioned regression model~\cite{riegler2015conditioned} was proposed for image super-resolution. Similar to CSF, Chen et al.~\cite{chen2017trainable} extended FoE priors and proposed a trainable nonlinear reaction diffusion model for both image denoising and super-resolution. By transforming MAP-based energy model into optimization with linear constraints, the discriminative priors can also be learned in the framework of alternating direction method with multipliers~\cite{sun2016deep}.
It can be observed that all these models are always over-reliant on the formats of complex statistic prior models, so that their propagation architectures are mostly intricate and inflexibility, which also increase the complexities of the training processes.

\subsubsection{Fully Data-dependent Neural Networks}

During the past few years, with the developments of deep neural networks~\cite{Simonyan2014Very,He2016Deep}, researchers have increasingly focused on designing deep end-to-end networks as image propagations for solving image-level applications~\cite{Ren2016Single,Fu2017Clearing}.
Different from prior-based approaches, those propagations do not have explicit descriptions of image priors and even do not directly express the generative modes of image-level tasks.
Instead, they learn the image propagations as deep networks, through minimizing the differences between the inputs and the desired solutions.
Convolutional neural network (CNN) is one of the most commonly-used network structures, which have been widely experienced and analyzed in image processing society~\cite{Cai2016DehazeNet,Zhang2016Beyond,Dong2016Image}.
For intricate image-level tasks, researchers usually design complicated neural networks which are specifically combined by multiple sub-networks~\cite{Yang2016Joint,Ren2016Single}.
Very recently, the authors of~\cite{Zhang2017Learning} proposed to learn denoiser networks to solve image restoration problems in the framework of optimization unrolling.
However, their iterations are proposed without any rigid theoretical analyses, thus their robust performances cannot be always guaranteed.

\begin{figure*}[t]
	\centering
	\includegraphics[width= 0.8\textwidth]{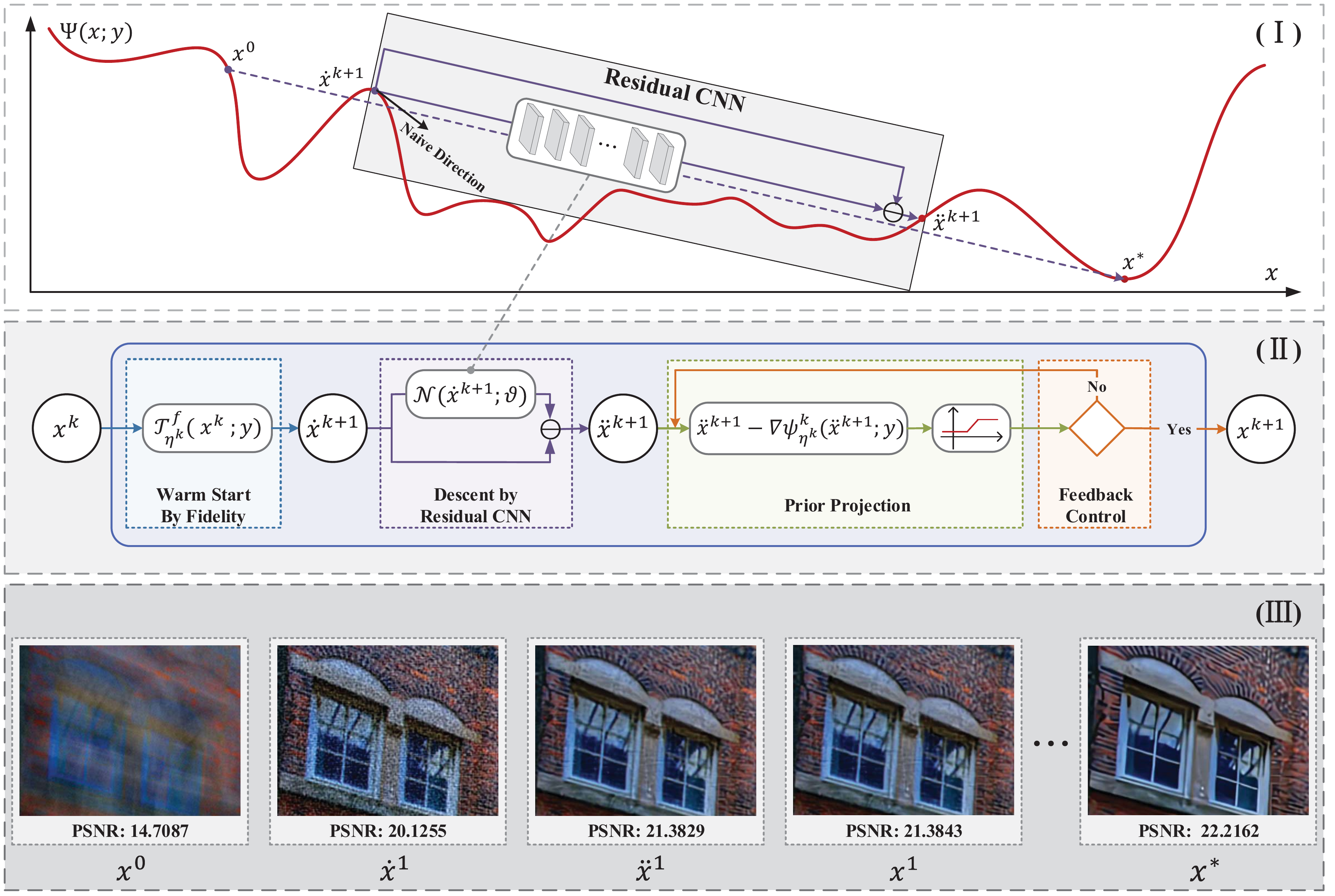}
	\caption{The illustration of DPE. In the top subfigure (I), it is illustrated that the residual
		CNN based propagative directions have the ability to avoid unwanted local minimums for our image propagation. Three basic building-blocks together with the feedback control strategy in $k$-th stage are demonstrated in the middle subfigure (II). Finally, we show the input, intermediate output of the first stage and the final result of DPE on the region of an example image in the bottom subfigure (III). The deconvolution results on the whole image with comparisons with existing approaches are also illustrated in Fig.~\ref{fig:stepfullres}.}
	\label{fig:framework}
\end{figure*}

\begin{figure*}[ht]
	\centering
	\begin{tabular}{c@{\extracolsep{0.1em}}c@{\extracolsep{0.1em}}c@{\extracolsep{0.1em}}c@{\extracolsep{0.1em}}c}
		\includegraphics[width=0.19\textwidth]{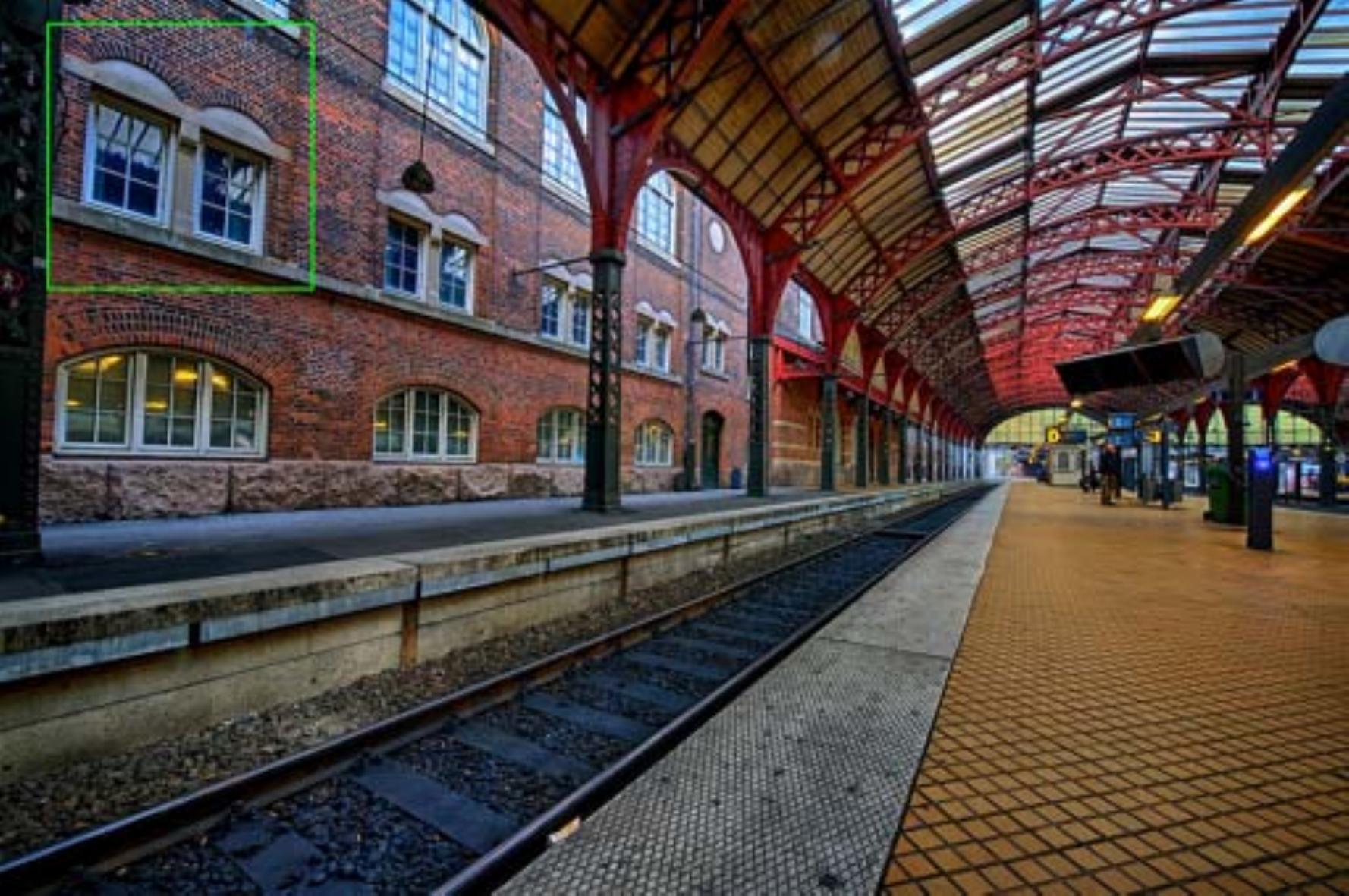}&
		\includegraphics[width=0.19\textwidth]{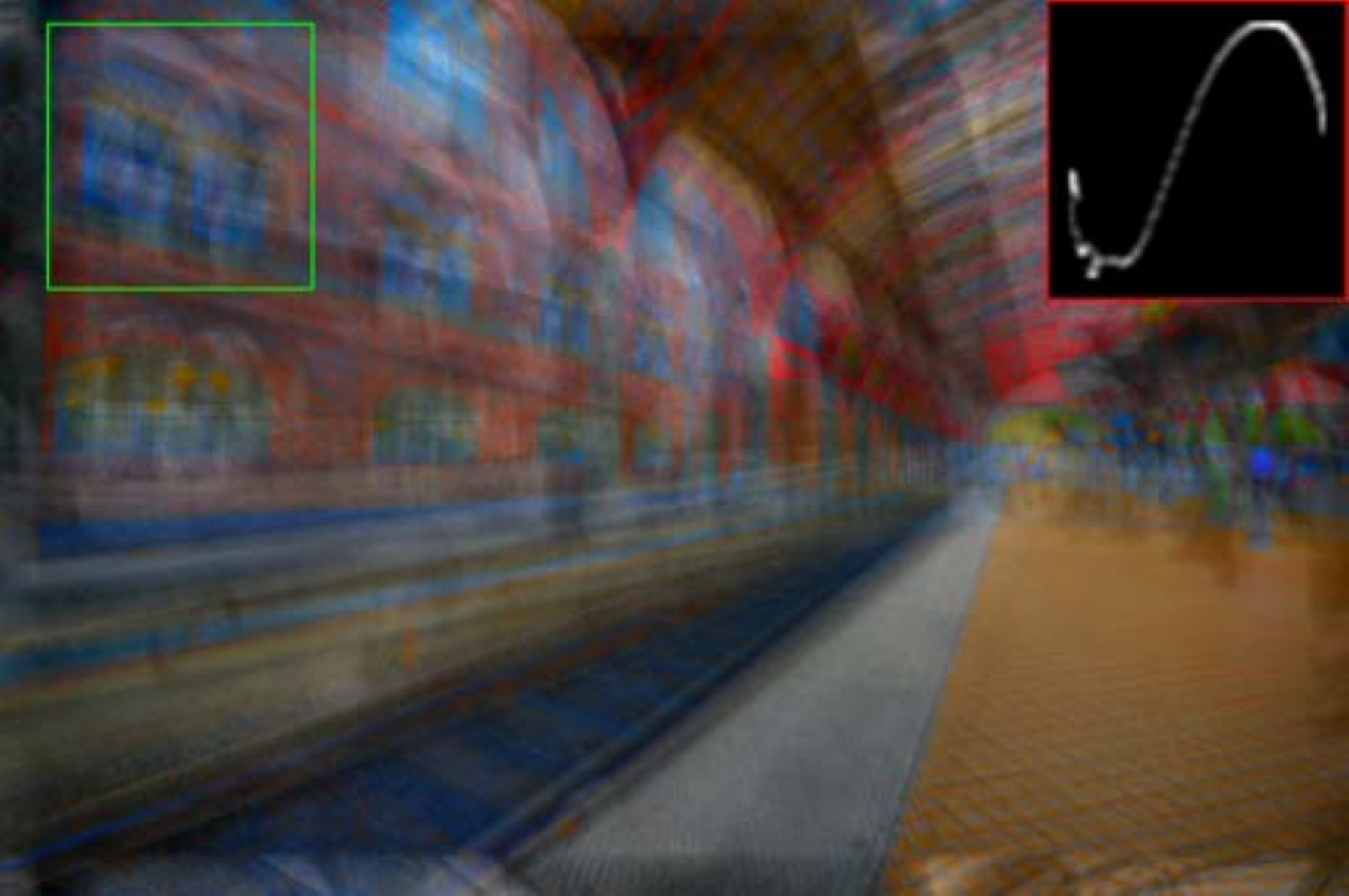}&
		\includegraphics[width=0.19\textwidth]{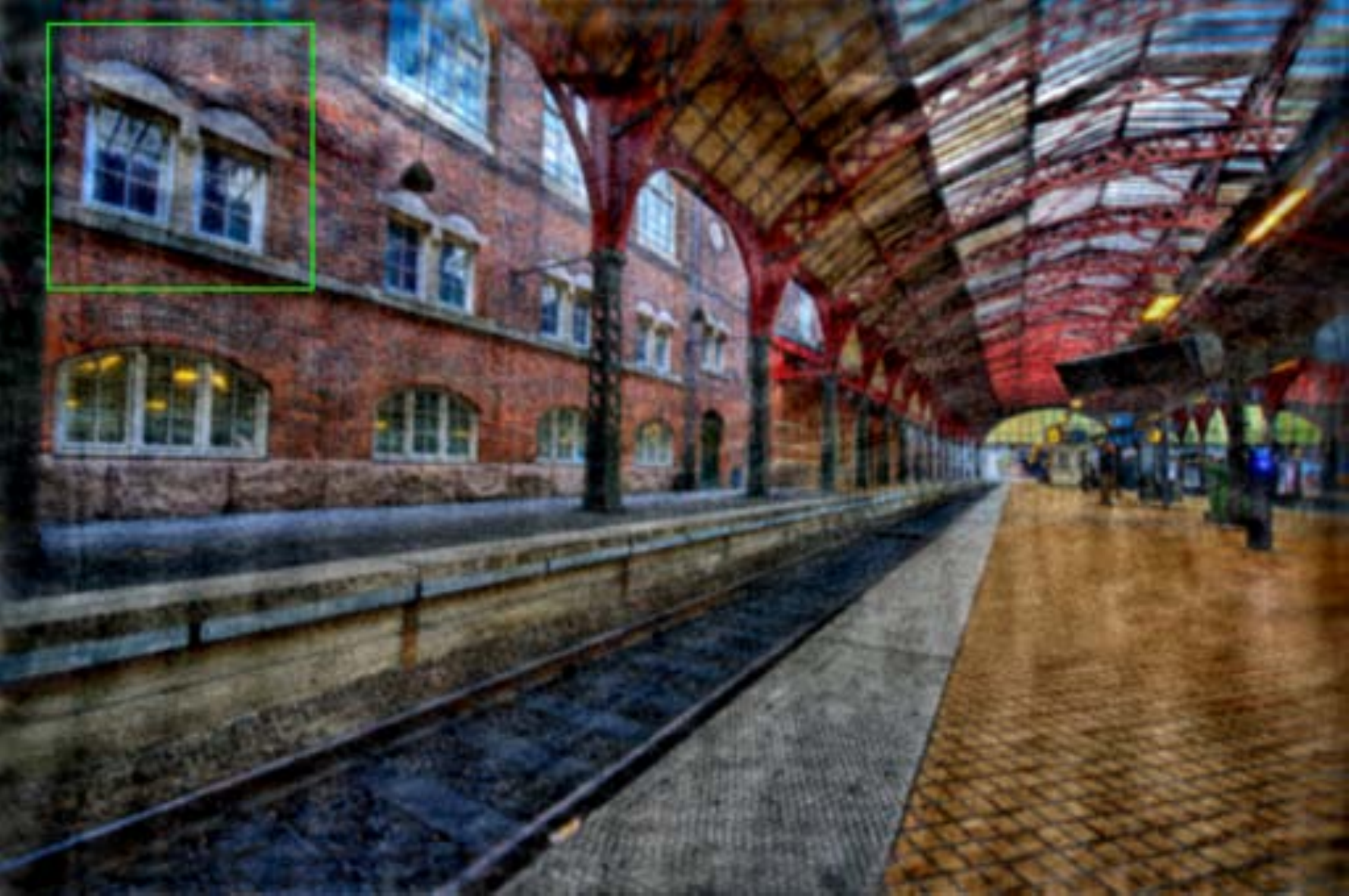}&
		\includegraphics[width=0.19\textwidth]{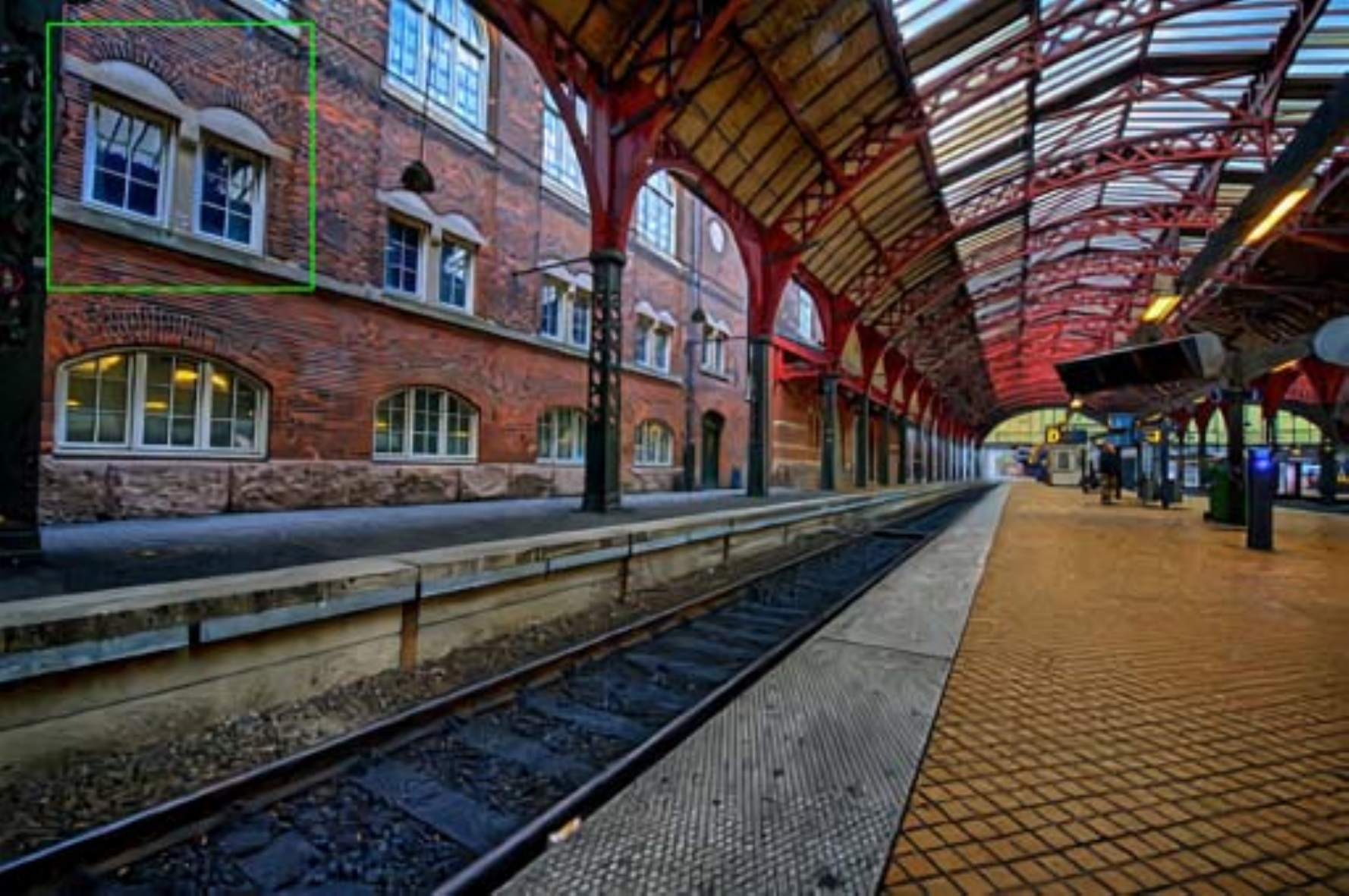}&
		\includegraphics[width=0.19\textwidth]{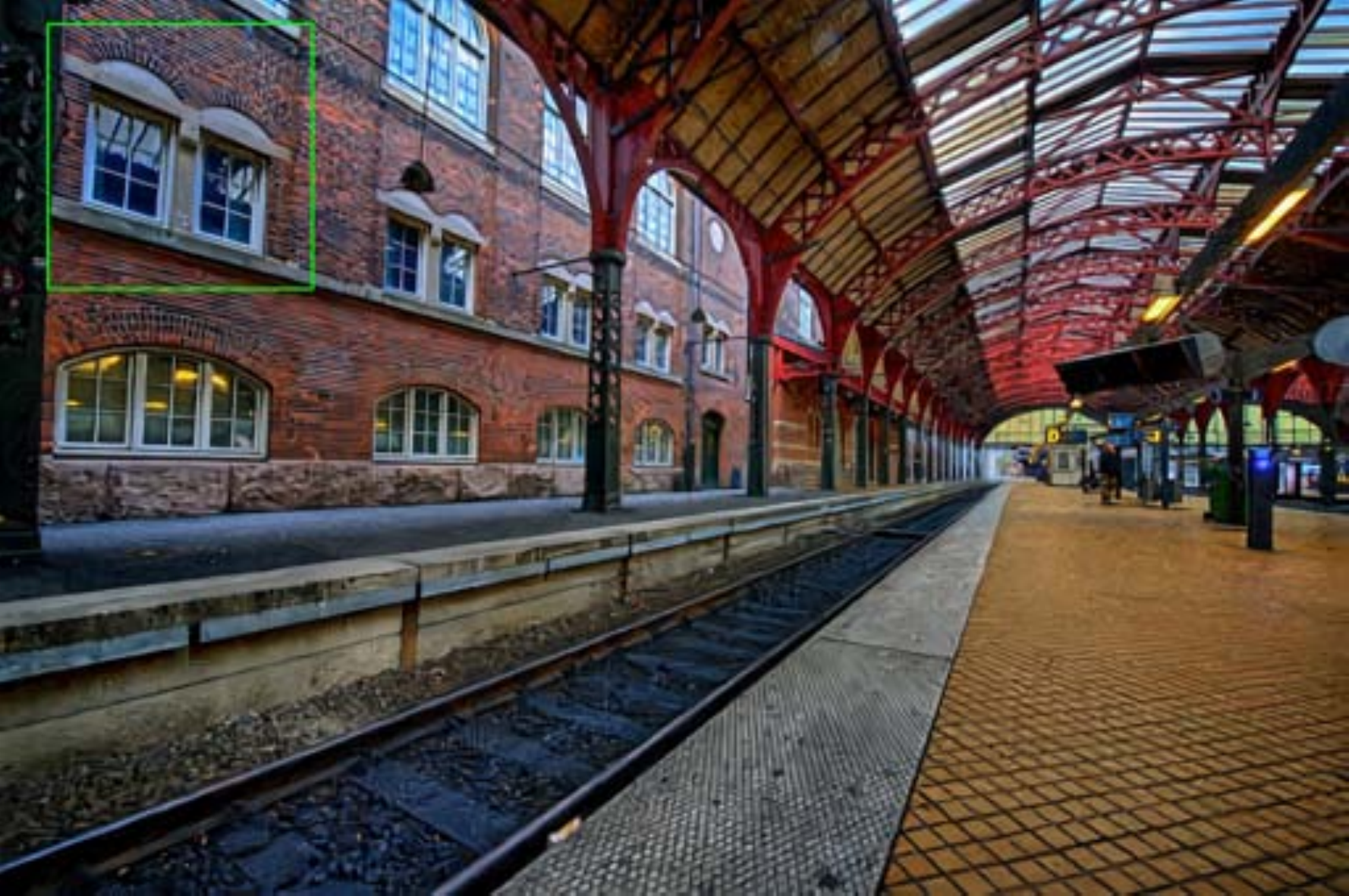}\\
		\textendash &\footnotesize 14.71&\footnotesize 18.78&\footnotesize 22.12 &\footnotesize \textbf{22.22}\\
		\footnotesize Ground Truth &\footnotesize Blurry Input&\footnotesize CSF & \footnotesize IRCNN & \footnotesize Ours
	\end{tabular}
	\caption{The performance of DPE on the whole image in Fig.~\ref{fig:framework} with comparisons to one prior-optimization (i.e., CSF~\cite{schmidt2016cascades}) and one deep network (i.e., IRCNN~\cite{Zhang2017Learning}) approaches. This blurry image is generated by a kernel with the size 75 $\times$ 75 and 1\% Gaussian noise. The PSNR scores are also reported accordingly.}
	\label{fig:stepfullres}
\end{figure*}

\subsection{Our Contributions}

As discussed above, most conventional approaches aim to design, optimize and learn different image priors based on particular understandings of the task. In contrast, recently proposed deep models are established in a heuristic manner and trained on a large number of data pairs. Though relatively good performances have been achieved, there still exist some important limitations in current deep visual enhancement models. For example, the existing network structures and architectures are mostly designed with engineering experiences and their performances are mainly dependent on the scale and quality of training data. However, there are actually rich domain knowledge and physical principles underline low-level vision tasks. On the other hand, the tight data-dependent nature of existing deep neural networks limits their application ranges in complex tasks, in which extremely less or even no high-quality training pairs are available. More importantly, till now it is still challenging to design and/or control the
feed-forward propagation behaviors of existing deep networks with solid theoretical manners.

To mitigate these issues, we in this paper propose a simple, flexible and generic framework, named deep prior ensemble (DPE),
to integrate both knowledge and data-based cues to build theoretically convergent visual propagation for different image enhancement problems.
Specifically, we first design three basic propagative building-blocks, i.e., task-aware warm start, data-dependent residual architecture and prior projection. By cascading these components
with a novel feedback control strategy, it is able to integrate the superiorities of both designed priors and learned descent directions into a unified framework for visual propagation.
Furthermore, we provide rigid theoretical analysis to demonstrate that the feed-forward propagation in {DPE is indeed converged to the desired task-related optimal solution.
In summary, we list our main contributions as the following four items.

\begin{enumerate}
	\item DPE investigates a novel visual propagation scheme to address different image enhancement tasks. At each stage, we can successfully integrate knowledge-driven priors (as warm starts) and fully data-dependent CNNs (as descent directions) for image propagation. A prior projection with error-based feedback control strategy is also introduced to guide the final propagation toward our desired output.
	
	\item We provide a rigid theoretical analysis for the feed-forward propagation behaviors of DPE and prove that even with experience-based network architectures, we can still guarantee the convergence of DPE (to the critical points of our fundamental image modeling energy), only under some mild conditions.
	
	\item On the one hand, we can interpret DPE as a
	data-dependent optimization scheme, in which network architectures are trained to adaptively predict descent directions for iterations, thus automatically avoid unwanted local minimums.
	On the other hand, DPE should also be understood as a theoretically converged recurrent network with prior-based feedback control.
	
	\item Finally, we demonstrate the efficiency of DPE by applying it to various image enhancement tasks, such as image deconvolution, interpolation, super-resolution, single image haze removal and underwater enhancement.
\end{enumerate}

The remainder of this paper is organized as follows. Section~\ref{sec:model} first designs three basic propagation building-blocks and then establishes DPE framework based on a cascade of these components with feedback control. In Section~\ref{sec:analysis}, we provide solid theoretical analysis to prove the convergence of DPE and discuss intrinsic relationships with existing knowledge-based and data-driven approaches. Extensive experiments on different image enhancement tasks are conducted in Section~\ref{sec:exp} to evaluate DPE. Finally, we conclude our work in Section~\ref{sec:con}.

\section{The Proposed Framework}\label{sec:model}

In this section, we first build a general but relatively simple MAP-type energy to unify our fundamental constraints on the problem of visual enhancement. Different from most existing approaches, which directly optimize the energy to obtain their solutions, we then develop a hybrid scheme to combine task-aware and data-dependent information for image propagations. The flowchart and core mechanism of our proposed framework is illustrated in Fig.~\ref{fig:framework}.

\subsection{Fundamental Energy for Visual Enhancement}

Most visual enhancement tasks involve the estimation of the latent image of interest given only an observation, that has been compromised by unknown corruptions. Because these problems are fundamentally ill-posed, conventional approaches often aim to design strong priors to regularize the solution space, resulting to MAP-type estimation energies. While possibly well-motived in principle, standard MAP approaches tightly rely on both correct prior selections and exact inference schemes, which may compromise their performances on real-world challenging tasks.

To mitigate these issues, we in this work first build a general but relatively simple MAP-type energy to unify our fundamental constraints on the problem of visual enhancement. Different from most existing approaches, which directly optimize the energy to obtain their solutions, here we develop a hybrid scheme to combine task-aware and data-dependent information for image propagations.

Specifically, the main purpose of image enhancement is to estimate the latent image $\mathbf{x}\in\mathbb{R}^d$ from an observation $\mathbf{y}\in\mathbb{N}^d$ that may possibly be degraded by corruptions, noises and blur kernels, etc. As mentioned above, we first build a unified MAP-type energy to enforce fundamental constraints on $\mathbf{x}$:
\begin{equation}
\max\limits_{\mathbf{x}} \log p(\mathbf{y}|\mathbf{x}) + \log p(\mathbf{x})
\Rightarrow \min\limits_{\mathbf{x}} f(\mathbf{x};\mathbf{y}) + g(\mathbf{x}).\label{eq:map}
\end{equation}
Here the fidelity term $f(\mathbf{x};\mathbf{y})$ is to measure the discrepancy between the estimated $\mathbf{x}$ and observed $\mathbf{y}$
and has a close relationship with the generative mode of the specific task, meanwhile, representing supposed noise type of specific task~\cite{Yuan2015L0}. The most common choice is $f(\mathbf{x};\mathbf{y})=\frac{1}{2}\|\mathcal{A}(\mathbf{x})-\mathbf{y}\|^2$ with a linear operator $\mathcal{A}$, which can be identity for denoising, mask operation for inpainting or convolution kernel for deblurring. On the other hand, $g(\mathbf{x})$ is known as the regularization term (derived from the image prior
model).
As mentioned above, the energy in Eq.~\eqref{eq:map} only aim to enforce fundamental constraints on the solution, here we just define it by the commonly used non-convex potential function on image gradient~\cite{Roth2009Fields}, i.e., $g(\mathbf{x})=\lambda\log(1+\theta \nabla\mathbf{x})$, with parameters $\lambda$ and $\theta$. Moreover, it is also necessary to consider the range restriction on the image, i.e., we define the feasible set of
$\mathbf{x}$ as
\begin{equation}
\Omega:=\left\{\mathbf{x}|\mathbf{x}\in\mathbb{R}^d, \alpha \leq [\mathbf{x}]_i\leq \beta, i=1,\cdots,d\right\},
\end{equation}
where $[\mathbf{x}]_i$ denotes the $i$-th element of the vectorized variable $\mathbf{x}$, $\mathbb{R}^d$ is the $d$-dimentional Euclidean space, $\alpha$ and $\beta$ are two constants that represent the lower and upper bounds of images, respectively.
Overall, we can summarize our fundamental cues of visual enhancement using the following non-convex and non-smooth energy minimization model:
\begin{equation}\label{eq:image-model}
\min\limits_{\mathbf{x}} \Psi(\mathbf{x};\mathbf{y}):= f(\mathbf{x};\mathbf{y}) + g(\mathbf{x}) + \mathcal{X}_{\Omega}(\mathbf{x}),
\end{equation}
where $\mathcal{X}_{\Omega}$ is the characteristic function of $\Omega$ defined as
\begin{equation}
\mathcal{X}_{\Omega}(\mathbf{x}):=\left\{
\begin{array}{cl}
0 & \mathbf{x}\in\Omega,\\
\infty & \mbox{otherwise}.
\end{array}
\right.
\end{equation}

Intuitively, one may adopt standard numerical algorithms to solve the model in
Eq.~\eqref{eq:image-model}.
Unfortunately, the performances of such direct strategy will be compromised by several issues, including local solutions stemming from non-convexity of the energy and relatively weak prior selections for complex tasks.
In fact, most generic numerical solvers are easily falling into unwanted local minimums and thus fail to find our task-related optimal solutions. Another possible idea is to train end-to-end deep networks to learn the underlying regression relationships between observed $\mathbf{y}$ and desired $\mathbf{x}$, while without taking the obvious characteristics of the task. Although straightforward, it is indeed inadvisable to completely discard the explicit and rich domain knowledge of the tasks, i.e., the MAP-inspired energy. Therefore, it is necessary to integrate the superiorities of both task-aware formulations and data-dependent networks to address visual enhancement tasks. 

\subsection{The Propagation Building-blocks}

In the following, we will design different building-blocks to establish our propagation scheme.
At $k$-th stage\footnote{$k\in\mathbb{N}$ where $\mathbb{N}$ denotes non-negative integers.} the purpose of our propagation is to update $\mathbf{x}^{k+1}$ from $\mathbf{x}^k$ based on $\Psi$. Therefore, we in this part consider the following proximal envelope of $\Psi$ with respect to $\mathbf{x}^k$, i.e.,
\begin{equation}
\Psi_{\eta^k}^k(\mathbf{x};\mathbf{y},\mathbf{x}^k):=\Psi(\mathbf{x};\mathbf{y}) + \frac{\eta^k}{2}\|\mathbf{x}-\mathbf{x}^k\|^2.\label{eq:proximal-model}
\end{equation}

\subsubsection{Warm Start by Fidelity}

we first generate a temporary variable $\dot{\mathbf{x}}^{k+1}$ by minimizing the task-related fidelity with the proximal envelope at $\mathbf{x}^k$ as follows:
\begin{equation}
\dot{\mathbf{x}}^{k+1} = \mathcal{T}^{f}_{\eta^k}(\mathbf{x}^k;\mathbf{y}):= \arg\min_{\mathbf{x}} f(\mathbf{x};\mathbf{y})+\frac{\eta^k}{2}\|\mathbf{x}-\mathbf{x}^k\|^2.\label{eq:warm-start}
\end{equation}
Indeed, by investigating the closed form solution of Eq.~\eqref{eq:warm-start} (with smooth $f$), we can also understand $\dot{\mathbf{x}}^{k+1}$ as a preliminary warm start, generated by fidelity based gradient descent with a step size parameter $\eta^k$.

\subsubsection{Descent by Residual CNN}

Different from conventional optimization techniques, which often perform descent updating based on the (sub)-gradient of the energy, here we adopt residual-type CNN, denoted as $\mathcal{N}(\mathbf{x};\bm{\vartheta})$, to predict directions for our image propagation
\begin{equation}
\ddot{\mathbf{x}}^{k+1} =\dot{\mathbf{x}}^{k+1} - \mathcal{N}(\dot{\mathbf{x}}^{k+1};\bm{\vartheta}),\label{eq:rcnn}
\end{equation}
where $\bm{\vartheta}$ are corresponding network parameters. Conducting in this way, we actually learn an adaptive direction from collected training date. Thus we can recognize this process as a data-dependent deep propagation. Here it is also necessary to emphasize that the role of $\mathcal{N}$ is slightly different from most existing end-to-end CNNs, which aim to directly address the particular enhancement task. In DPE, $\mathcal{N}$ is only a general predictor for propagative direction estimation, thus its training should not be sensitive to specific types of tasks. More details of this architecture will be discussed in Section~\ref{sec:exp}.

\subsubsection{Prior Projection}
Notice that though the principles of the task (by fidelity) and information from training data (by residual CNN) have been used, we still cannot guarantee that the output of our propagation is always in the feasible solution space.
To address this issue, we design a prior projection process to control propagation toward desired outputs:
\begin{equation}
\mathbf{x}^{k+1}=\mathtt{prox}_{\mathcal{X}_{\Omega}}(\ddot{\mathbf{x}}^{k+1}-\nabla\psi_{\eta^k}^k(\ddot{\mathbf{x}}^{k+1};\mathbf{y},\mathbf{x}^k)),\label{eq:prior-project}
\end{equation}
where $\psi_{\eta^k}^k(\mathbf{x};\mathbf{y},\mathbf{x}^k) = f(\mathbf{x};\mathbf{y}) + g(\mathbf{x})+\frac{\eta^k}{2}\|\mathbf{x}-\mathbf{x}^k\|_2^2$ and
$\mathtt{prox}_{\mathcal{X}_{\Omega}}(\cdot)$ denotes the projection operator on $\Omega$.
It will be theoretically verified that this step do can help
bring the output of deep architecture (i.e., $\ddot{\mathbf{x}}^{k+1}$) back to the feasible region of the MAP-inspired formulation, i.e., Eq.~\eqref{eq:image-model}.

\subsection{Deep Prior Ensemble}

Now we are ready to design the formal deep prior ensemble (DPE) scheme. The most straightforward way seems to be cascading all above designed building-blocks as follows
\begin{equation}
\left\{
\begin{array}{l}
\dot{\mathbf{x}}^{k+1}=\mathcal{T}^{f}_{\eta^k}(\mathbf{x}^k;\mathbf{y}),\\
\ddot{\mathbf{x}}^{k+1}=\dot{\mathbf{x}}^{k+1}-\mathcal{N}^k(\dot{\mathbf{x}}^{k+1};\bm{\vartheta}^k), \\
\mathbf{x}^{k+1}=\mathtt{prox}_{\mathcal{X}_{\Omega}}(\ddot{\mathbf{x}}^{k+1}-\nabla\psi_{\eta^k}^k(\ddot{\mathbf{x}}^{k+1};\mathbf{y})).\\
\end{array}
\right.\label{eq:iter}
\end{equation}
From the optimization perspective, Eq.~\eqref{eq:iter} can also be understood as network-incorporated iterations for solving Eq.~\eqref{eq:image-model}. But due to the inexactness in each iteration, it is challenging to directly analyze their propagative behaviors. So a natural question arises: can we design a more deliberate iterative scheme to generate theoretically better (e.g., convergent) propagations for visual enhancement?
By introducing the following sub-gradient bound condition, we can give a positive answer to above question\footnote{Detailed analysis can be found in the following section.}.
\begin{condition}\label{condition}(Sub-gradient Error Bound Condition)
	For $\forall k$, denote the sub-gradient error of $\Psi_{\eta^k}^k$ in Eq.~\eqref{eq:proximal-model} w.r.t. $\mathbf{x}^{k+1}$ as $\mathbf{m}^{k+1} = \mathbf{g}^{k+1} + \nabla\psi_{\eta^{k}}^{k}(\mathbf{x}^{k+1};\mathbf{y},\mathbf{x}^k)$, where $\mathbf{g}^{k+1} \in \partial \mathcal{X}_{\Omega}(\mathbf{x}^{k+1})$. Then we consider the following condition:
	\begin{equation}
	\|\mathbf{m}^{k+1}\| \leq c^{k}\|\mathbf{x}^{k+1}-\mathbf{x}^{k}\|,\label{eq:error-condition}
	\end{equation}
	where $c^{k}\in(0, \eta^{k}/2)$ is a constant.
\end{condition}
In fact, this sub-gradient error can be used as a controller to regularize our propagation behavior. But due to the non-uniqueness of $\mathbf{g}^{k+1}$ for $\mathcal{X}_{\Omega}(\mathbf{x})$, it is hard to check Condition~\ref{condition} in practice.
Therefore, we provide an equivalent expression of $\mathbf{m}^{k+1}$ in the following proposition\footnote{Please refer to Appendix for all the proofs of proposition, theorem and corollary in this paper.}.
\begin{prop}\label{prop:error}
	The error $\mathbf{m}^{k+1}$ can be reformulated as
	\begin{equation}\label{eq:m_form}
	\mathbf{m}^{k+1} = \ddot{\mathbf{x}}^{k+1}-\mathbf{x}^{k+1} + \nabla \psi_{\eta^k}^k(\mathbf{x}^{k+1};\mathbf{y}) - \nabla \psi_{\eta^k}^k(\ddot{\mathbf{x}}^{k+1};\mathbf{y}),
	\end{equation}
	which is more implementable for practical use.
\end{prop}

Finally, we summarize the formal scheme of DPE in Alg.~\ref{alg:DPE}. It can be seen that we just adopt Condition~\ref{condition} as feedbacks to control the prior projection process. Theoretical analysis in the following section will provide nice convergence properties for our proposed DPE.
\begin{algorithm}[!t]
	\caption{Deep Prior Ensemble (DPE)}
	\begin{algorithmic}[1]
		\STATE Given observation $\mathbf{y}$ and parameters $t_{\max}>1$,
		$\{\eta^k\}_{k\in\mathbb{N}}$, and $\{c^k|c^k\in(0, \eta^{k}/2)\}_{k\in\mathbb{N}}$.
		\STATE Initialize: $\mathbf{x}^0=\mathbf{y}$.
		\WHILE {not converged}
		\STATE
		$\dot{\mathbf{x}}^{k+1}\leftarrow\mathcal{T}^{f}_{\eta^k}(\mathbf{x}^k;\mathbf{y})$.
		\STATE
		$\ddot{\mathbf{x}}^{k+1}\leftarrow\dot{\mathbf{x}}^{k+1}-\mathcal{N}^k(\dot{\mathbf{x}}^{k+1};\bm{\vartheta}^k)$.	
		\FOR{$t=1, \cdots, t_{\max}$}
		\STATE $\mathbf{x}_t^{k+1}\leftarrow\mathtt{prox}_{\mathcal{X}_{\Omega}}(\ddot{\mathbf{x}}^{k+1}-\nabla\psi_{\eta^k}^k(\ddot{\mathbf{x}}^{k+1};\mathbf{y}))$.
		\STATE Calculate $\mathbf{m}_t^{k+1}$ using Eq.~\eqref{eq:m_form} with $\mathbf{x}_t^{k+1}$.
		\IF{$\mathbf{m}_t^{k+1} \leq c^{k}\|\mathbf{x}_t^{k+1}-\mathbf{x}^{k}\|$ or $t=t_{\max}$}
		\STATE $\mathbf{x}^{k+1}\leftarrow\mathbf{x}_t^{k+1}$ and break.
		\ELSE
		\STATE $\ddot{\mathbf{x}}^{k+1}\leftarrow\mathbf{x}_t^{k+1}$.
		\ENDIF
		\ENDFOR
		\ENDWHILE
	\end{algorithmic}\label{alg:DPE}
\end{algorithm}

\section{Theoretical Analysis and Discussions}\label{sec:analysis}

\subsection{Propagative Behaviors Analysis}
We first analyze the propagative behaviors of DPE from an optimization perspective (e.g., convergence and rate) based on the fundamental energy model in Eq.~\eqref{eq:image-model}.
Before providing the main convergence results, we in the following proposition shows that the $\mathbf{x}^{k+1}$ propagated by our proposed DPE, i.e., the highly nonlinear operations in Eq. \eqref{eq:iter}, can be regarded as the solution of an error-penalized abstract optimization model.

\begin{prop}\label{prop:equiv}
The $\mathbf{x}^{k+1}$ calculated by Eq. \eqref{eq:iter}, can be regarded as the solution of an error-penalized abstract optimization model, that is,
\begin{equation}\label{eq:equiv_form}
\mathbf{x}^{k+1} \in \arg\min_{\mathbf{x}} \psi_{\eta^k}^k(\mathbf{x};\mathbf{y}) + \mathcal{X}_{\Omega}(\mathbf{x}) - (\mathbf{m}^{k+1})^{\top}\mathbf{x}.
\end{equation}
\end{prop}
Here we must emphasize that this equivalent reformulation will only be used for theoretical analysis, and thus it will not be practically solved by our proposed method.

\begin{thm}\label{thm:converge}
	Suppose that $\{\mathbf{x}^k\}_{k\in\mathbb{N}}$ is a sequence generated by Alg.~\ref{alg:DPE}.
	Then we have the following assertions.
	\begin{enumerate}
		\item The energy in Eq.~\eqref{eq:image-model} is sufficient descent, i.e.,
		\begin{equation}\label{eq:suff_des}
		\Psi(\mathbf{x}^k;\mathbf{y}) - \Psi(\mathbf{x}^{k+1};\mathbf{y}) \geq \alpha_1\|\mathbf{x}^{k+1}-\mathbf{x}^k\|^2,
		\end{equation}
		with $\alpha_1 := \min_{k\in\mathbb{N}}\{\frac{\eta^k}{4}-\frac{(c^k)^2}{\eta^k}\}$.
		Moreover, the propagation process is bounded during iterative stages, namely, the sequence $\{\mathbf{x}^k\}_{k\in\mathbb{N}}$ propagated by DPE is bounded.
		
		\item The sub-gradient of $\Psi(\mathbf{x}^k;\mathbf{y})$ satisfies
		\begin{equation}\label{eq:subgrad_bound}
		\|\partial \Psi(\mathbf{x}^k;\mathbf{y})\|\leq\alpha_2\|\mathbf{x}^k-\mathbf{x}^{k-1}\|,
		\end{equation}
		with $\alpha_2 := \max_{k\in\mathbb{N}}\{\eta^k+c^k\}$.
		
		\item Any limit point $\mathbf{x}^{\ast}$ of the propagation sequence $\{\mathbf{x}^k\}_{k\in\mathbb{N}}$ is a critical point of the objective function $\Psi(\mathbf{x};\mathbf{y})$, i.e., $\mathbf{0} \in \partial \Psi(\mathbf{x}^{\ast};\mathbf{y})$; furthermore, $\Psi(\mathbf{x};\mathbf{y})$ is constant on the set of all limit points of the sequence $\{\mathbf{x}^k\}_{k\in\mathbb{N}}$.
	\end{enumerate}
\end{thm}

Notice that in nonconvex scenario, the critical point is just the necessary condition to the local optimal solution.
Based on this theorem, we can further prove in the following corollary that the propagations generated by DPE can obtain a preferable solution of our fundamental visual enhancement energy.
\begin{corollary}\label{cor:converge}
	By verifying the semi-algebraic property of $\Psi(\mathbf{x};\mathbf{y})$~\cite{Bolte2014Proximal}, we have that $\{\mathbf{x}^k\}_{k\in\mathbb{N}}$ is a Cauchy sequence, thus the whole sequence converges to the critical point of our fundamental visual enhancement energy. Furthermore, we can obtain the convergence rate of DPE, based on a particular desingularizing function $\psi(s)=cs^{1-\theta}$ with $c>0$ and $\theta\in[0,1)$.
	Specifically, the sequence converges after finite iterations if $\theta = 0$; the linear and sub-linear rates can be obtained if function $\psi(s)$ faces the case of $\theta \in (0, 1/2]$ and $\theta \in (1/2, 1)$, respectively.
\end{corollary}

In summary, the analyses in Theorem~\ref{thm:converge} and Corollary~\ref{cor:converge} actually verify the core mechanism of our hybrid propagation. That is, we integrate both domain-knowledge (i.e., task-aware warm start in Eq.~\eqref{eq:warm-start}) and information from training data (i.e., descent by residual CNN in Eq.~\eqref{eq:rcnn}) to generate a propagation scheme. By adaptive controlling the prior projection (i.e., prior projection in Eq.~\eqref{eq:prior-project} and Condition~\ref{condition}), we can guarantee that the final output of DPE always satisfies our
fundamental constrains on visual enhancement tasks (i.e., the critical point of energy in Eq.~\eqref{eq:image-model}).

\subsection{Relationships with Existing Approaches}
As discussed above, our DPE actually provides a flexible ensemble framework to integrate both conventional knowledge-driven cues (MAP-type models) and data-based priors (deep networks) for image propagation.
Actually, DPE should be understood as either a network driven prior optimization scheme (as an analogy to
conventional optimization algorithms) or knowledge guided deep model (as an analogy to standard experience-based CNNs). In the following, we will discuss these relationships in detail.

\subsubsection{Network Driven Prior Optimization}

Conventional prior-based approaches often design and optimize MAP-type models to generate propagations for visual enhancement.
From this perspective, DPE can also be understood as an inexact, network-guided iterative scheme for minimizing the energy formulation in Eq.~\eqref{eq:image-model}.
It seems like that standard optimization techniques~\cite{Attouch2010Proximal}\cite{lin2011linearized} can also be used for solving Eq.~\eqref{eq:image-model}. However,
most of their updating schemes are designed based on the condition $\mathbf{m}^{k+1}=\mathbf{g}^{k+1} + \nabla\psi_{\eta^{k}}^{k}(\mathbf{x}^{k+1};\mathbf{y}) = 0$. In contrast, recall that DPE actually enforce the condition $\|\mathbf{m}^{k+1}\|\leq c^{k}\|\mathbf{x}^{k+1}-\mathbf{x}^{k}\|$ for our propagation. Hence these standard prior optimization approaches should be regarded as special cases of DPE with more strict assumptions, since the inequality in Condition~\ref{condition} is invariably satisfied during stages of propagation. Taken in this sense, it is also quite obvious that our propagation is more flexible than the standard exact optimization schemes.
More importantly, as illustrated in Figs.~\ref{fig:framework} and \ref{fig:stepfullres}, thanks to such flexibility, we can introduce residual CNN to generate mappings from current stage to the next stage, so that directly learn descent directions from training data for our propagation. Notice that such data-dependent iteration scheme is
still with nice convergence properties.

\subsubsection{Knowledge Guided Deep Model}
Though end-to-end deep learning approaches have obtained promising performance for relative simple image enhancement tasks, such as denoising and super-resolution~\cite{Xu2014Deep,Kim2016Accurate,Yan2016Blind,Yang2016Joint}, their performances are tightly depended on the training data. This is because little or no task information can be revealed in their designed networks.
Moreover, it is also challenging to establish network structures to directly learn end-to-end mappings for complex physical principles in image enhancement tasks, e.g., image deconvolution with large size blur kernel.
In contrast, our DPE actually provide a new perspective to build deep models using domain knowledge for enhancement tasks. In other words, task-aware processes in DPE can not only significantly reduce the training complexity of the network, but also control the convergence for our final propagation.
\begin{table}[t]
	\centering
	\caption{Summary of image enhancement tasks. Here (A)-(E) denote non-blind deconvolution, image interpolation, super-resolution, single image haze remove and underwater image enhancement, respectively.}
	\begin{tabular}{|c|c|c|}
		\hline
		Task & Problem Formulation & Fidelity\\\hline
		(A) & $\mathbf{y}=\mathbf{k}\otimes\mathbf{x}+\mathbf{e}$ & $\|\mathbf{k}\otimes\mathbf{x}-\mathbf{y}\|^2$\\\hline
		(B) & $\mathbf{y}=\mathbf{M}\odot\mathbf{x}+\mathbf{e}$ & $\|\mathbf{M}\odot\mathbf{x}-\mathbf{y}\|^2$\\\hline
		(C) & $\mathbf{y}=\mathbf{S}\mathbf{H}\mathbf{x}+\mathbf{e}$ & $\|\mathbf{S}\mathbf{H}\mathbf{x}-\mathbf{y}\|^2$\\\hline
		(D) & \multirow{2}*{$\begin{array}{c}
			\mathbf{I}=\mathbf{x}\odot\mathbf{J} + (\mathbf{1}-\mathbf{x})\odot\mathbf{A}\\
			\mathbf{y}=\mathbf{x}+\mathbf{e}
			\end{array}$}&\multirow{2}*{$\|\mathbf{x}-\mathbf{y}\|^2$}\\\cline{1-1}
		(E) &&\\\hline
	\end{tabular}
	\label{tab:summary}
\end{table}

\section{Experimental Results}\label{sec:exp} 	
We first verify our developed theoretical results on visual propagation and then compare the proposed DPE framework with state-of-the-art approaches on various image enhancement tasks, including non-blind deconvolution, image interpolation, super-resolution, single image haze removal and underwater enhancement. All these experiments are conducted on a PC with Intel Core i7 CPU at 3.6GHz, 32 GB RAM and a NVIDIA GeForce GTX 1050 Ti GPU.

\subsection{Experimental Setting}

\subsubsection{Summary of Enhancement Tasks}
Table~\ref{tab:summary} summarizes the formulations and fidelities for different enhancement tasks considered in this work. Specifically, $\mathbf{x}$, $\mathbf{y}$ and $\mathbf{e}$ are denoted as latent image, corrupted observation and errors/noises for non-blind deconvolution (A), image interpolation (B)
and super-resolution (C), respectively. Here $\mathbf{k}$ in (A) denotes the blur kernel and $\otimes$ is the convolution. As for (B), $\mathbf{M}$ denotes the mask and $\odot$ represents the dot product, while in (C) $\mathbf{S}$ and $\mathbf{H}$ are the binary sampling matrix and circulant matrix representing the convolution for the anti-aliasing filter, respectively. Notice that both haze removal and underwater enhancement tasks are based on the atmospheric scattering model (i.e., $\mathbf{I}=\mathbf{x}\odot\mathbf{J} + (\mathbf{1}-\mathbf{x})\odot\mathbf{A}$), in which $\mathbf{A}$ is the global atmospheric light, $\mathbf{I}$, $\mathbf{J}$ and $\mathbf{x}$ are
the color observation, the latent scene radiance and the medium transmission, respectively. So we consider $\mathbf{y}$ as the initial transmission estimation and $\mathbf{e}$ the errors in these two tasks.

\begin{figure}[t]
	\centering
	\begin{tabular}{c@{\extracolsep{0.05em}}c@{\extracolsep{0.05em}}c@{\extracolsep{0.05em}}c}
		\includegraphics[width=0.118\textwidth]{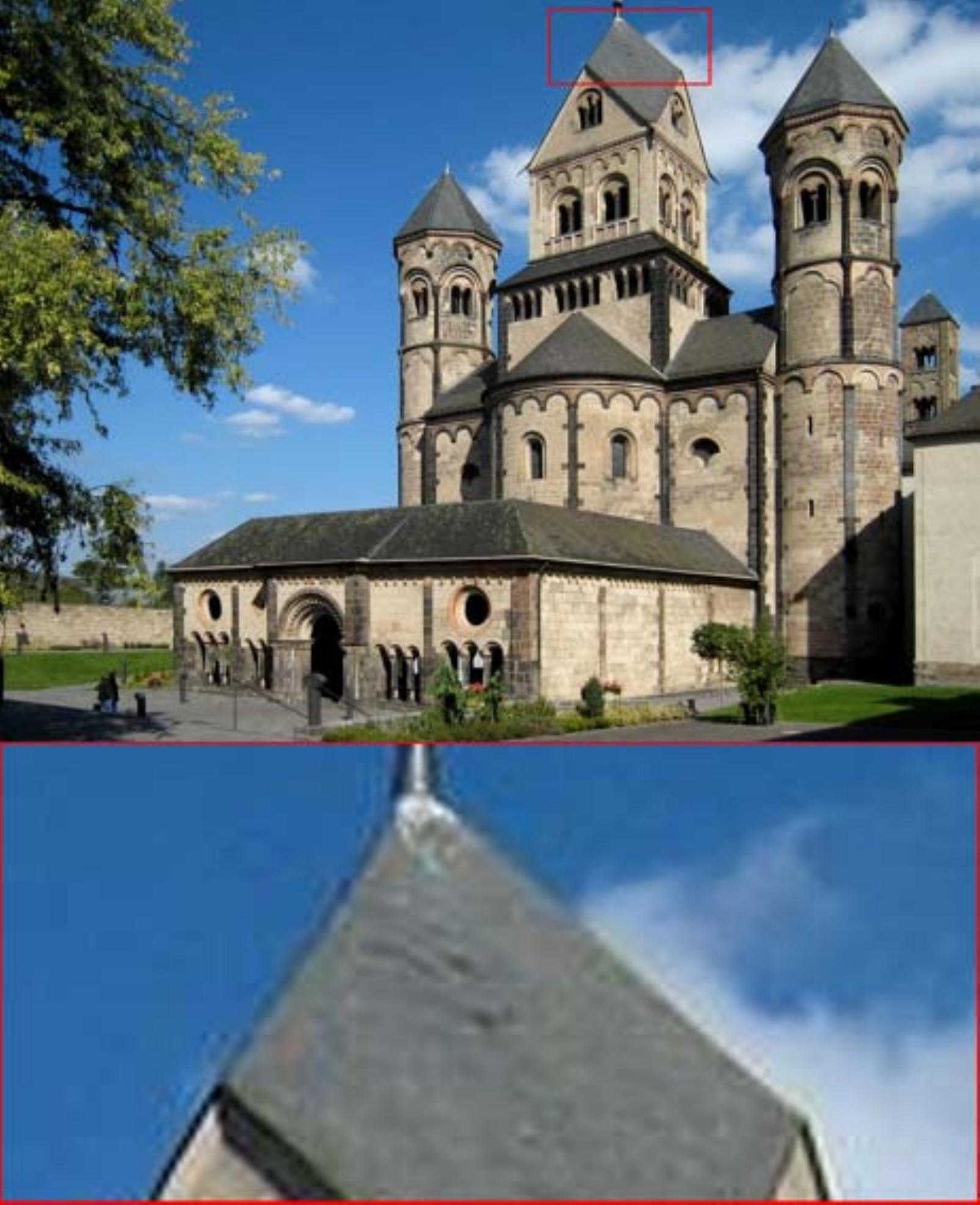}&
		\includegraphics[width=0.118\textwidth]{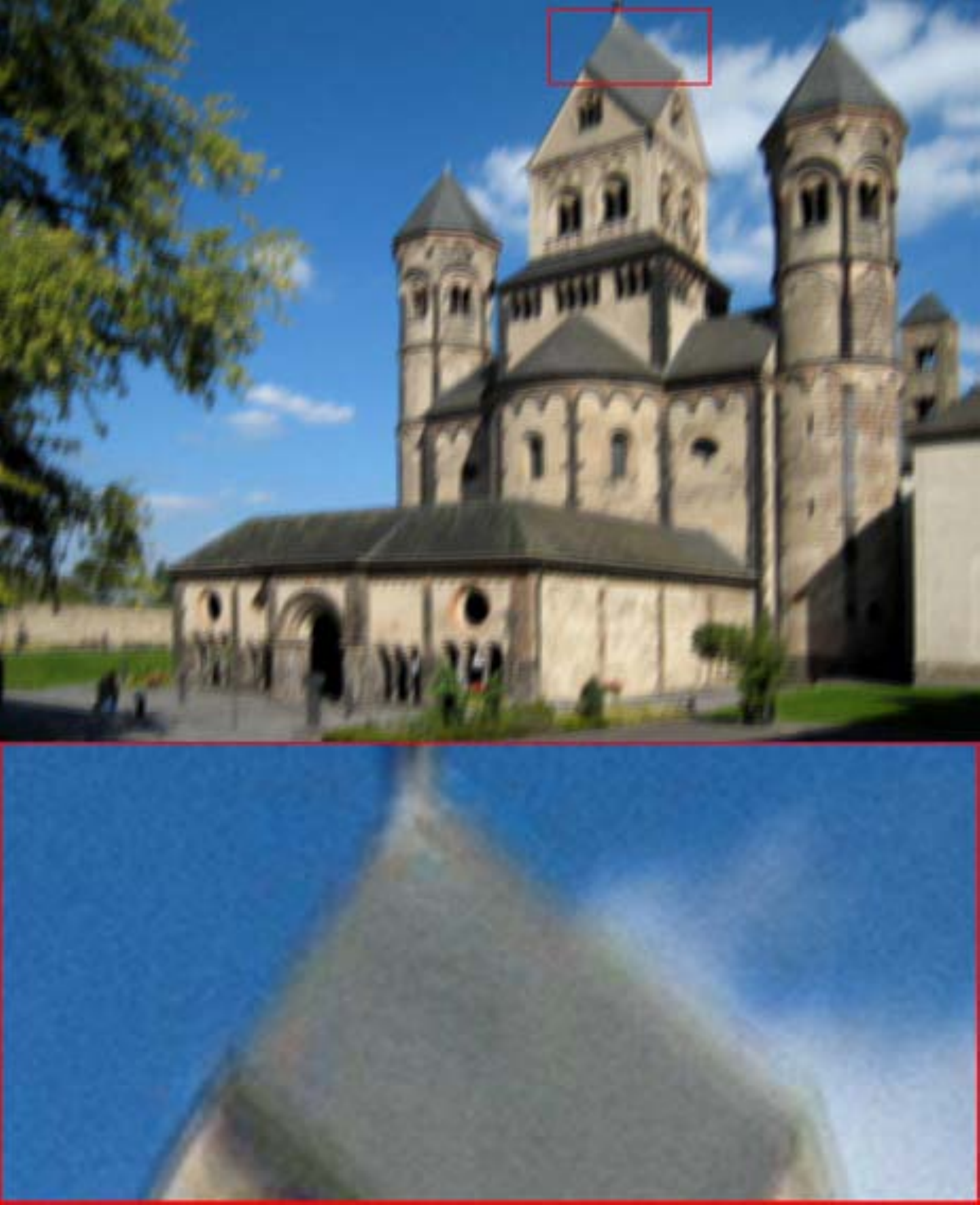}&
		\includegraphics[width=0.118\textwidth]{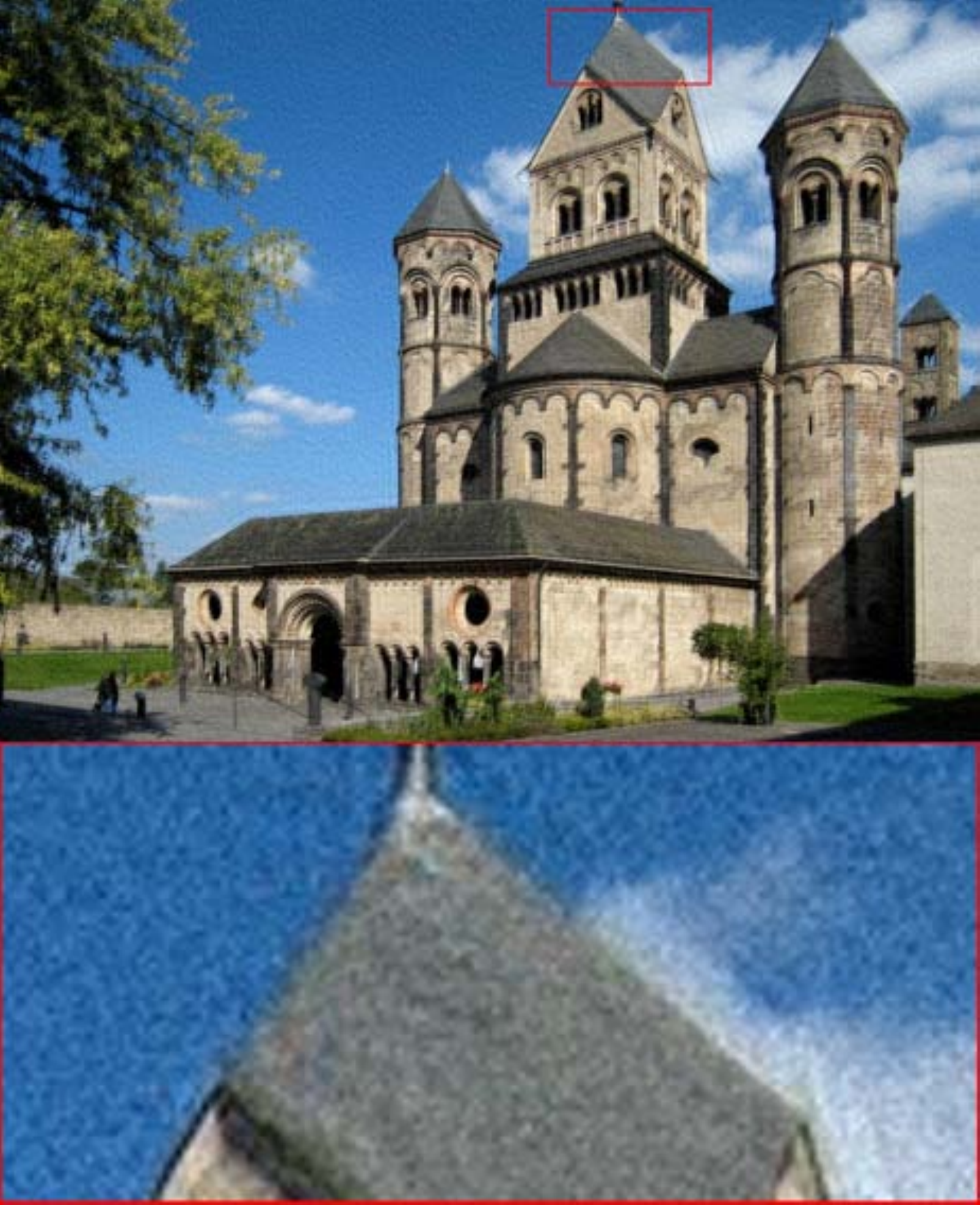}&
		\includegraphics[width=0.118\textwidth]{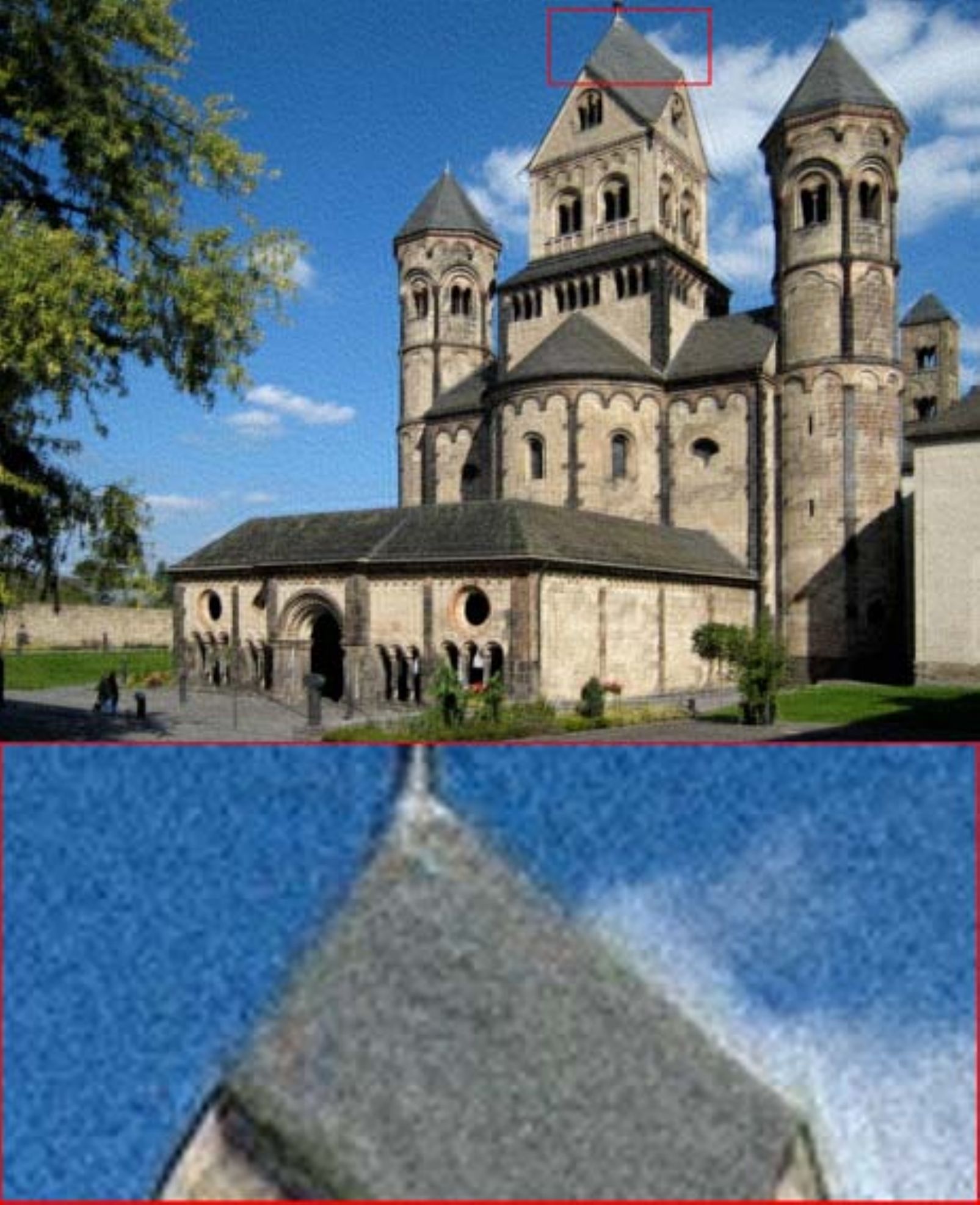}\\
		\textendash &\footnotesize 18.49&\footnotesize 26.23&\footnotesize 26.34\\
		\footnotesize Ground Truth & \footnotesize Blurry Input & \footnotesize GD &\footnotesize PG \\
		\includegraphics[width=0.118\textwidth]{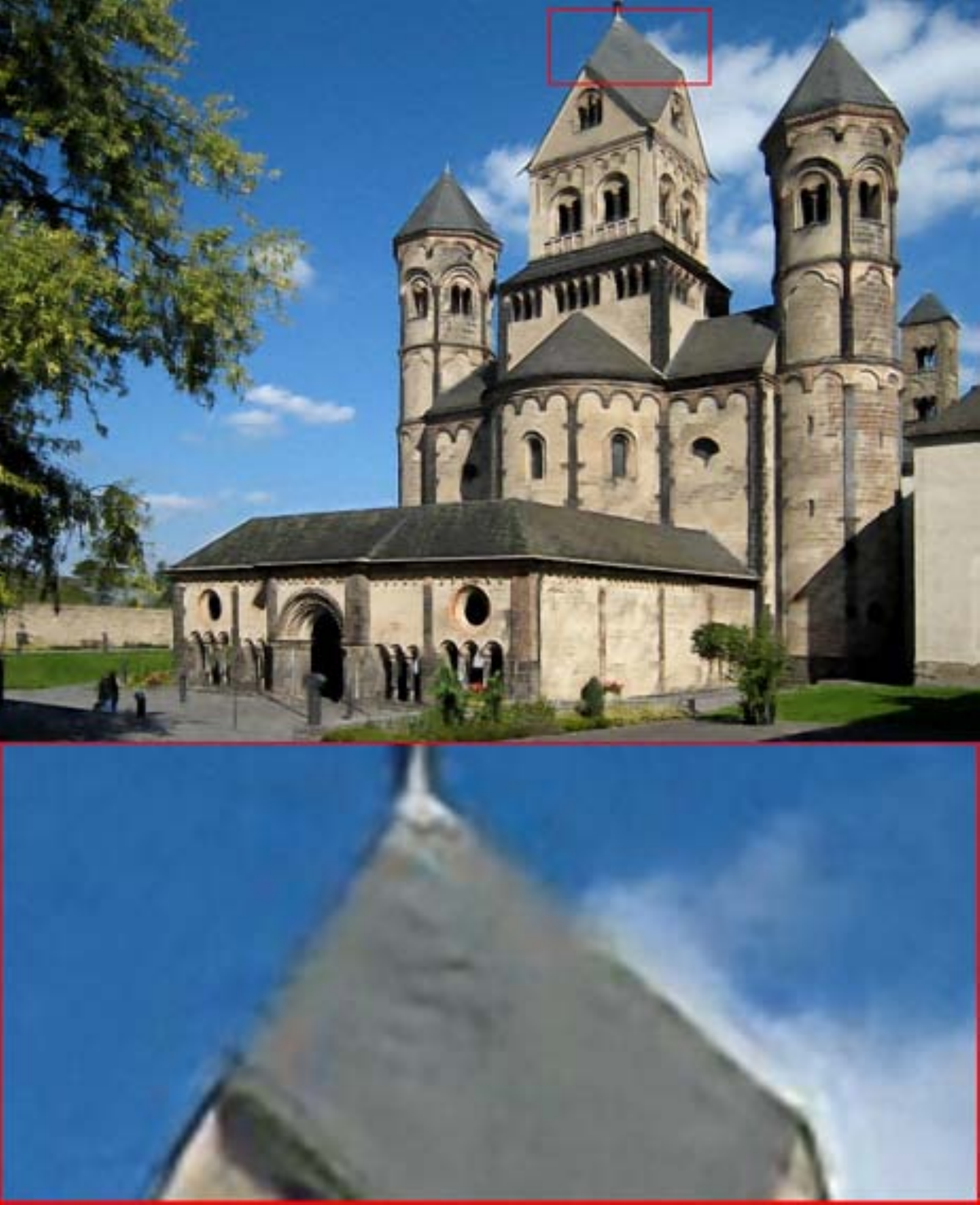}&
		\includegraphics[width=0.118\textwidth]{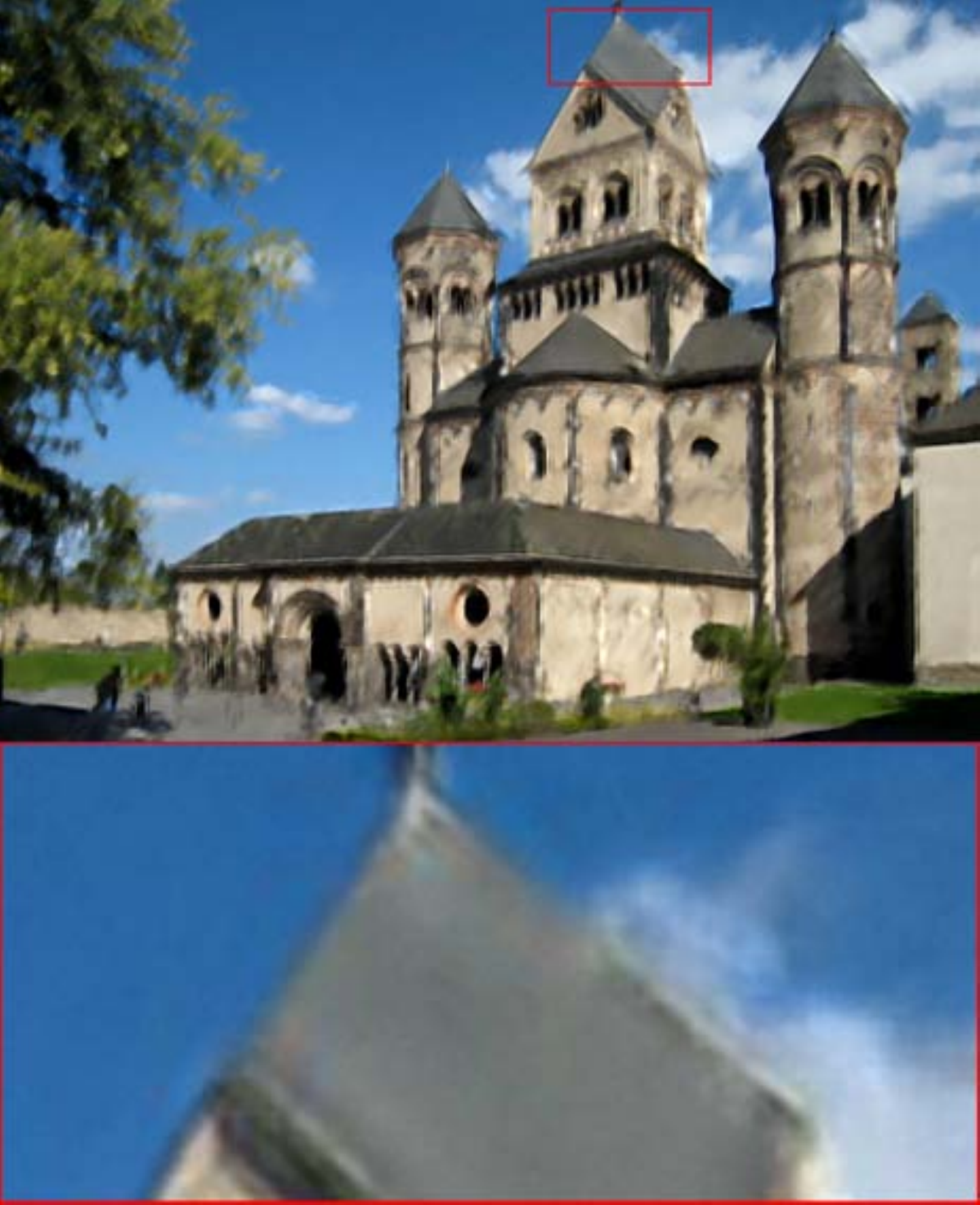}&
		\includegraphics[width=0.118\textwidth]{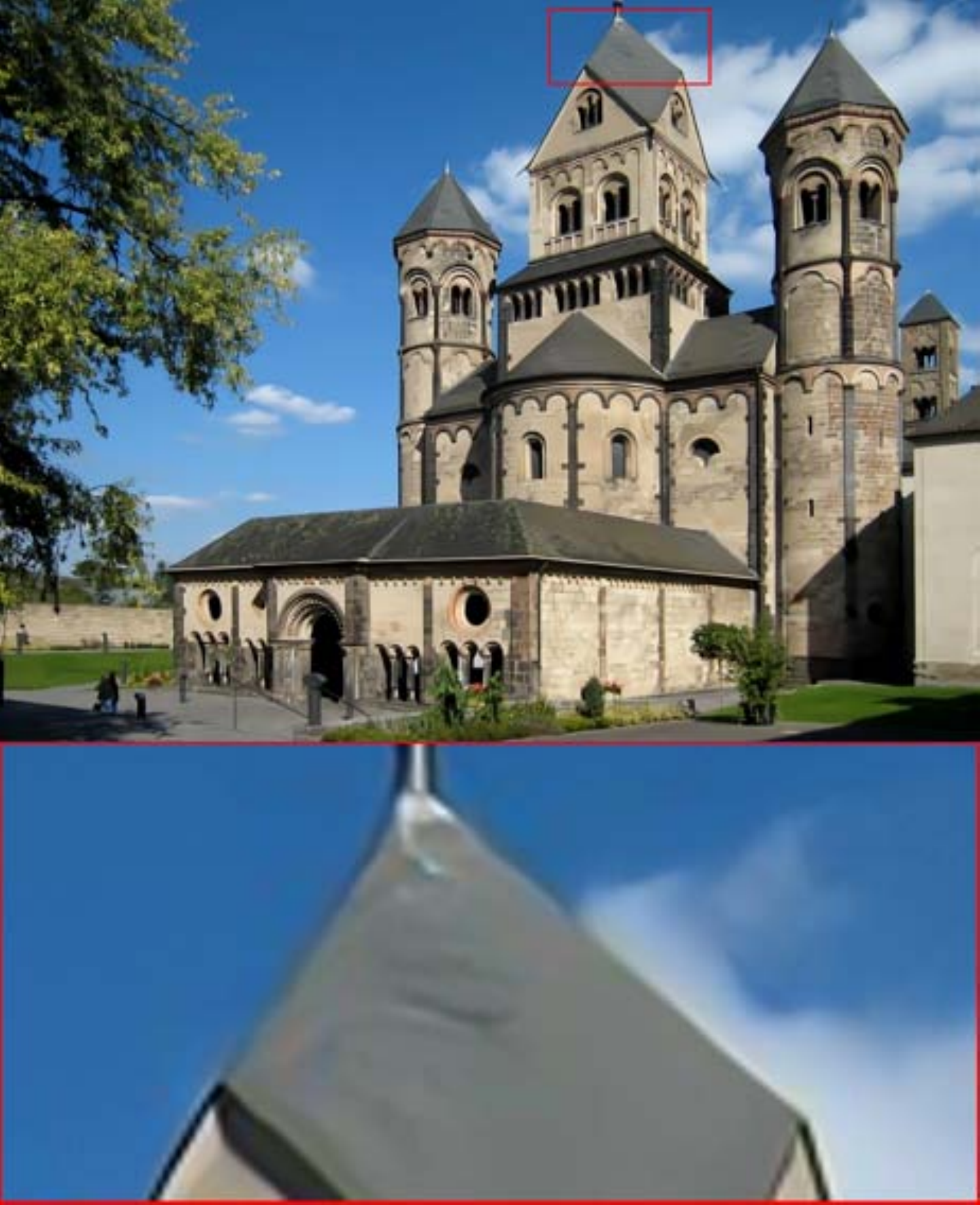}&
		\includegraphics[width=0.118\textwidth]{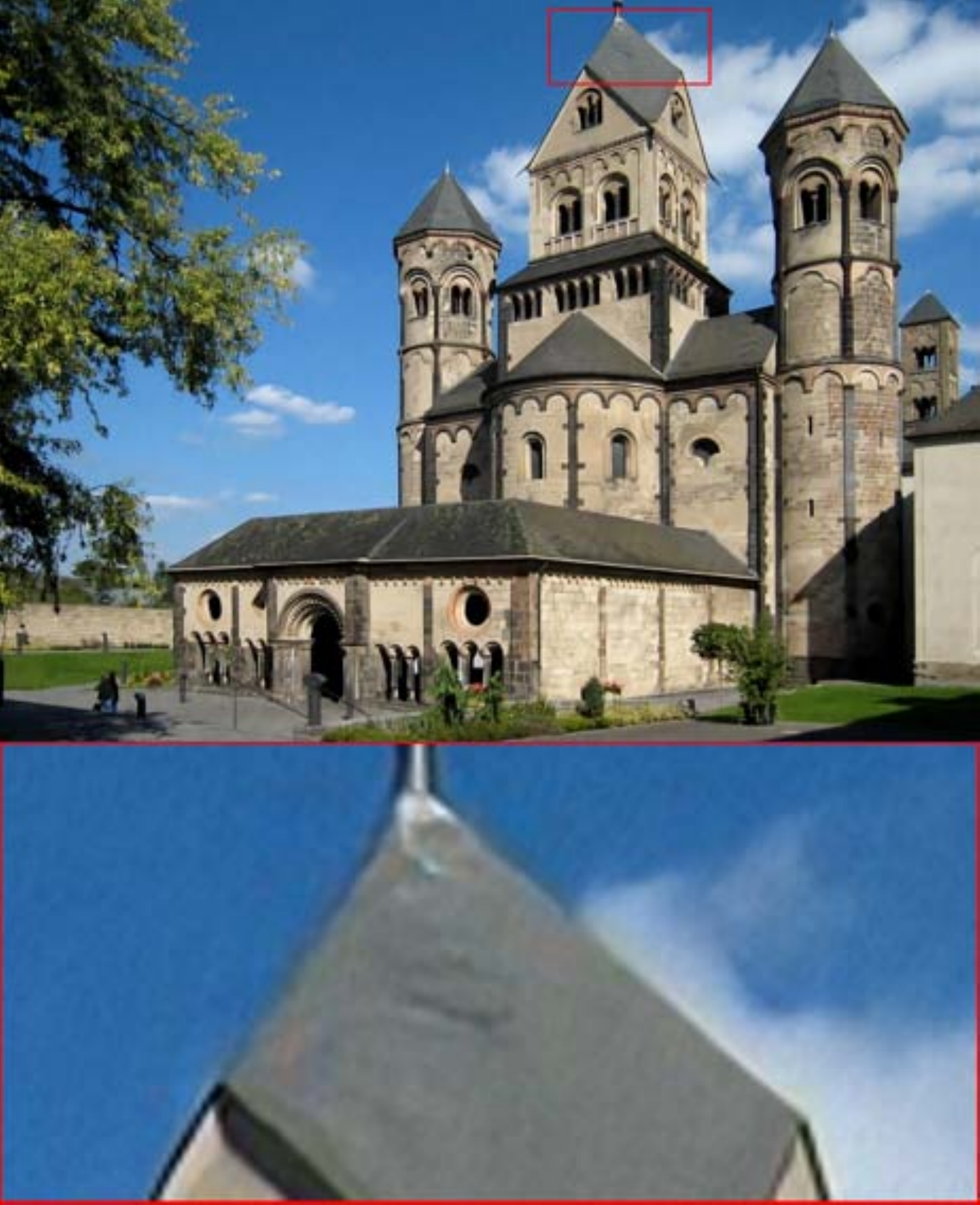}\\
		\footnotesize 25.79&\footnotesize 20.88&\footnotesize 29.95&\footnotesize 30.03\\
		\footnotesize SK-Net &\footnotesize MK-Net &\footnotesize S-DPE & \footnotesize C-DPE\\
	\end{tabular}
	\caption{Visual comparisons of different image propagation mechanisms. The PSNR scores are also reported accordingly. All the abbreviations are the same as that in Table~\ref{tab:nbcompare1}.}			
	\label{fig:effeciency}
\end{figure}

\subsubsection{Propagative Architectures and Training Strategies}
As discussed above, different from most existing end-to-end networks, which directly address the enhancement tasks, here our architectures are established to discover the propagative directions from collected dataset. Therefore, we build a negative type residual block (i.e., $\mathbf{x}-\mathcal{N}(\mathbf{x})$), in which $\mathcal{N}$ consists of $7$ dilated convolution layers (with filter size $3\times 3$). The ReLU nonlinearities are added between each two linear layers accordingly. We also introduce batch normalizations for convolution operations from $2$-nd to $6$-th linear layers.

In general, our propagations should be always toward natural image/transmission distributions. Moreover, the descent directions should also have the ability to removal propagative errors during iterations. Therefore, we would like to add different levels of Gaussian noise to our desired image/transmission to synthesize these propagative errors. Specifically, we train $\mathcal{N}$ on noise level range $[0, 20]$ (divided by a step size of $2$), resulting in a set of $10$ candidate propagative architectures. Then we incorporate these $\mathcal{N}$ into DPE following criterion designed in Alg.~\ref{alg:DPE} to build our formal propagation scheme.

As for our particular training data, they are generated based on the propagation properties of these enhancement tasks. That is, the applications considered in the following can be divided into two categories: image propagation (i.e., non-blind image deconvolution, image interpolation and super-resolution) and transmission propagation (i.e., single image haze removal and underwater image enhancement). 
For the first category of tasks, we collect 800 images, in which 400 are from~\cite{chen2017trainable} and the other 400 are from ImageNet database~\cite{5206848}. We crop them into small patches of size $35\times 35$ and select $N=256\times4000$ patches for training. While 
for transmission-based applications, we utilize NYU dataset~\cite{Silberman}, including 1449 depth images, to synthesize transmission maps. Then we crop them into patches of size $80\times80$ and select $N = 128\times4000$ patches as our training data.

\subsection{Model Analysis and Verification}

We in this section first conduct experiments on standard non-blind deconvolution task to investigate the properties of DPE and verify our theoretical results.

\subsubsection{Mechanism Comparisons}

Here we consider different image propagation mechanisms, including conventional prior-optimization strategies and fully-data-dependent deep networks, to address non-blind image deconvolution problems. These experiments are conducted on the most widely used Levin et al. dataset, with 32 blurry images of size $255\times 255$~\cite{Levin2009Understanding} and Sun et al.' dataset, with 640 blurry images with $1\%$ Gaussian noises, sizes range from $620\times1024$ to $928\times1024$~\cite{Sun2013Edge}.
As for prior-optimization, we adopt two different strategies, including gradient descent iterations for the smooth prior model in Eq.~\eqref{eq:map} (denoted as GD) and proximal gradient scheme for the non-smooth model in Eq.~\eqref{eq:image-model} (denoted as PG). In contrast, we also train our residual network building-blocks $\mathcal{N}$ on different blurry datasets. That is, we generate blurry training data using a single kernel, which is the same as the test one (denoted as SK-Net) or multiple kernels (denoted as MK-Net). In this experiment, we also consider our DPE with two different settings, i.e., propagation without prior projection (denoted as S-DPE) and with prior projection (denoted as C-DPE).

\begin{figure}[t]
	\centering
	\begin{tabular}{c@{\extracolsep{0.4em}}c}
		\includegraphics[height=0.17\textwidth]{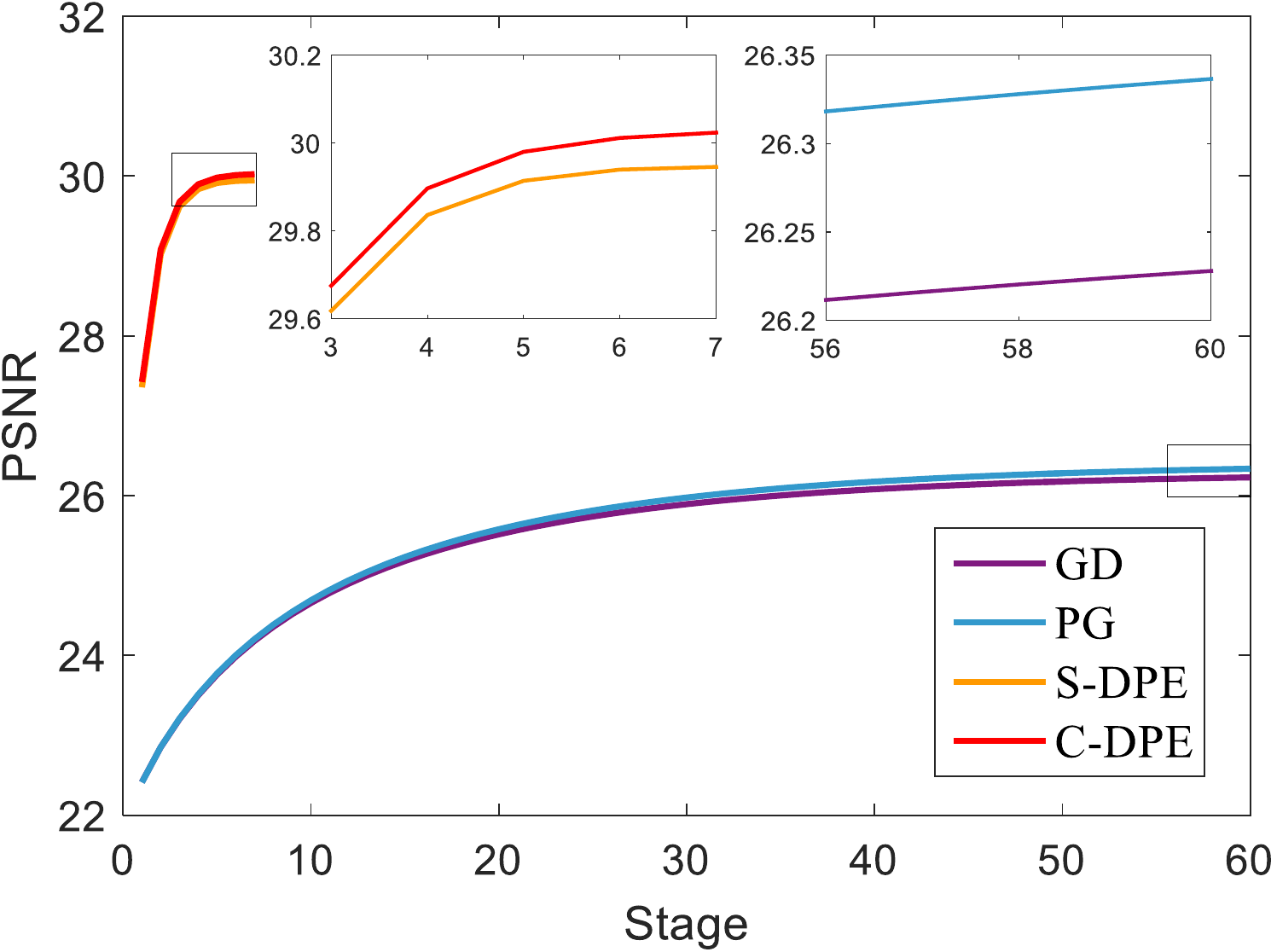}
		&\includegraphics[height=0.17\textwidth]{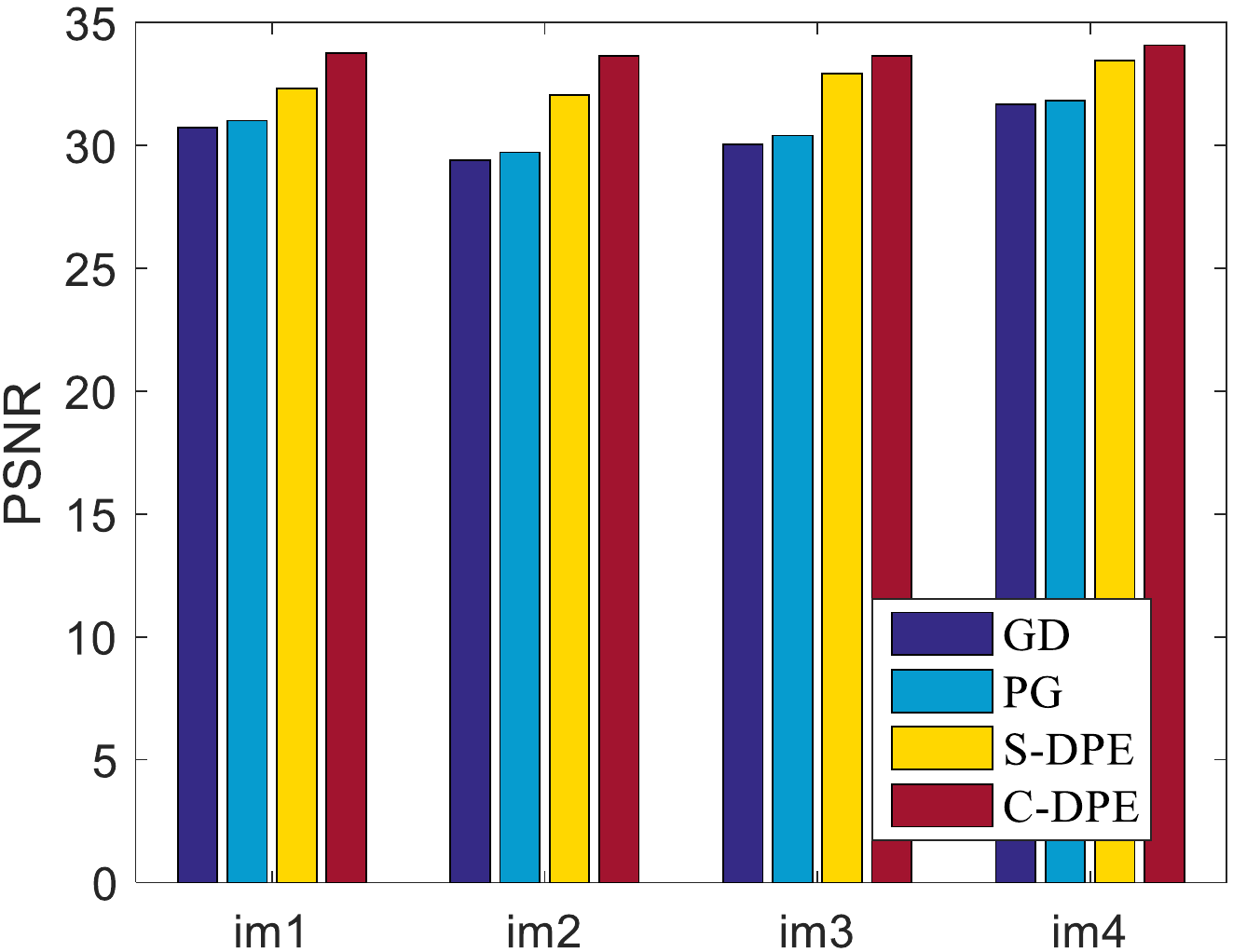}\\
		(a) & (b)
	\end{tabular}
	\caption{Quantitative comparison on Levin et al.' benchmark. (a) PSNR vs iteration numbers on the image in Fig.~\ref{fig:effeciency}. (b) PSNR scores of all the images with the first kernel in Levin et al.' benchmark.}
	\label{fig:quancomp}
\end{figure}
\begin{table}[t]
	\centering
	\caption{Average non-blind image deconvolution results (i.e., PSNR and SSIM) on Sun et al.' benchmark.}
	\begin{tabular}{|c|c|c|c|c|c|c|}
		\hline
		Metric	& GD &PG&SK-Net&MK-Net&S-DPE &C-DPE\\ \hline
		PSNR    & 27.64  & 27.71  & 26.04 & 23.40& 32.54& \textbf{32.70}\\
		\hline
		SSIM     & 0.65   &  0.66 &0.72& 0.59&0.89 &\textbf{0.90}\\ \hline
	\end{tabular}
	\label{tab:nbcompare1}
\end{table}
\begin{table*}[t]
	\centering
	\caption{Average non-blind deconvolution results on Levin et al.' and Sun et al.' benchmarks.}
	\begin{tabular}{|c|c|c|c|c|c|c|c|c|c|c|}
		\hline
		Dataset	&	Metric        & TV &HL&CSF&IDDBM3D&EPLL&RTF&MLP&IRCNN& Ours\\ \hline
		\multirow{3}{*}{Levin et al.'\cite{Levin2009Understanding}}&PSNR    & 29.38    & 30.12  & 32.74  & 31.53 & 31.65 & 33.26 & 31.32 & 32.51 & \textbf{33.44}\\
		\cline{2-11}
		~&SSIM     & 0.88   &  0.90  &  0.93 & 0.90& 0.93 & 0.94 & 0.90 & 0.92 & \textbf{0.95}\\ \cline{2-11}
		~&Time &1.22  &  0.10	 & 0.12  & 0.43 & 70.32 & 26.63& 0.49 & 2.85 &6.03\\ \hline
		\multirow{3}{*}{Sun et al.'\cite{Sun2013Edge}}&PSNR    & 30.67    & 31.03  & 31.55 & 30.79 & 32.44 & 32.45 & 31.47 & 32.61 & \textbf{32.82} \\
		\cline{2-11}
		~&SSIM     & 0.85   &  0.85  &  0.87 & 0.87 & 0.88 & 0.89 & 0.86 & 0.89 & \textbf{0.90}\\\cline{2-11}
		~&Time &6.38  &  0.49 & 0.50& 48.66 & 721.98 & 240.98& 4.59 & 16.67 &13.20\\ \hline
	\end{tabular}
	\label{tab:nbcompare2}
\end{table*}
Fig.~\ref{fig:effeciency} first illustrated visual comparisons of these different approaches on an example image from Sun et al.' benchmark. We observed that both of our hybrid propagations can achieve better qualitative performance. We then plotted the convergence behaviors of conventional iteration algorithms (i.e., GD and PG)
and our propagations (i.e., S-DPE and C-DPE) on this example image in Fig.~\ref{fig:quancomp} (a). The PSNR scores of all the images with the first blur kernel are also shown in Fig.~\ref{fig:quancomp} (b). We can see that our proposed hybrid schemes perform consistently better than conventional optimization algorithms, in which C-DPE is the best among all the compared strategies.


Table~\ref{tab:nbcompare1} further reported average quantitative performances on Sun et al.' benchmark. It can be seen that PG achieves better performance as it is based on a more accurate prior model. We also observed that directly performing propagations using the networks cannot obtain good performance, even trained on data generated by test kernel (i.e., SK-Net). This is mainly because that relative simple architectures may have difficulty in fitting the deconvolution process. In contrast, our hybrid propagation schemes (i.e., S-DPE and C-DPE) perform much better than these approaches. Moreover, our designed prior projection based feedback strategy (i.e., C-DPE) can further improve the performance of the simplified DPE (i.e., S-DPE).

\subsubsection{Propagative Behaviors Analysis}

We then verify the sub-gradient error bound (Condition~\ref{condition}) in our theoretical part. That is, we plot curves of sub-gradient error (orange) and relative error (blue) on example images from Levin et al.' and Sun et al.' benchmarks in Fig.~\ref{fig:convergence}. It can be seen that the error conditions in Eq.~\eqref{eq:error-condition} are always satisfied during the propagations, which verify our theoretical results in Section~\ref{sec:analysis}.

\begin{figure}[t]
	\centering
	\begin{tabular}{c@{\extracolsep{0.4em}}c}
		\includegraphics[height=0.17\textwidth]{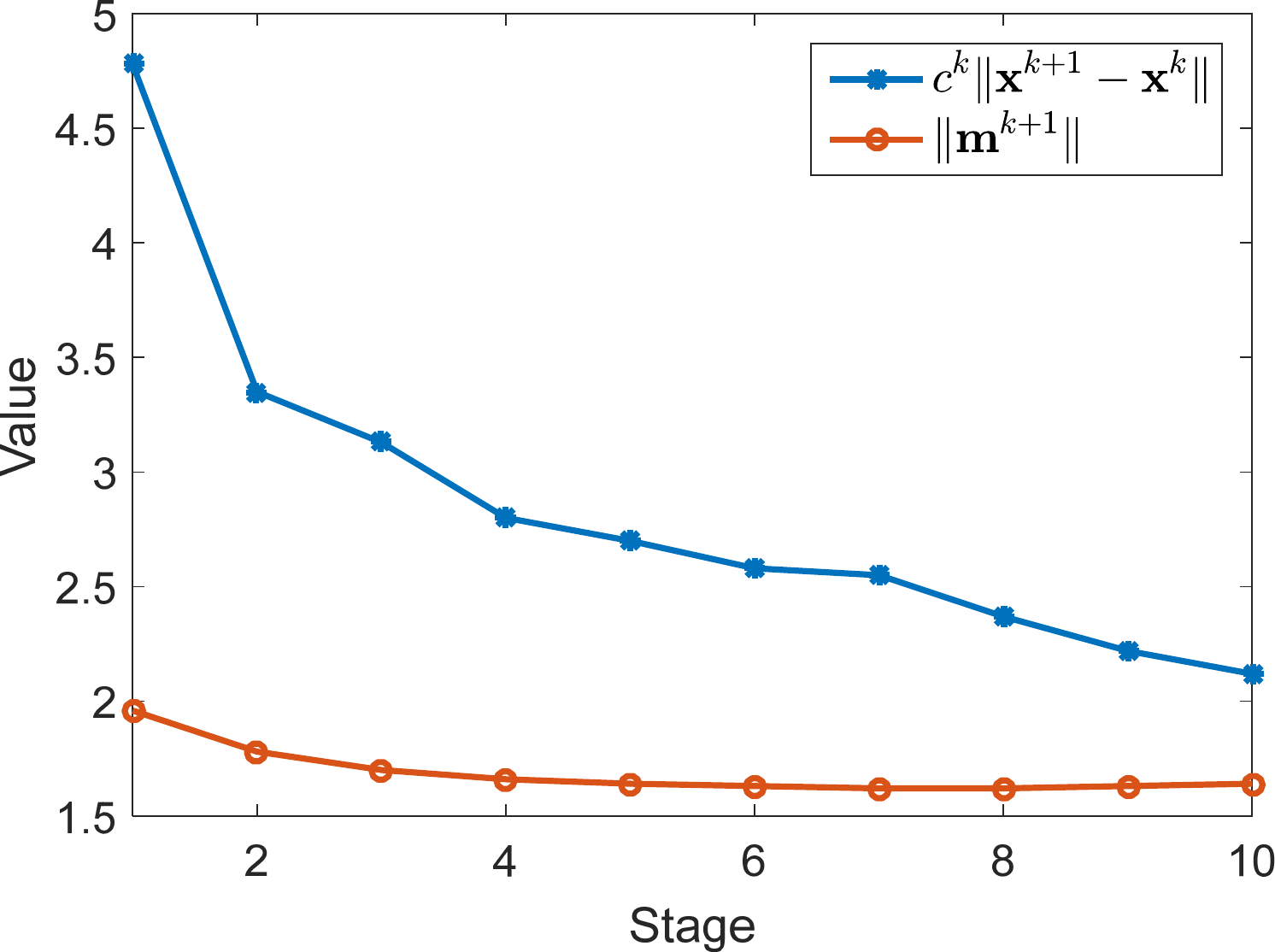}
		&\includegraphics[height=0.17\textwidth]{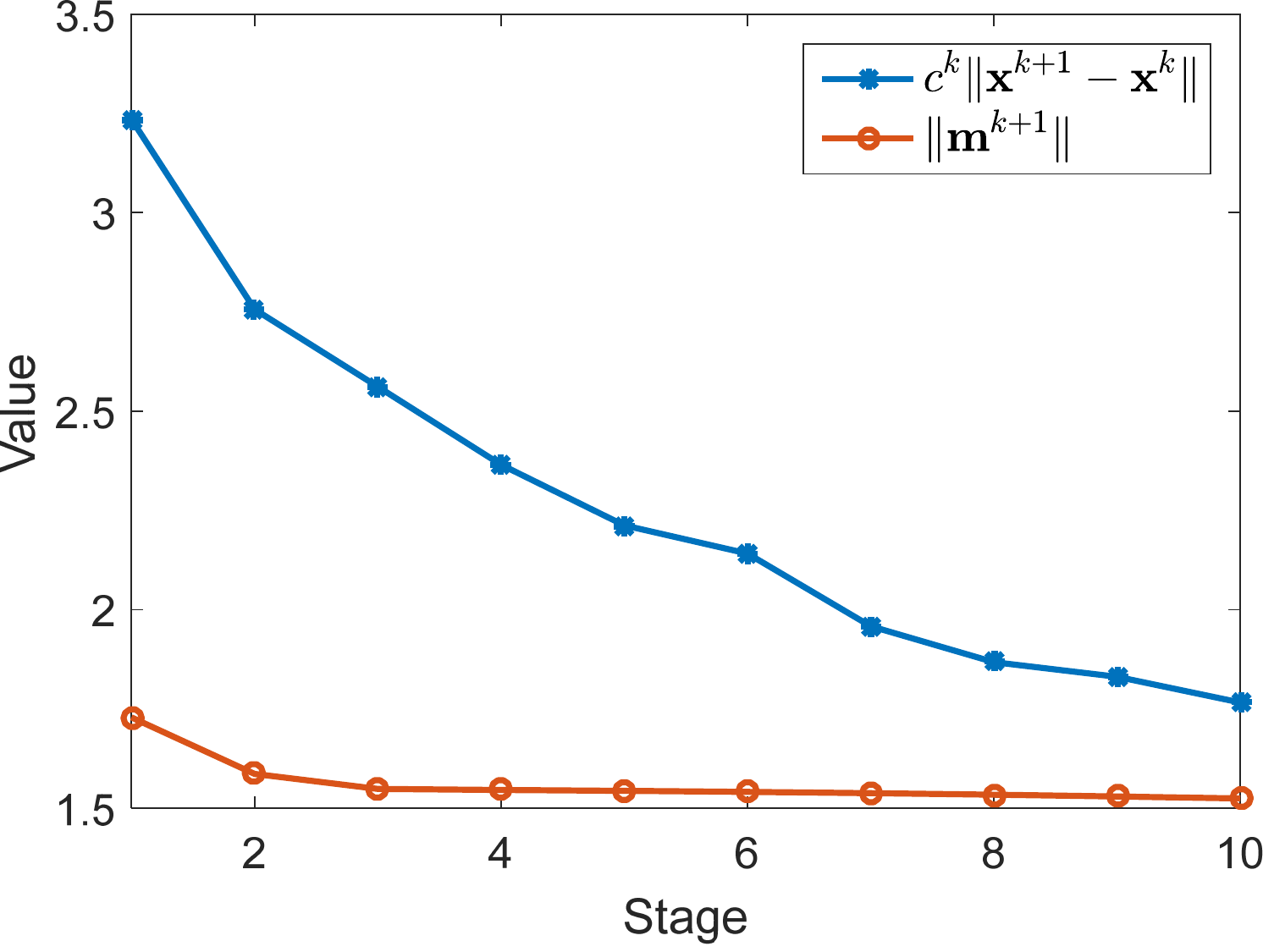}\\
		Levin et al. & Sun et al.
	\end{tabular}
	\caption{The curves of sub-gradient error (orange) and relative error (blue) on example images in Levin et al.' and Sun et al.' benchmarks.}
	\label{fig:convergence}
\end{figure}

We also plotted the intermediate results of DPE (i.e., PSNR scores of the propagative variables $\{\dot{\mathbf{x}}^k, \ddot{\mathbf{x}}^k, \mathbf{x}^k\}$) on example images from both Levin et al.'s and Sun et al.'s benchmarks in Fig.~\ref{fig:trend}. We observed that the fidelity based warm start provided relative good initial values (i.e., $\dot{\mathbf{x}}^k$) at each stage. The network based descent direction (i.e., $\ddot{\mathbf{x}}^k$) can significantly improve the performance, especially on challenging image in Sun et al.' dataset. Finally, our prior projection can fine tune the deep prior ensemble and result to a more stable image propagation (i.e., $\mathbf{x}^k$).

\subsection{Real-world Image Enhancement Tasks}

In this subsection, we evaluate DPE on various image enhancement tasks, including non-blind image deconvolution, image interpolation, super-resolution, single image haze removal and underwater image enhancement, with comparisons to state-of-the-art approaches for these problems.

\begin{figure}[t]
	\centering
	\begin{tabular}{c@{\extracolsep{0.2em}}c}
		\includegraphics[height=0.17\textwidth]{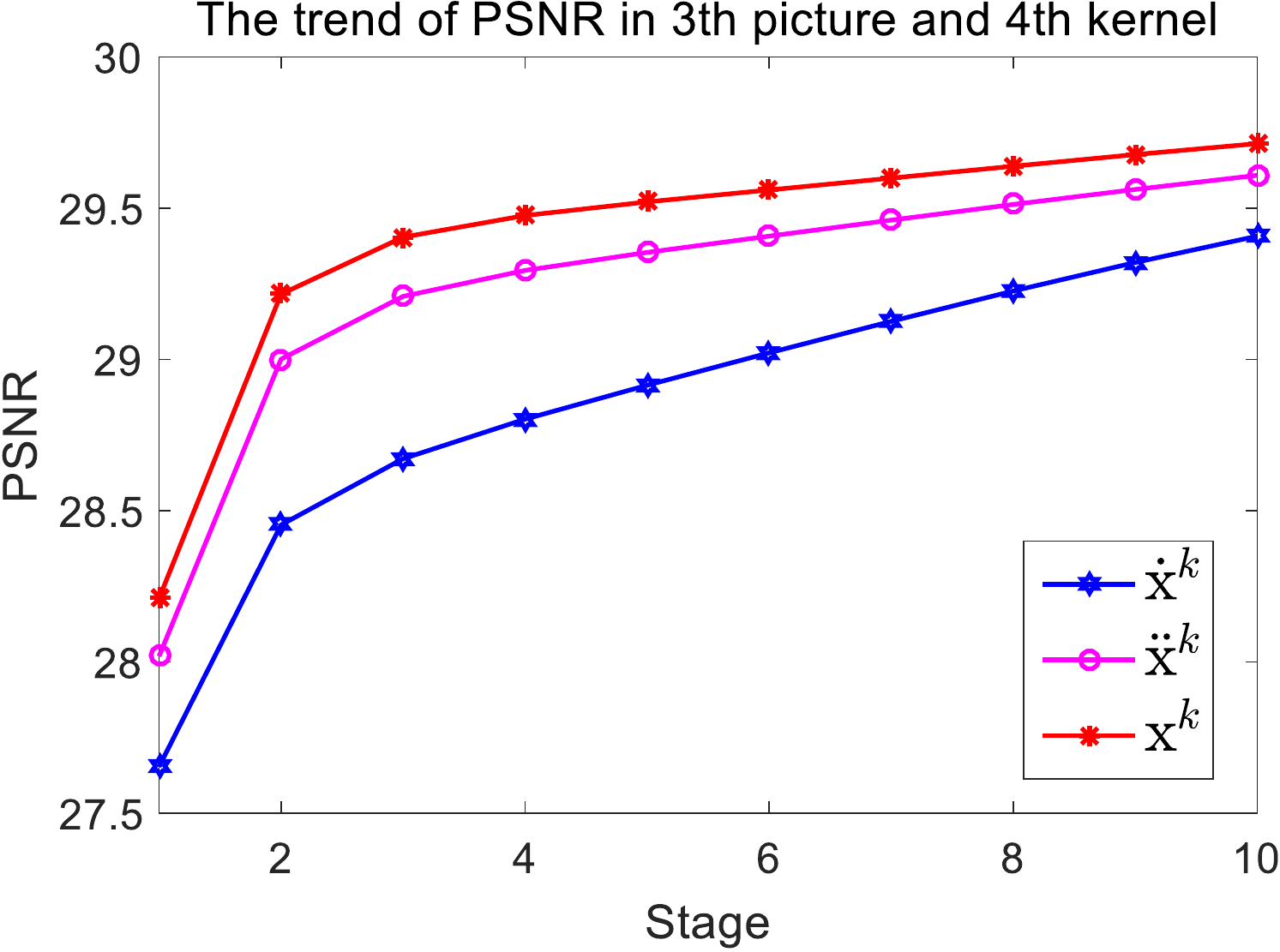}
		&\includegraphics[height=0.17\textwidth]{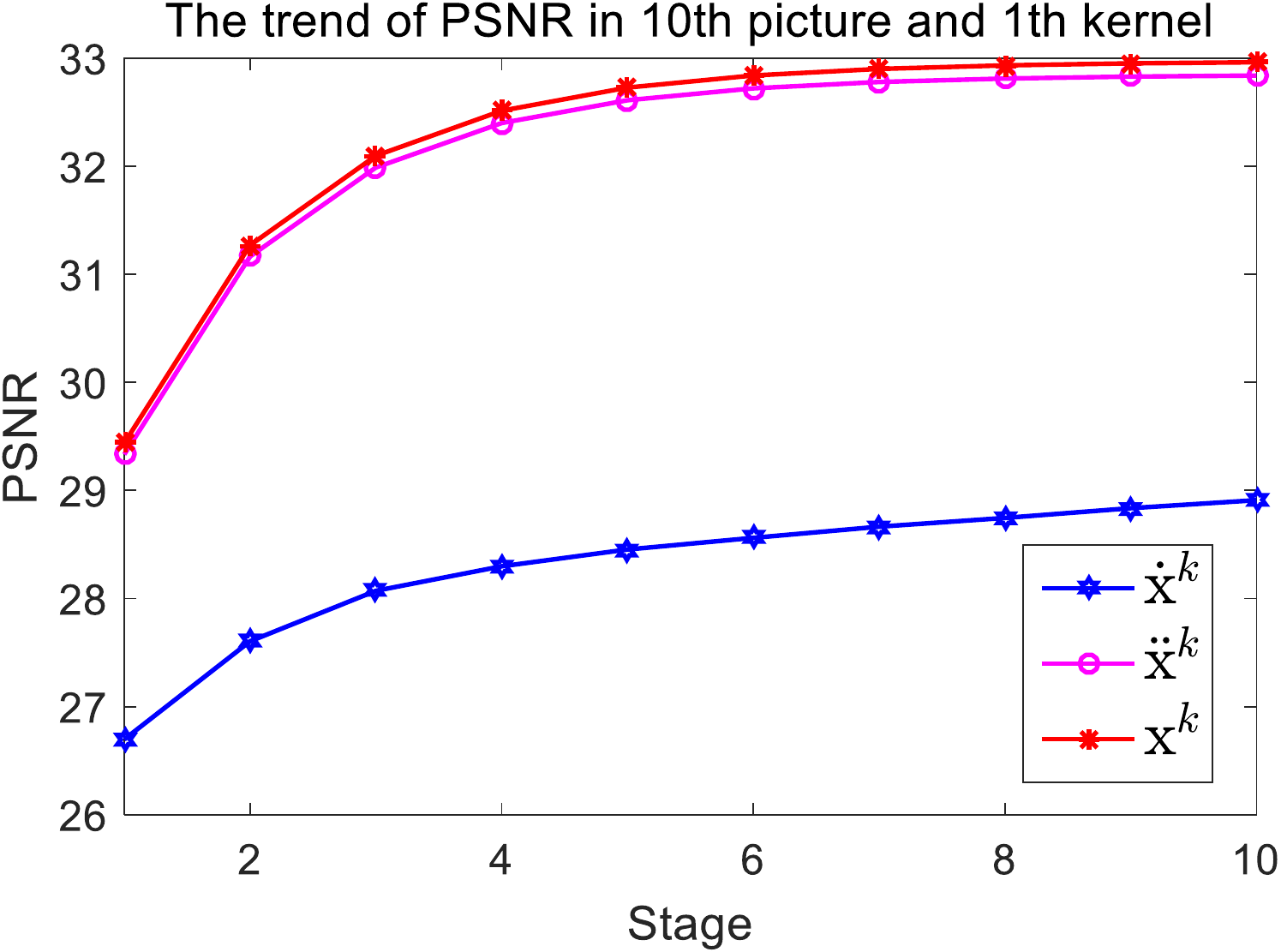}\\
		Levin et al. & Sun et al.
	\end{tabular}
	\caption{The curves of PSNR with respect to intermediate results of DPE on example images in Levin et al.'s and Sun et al.'s benchmarks.}
	\label{fig:trend}
\end{figure}

\subsubsection{Non-blind Image Deconvolution}
As for this task, we compared the proposed framework with several state-of-the-art algorithms, including TV~\cite{Li2013An}, HP~\cite{Krishnan2009Fast}, CSF~\cite{Schmidt2014Shrinkage}, IDDBM3D~\cite{danielyan2012bm3d}, EPLL~\cite{zoran2011learning}, RTF~\cite{schmidt2016cascades}, MLP\cite{Schuler2013A}, and IRCNN~\cite{Zhang2017Learning} on both Levin et al.' and Sun et al.' datasets. In Table~\ref{tab:nbcompare2}, we can see that the PSNR and SSIM scores of the proposed DPE are significantly better than the other deconvolution methods. It is observed that the speed of DPE is slower than some simple prior optimization techniques (e.g., TV, HL and CSF), which have very poor restoration performance. But fortunately, our propagation is much faster than the CNN based approaches (e.g., IRCNN)
and other high-performance approaches (e.g., IDDBM3D, EPLL and RTF).
Fig.~\ref{fig:nbcomp} then compared the visual performances of DPE against approaches with relatively high quantitative scores in Table~\ref{tab:nbcompare2} on an example image in Sun et al' benchmark, Notice that this image is corrupted not only by blur kernels, but also 5\% additional Gaussian noise. It is easy to conclude that our method achieved both qualitative enhanced results (e.g., generates the much clearer image with fine texture) and better quantitative performance.

\begin{figure*}[ht]
	\centering
	\begin{tabular}{c@{\extracolsep{0.1em}}c@{\extracolsep{0.1em}}c@{\extracolsep{0.1em}}c@{\extracolsep{0.1em}}c@{\extracolsep{0.1em}}c}
		\includegraphics[width = 0.163\textwidth]{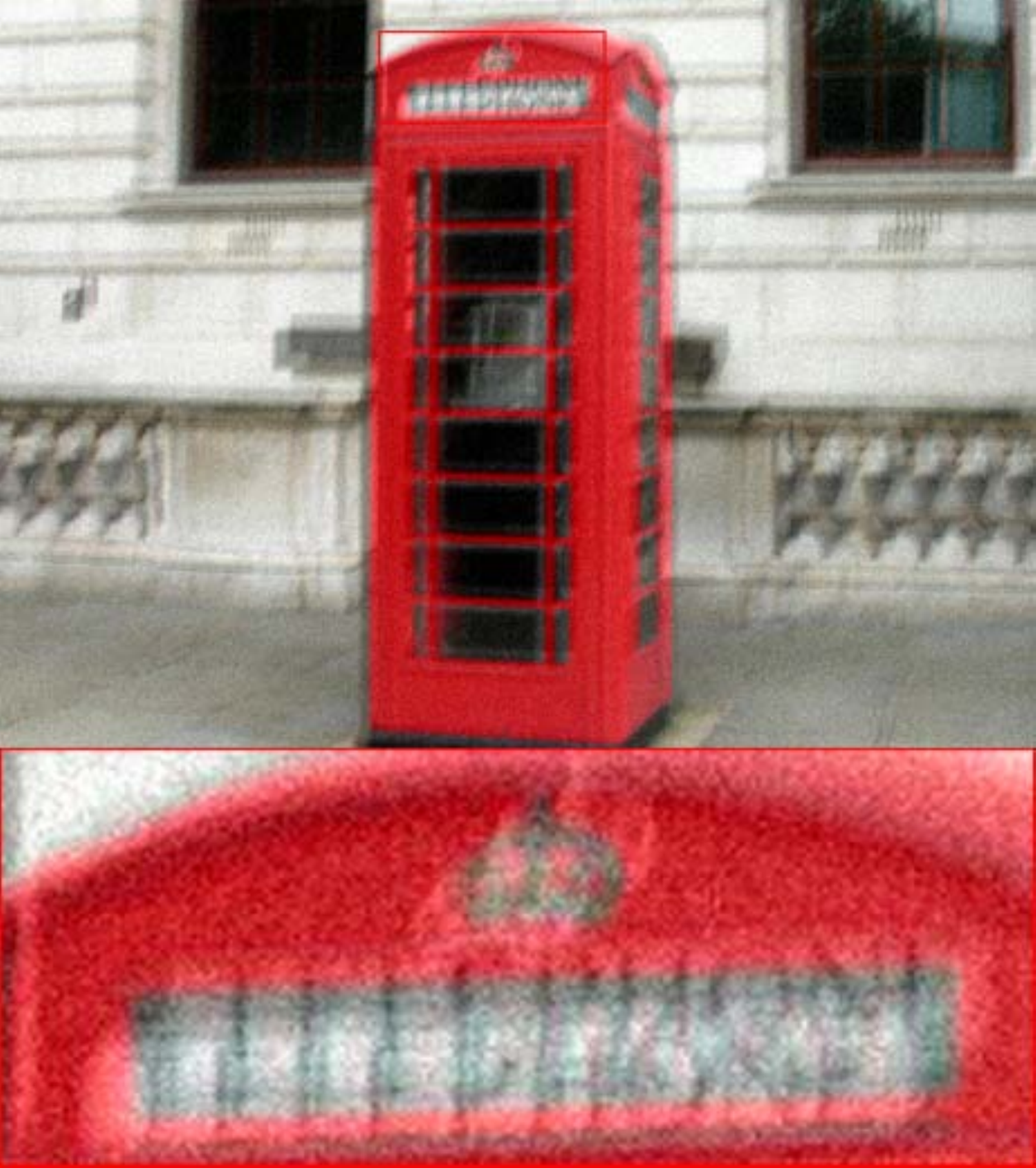}&
		\includegraphics[width = 0.163\textwidth]{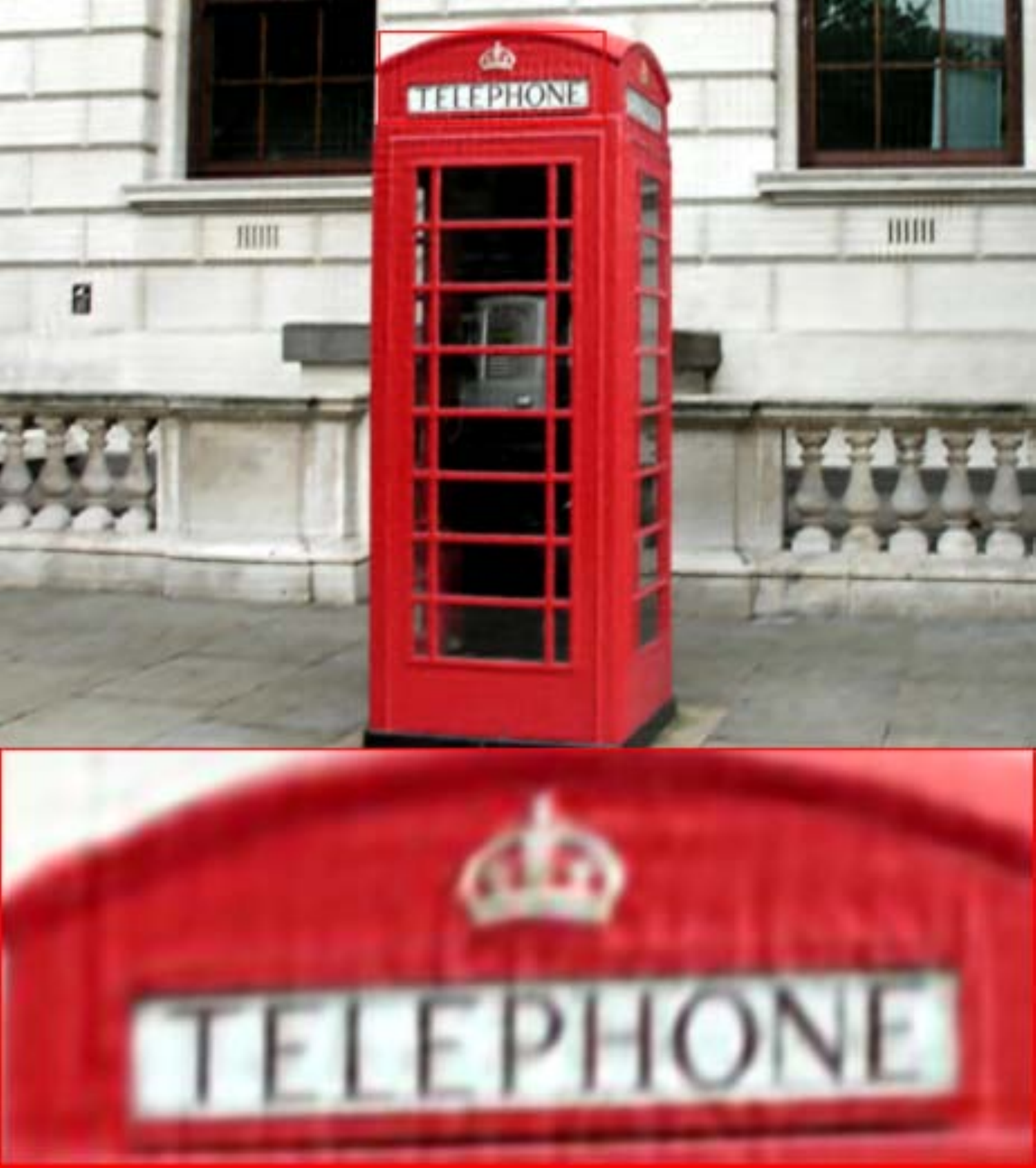}&
		\includegraphics[width = 0.163\textwidth]{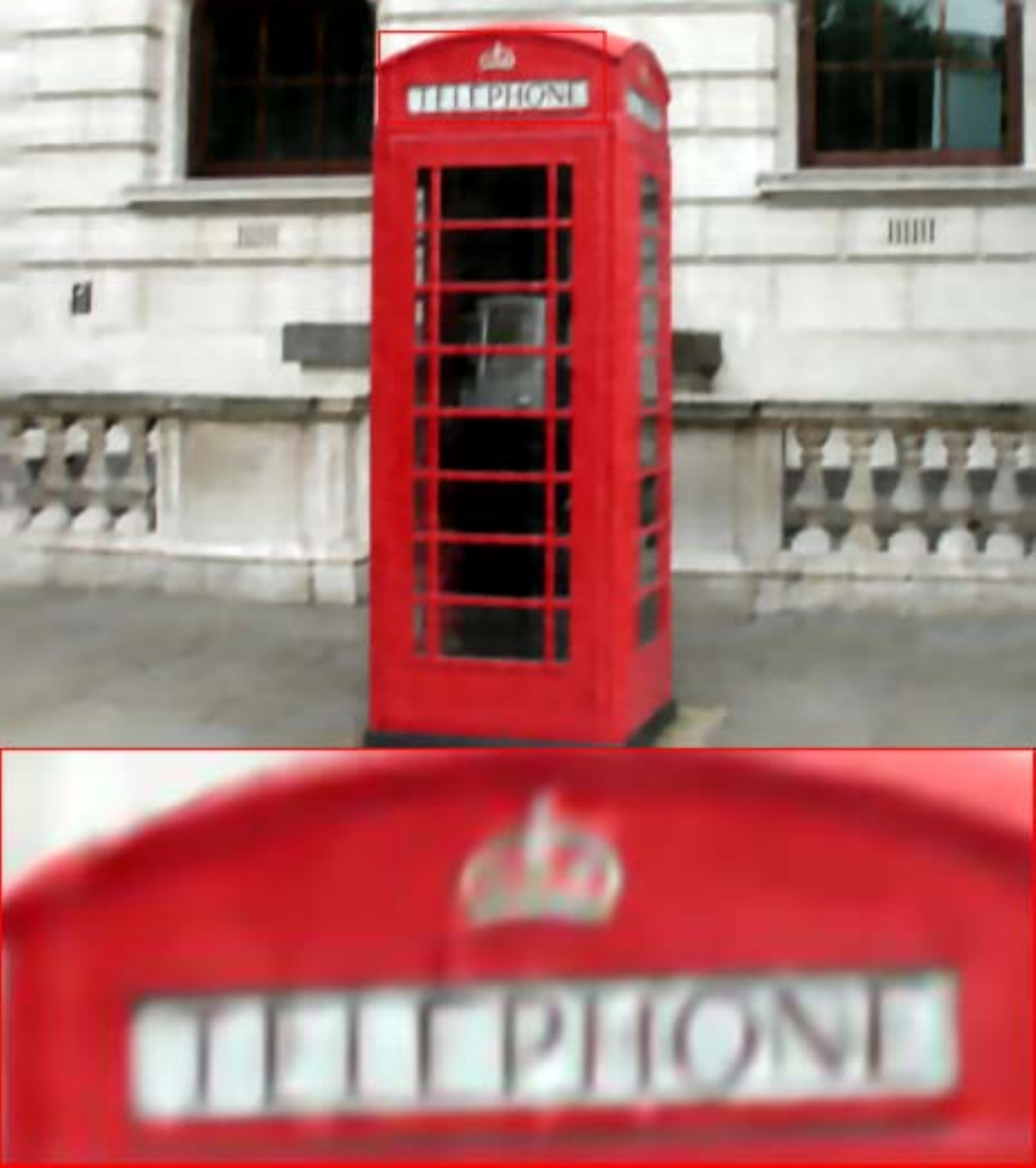}&
		\includegraphics[width = 0.163\textwidth]{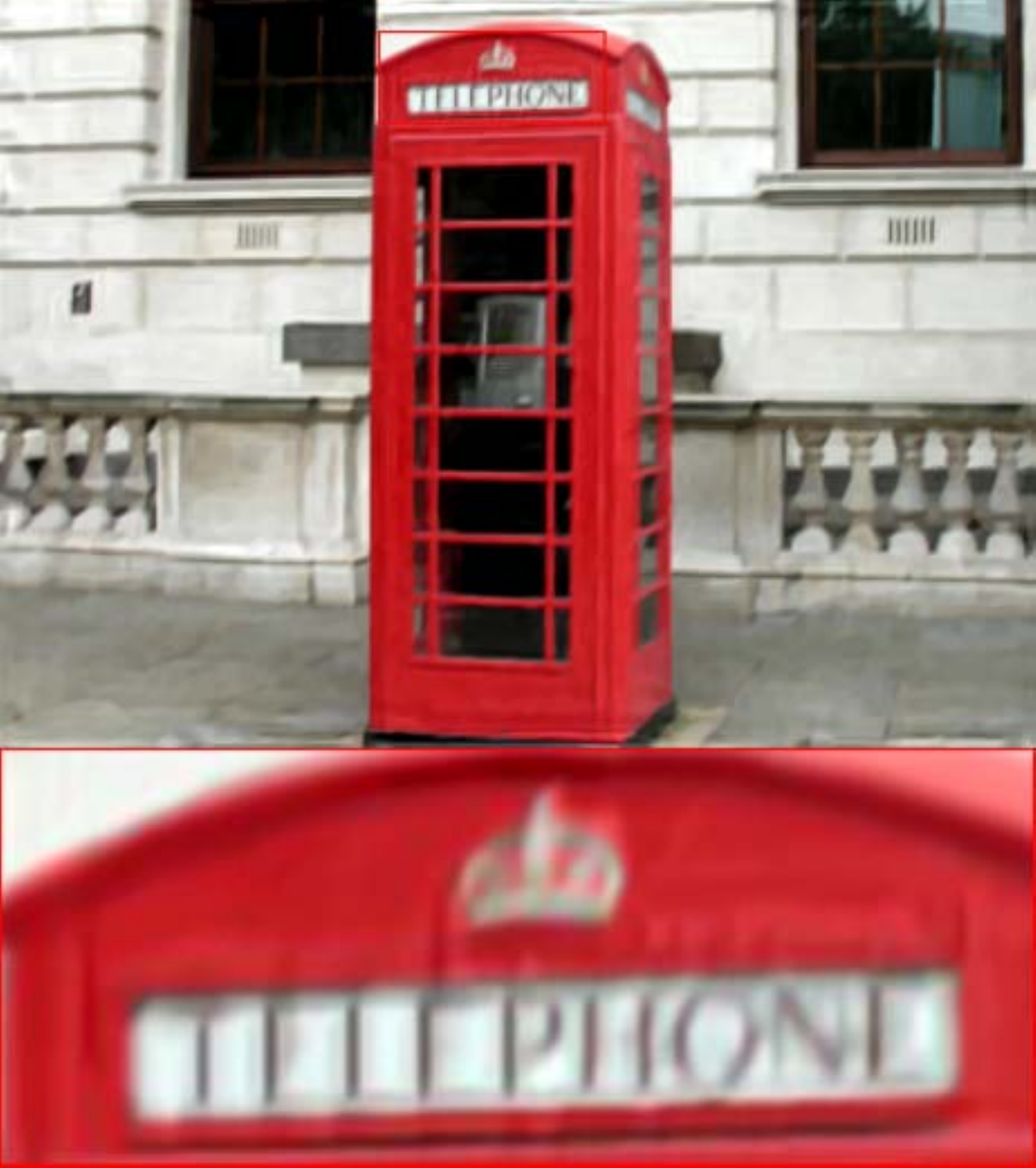}&
		\includegraphics[width = 0.163\textwidth]{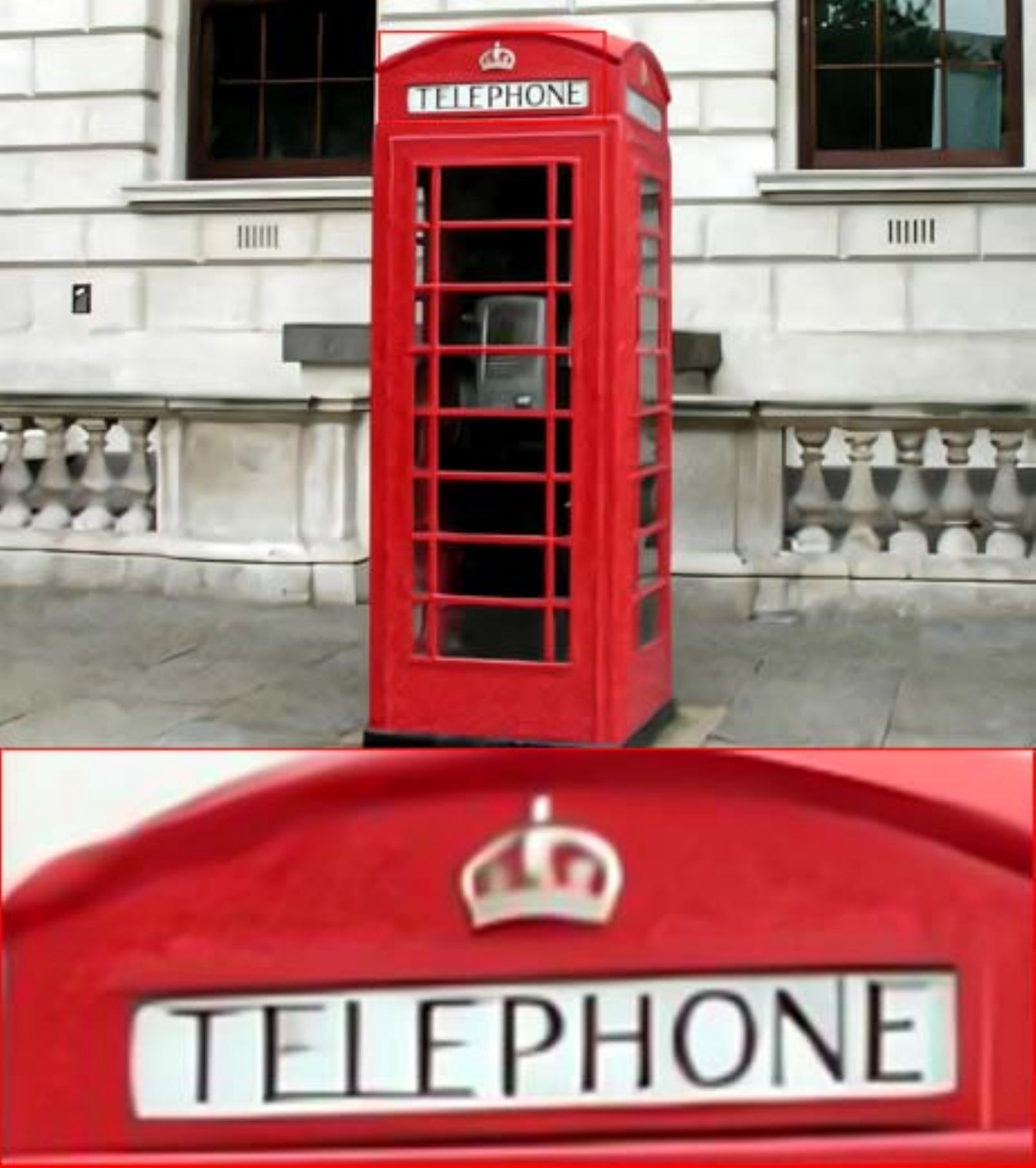}&
		\includegraphics[width = 0.163\textwidth]{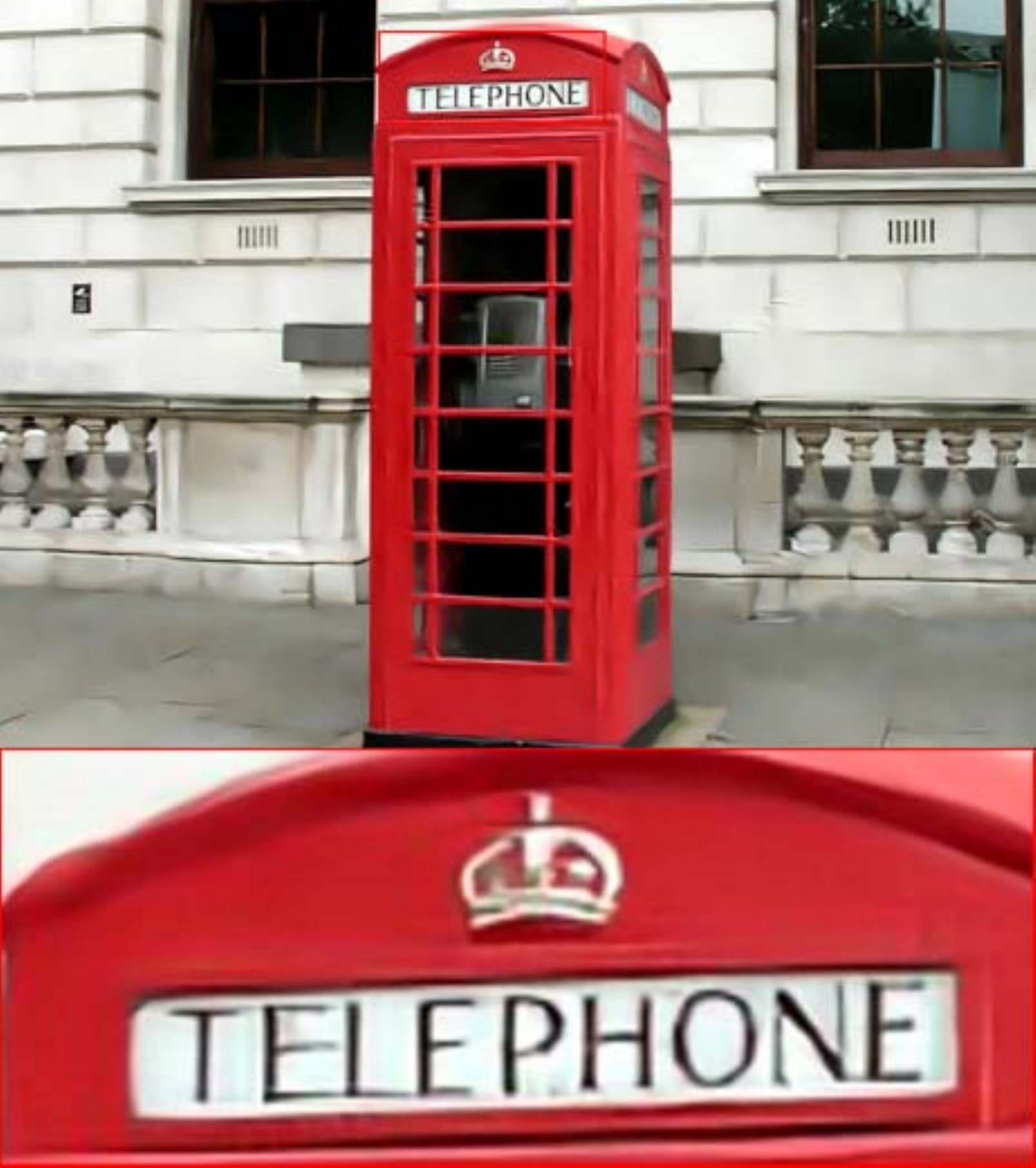}\\
		18.62 / 0.23 & \footnotesize 24.14 / 0.71&\footnotesize 27.46 / 0.81 & \footnotesize 25.74 / 0.76 & \footnotesize 28.70 / 0.82 & \footnotesize \textbf{28.83 / 0.83}\\
		\footnotesize  Blurry Input &\footnotesize MLP&\footnotesize EPLL&\footnotesize IDDBM3D&\footnotesize  IRCNN&\footnotesize  Ours\\
	\end{tabular}
	\caption{The performance on non-blind deconvolution with zoomed in comparisons. The quantitative scores (i.e., PSNR / SSIM) are also reported accordantly.}
	\label{fig:nbcomp}
\end{figure*}
\begin{table}[h]
	\centering
	\caption{Average PSNR scores of different image interpolation methods on CBSD68 benchmark. }
	\begin{tabular}{|c|c|c|c|c|c|}
		\hline
		Mask& TV&FoE&VNL&ISDSB&Ours\\
		\hline
		20\% &36.30&38.23&28.87&35.20&\textbf{39.31}\\
		\hline
		40\%&32.22&34.01&27.55&31.32&\textbf{34.74}\\
		\hline
		60\%&29.20&{30.81}&26.13&28.23&\textbf{31.25}\\
		\hline
		80\%&26.07&{27.64}&24.23&24.92&\textbf{27.74}\\
		\hline
		Text&35.29&{37.05}&28.58&34.91&\textbf{37.33}\\
		\hline
	\end{tabular}
	\label{tab:inpaintpsnr}
\end{table}
\begin{figure*}[htb]
	\centering
	\begin{tabular}{c@{\extracolsep{0.1em}}c@{\extracolsep{0.1em}}c@{\extracolsep{0.1em}}c@{\extracolsep{0.1em}}c@{\extracolsep{0.1em}}c@{\extracolsep{0.1em}}c}
		\includegraphics[width=0.14\textwidth]{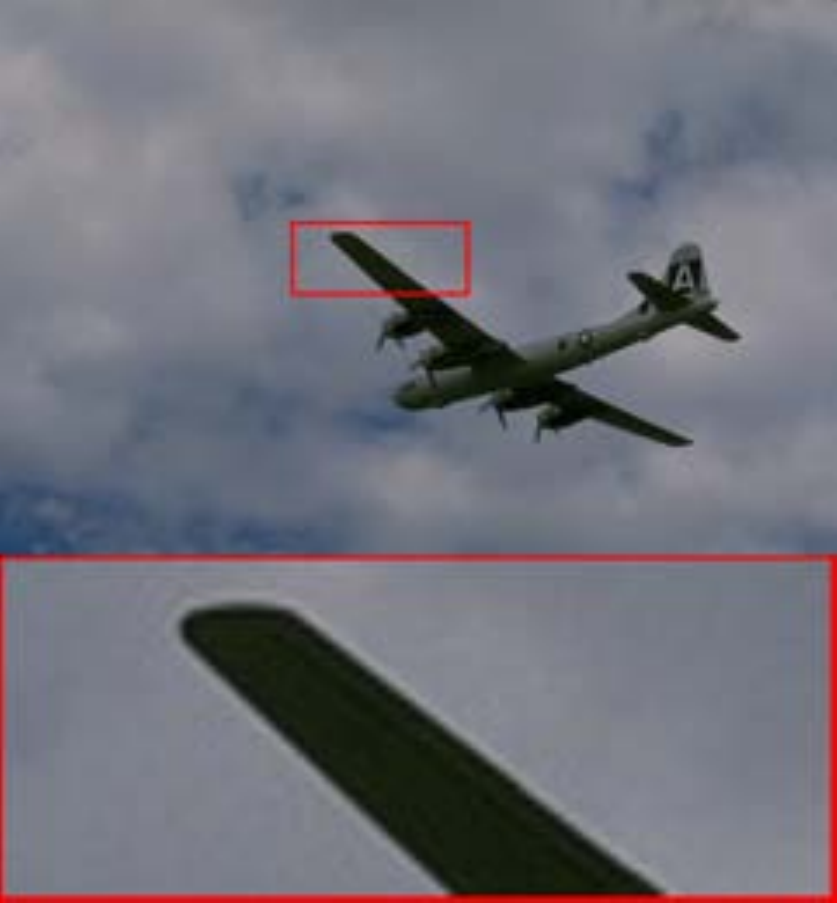}&
		\includegraphics[width=0.14\textwidth]{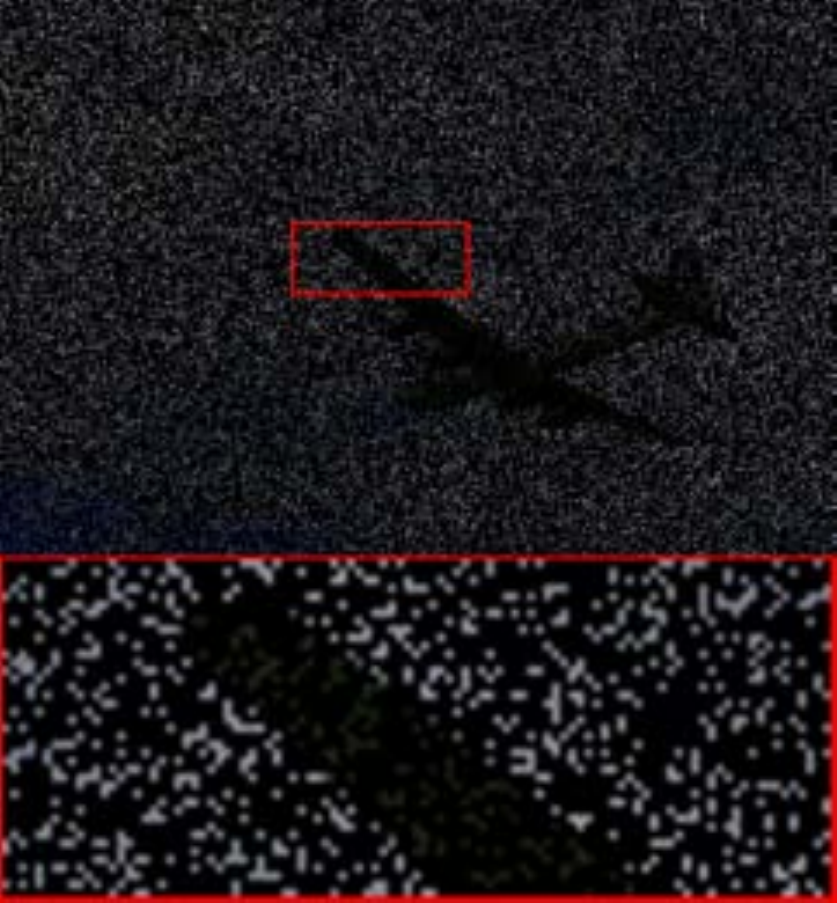}&
		\includegraphics[width=0.14\textwidth]{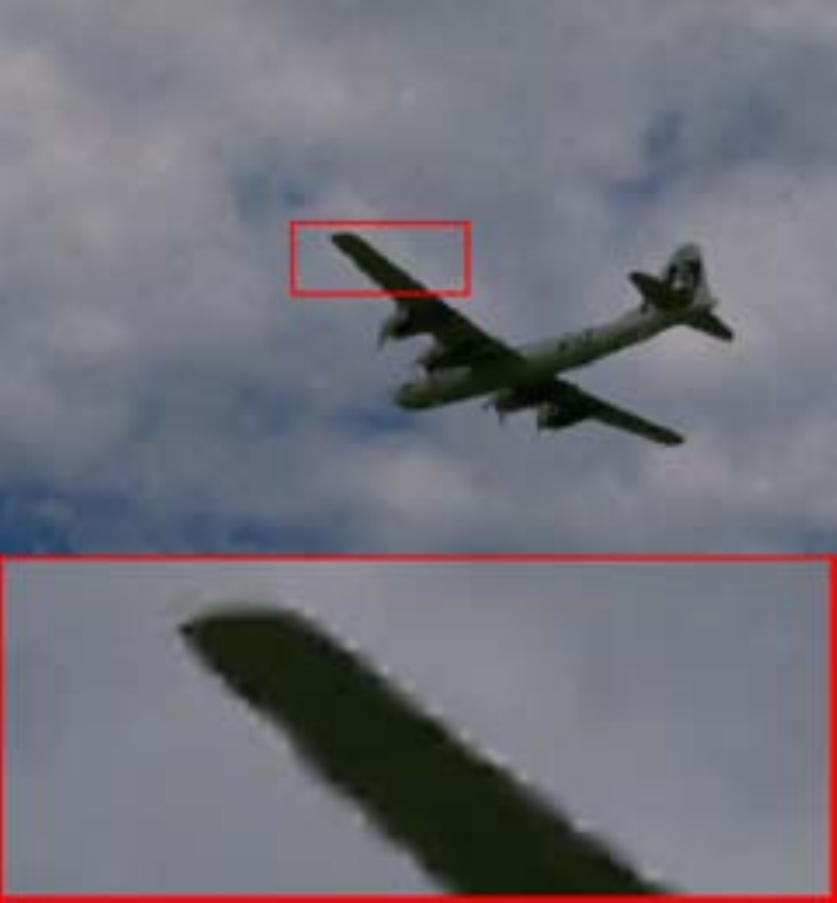}&
		\includegraphics[width=0.14\textwidth]{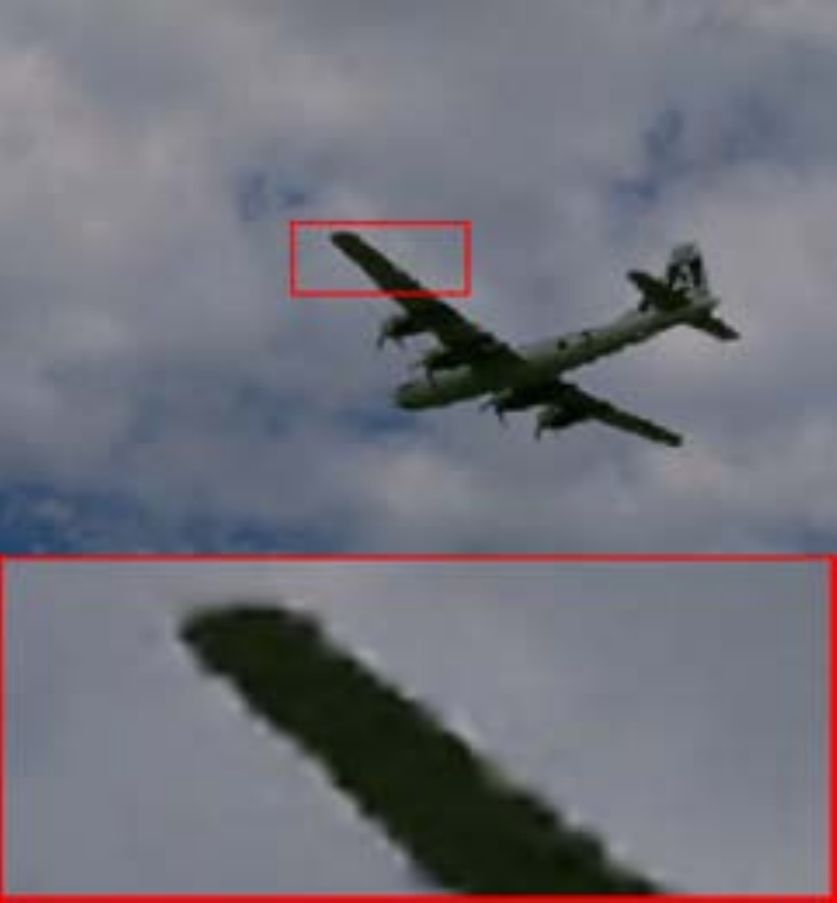}&
		\includegraphics[width=0.14\textwidth]{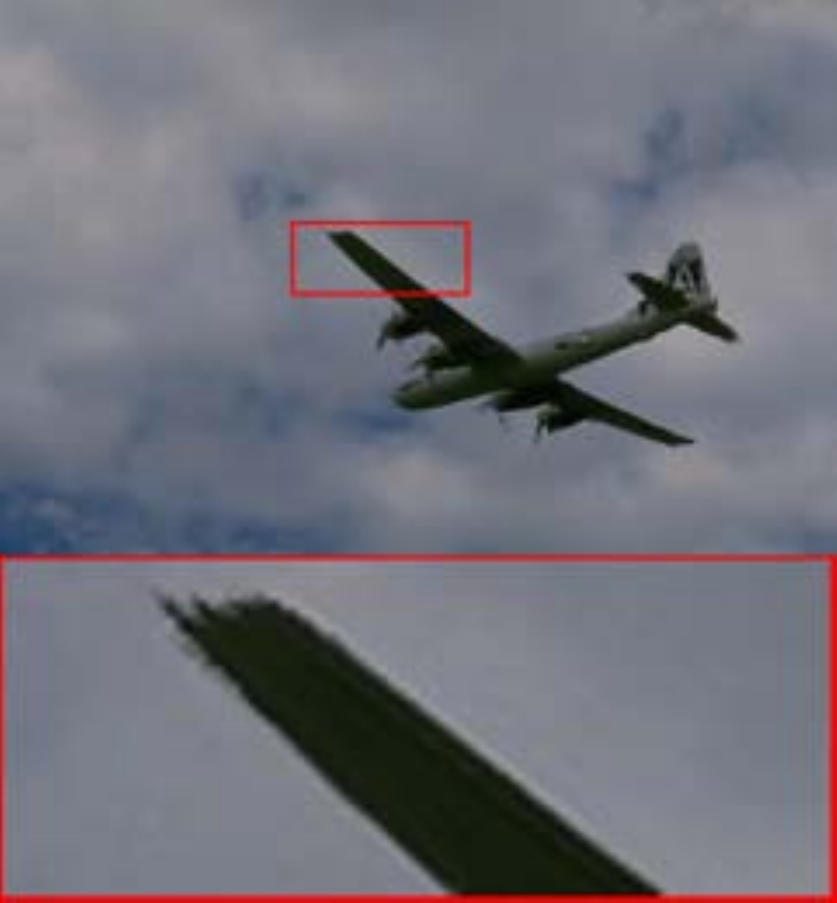}&
		\includegraphics[width=0.14\textwidth]{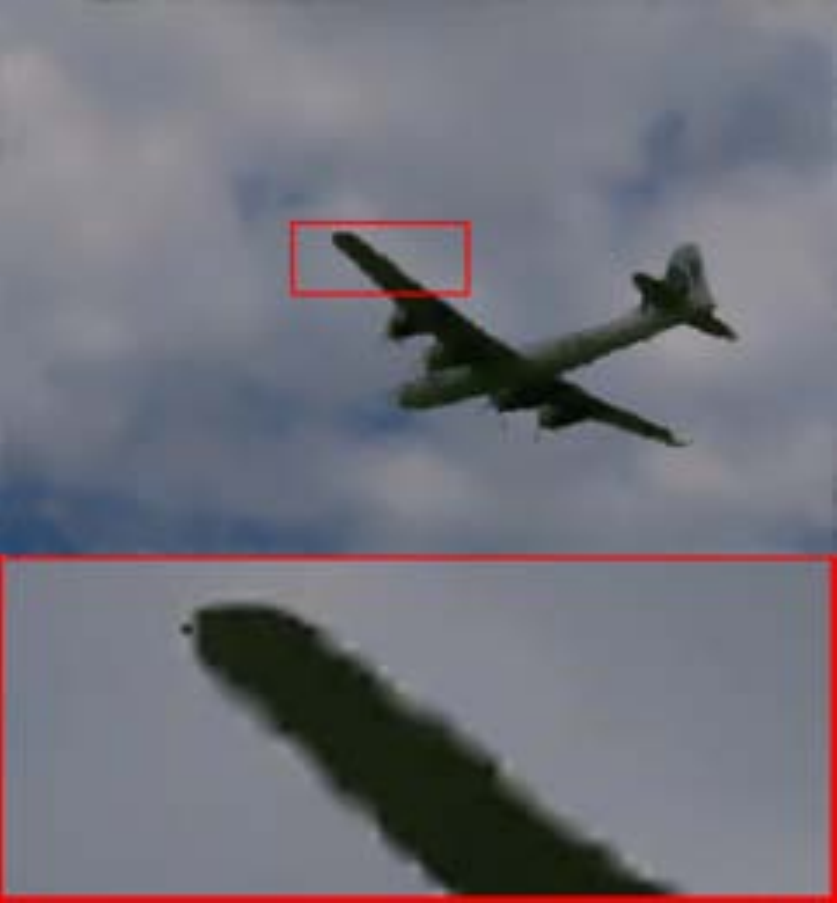}&
		\includegraphics[width=0.14\textwidth]{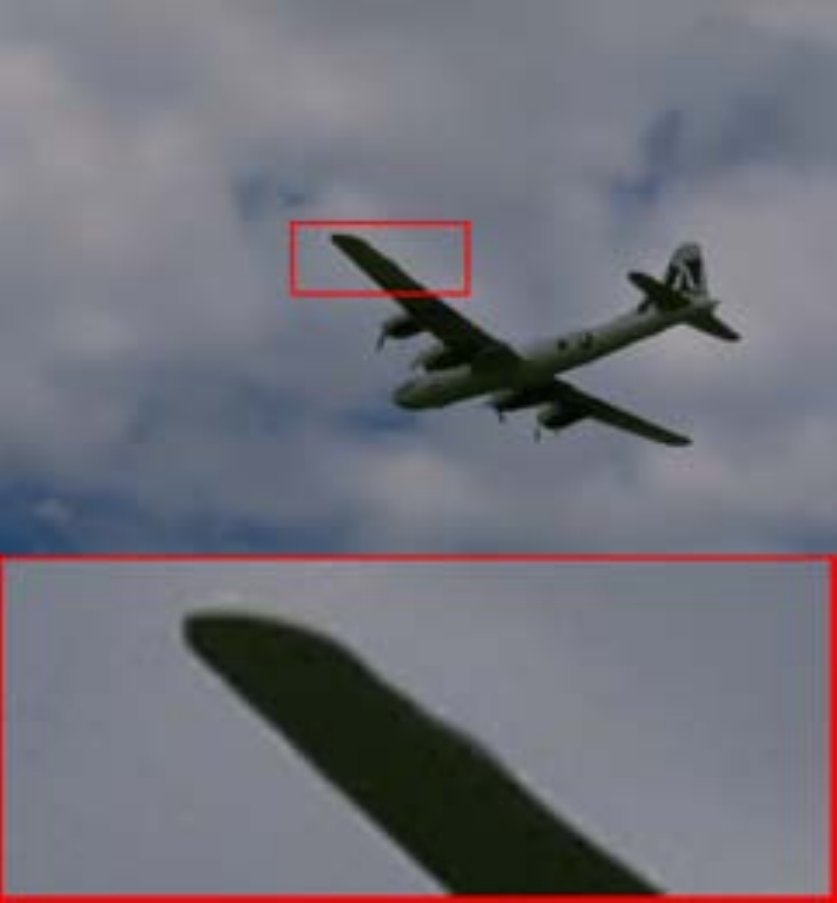}\\
		--& \footnotesize 7.55&\footnotesize 35.87&\footnotesize 37.12&\footnotesize 33.50&\footnotesize 34.52&\footnotesize \textbf{38.05}\\
		\includegraphics[width=0.14\textwidth]{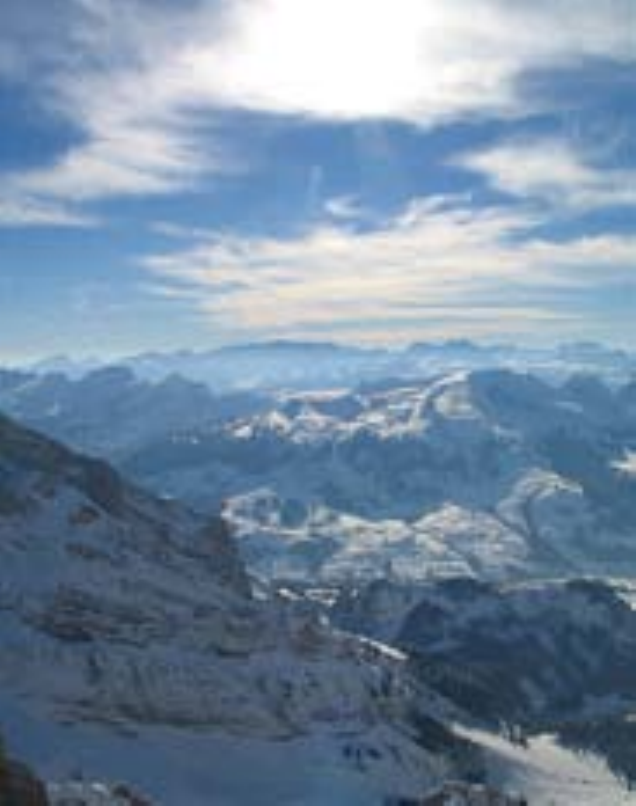}&
		\includegraphics[width=0.14\textwidth]{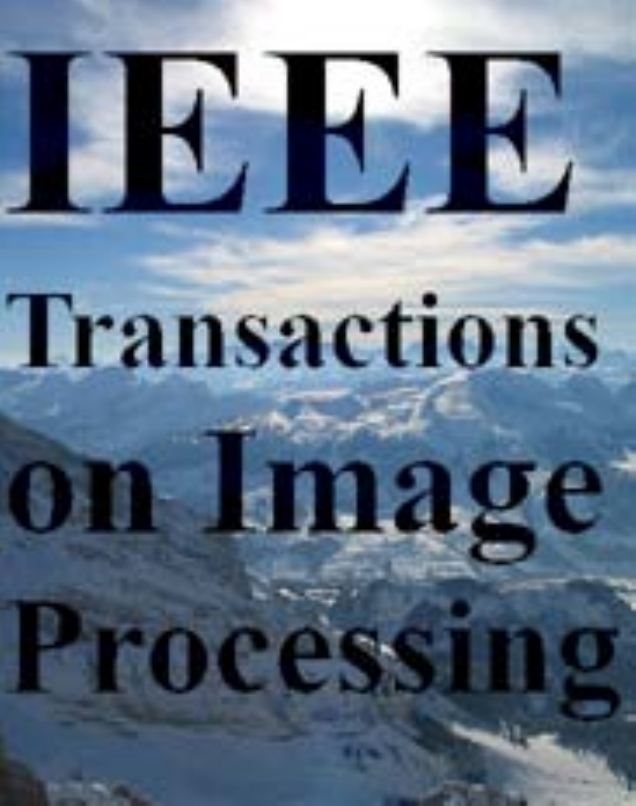}&
		\includegraphics[width=0.14\textwidth]{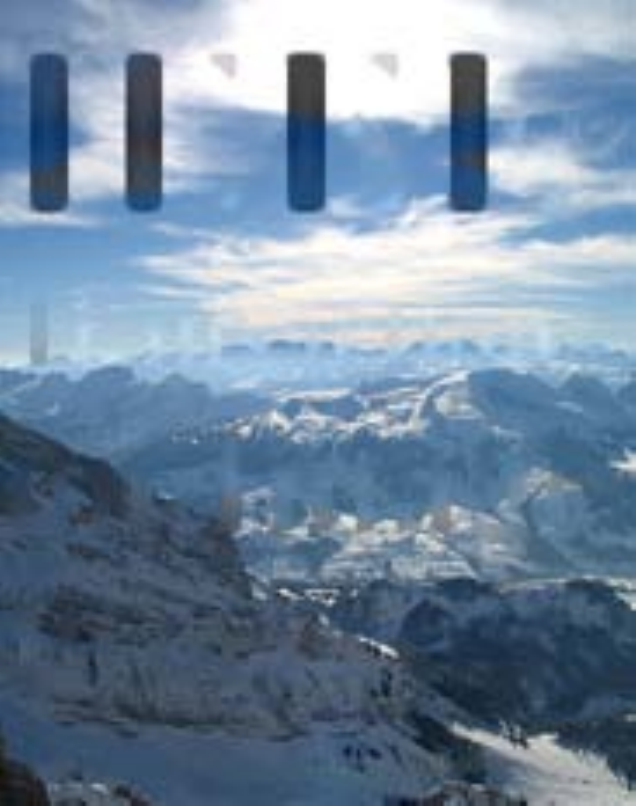}&
		\includegraphics[width=0.14\textwidth]{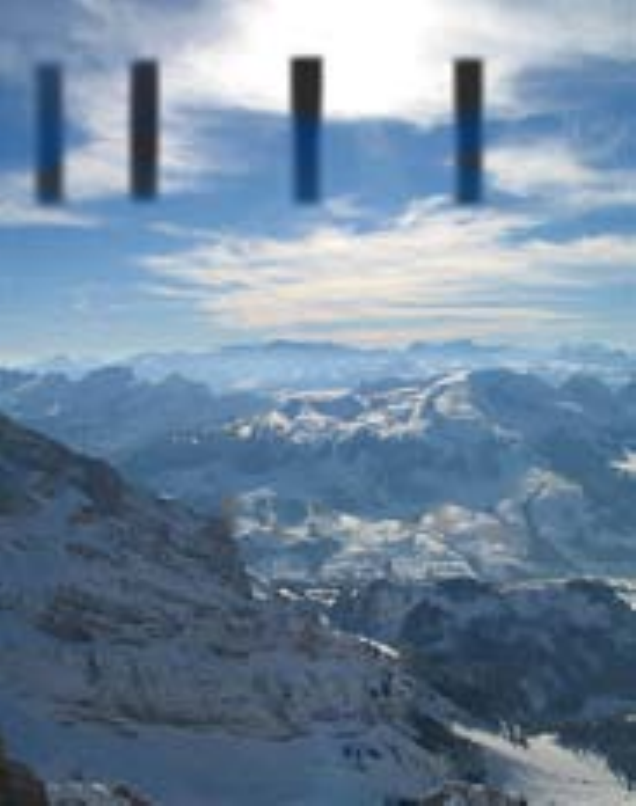}&
		\includegraphics[width=0.14\textwidth]{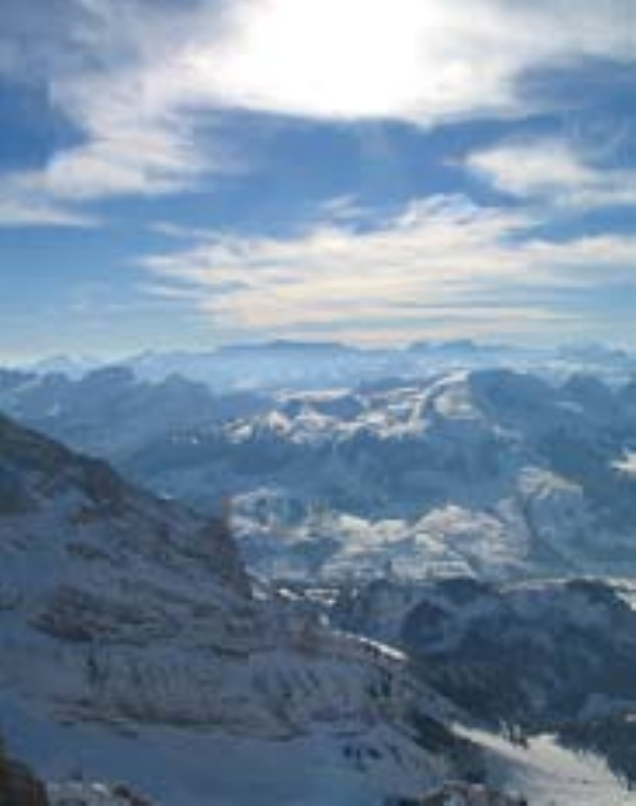}&
		\includegraphics[width=0.14\textwidth]{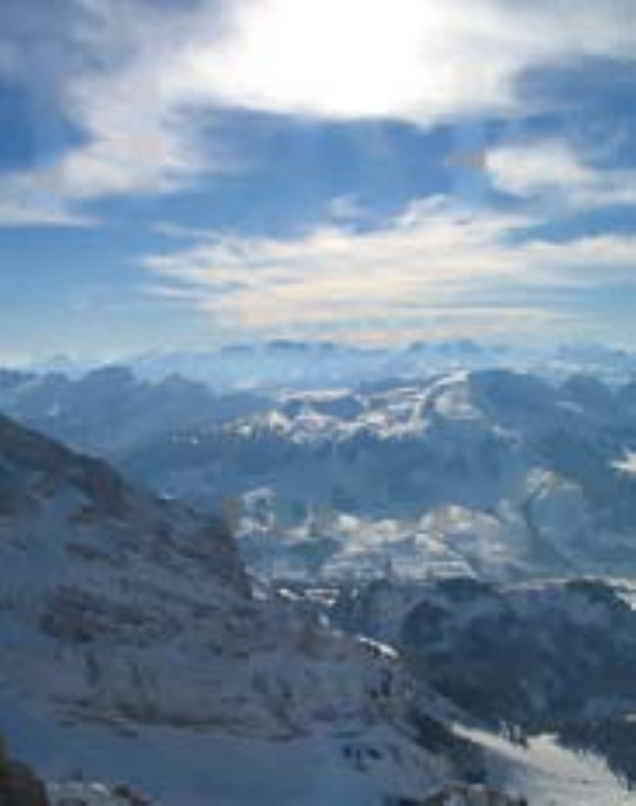}&
		\includegraphics[width=0.14\textwidth]{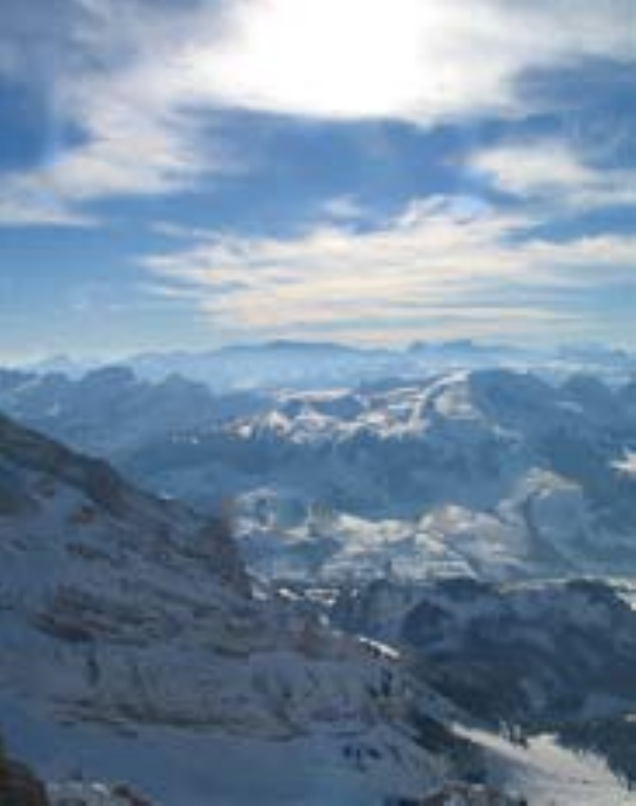}\\
		--& \footnotesize 12.62&\footnotesize 20.27&\footnotesize 22.51&\footnotesize 33.90&\footnotesize 33.20&\footnotesize \textbf{34.08}\\
		\footnotesize Ground Truth &\footnotesize Corrupted Input &\footnotesize TV&\footnotesize FoE&\footnotesize VNL&\footnotesize ISDSB&\footnotesize Ours\\
	\end{tabular}
	\caption{The performances on image interpolation with zoomed in comparisons. Both random masks with 80\% missing values (top row) and text masks (bottom row) are considered. The PSNR scores are also reported below each subfigure.}
	\label{fig:inpaintres}
\end{figure*}
\begin{table*}[htb]
	\centering
	\caption{Average super-resolution performance on Set14 benchmark.}
	\begin{tabular}{|c|c|c|c|c|c|c|c|}
		\hline
		Scale& Bicubic&A+&TNRD&IRCNN&SRCNN&VDSR &Ours\\ \hline
		$\times $2&30.24\,/\,0.8688&32.28\,/\,0.9056&32.54\,/\,0.9069&32.88\,/\,0.9114&32.42\,/\,0.9063& \textbf{33.03}\,/\,0.9124 &32.94\,/\,\textbf{0.9127}\\
		$\times $3&27.55\,/\,0.7742&29.13\,/\,0.8188&29.46\,/\,0.8236&29.61\,/\,0.8285&29.28\,/\,0.8209&\textbf{29.77}\,/\,0.8314&29.69\,/\,\textbf{0.8328}\\
		$\times $4&26.00\,/\,0.7027&27.32\,/\,0.7491&27.68\,/\,0.7570&27.72\,/\,0.7620&27.49\,/\,0.7503&\textbf{28.01}\,/\,0.7674&27.83\,/\,\textbf{0.7702}\\ \hline
	\end{tabular}
	\label{tab:srquanres}
\end{table*}
\begin{figure*}[t]
	\centering
	\begin{minipage}{0.406\textwidth}
		\subfigure{
			\begin{minipage}{1\textwidth}
				\includegraphics[width=1\textwidth]{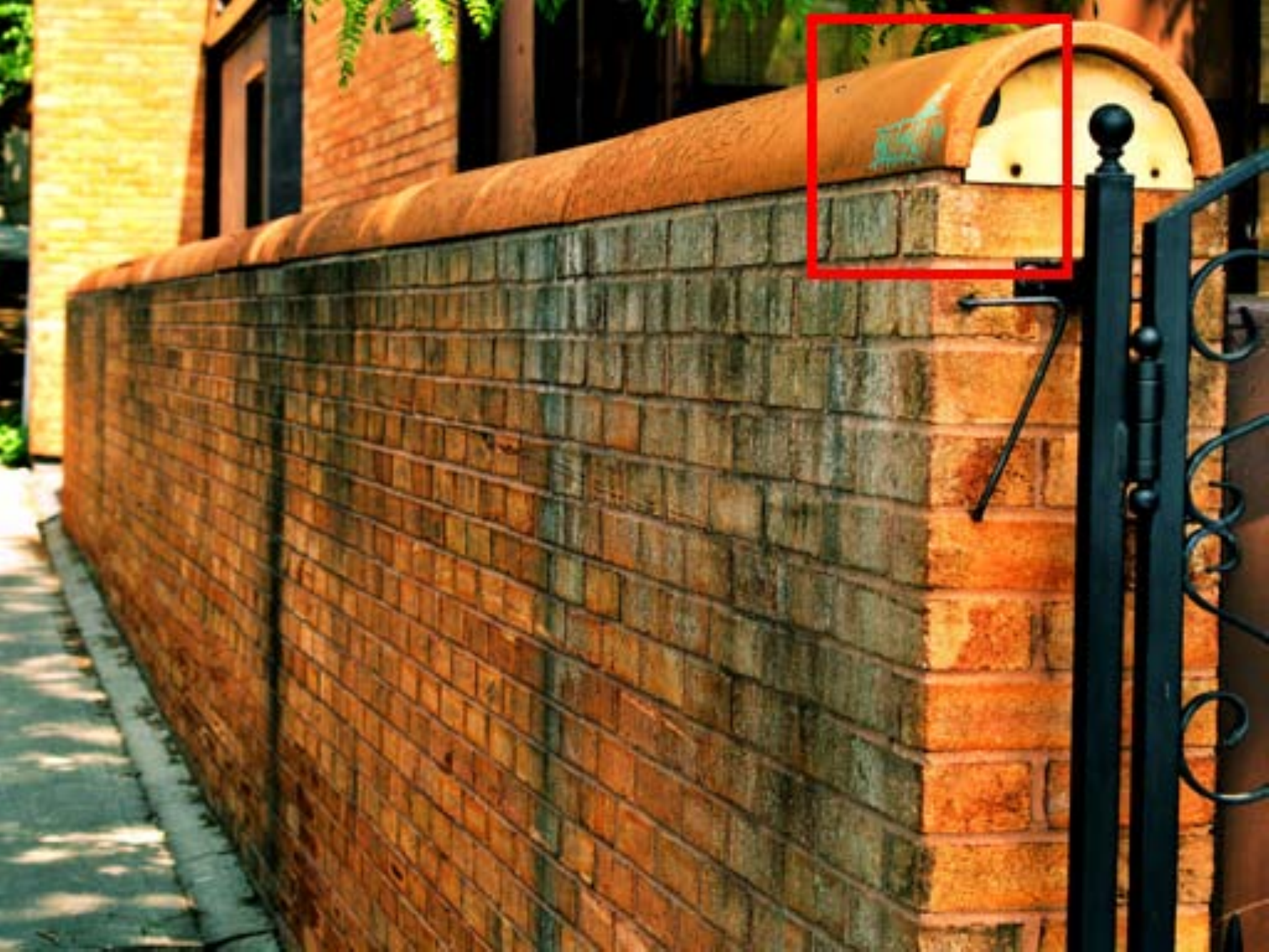}
				\centering  ``img\_018'' from Urban100~\cite{Huang2015Single} \vspace{1ex}\\
			\end{minipage}
		}
	\end{minipage}
	\begin{minipage}{0.13\textwidth}
		\subfigure{
			\begin{minipage}{1\textwidth}
				\includegraphics[width=1\textwidth]{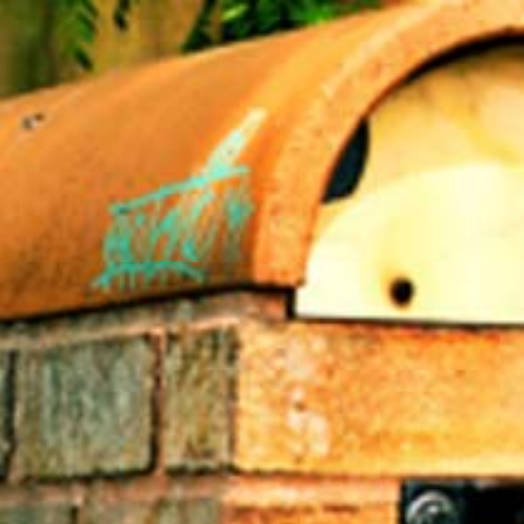}
				\centering \footnotesize \textendash \\ \footnotesize Ground Truth\\
			\end{minipage}
		}
		\subfigure{
			\begin{minipage}{1\textwidth}
				\includegraphics[width=1\textwidth]{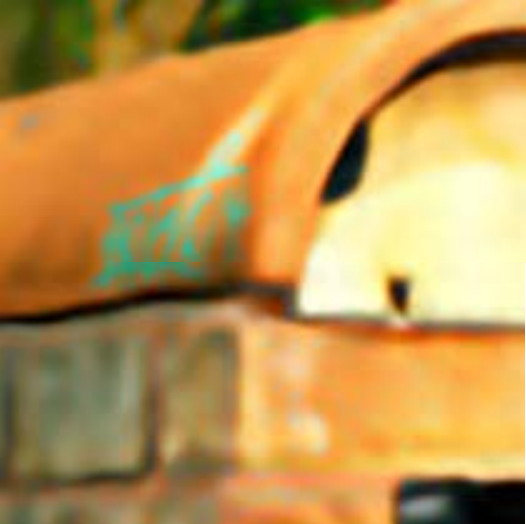}
				\centering \footnotesize 25.38 / 0.6402 \\ TNRD\\
			\end{minipage}
		}
	\end{minipage}
	\begin{minipage}{0.13\textwidth}
		\subfigure{
			\begin{minipage}{1\textwidth}
				\includegraphics[width=1\textwidth]{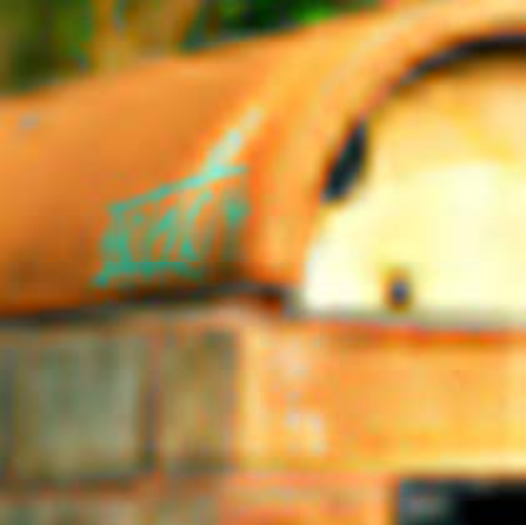}
				\centering \footnotesize 24.11 / 0.5666 \\ Bicubic \\
			\end{minipage}
		}
		\subfigure{
			\begin{minipage}{1\textwidth}
				\includegraphics[width=1\textwidth]{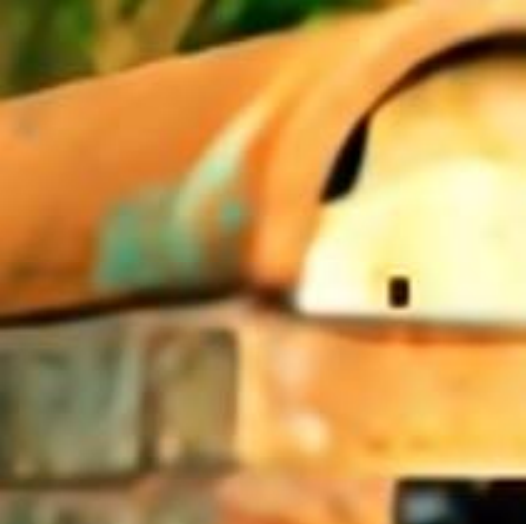}
				\centering \footnotesize 25.52 / 0.6488 \\ IRCNN\\
			\end{minipage}
		}
	\end{minipage}
	\begin{minipage}{0.13\textwidth}
		\subfigure{
			\begin{minipage}{1\textwidth}
				\includegraphics[width=1\textwidth]{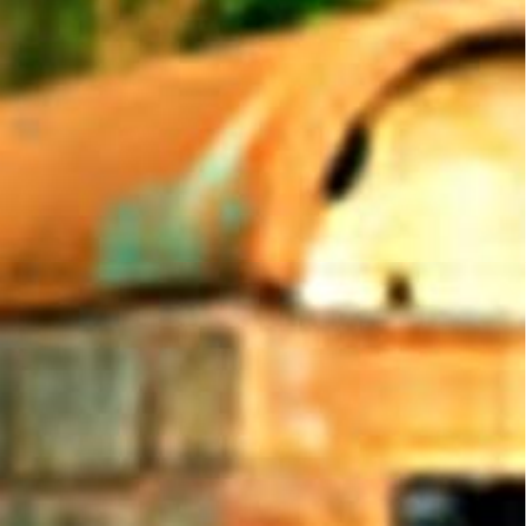}
				\centering \footnotesize 24.79 / 0.6133 \\ \footnotesize A$+$\\
			\end{minipage}
		}
		\subfigure{
			\begin{minipage}{1\textwidth}
				\includegraphics[width=1\textwidth]{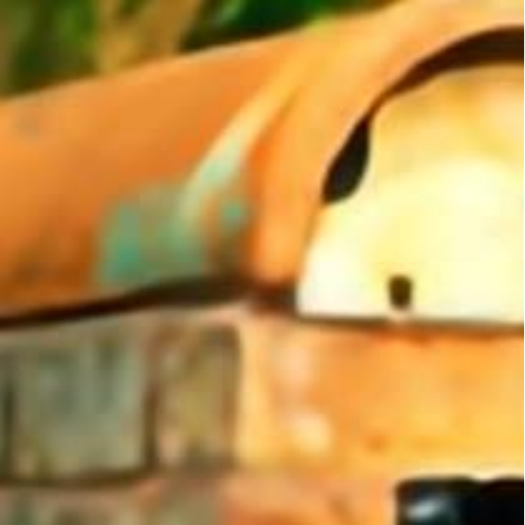}
				\centering \footnotesize 25.47 / 0.6464 \\ VDSR\\
			\end{minipage}
		}
	\end{minipage}
	\begin{minipage}{0.13\textwidth}
		\subfigure{
			\begin{minipage}{1\textwidth}
				\includegraphics[width=1\textwidth]{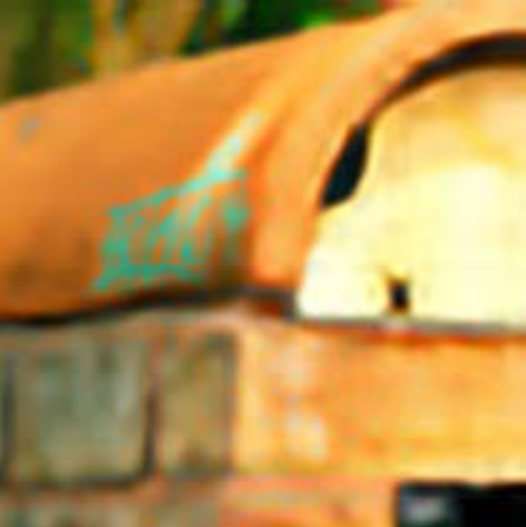}
				\centering \footnotesize 25.27 / 0.6322 \\ SRCNN\\
			\end{minipage}
		}
		\subfigure{
			\begin{minipage}{1\textwidth}
				\includegraphics[width=1\textwidth]{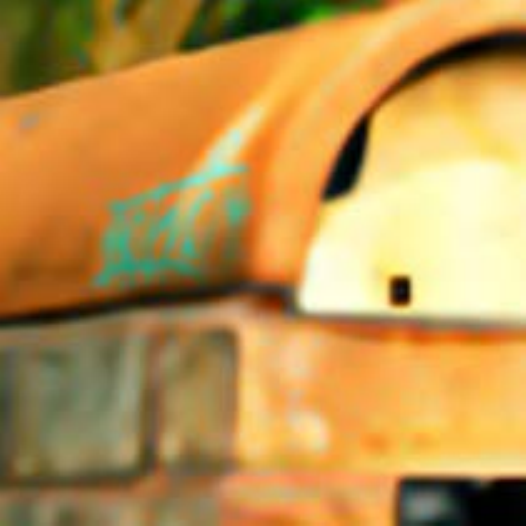}
				\centering \footnotesize \textbf{25.58 / 0.6501} \\ Ours\\
			\end{minipage}
		}
	\end{minipage}
	\caption{The performances on $\times$4 super-resolution with zoomed in comparisons. The quantitative scores (i.e., PSNR / SSIM) are also reported accordantly.}
	\label{fig:srvisres}
\end{figure*}
\begin{table*}[htb]
	\centering
	\caption{Average dehazing performance on Fattal's and D-Hazy benchmarks. }
	\begin{tabular}{|c|c|c|c|c|c|c|c|c|c|c|c|c|}
		\hline
		~&\multicolumn{6}{|c|}{Fattal's benchmark~\cite{Fattal2014}}&\multicolumn{6}{|c|}{D-Hazy benchmark~\cite{AncutiDHazy}}\\
		\hline
		Metric&~\cite{He2011Single}&\cite{Meng2014Efficient}&\cite{Cai2016DehazeNet}&\cite{Berman2016Non}&\cite{Ren2016Single}&Ours&~\cite{He2011Single}&\cite{Meng2014Efficient}&\cite{Cai2016DehazeNet}&\cite{Berman2016Non}&\cite{Ren2016Single}&Ours\\
		\hline
		PSNR&27.11&26.13&21.63&26.09&24.40&\textbf{28.61}&16.32&15.82&15.61&16.44&13.91&\textbf{17.10}\\
		\hline
		SSIM&0.958&0.951&0.891&0.954&0.934&\textbf{0.964}&0.841&0.850&0.839&0.839&0.820&\textbf{0.866}\\
		\hline
		$L_1$ Error&0.034&0.040&0.073&0.041&0.047&\textbf{0.030}&0.123&0.146&0.145&0.124&0.184&\textbf{0.111}\\
		\hline
	\end{tabular}
	\label{tab:dehazquanres}
\end{table*}

\begin{figure*}[t]
	\centering
	\begin{tabular}{c@{\extracolsep{0.1em}}c@{\extracolsep{0.1em}}c@{\extracolsep{0.1em}}c@{\extracolsep{0.1em}}c@{\extracolsep{0.1em}}c@{\extracolsep{0.1em}}c@{\extracolsep{0.1em}}c}	
		\includegraphics[width=0.122\textwidth]{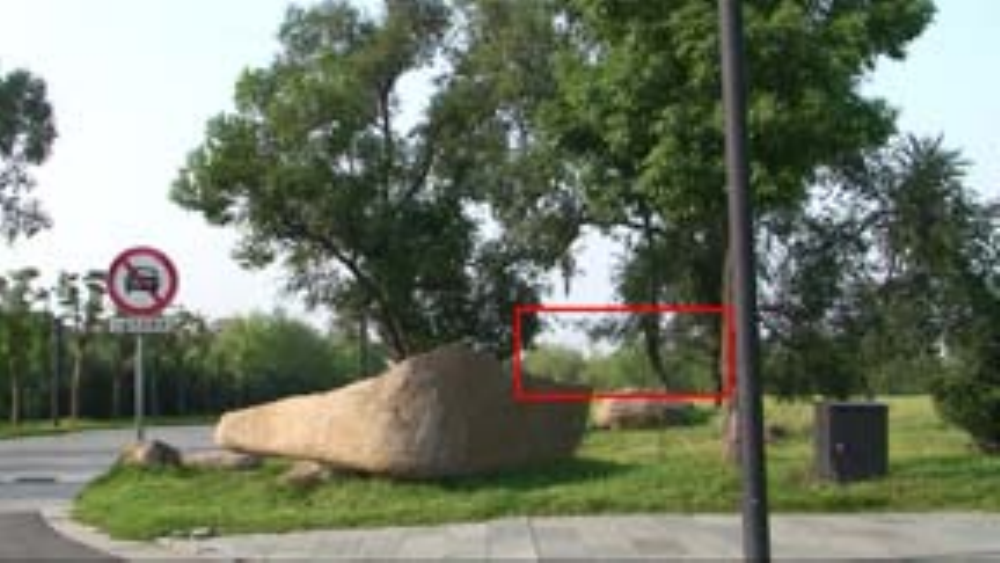}&	
		\includegraphics[width=0.122\textwidth]{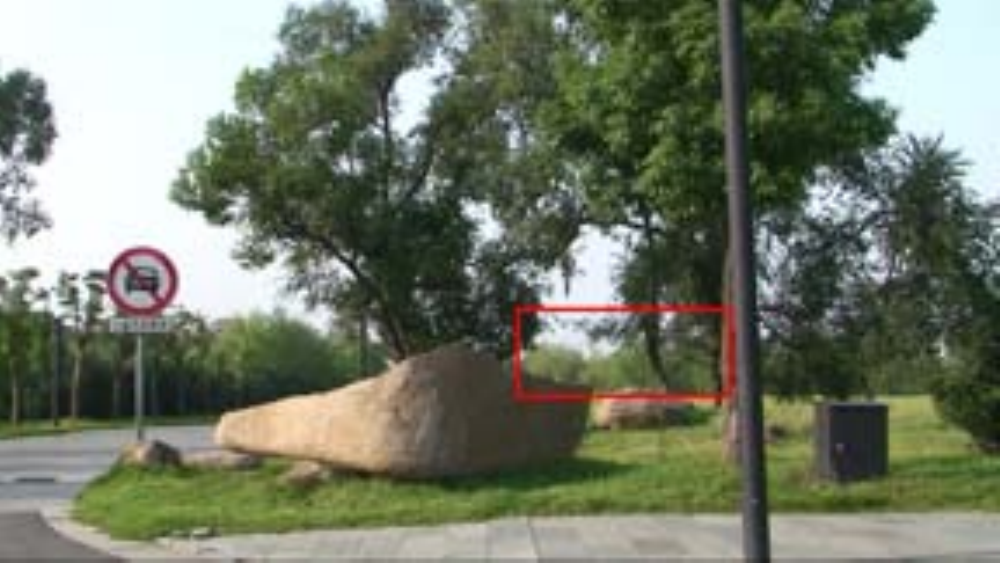}&	
		\includegraphics[width=0.122\textwidth]{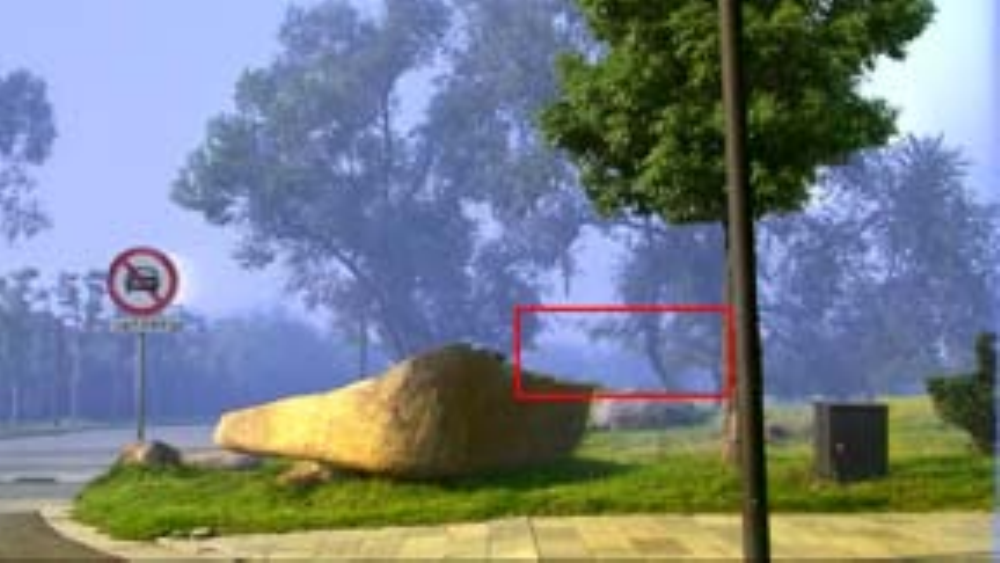}&
		\includegraphics[width=0.122\textwidth]{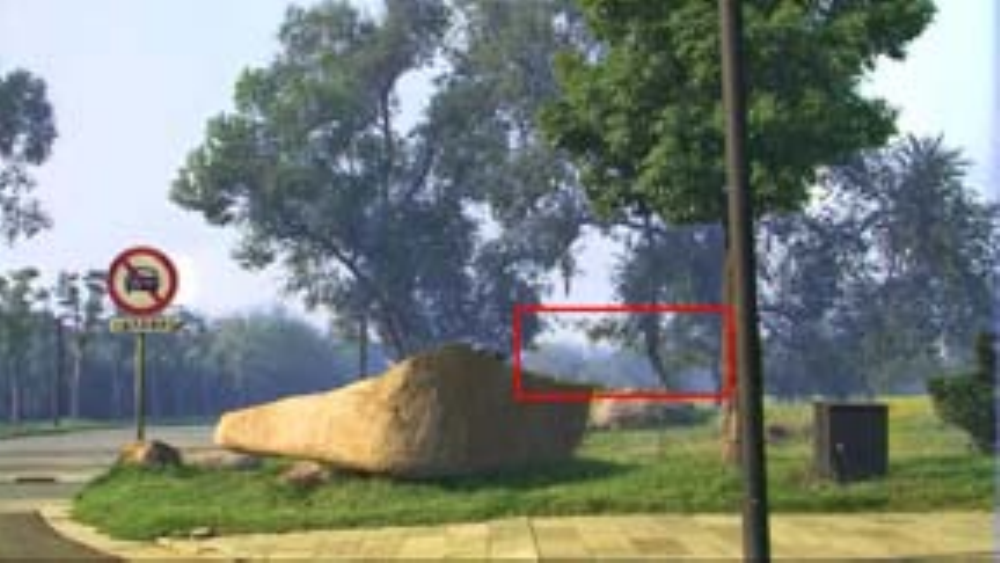}&
		\includegraphics[width=0.122\textwidth]{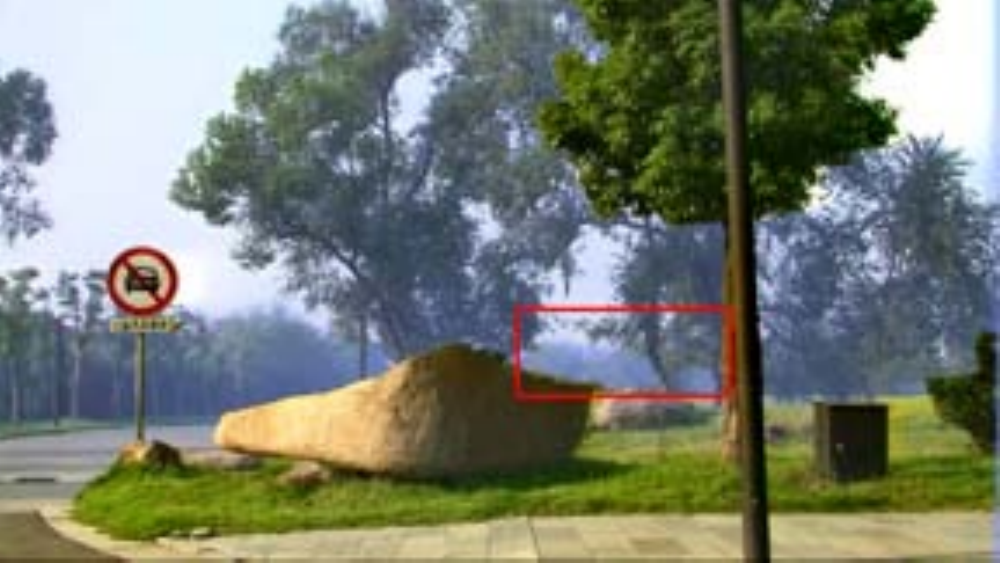}&
		\includegraphics[width=0.122\textwidth]{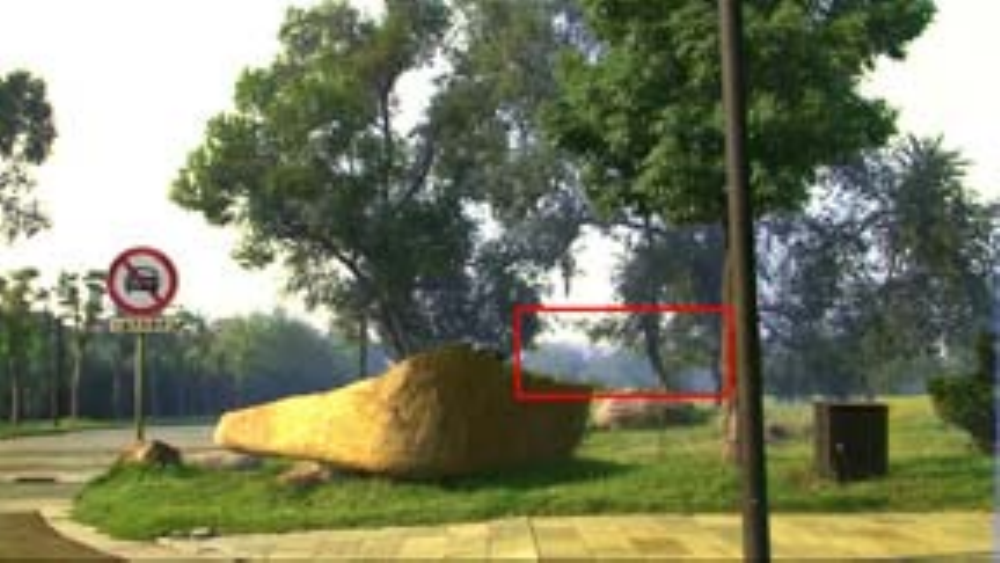}&
		\includegraphics[width=0.122\textwidth]{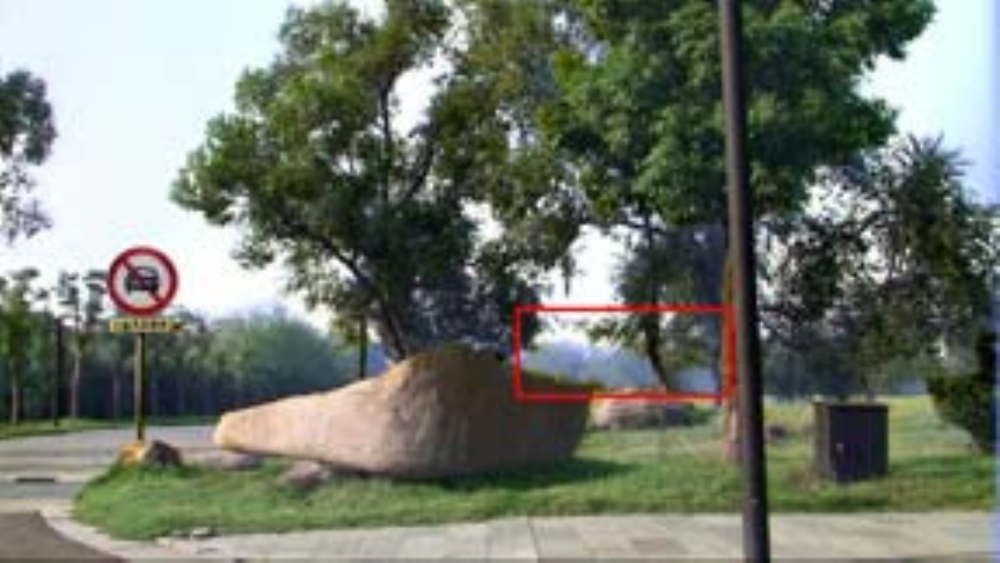}&
		\includegraphics[width=0.122\textwidth]{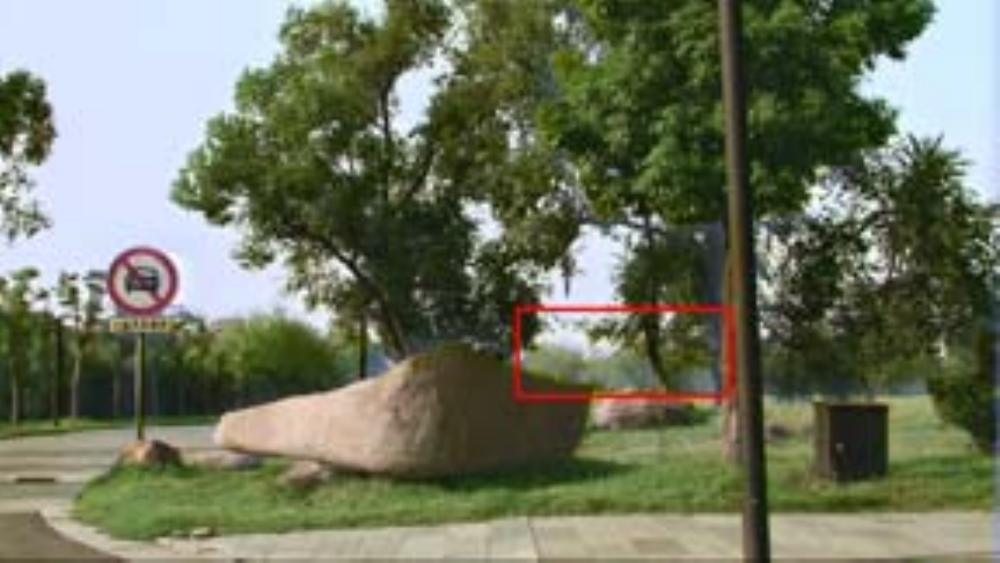}\\
		--& \footnotesize 0.8122 / 0.1304
		&  \footnotesize 0.9680 / 0.0313 & \footnotesize 0.9463 / 0.0538	&\footnotesize 0.8420 / 0.1135 & \footnotesize 0.9699 / 0.0313 & \footnotesize 0.9252 / 0.0632 & \footnotesize \textbf{0.9786 / 0.0300}\\
		\includegraphics[width=0.122\textwidth]{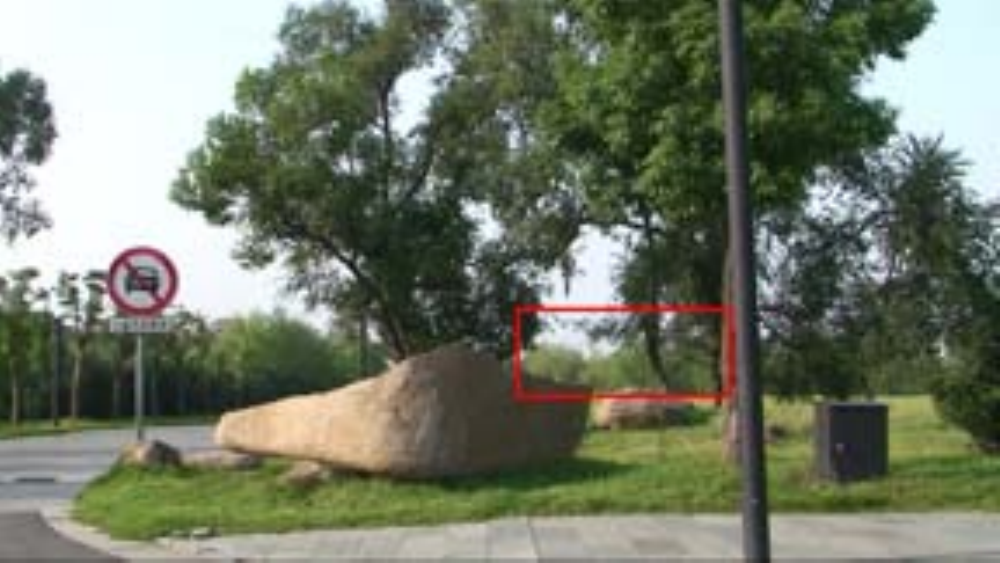}&	
		\includegraphics[width=0.122\textwidth]{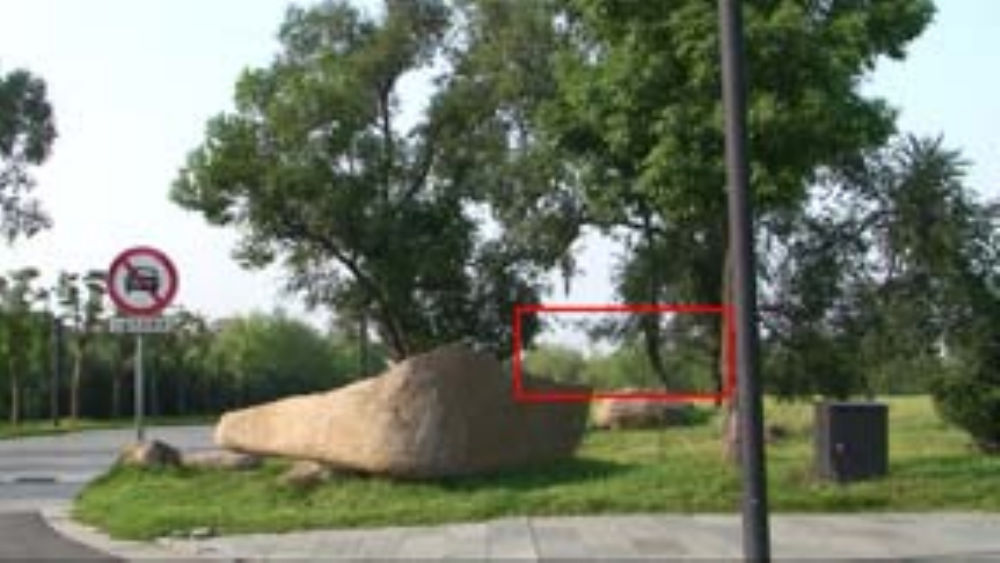}&
		\includegraphics[width=0.122\textwidth]{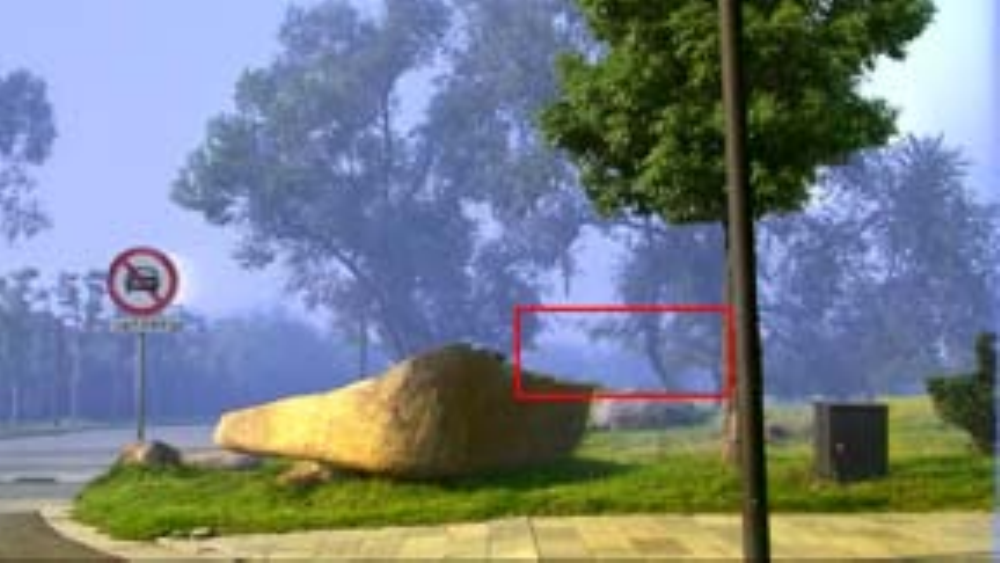}&
		\includegraphics[width=0.122\textwidth]{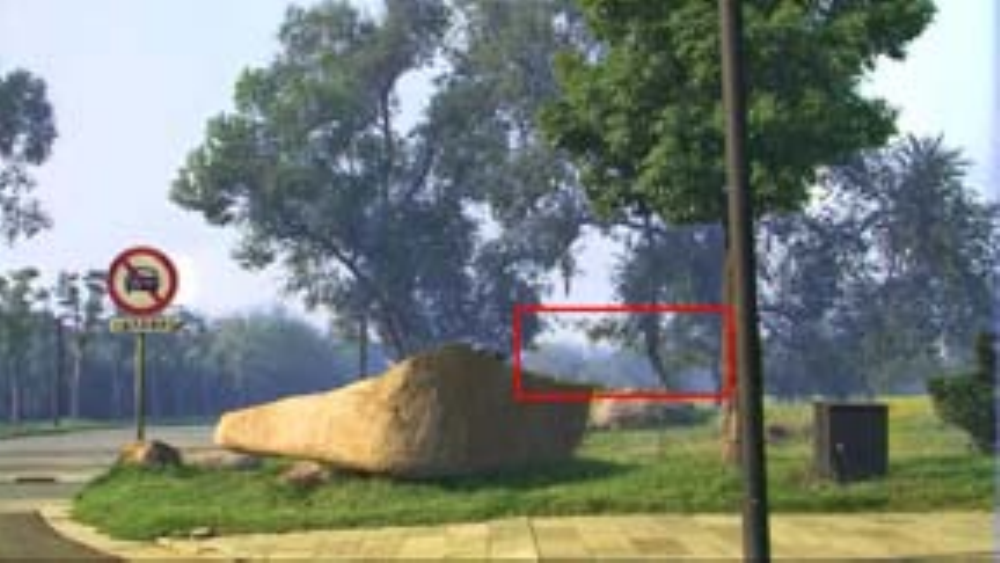}&
		\includegraphics[width=0.122\textwidth]{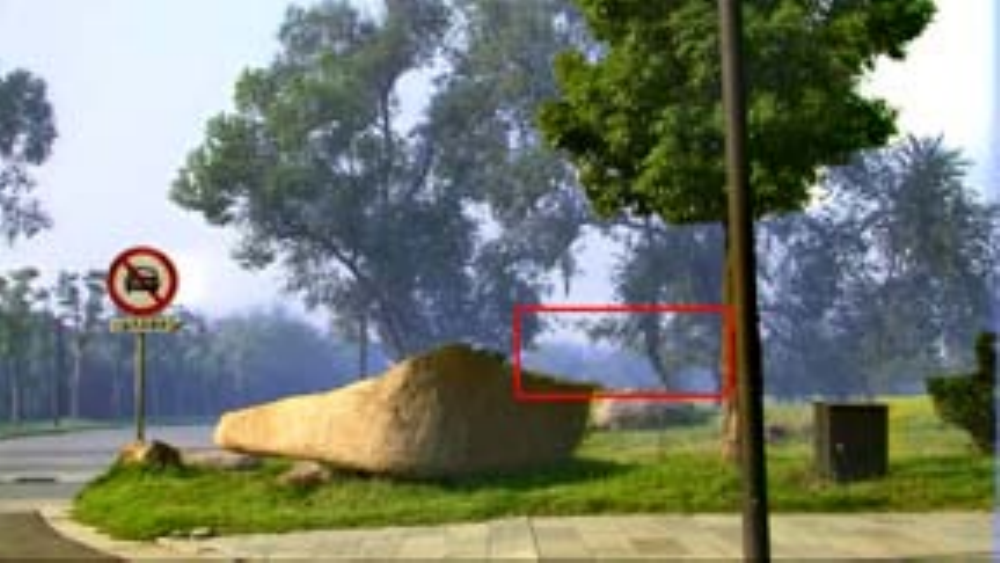}&
		\includegraphics[width=0.122\textwidth]{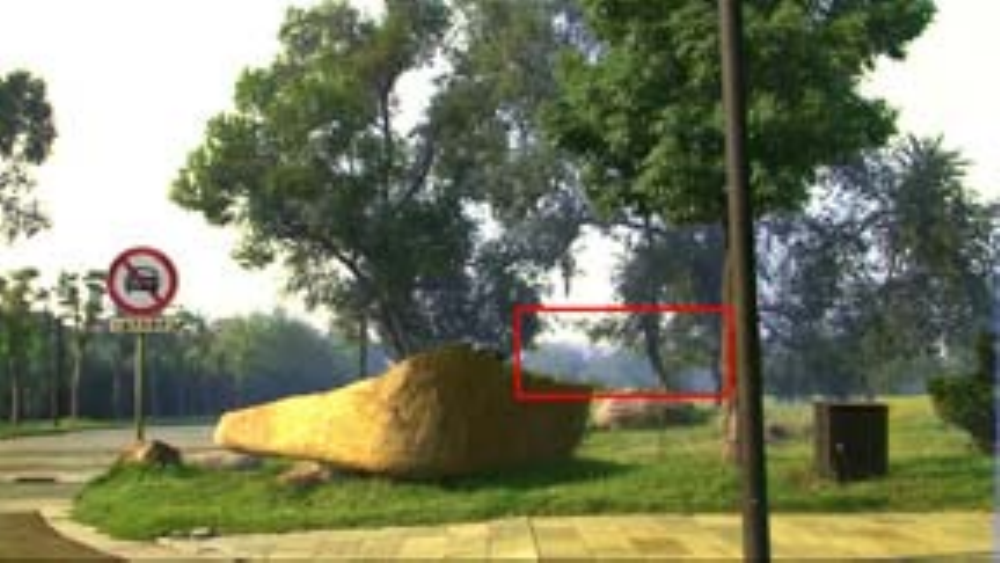}&
		\includegraphics[width=0.122\textwidth]{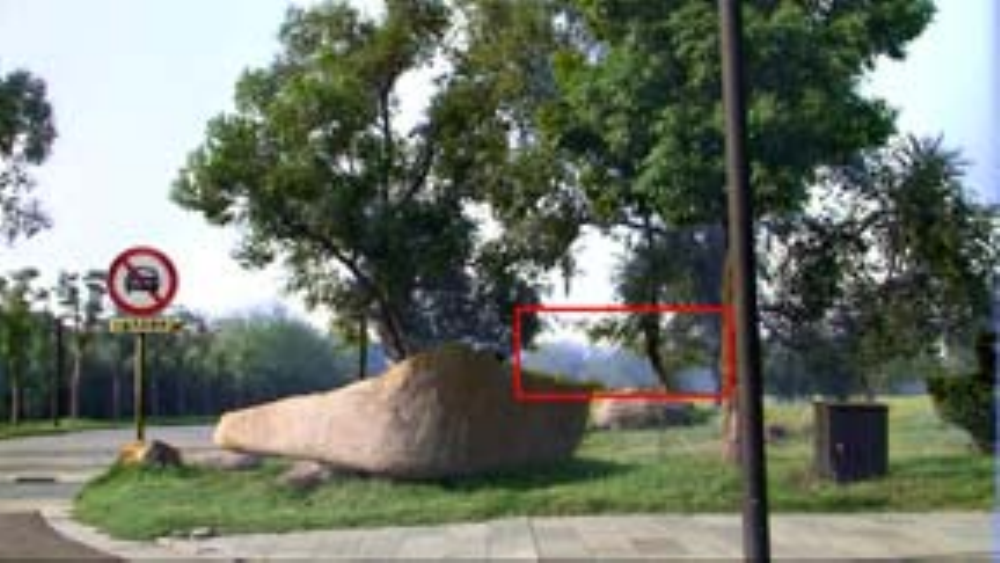}&
		\includegraphics[width=0.122\textwidth]{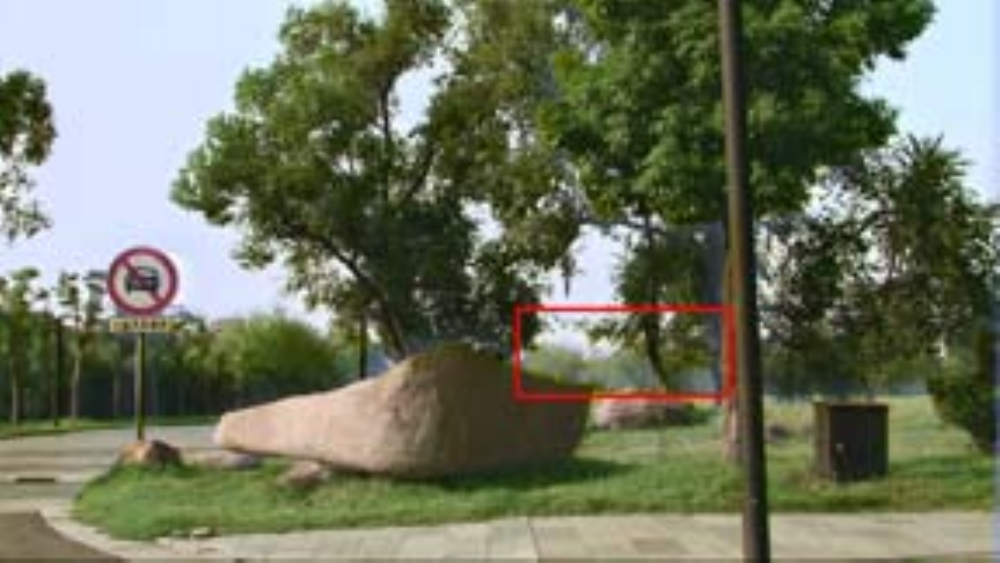}\\
		--& \footnotesize 0.8487 / 0.1120
		&\footnotesize 0.8514 / 0.1112 & \footnotesize 0.9579 / 0.0472 & \footnotesize 0.9334 / 0.0613 & \footnotesize 0.9754 / 0.0363 & \footnotesize 0.9690 / 0.0315 & \footnotesize \textbf{0.9815 / 0.0308}\\
		\footnotesize Ground Truth &\footnotesize Hazy Input &~\cite{Cai2016DehazeNet} & \footnotesize~\cite{Meng2014Efficient}& \footnotesize~\cite{Ren2016Single}  & \footnotesize\cite{He2011Single}  & \footnotesize~\cite{Berman2016Non}& \footnotesize Ours
	\end{tabular}
	\caption{The enhancement performances on example hazy images in Fattal's synthetic benchmark~\cite{Fattal2014}. The quantitative scores (i.e., SSIM / $L_1$ Error) are also reported accordantly.}
	\label{fig:sydehazvisres}
\end{figure*}

\begin{figure*}[t]
	\centering
	\begin{tabular}{c@{\extracolsep{0.1em}}c@{\extracolsep{0.1em}}c@{\extracolsep{0.1em}}c@{\extracolsep{0.1em}}c@{\extracolsep{0.1em}}c@{\extracolsep{0.1em}}c}	
		\includegraphics[width=0.122\textwidth]{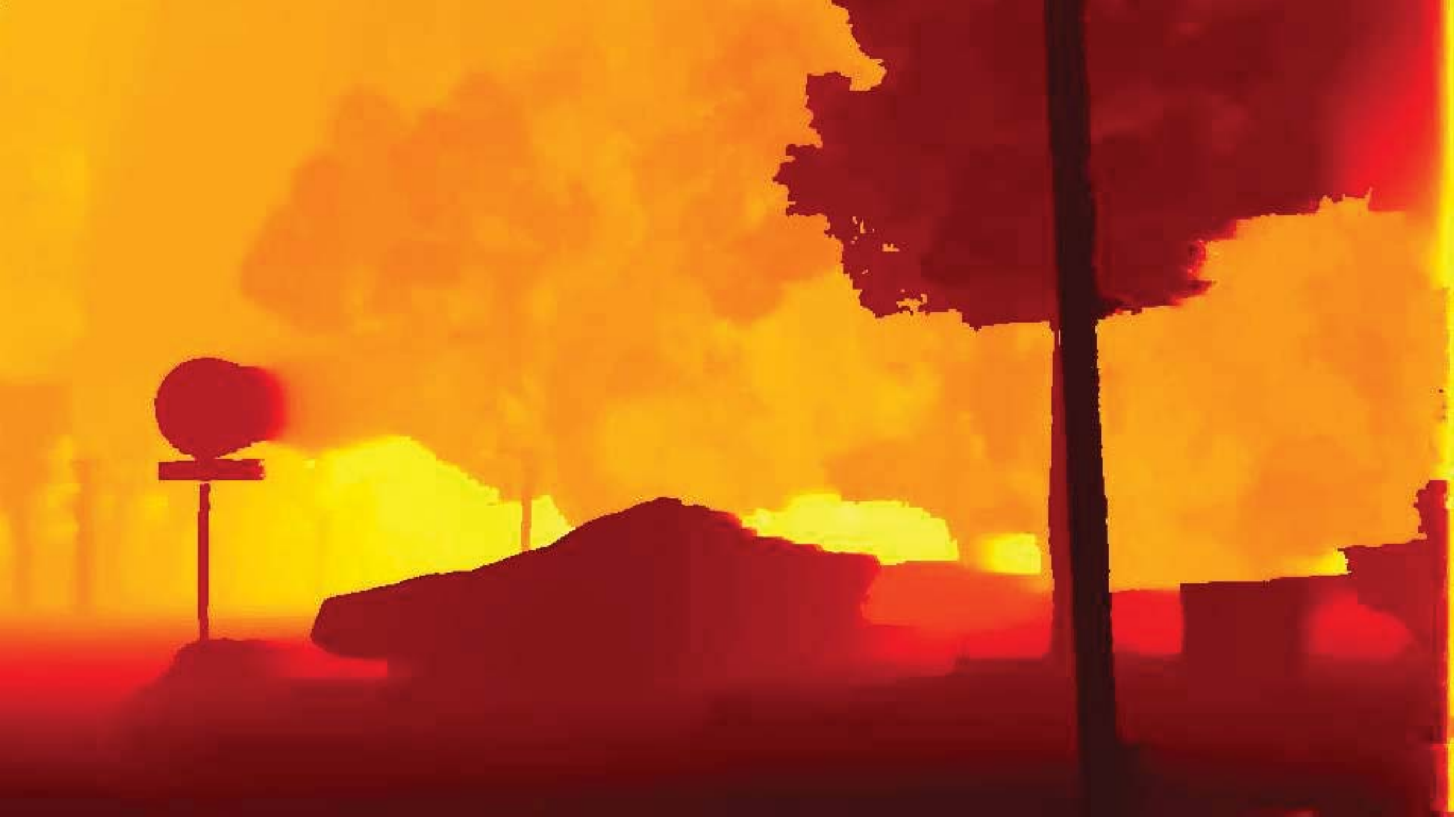}&
		\includegraphics[width=0.122\textwidth]{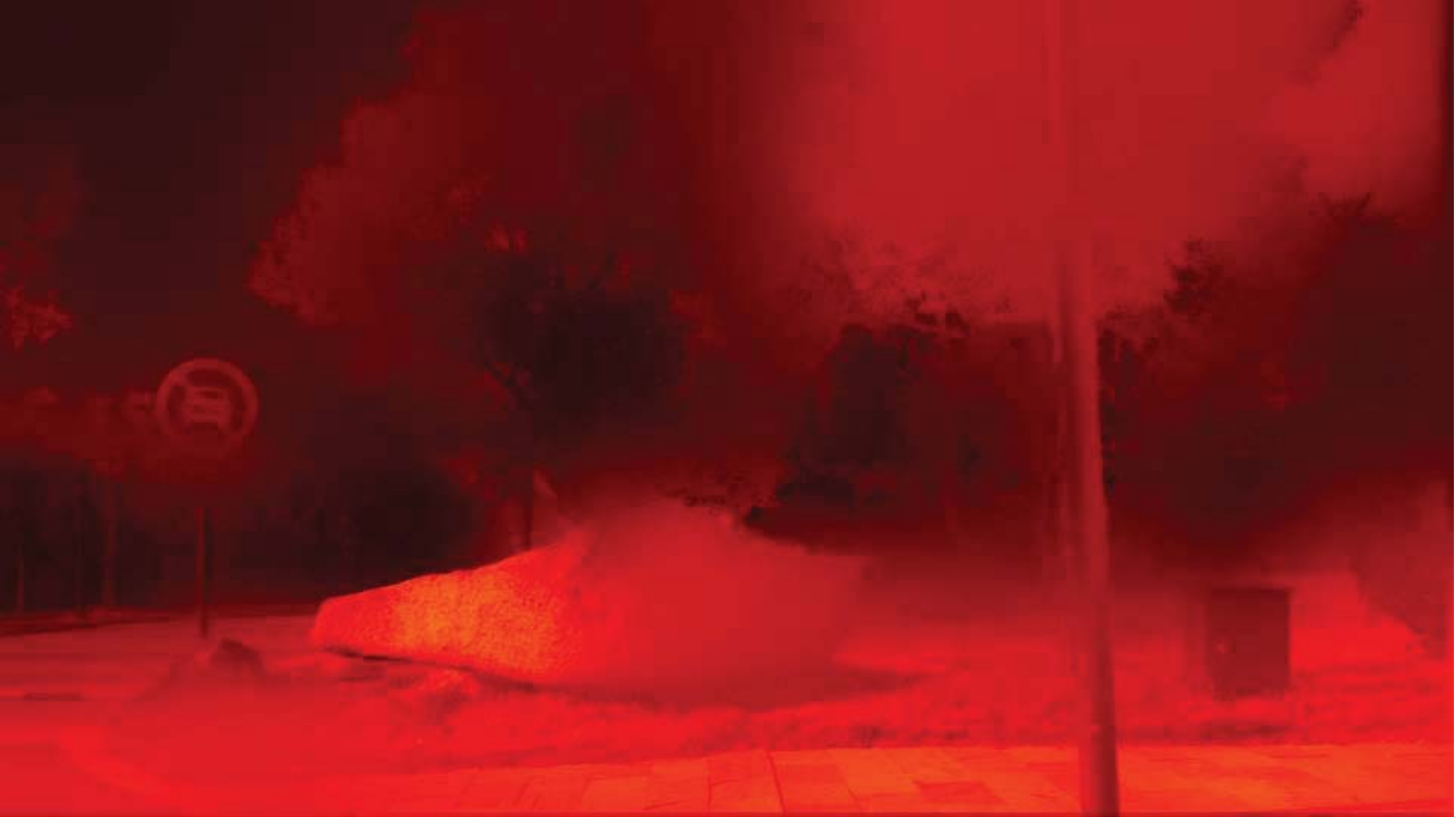}&
		\includegraphics[width=0.122\textwidth]{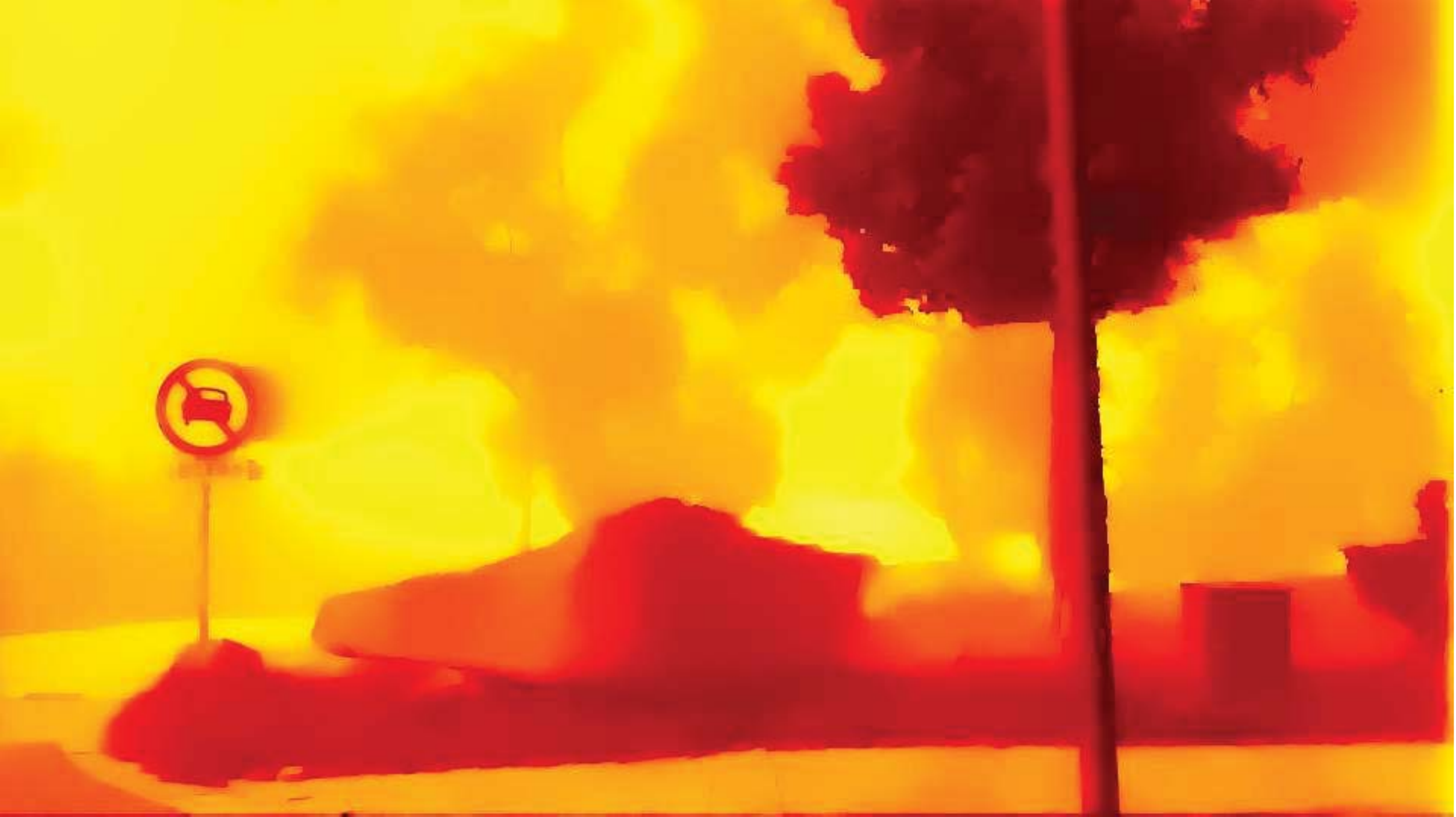}&		
		\includegraphics[width=0.122\textwidth]{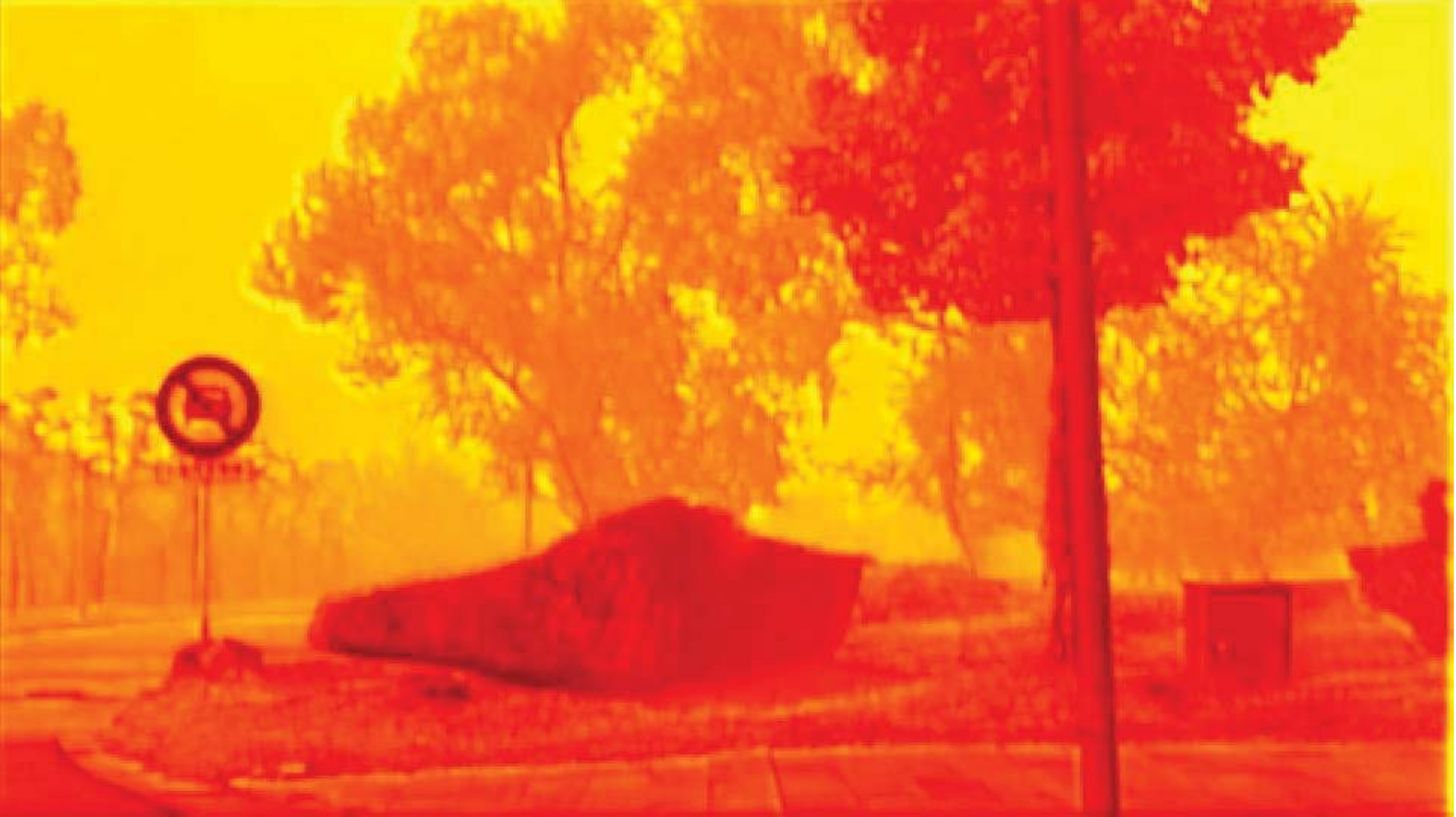}&
		\includegraphics[width=0.122\textwidth]{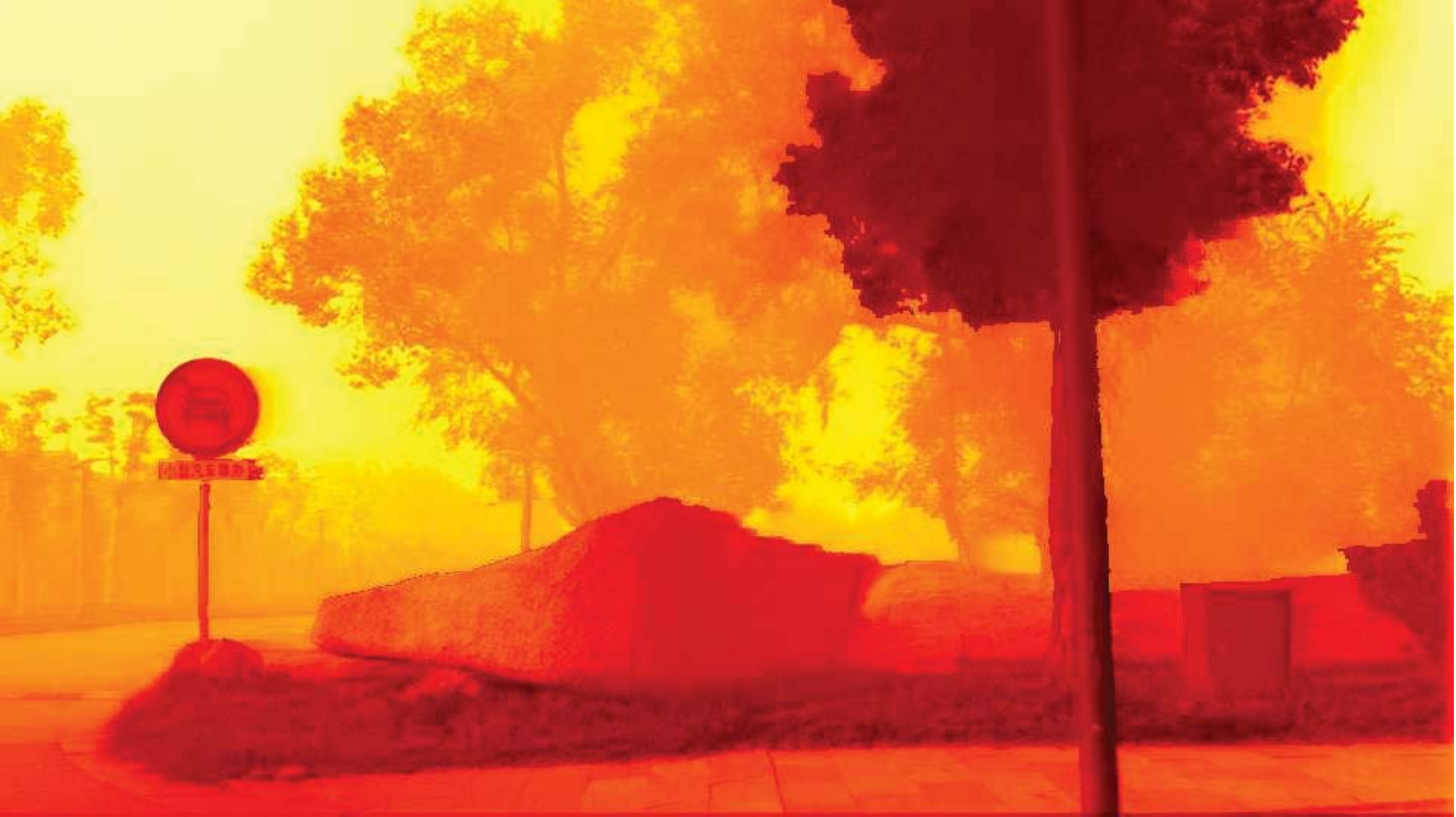}&
		\includegraphics[width=0.122\textwidth]{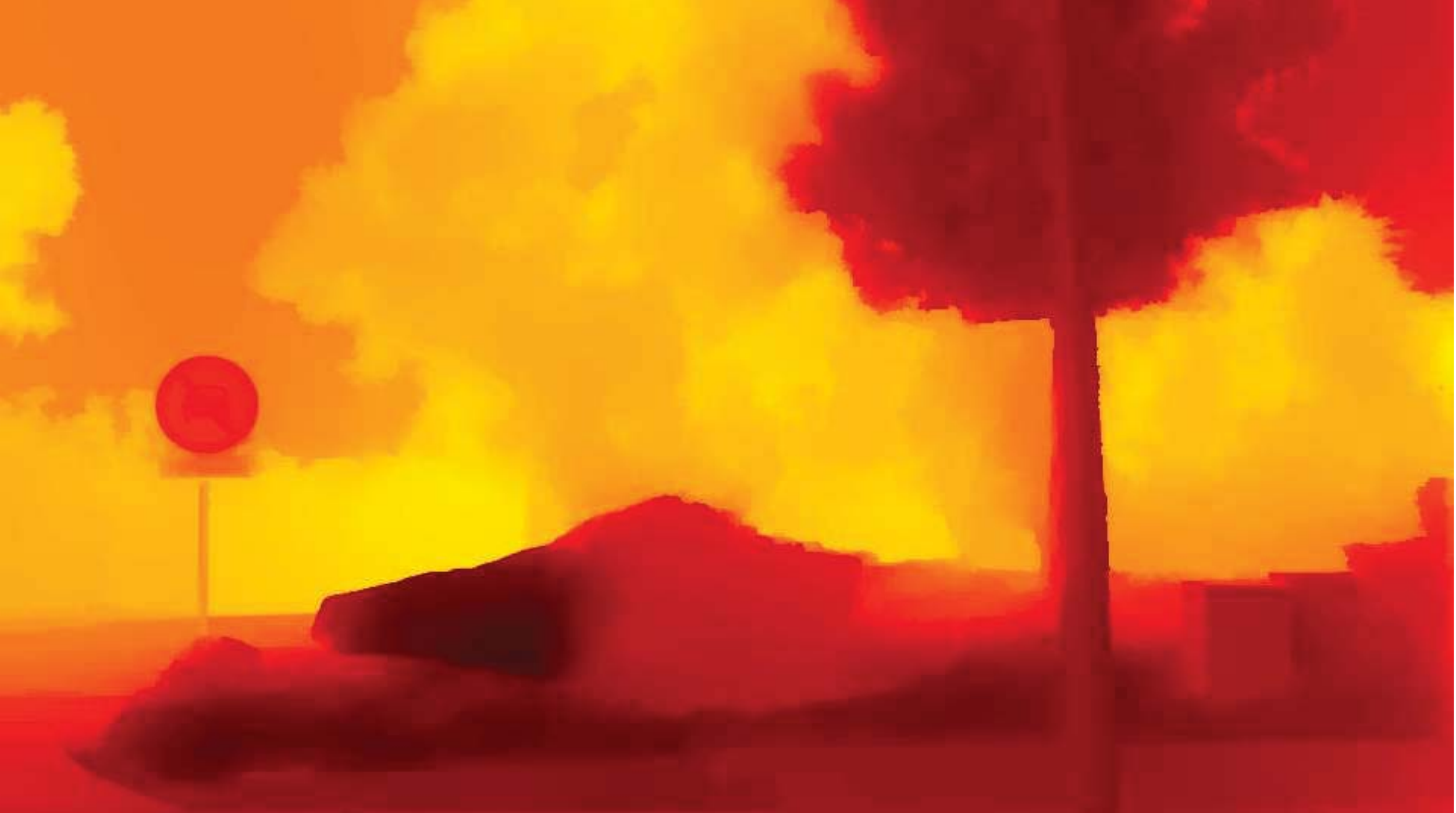}&
		\includegraphics[width=0.122\textwidth]{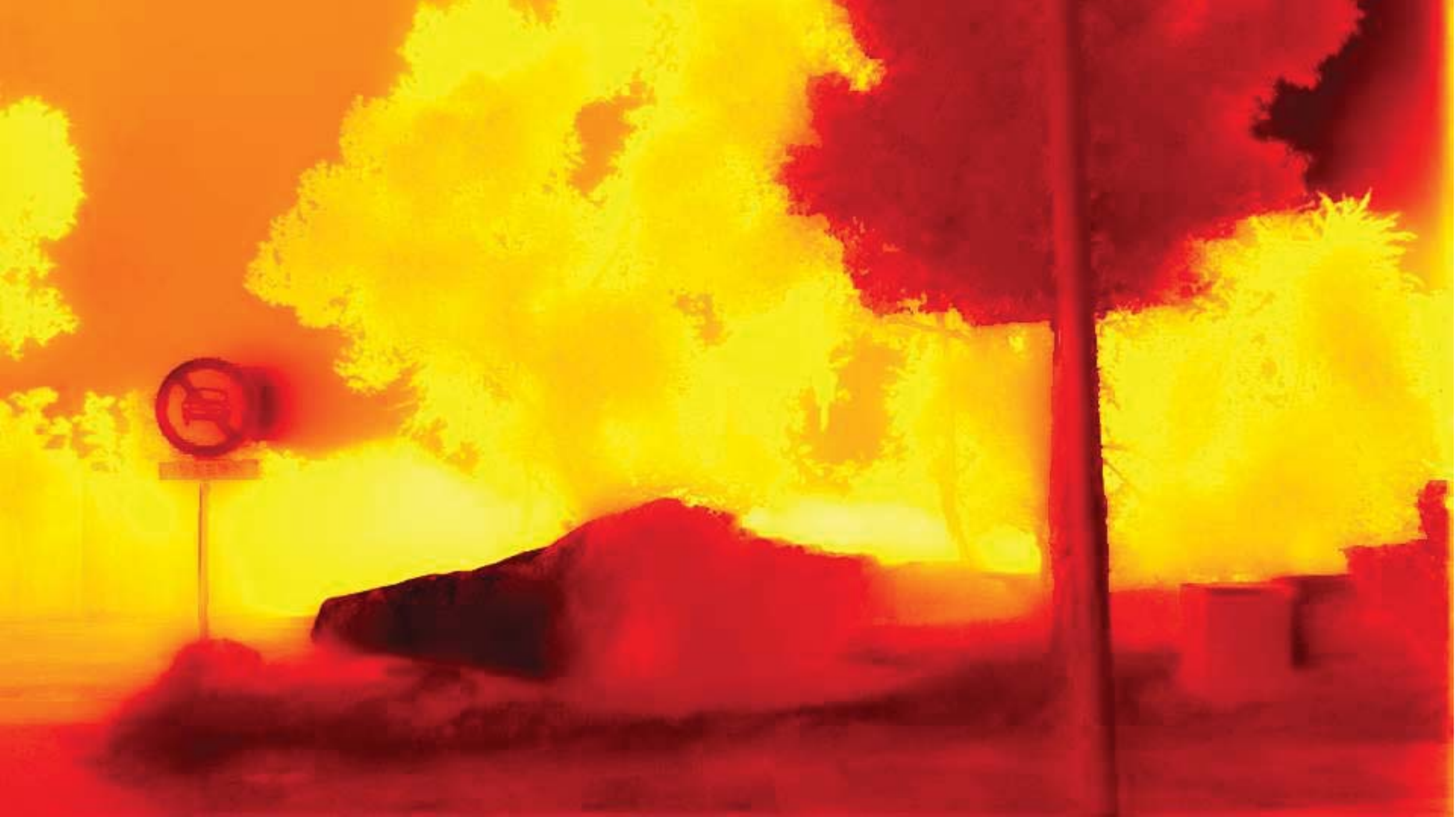}\\
		\includegraphics[width=0.122\textwidth]{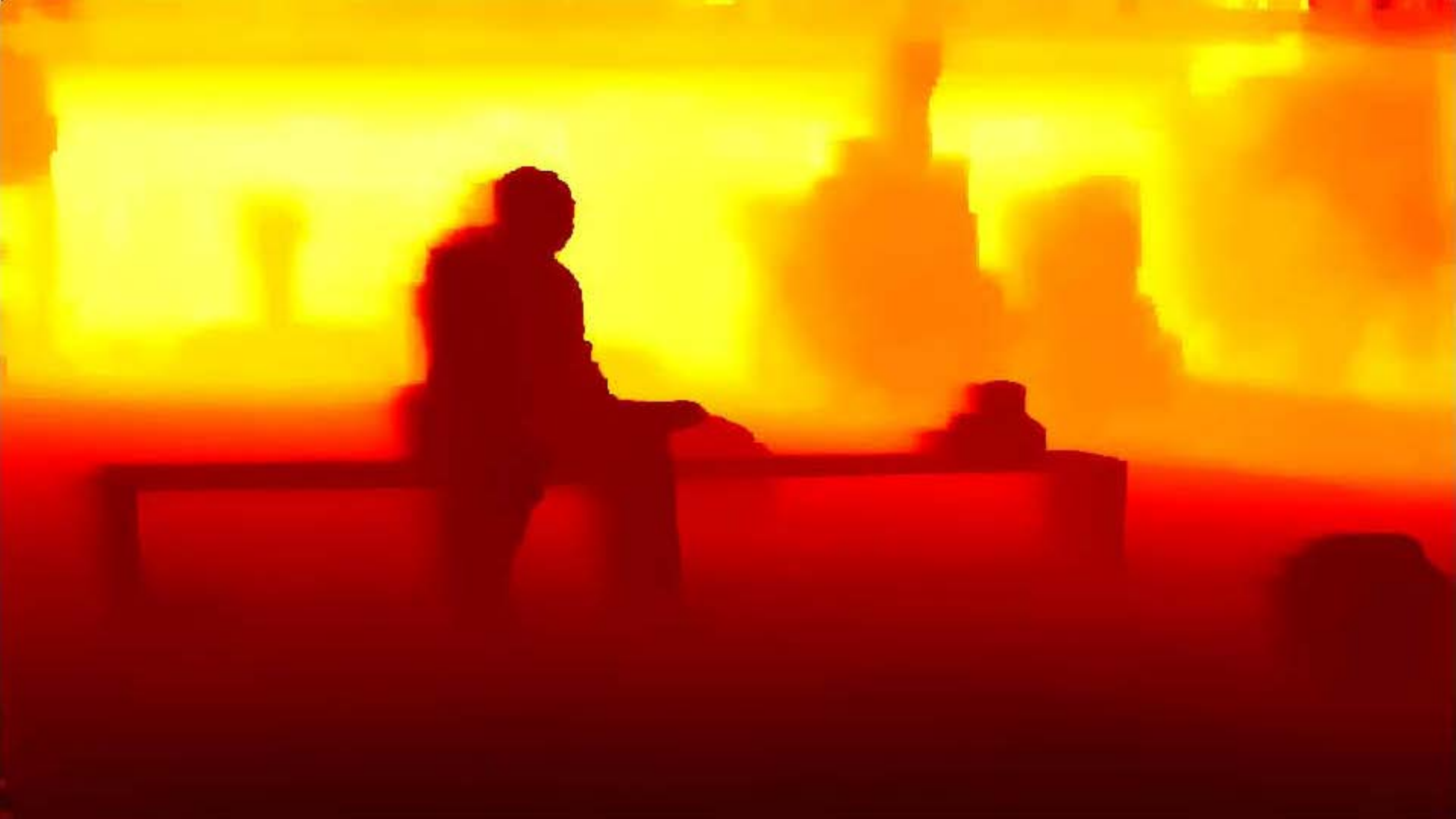}&
		\includegraphics[width=0.122\textwidth]{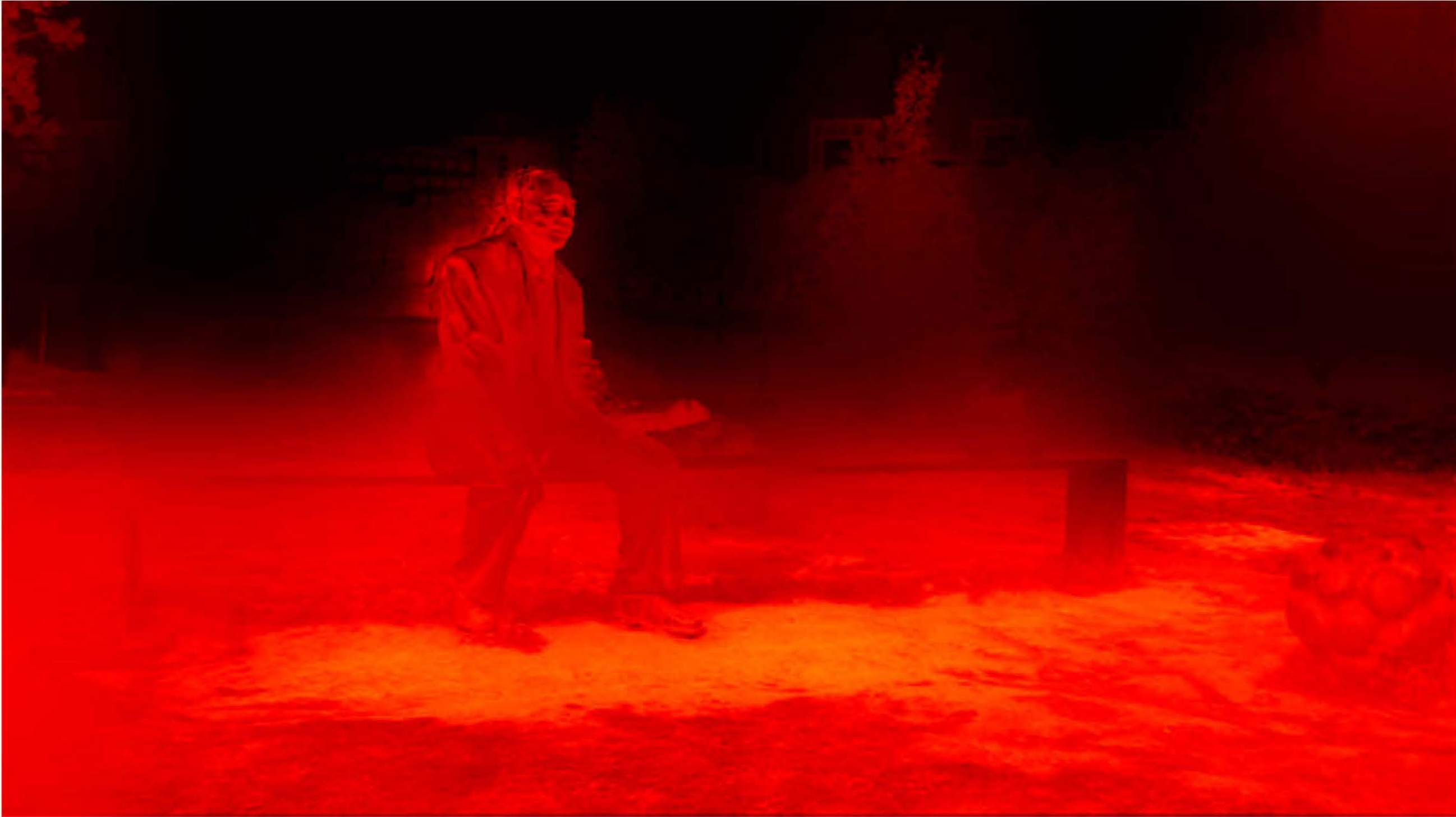}&
		\includegraphics[width=0.122\textwidth]{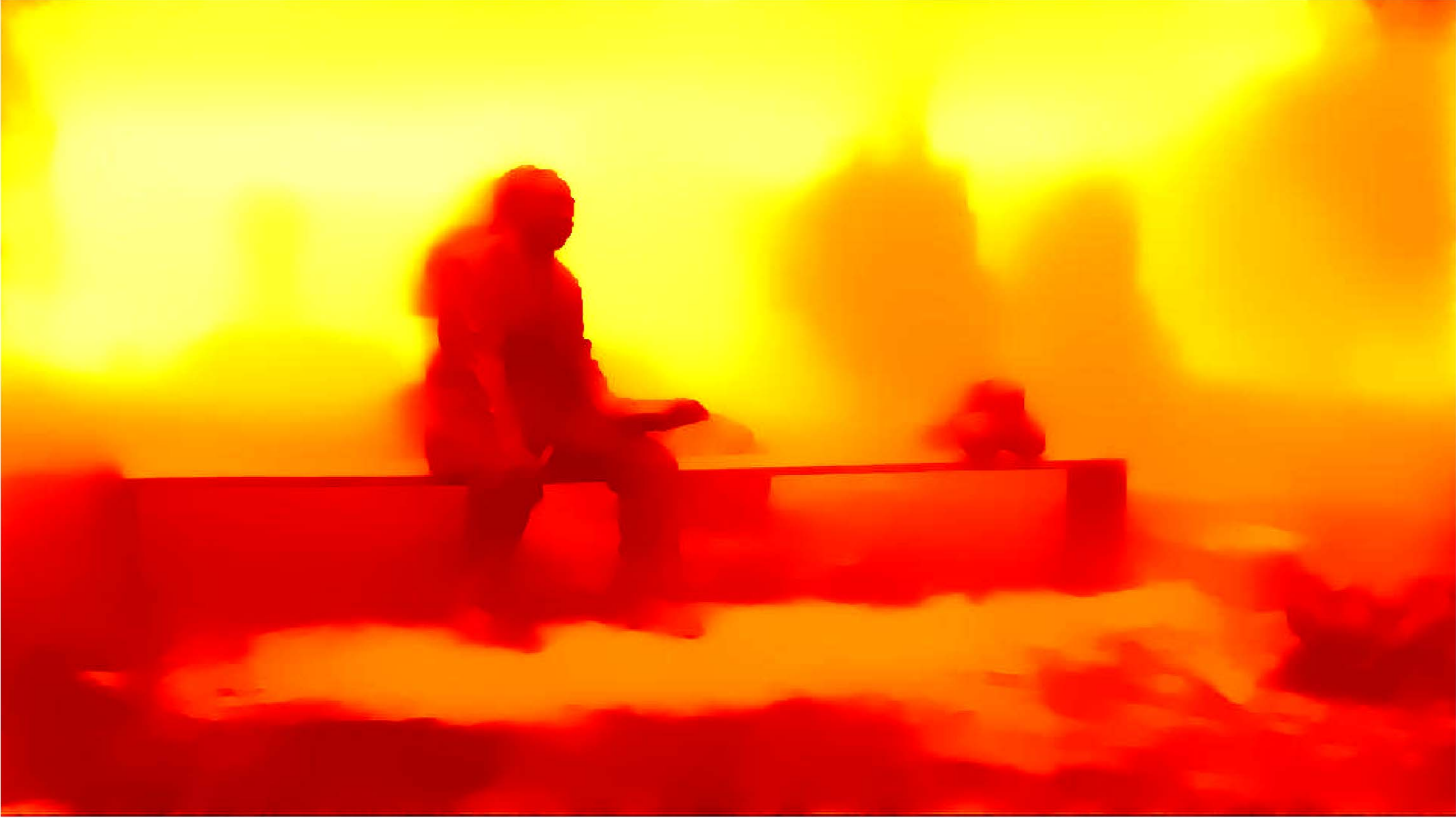}&	
		\includegraphics[width=0.122\textwidth]{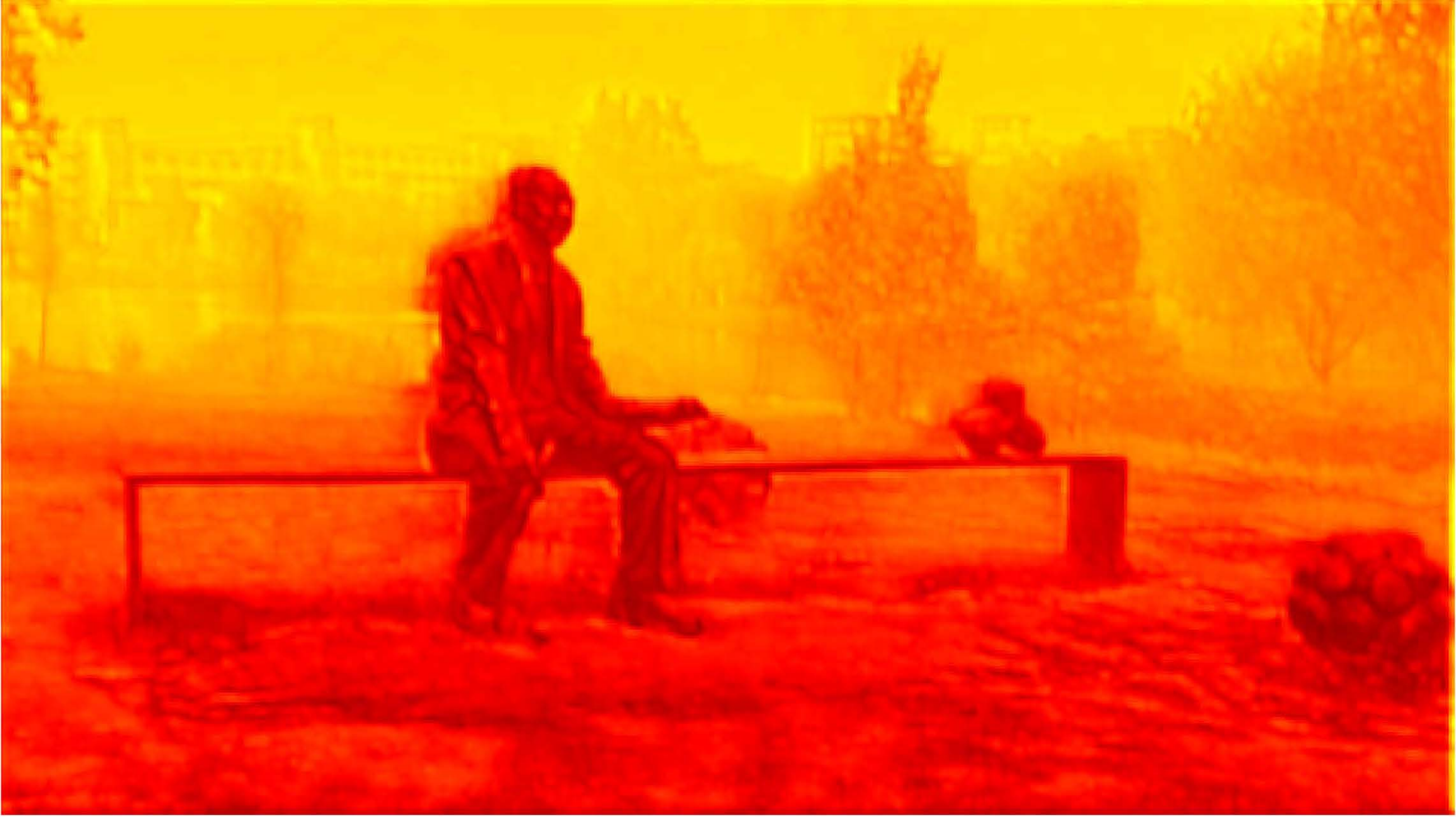}&
		\includegraphics[width=0.122\textwidth]{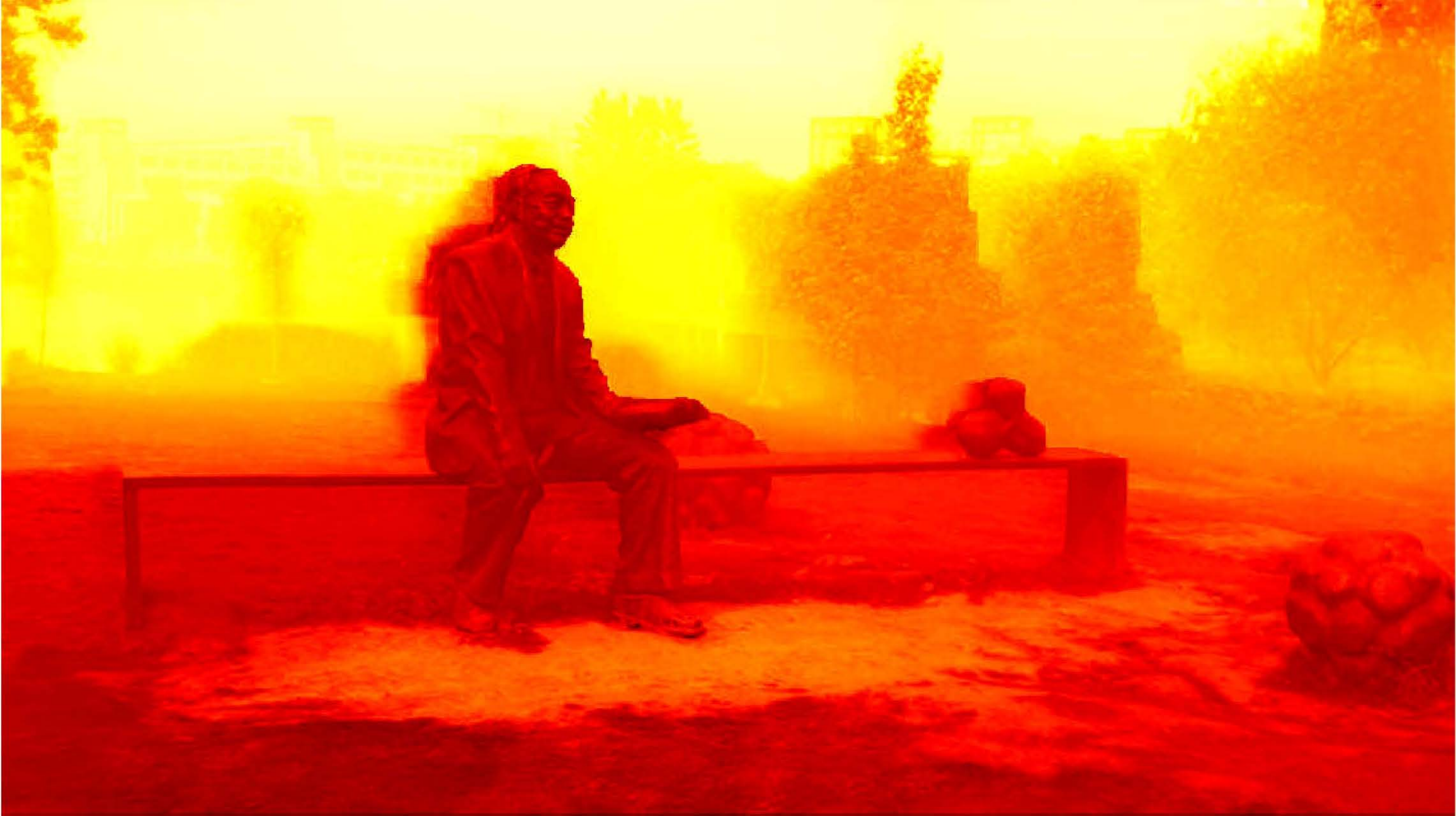}&
		\includegraphics[width=0.122\textwidth]{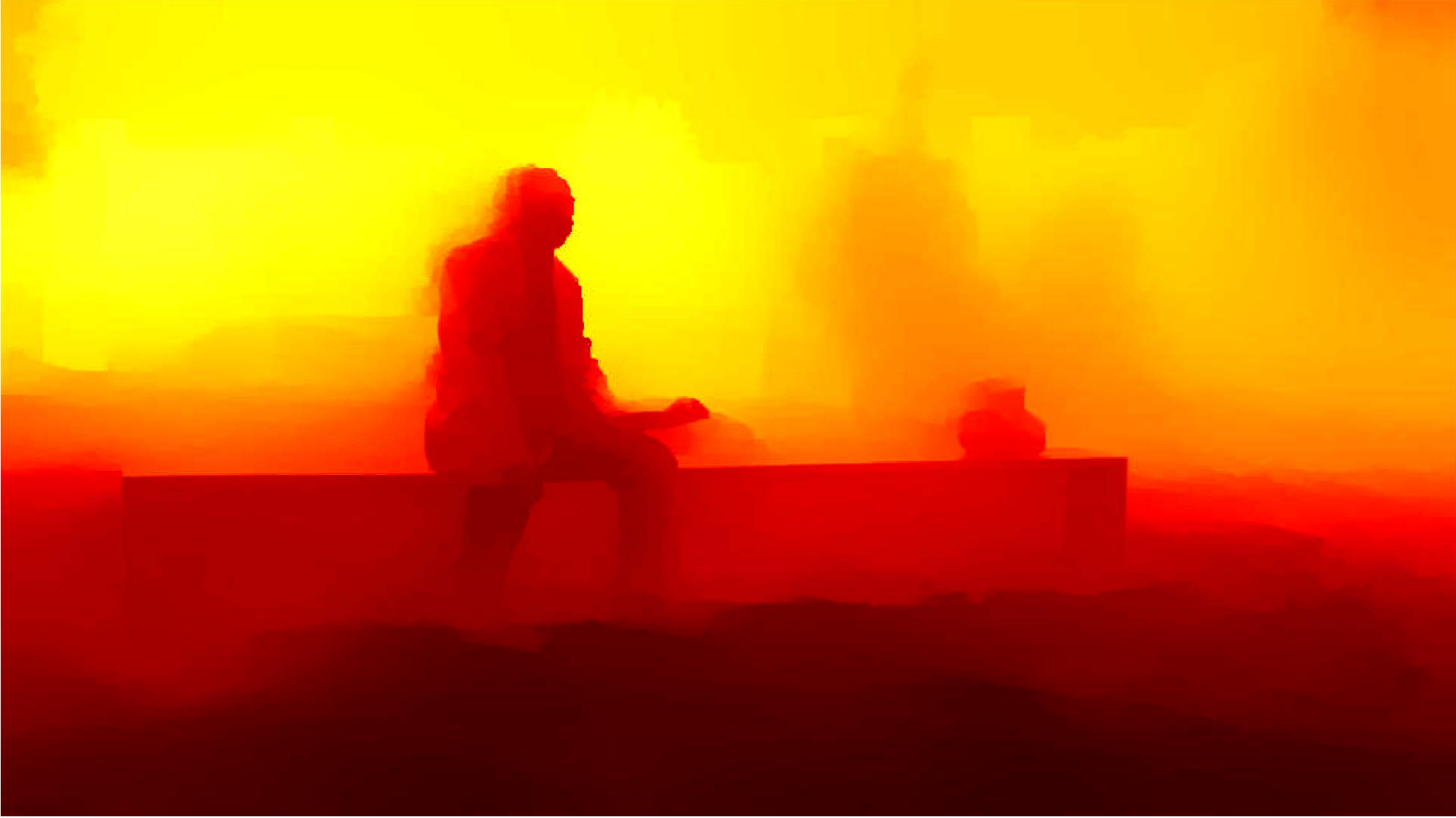}&
		\includegraphics[width=0.122\textwidth]{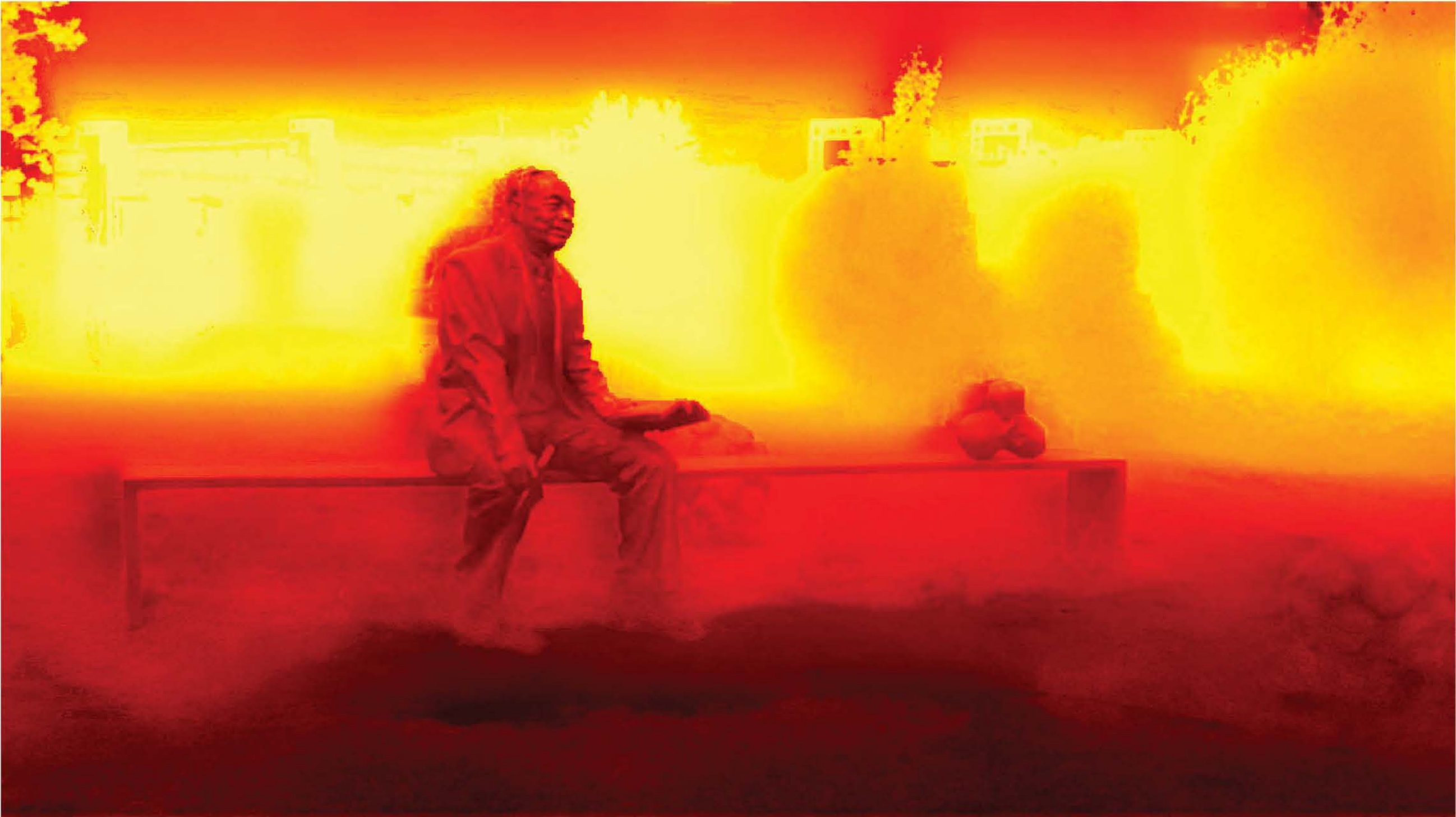}\\
		\footnotesize Ground Truth &~\cite{Cai2016DehazeNet} & \footnotesize~\cite{Meng2014Efficient}& \footnotesize~\cite{Ren2016Single}  & \footnotesize\cite{He2011Single}  & \footnotesize~\cite{Berman2016Non}& \footnotesize Ours
	\end{tabular}
	\caption{The transmission estimation results on example hazy images in Fattal's synthetic benchmark~\cite{Fattal2014}, which are corresponding to the results in Fig.~\ref{fig:sydehazvisres}.}
	\label{fig:transmission-sydehazvisres}
\end{figure*}
\subsubsection{Image Interpolation}
For the task of image interpolation (a.k.a. inpainting), we generated two types of corruptions, i.e., random masks with 20\%, 40\%, 60\% and 80\% missing pixels and text masks with either English or Chinese characters on the CBSD68 dataset~\cite{Zhang2016Beyond}, which contains 68 images with the size of 481 $\times$ 321.
We compared DPE with some state-of-the-art methods, including TV~\cite{Getreuer2012Total}, FoE~\cite{Roth2009Fields}, VNL~\cite{Arias2011A} and ISDSB~\cite{He2014Iterative} on this task.
Table~\ref{tab:inpaintpsnr} shows the quantitative results on image interpolation task. It is obvious that our method performs pretty well in terms of both PSNR and SSIM on different rates of missing pixels and text masks. Fig.~\ref{fig:inpaintres} then compared the visual results of these approaches.
The top row of Fig.~\ref{fig:inpaintres} illustrated the results of image with 80\% missing pixels. It is not hard to see from the zoomed in comparisons that the edge of the object can be successfully preserved in our image propagation. The bottom row of Fig.~\ref{fig:inpaintres} showed the results of text removal. We found that existing approaches either failed to remove the bold English characters or over smooth image details. In contrast, DPE achieved the best quantitative and qualitative performances.

\subsubsection{Super-Resolution}
The task of super-resolution is another important image enhancement task and has received much attention in the past few years. In this experiment, we compare DPE with several state-of-the-art methods including two conventional approaches (i.e., A$+$~\cite{Timofte2014A}, TNRD~\cite{chen2017trainable}) and three deep networks (i.e., IRCNN~\cite{Zhang2017Learning}, SRCNN~\cite{Dong2016Image}, VDSR~\cite{Kim2016Accurate}). For quantitative comparisons, we reported PSNR and SSIM on Set14 benchmark~\cite{Kim2016Accurate} in Table~\ref{tab:srquanres}. We observed that the PSNR score of VDSR is a little bit higher than ours. This is mainly because its network is particularly designed for super-resolution task. Moreover, they first collect training datasets for several specified scales and then combine
them into one big dataset for network training. Fortunately,
it can be seen that DPE achieved higher SSIM score, which is more convincing to measure the image structure information. We also plotted super-resolution results of an example image from Urban100 dataset~\cite{Huang2015Single} in Fig.~\ref{fig:srvisres}. It is easy to see that our method can generate clearer texture than other state-of-the-art methods.
\begin{figure*}[htb]
	\centering
	\begin{tabular}{c@{\extracolsep{0.1em}}c@{\extracolsep{0.1em}}c@{\extracolsep{0.1em}}c@{\extracolsep{0.1em}}c@{\extracolsep{0.1em}}c@{\extracolsep{0.1em}}c@{\extracolsep{0.1em}}c}		
		\includegraphics[width=0.122\textwidth]{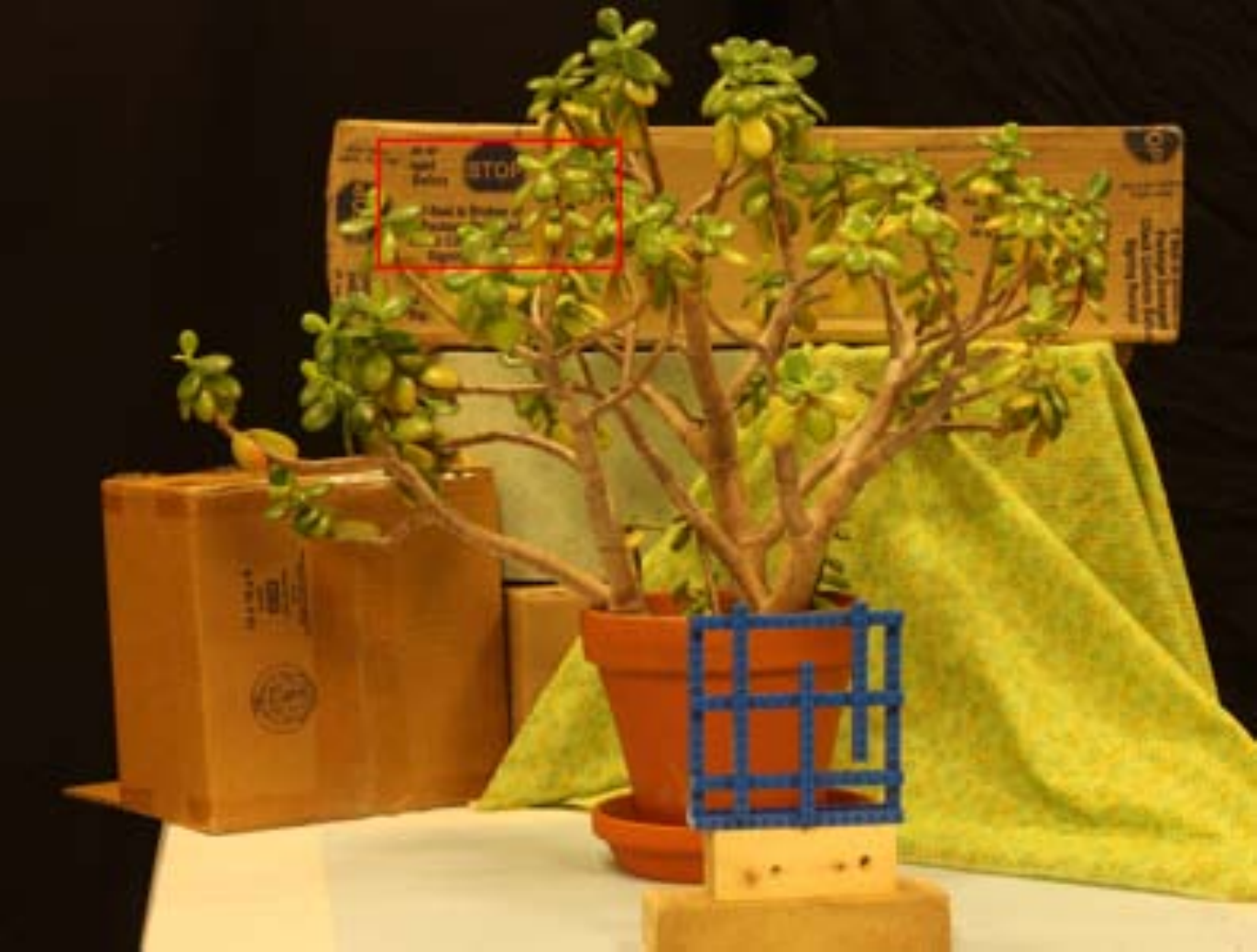}&
		\includegraphics[width=0.122\textwidth]{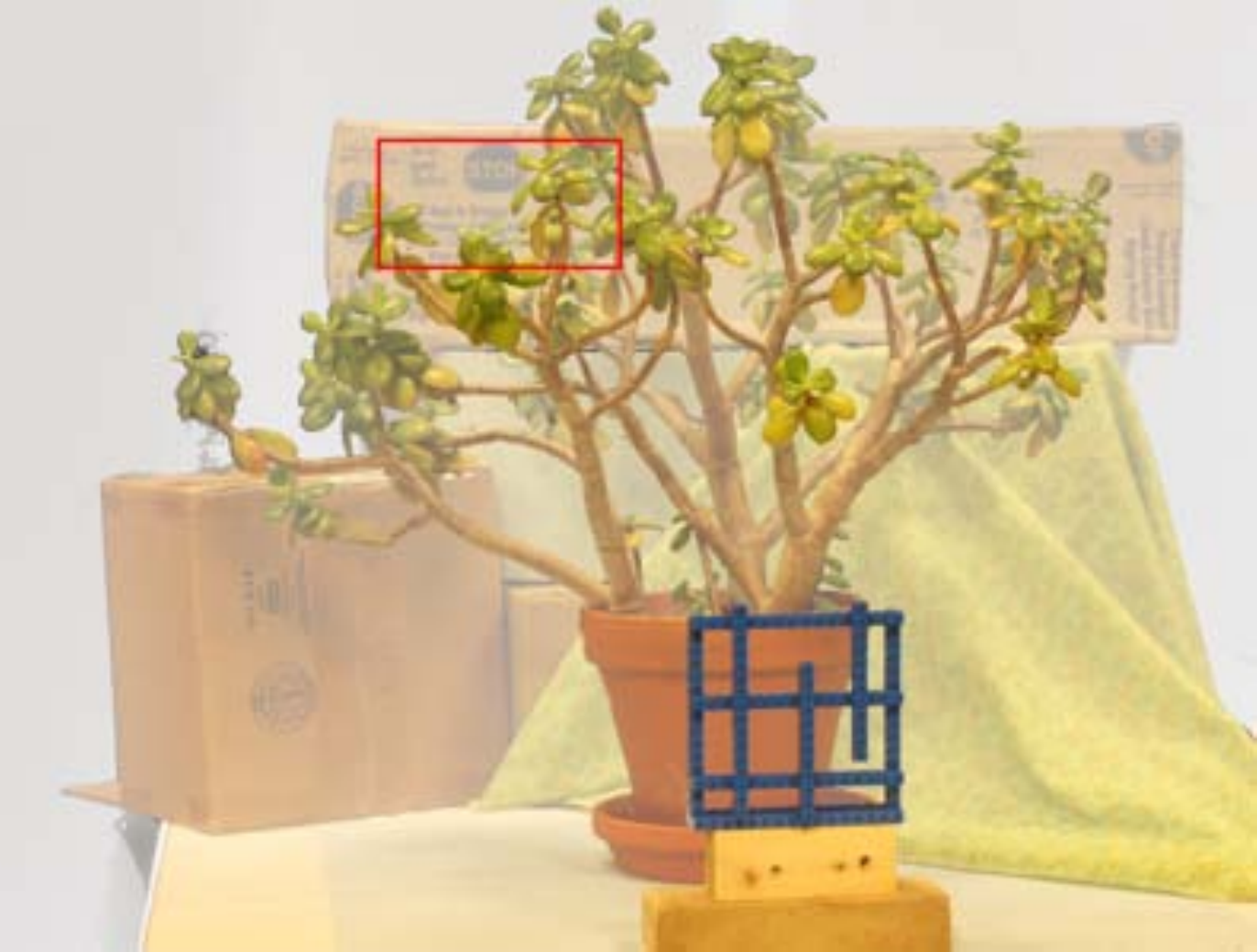}&
		\includegraphics[width=0.122\textwidth]{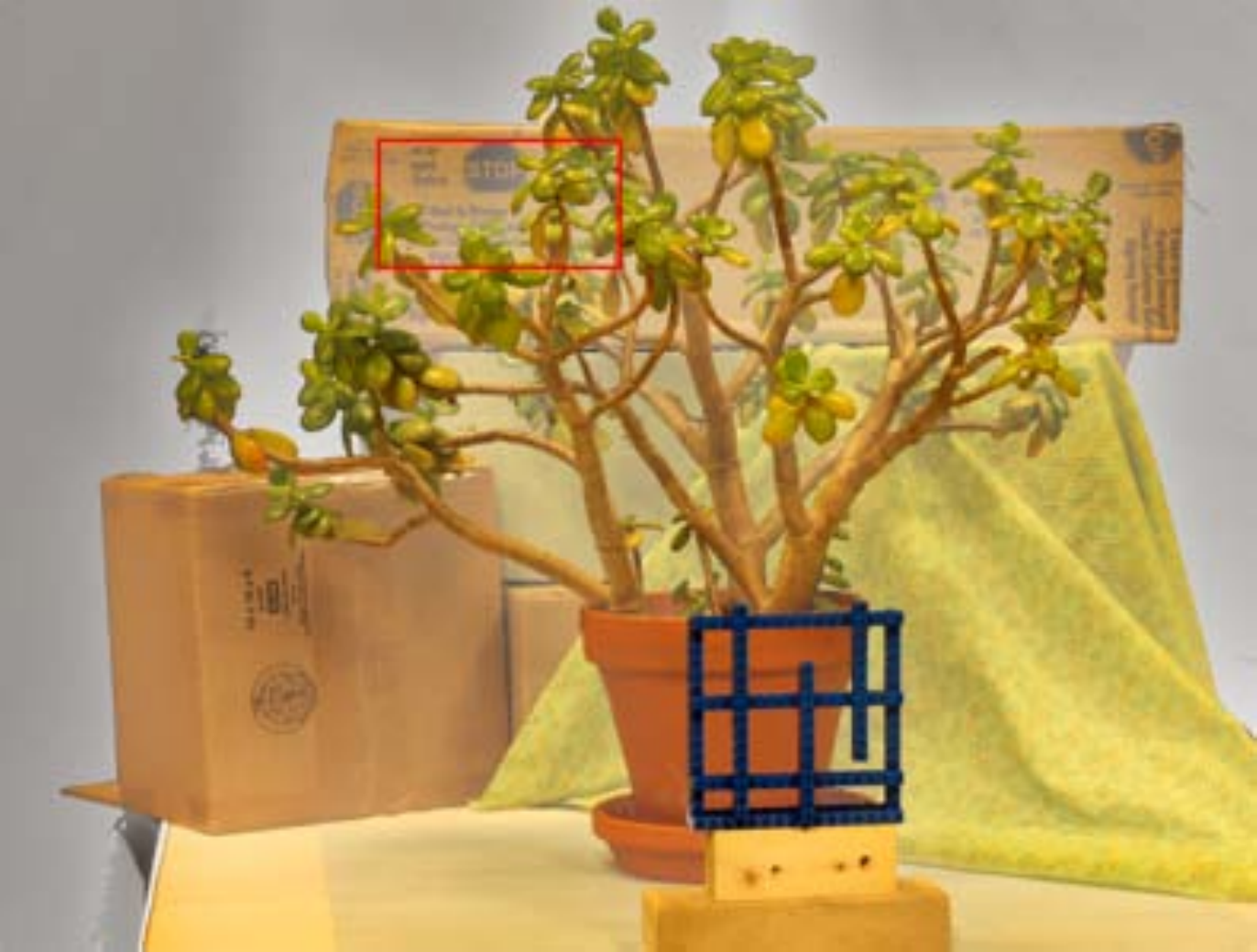}&
		\includegraphics[width=0.122\textwidth]{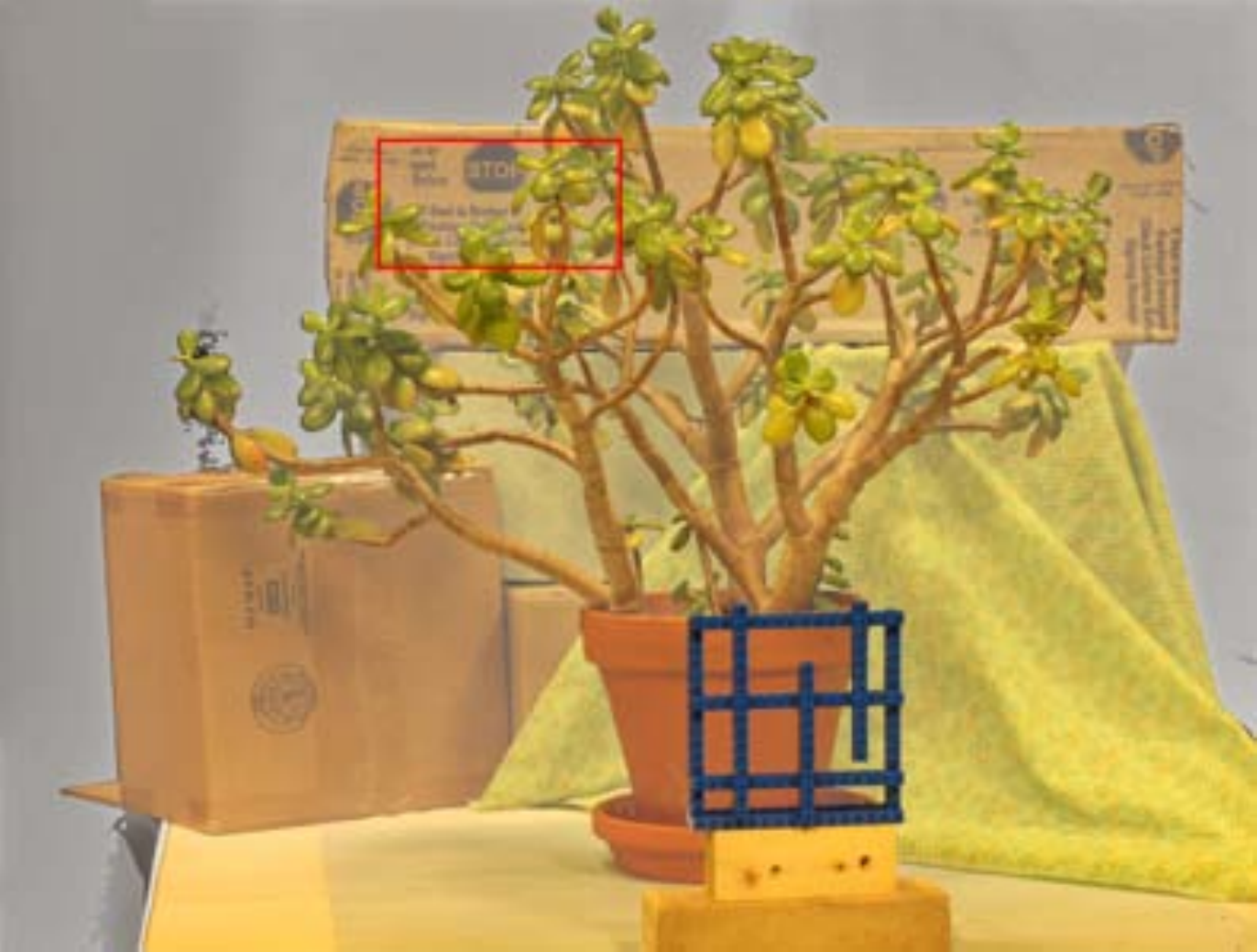}&
		\includegraphics[width=0.122\textwidth]{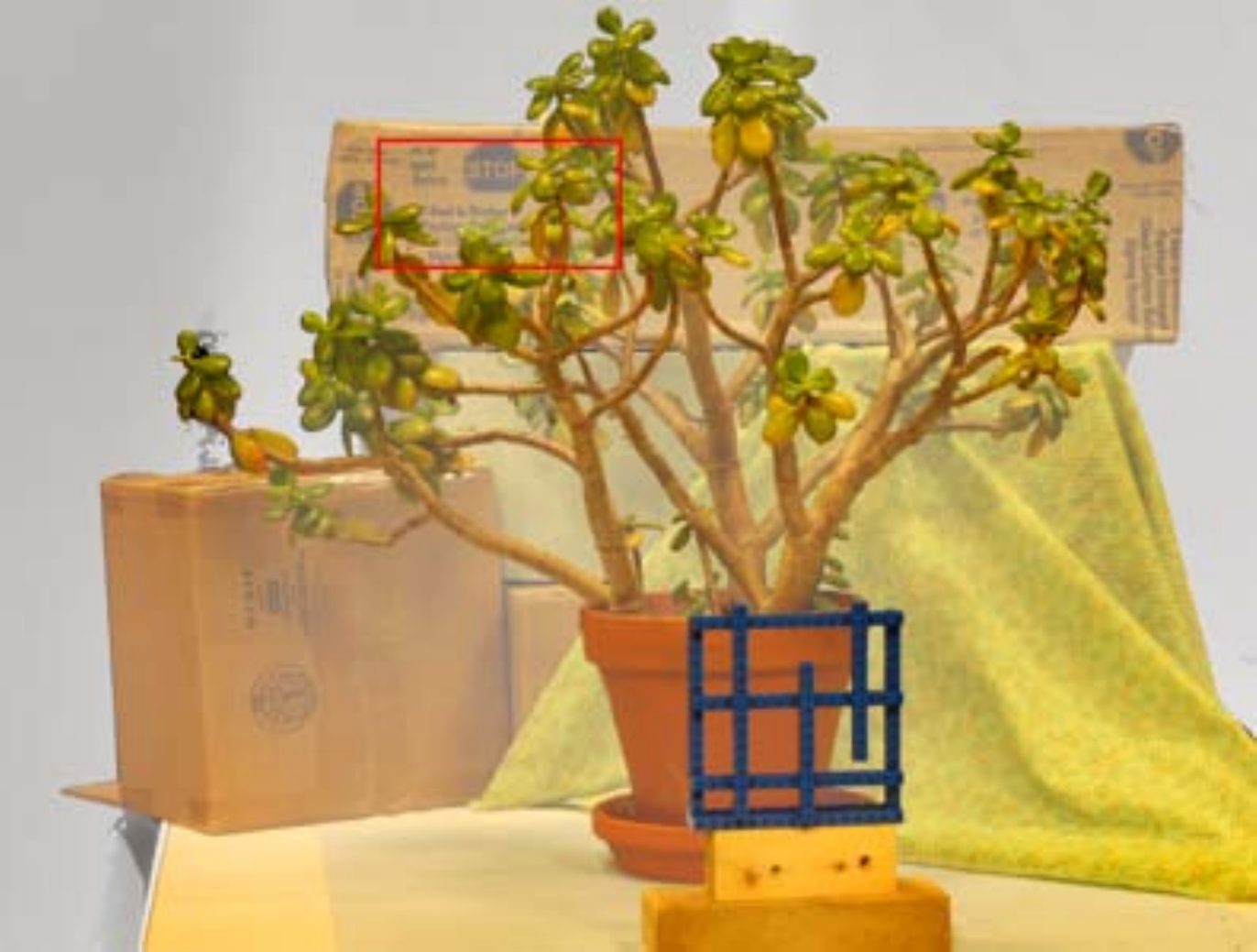}&
		\includegraphics[width=0.122\textwidth]{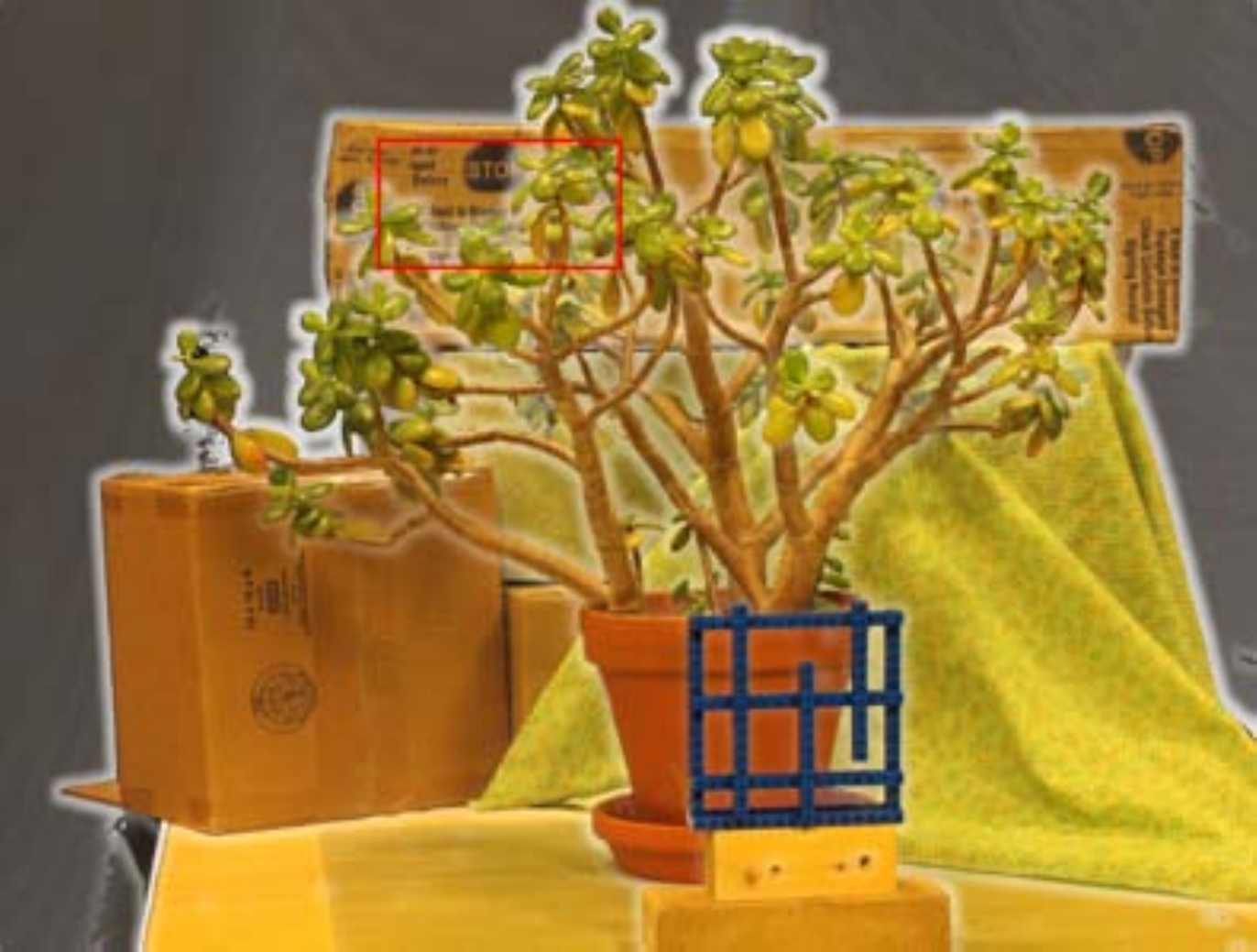}&
		\includegraphics[width=0.122\textwidth]{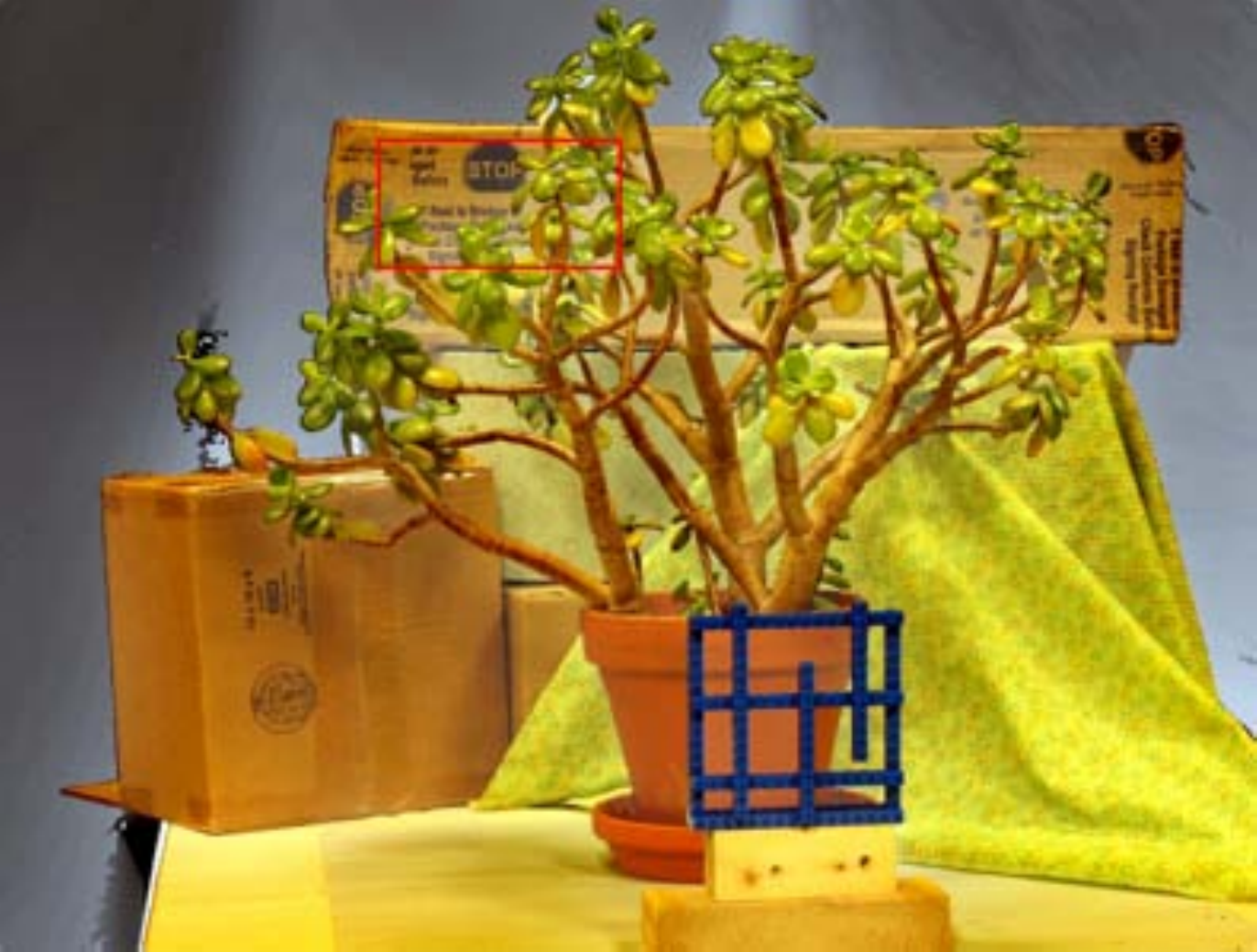}&
		\includegraphics[width=0.122\textwidth]{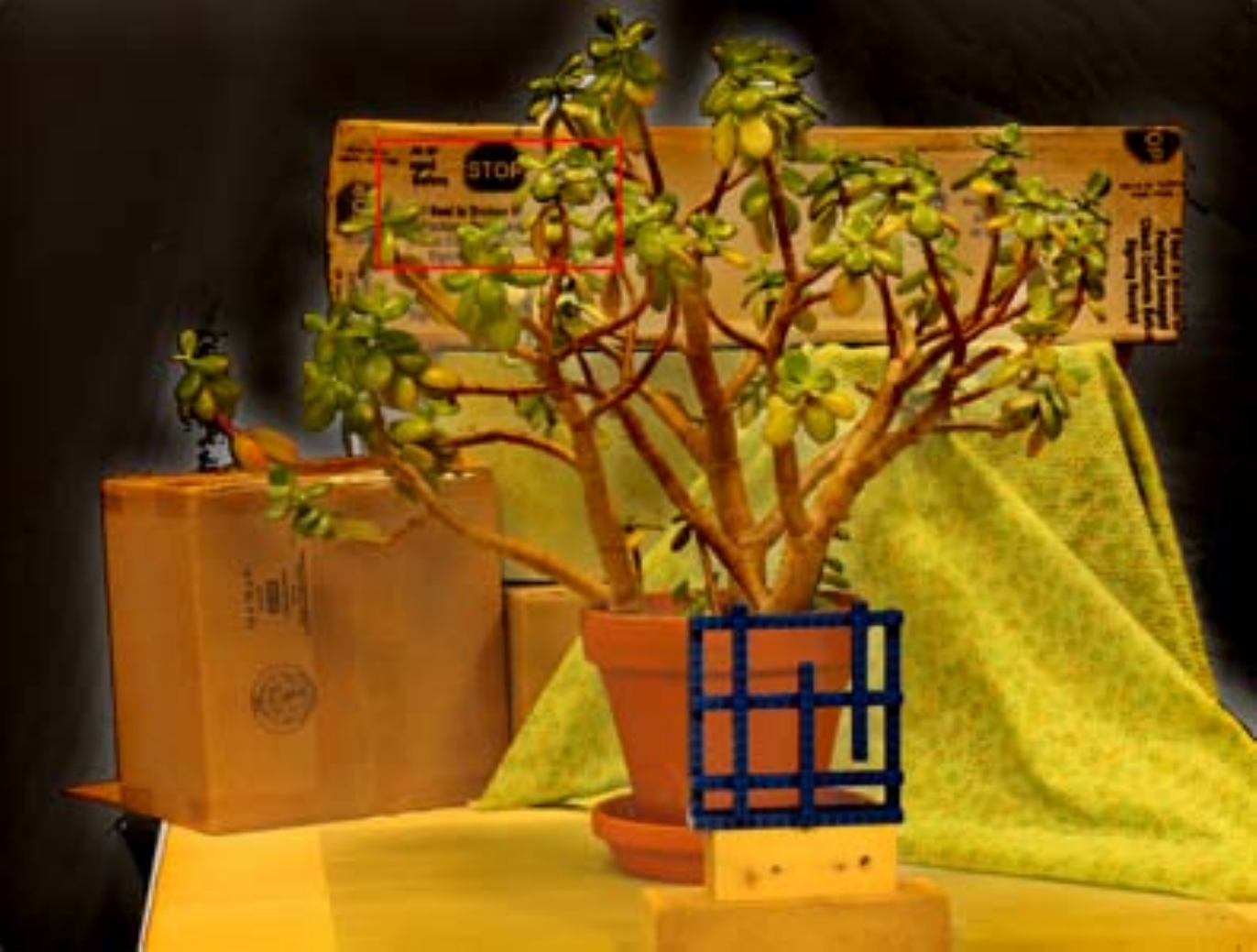}\\
		--& \footnotesize 0.6283 / 0.3830
		&\footnotesize 0.6613 / 0.2466 & \footnotesize 0.6801 / 0.2479&\footnotesize 0.6510 / 0.1249& \footnotesize 0.7196 / 0.1330 & \footnotesize 0.7234 / 0.1501 & \footnotesize \textbf{0.8443 / 0.0622}\\
		\includegraphics[width=0.122\textwidth]{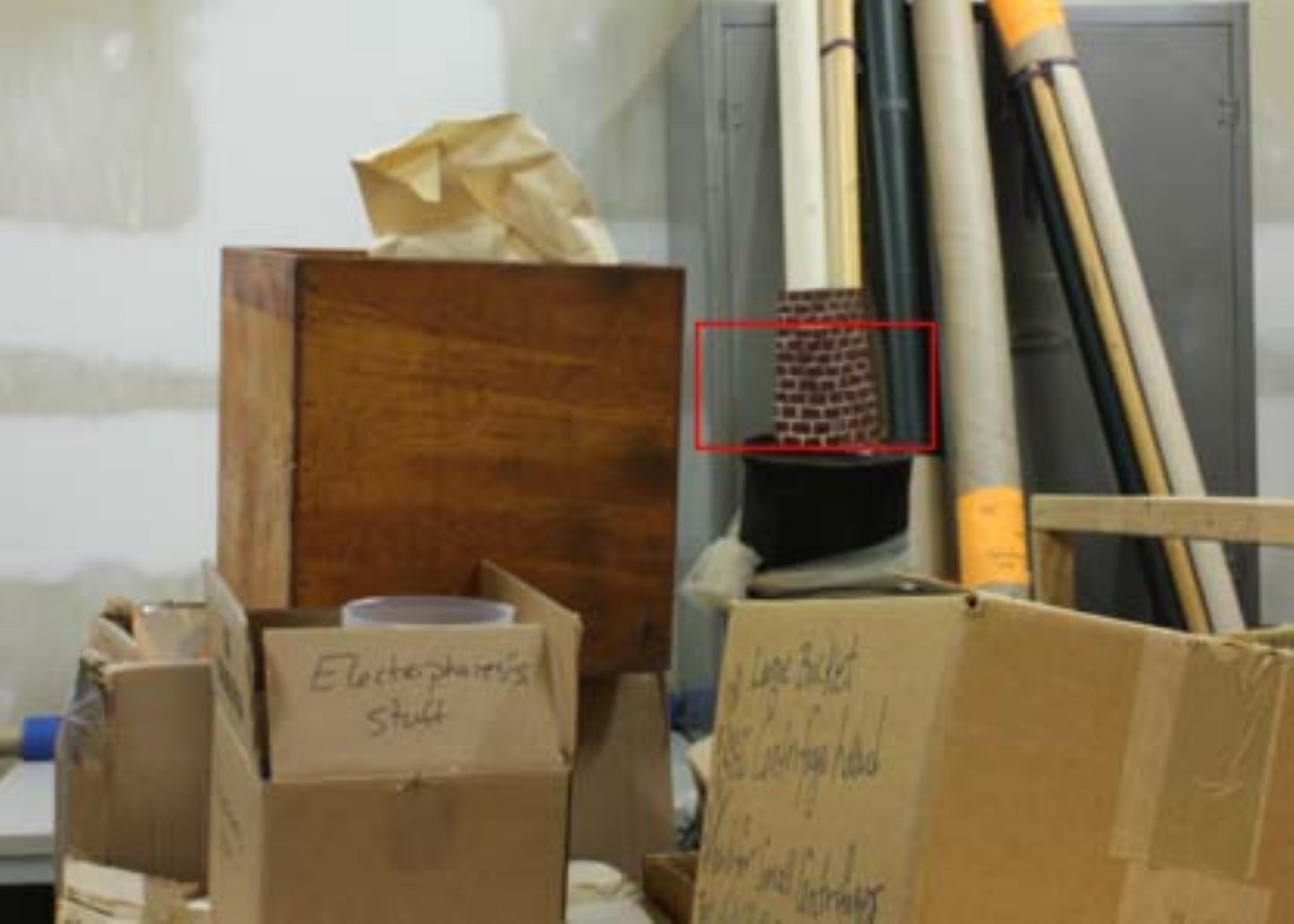}&
		\includegraphics[width=0.122\textwidth]{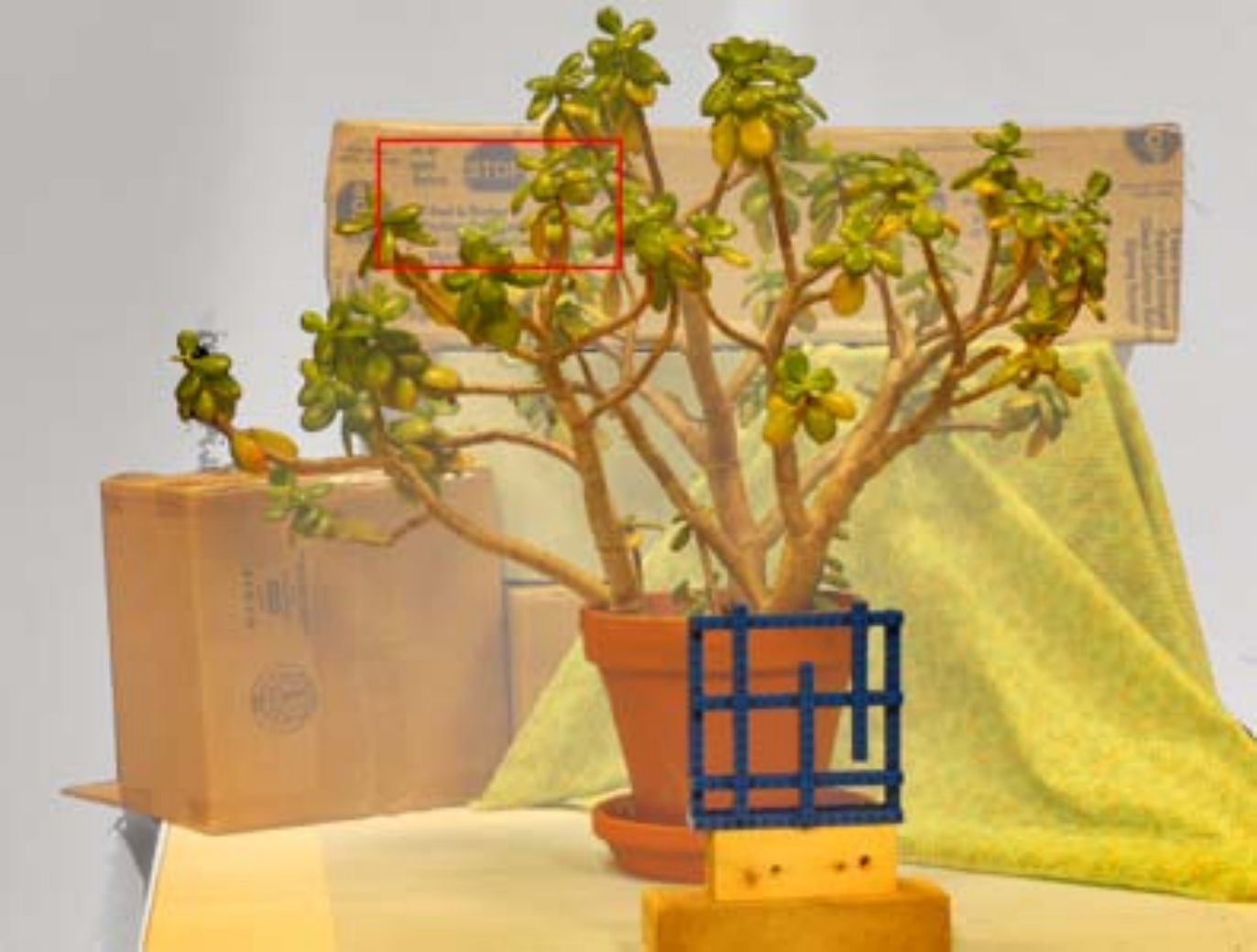}&
		\includegraphics[width=0.122\textwidth]{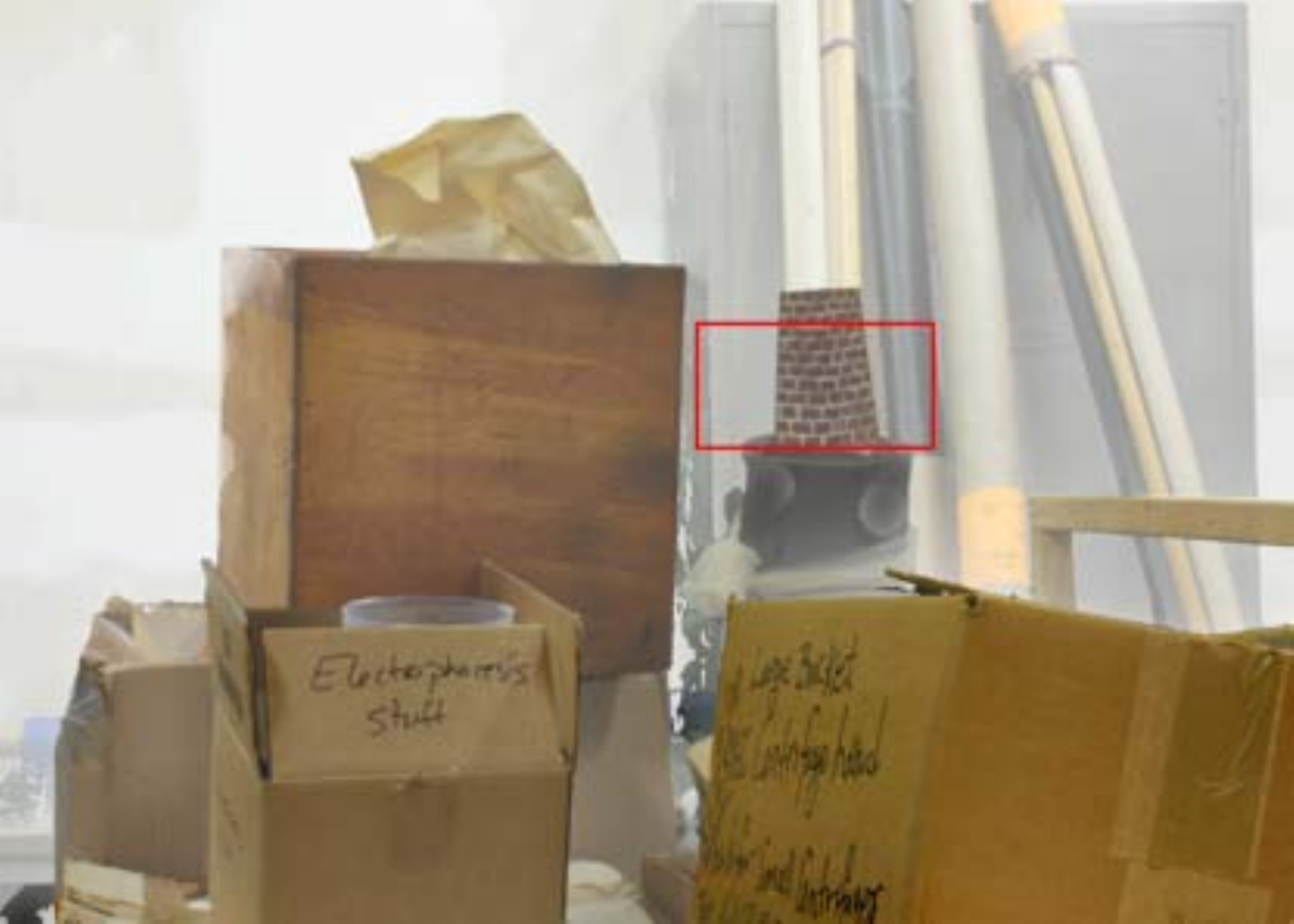}&
		\includegraphics[width=0.122\textwidth]{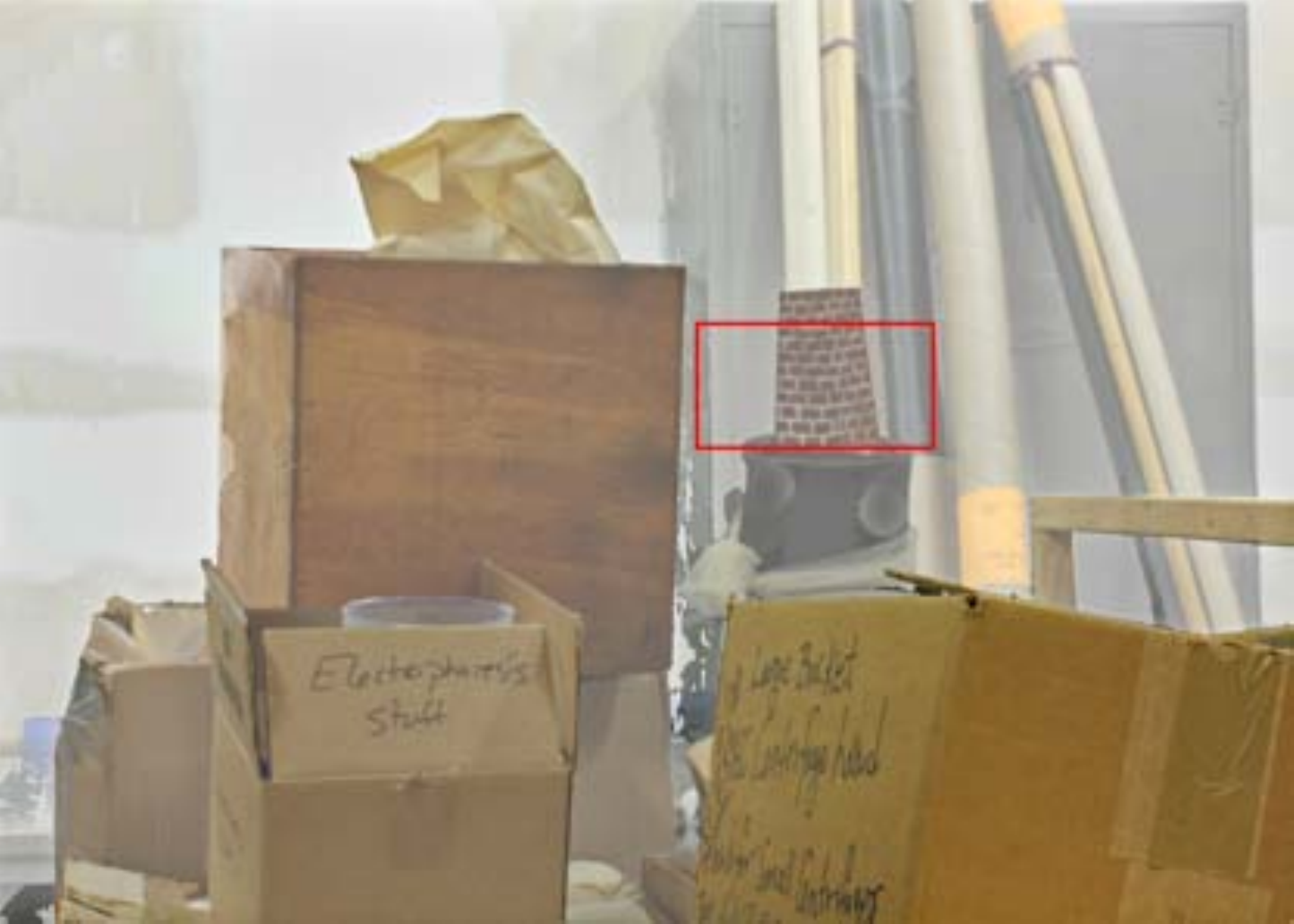}&
		\includegraphics[width=0.122\textwidth]{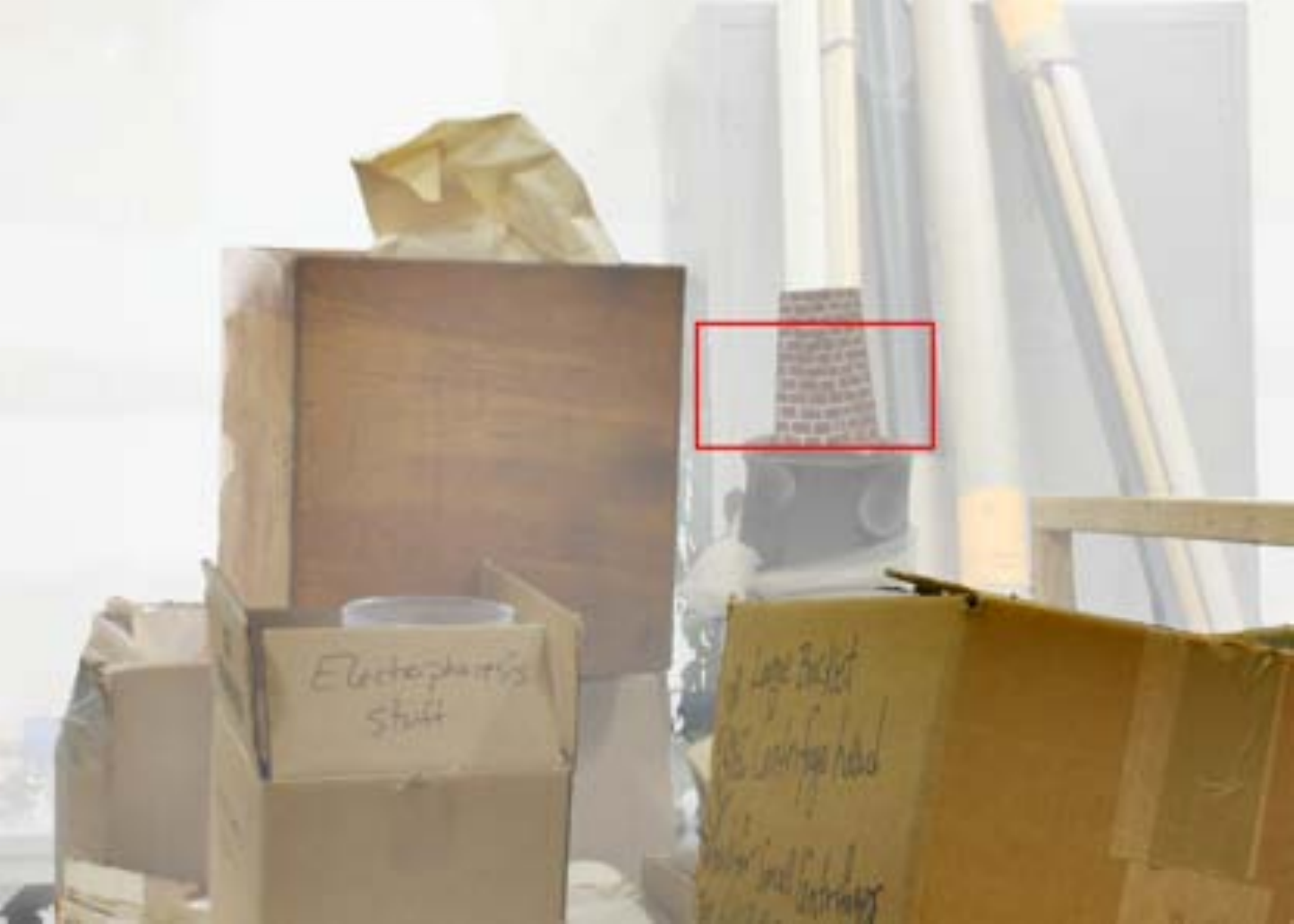}&
		\includegraphics[width=0.122\textwidth]{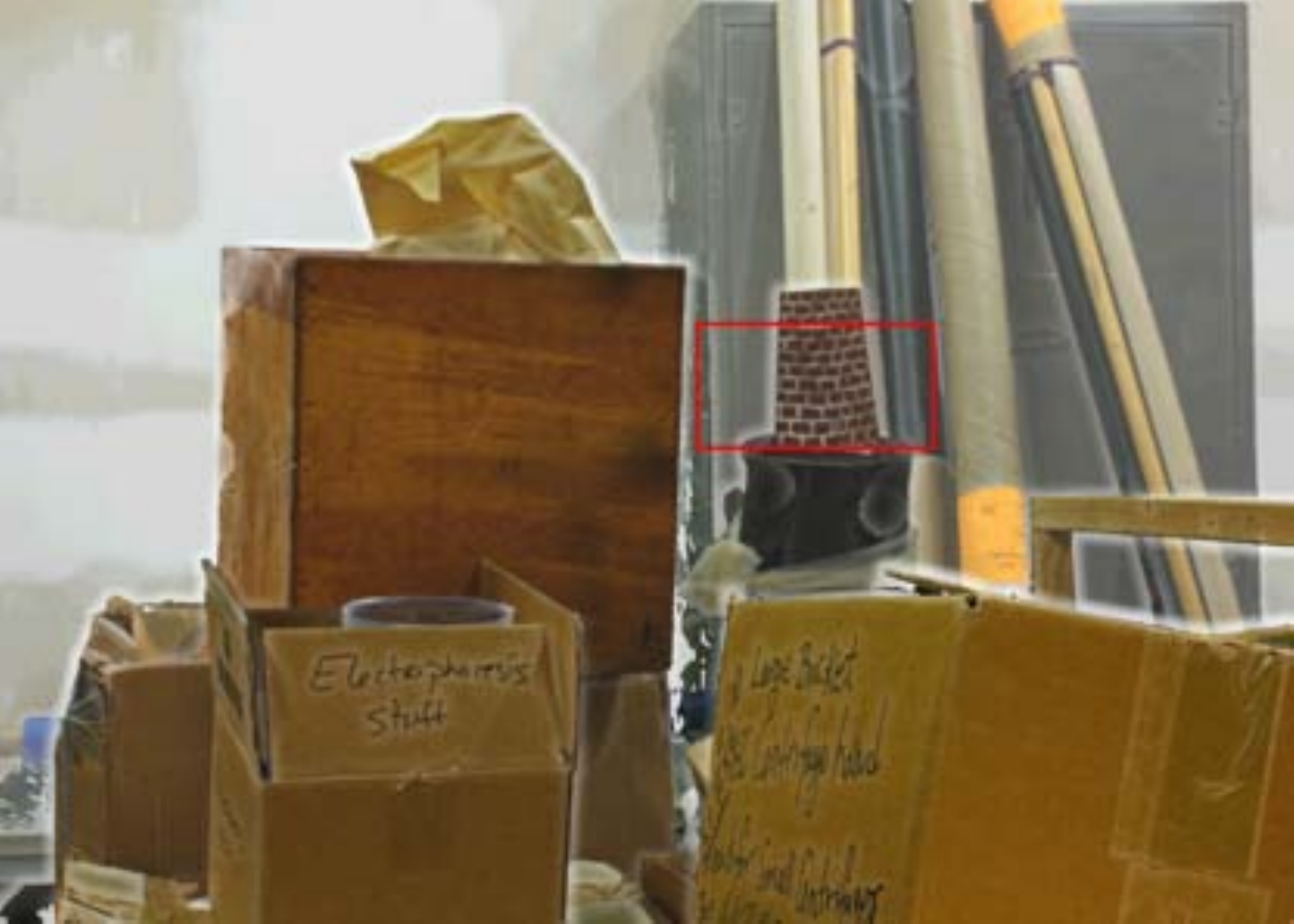}&
		\includegraphics[width=0.122\textwidth]{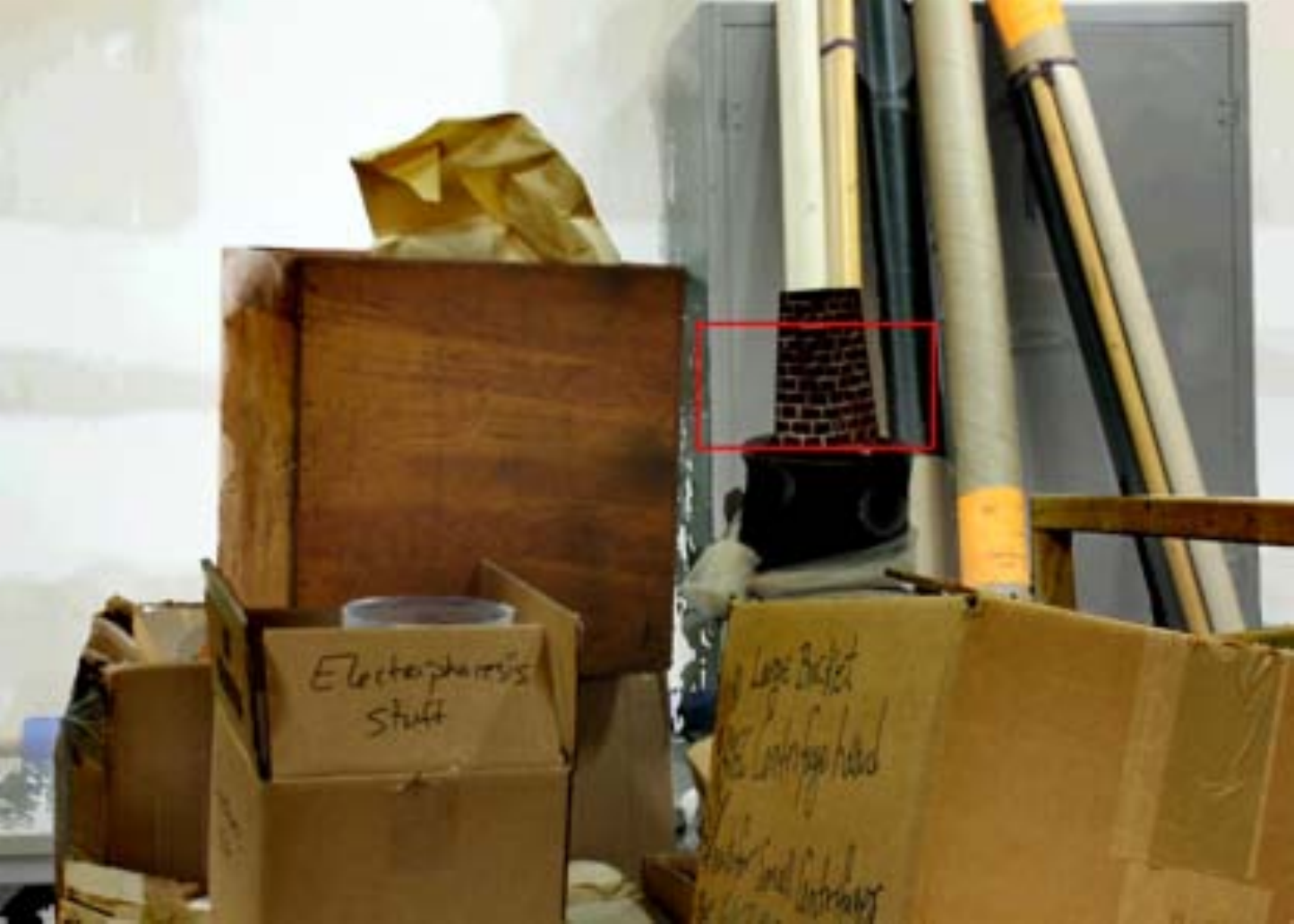}&
		\includegraphics[width=0.122\textwidth]{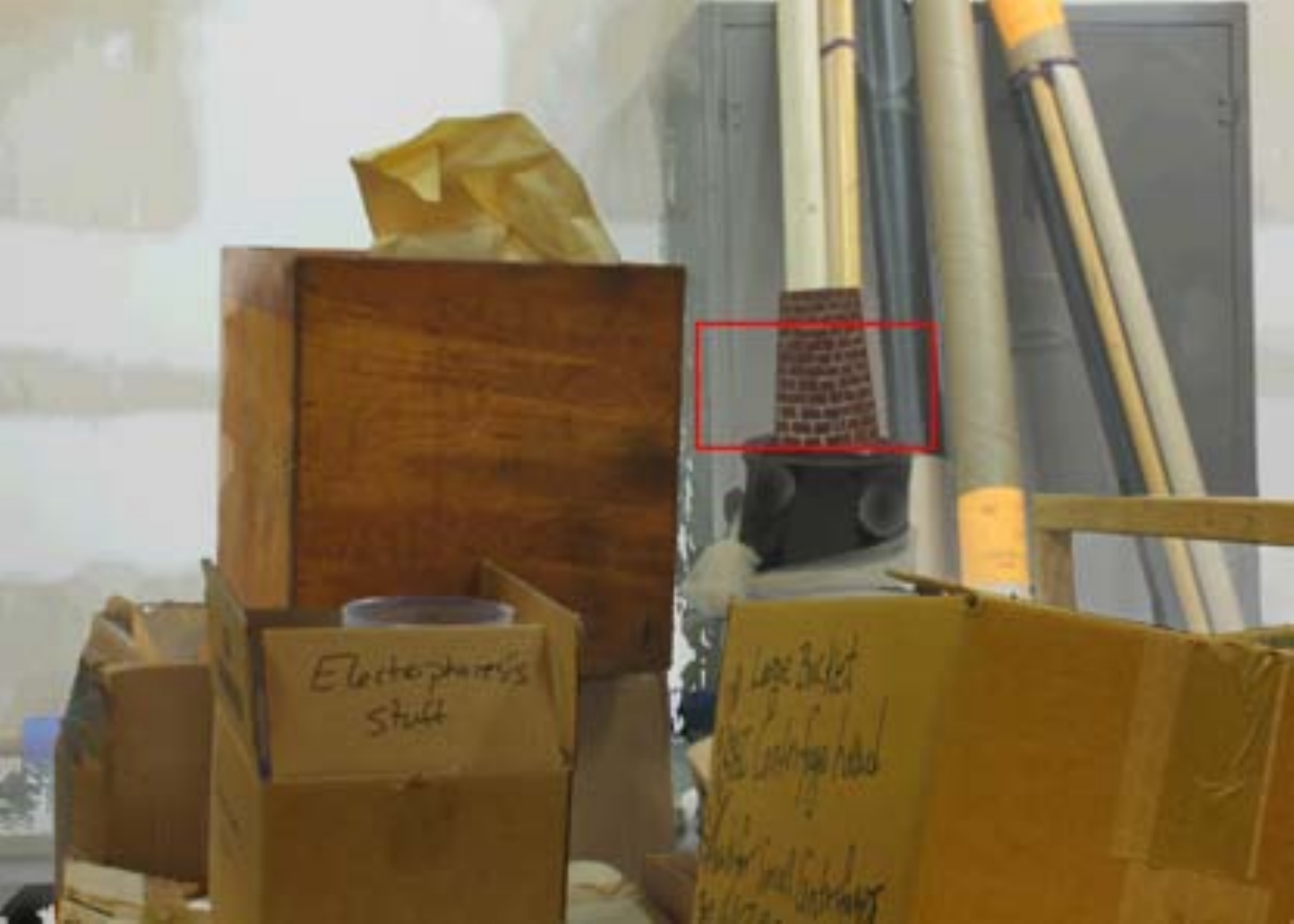}\\
		--& \footnotesize 0.7720  / 0.3260
		&	\footnotesize 0.8494 / 0.1756 & \footnotesize 0.8503 / 0.1605&\footnotesize 0.8014 / 0.1314& \footnotesize 0.8910 / 0.0868 & \footnotesize 0.8777 / 0.1013 & \footnotesize \textbf{0.9057 / 0.0801}\\
		\footnotesize Ground Truth &\footnotesize Hazy Input &\footnotesize~\cite{Cai2016DehazeNet} & \footnotesize~\cite{Meng2014Efficient} &\footnotesize~\cite{Ren2016Single}  & \footnotesize\cite{He2011Single}  & \footnotesize~\cite{Berman2016Non}& \footnotesize Ours
	\end{tabular}
	\caption{The enhancement performances on example hazy images in D-Hazy benchmark~\cite{AncutiDHazy}. The quantitative scores (i.e., SSIM / $L_1$ Error) are also reported accordantly.}
	\label{fig:sydehazvisres2}
\end{figure*}

\begin{figure*}[htb]
	\centering
	\begin{tabular}{c@{\extracolsep{0.1em}}c@{\extracolsep{0.1em}}c@{\extracolsep{0.1em}}c@{\extracolsep{0.1em}}c}
		\includegraphics[width=0.195\textwidth]{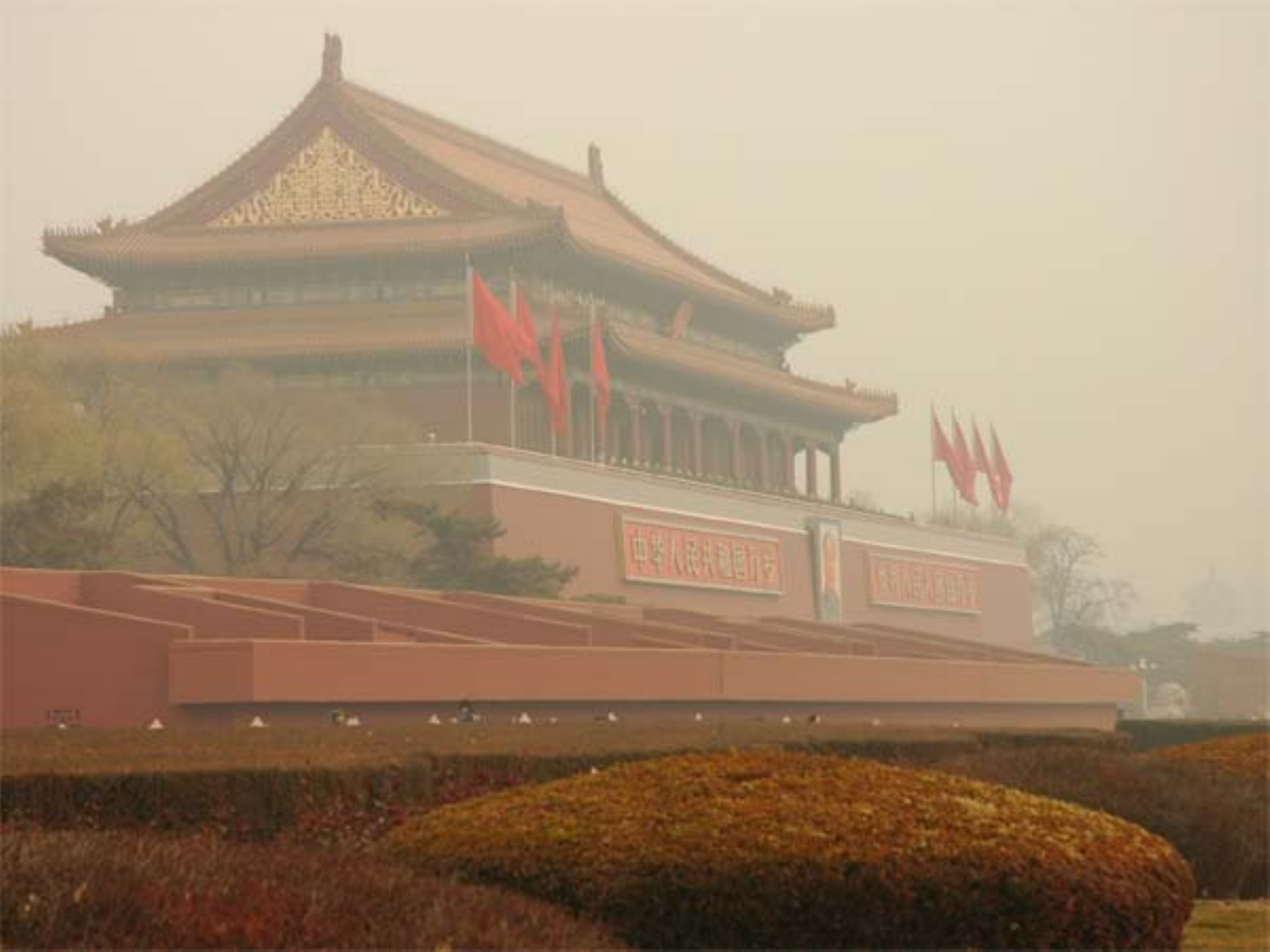}&
		\includegraphics[width=0.195\textwidth]{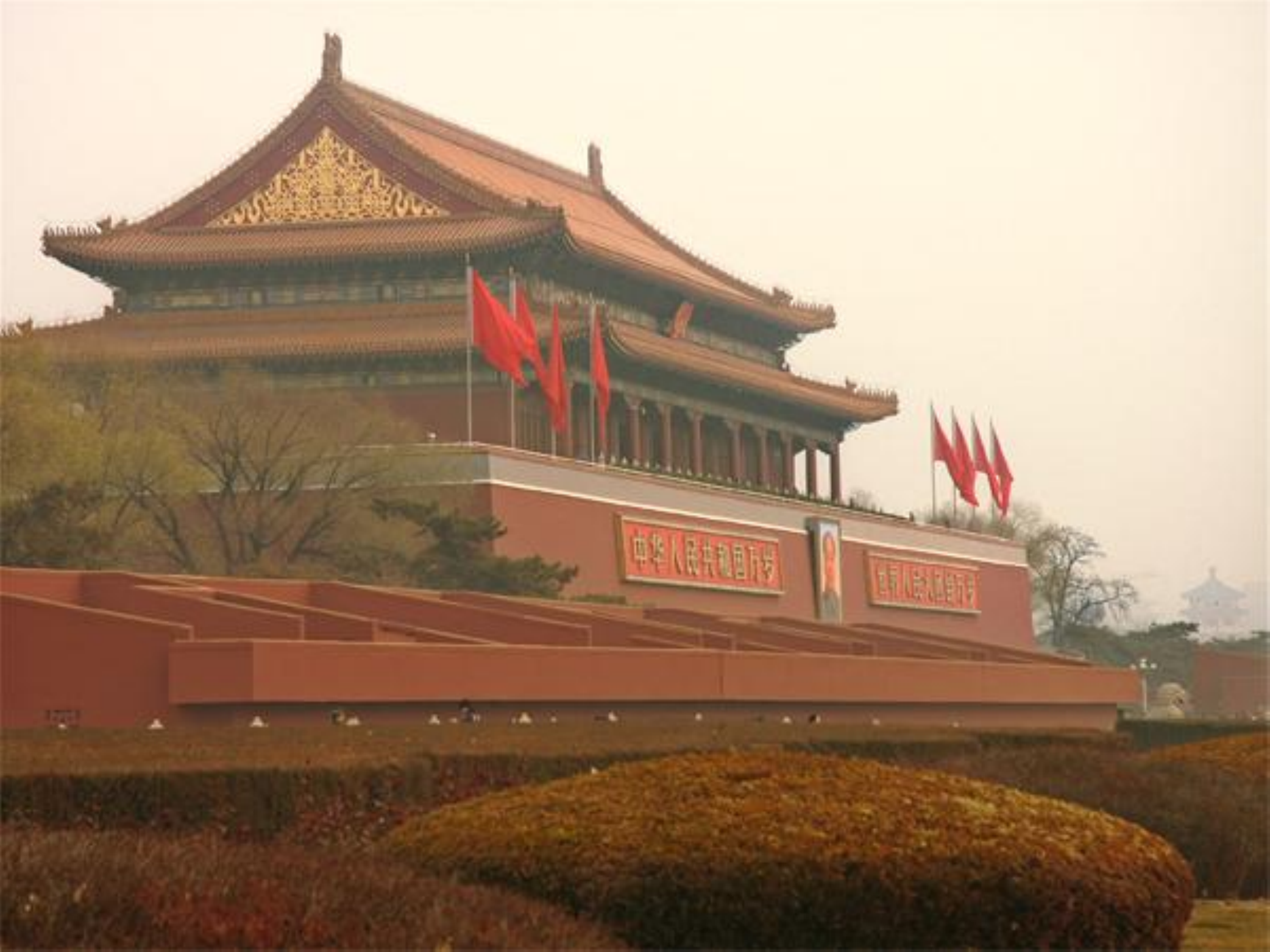}&
		\includegraphics[width=0.195\textwidth]{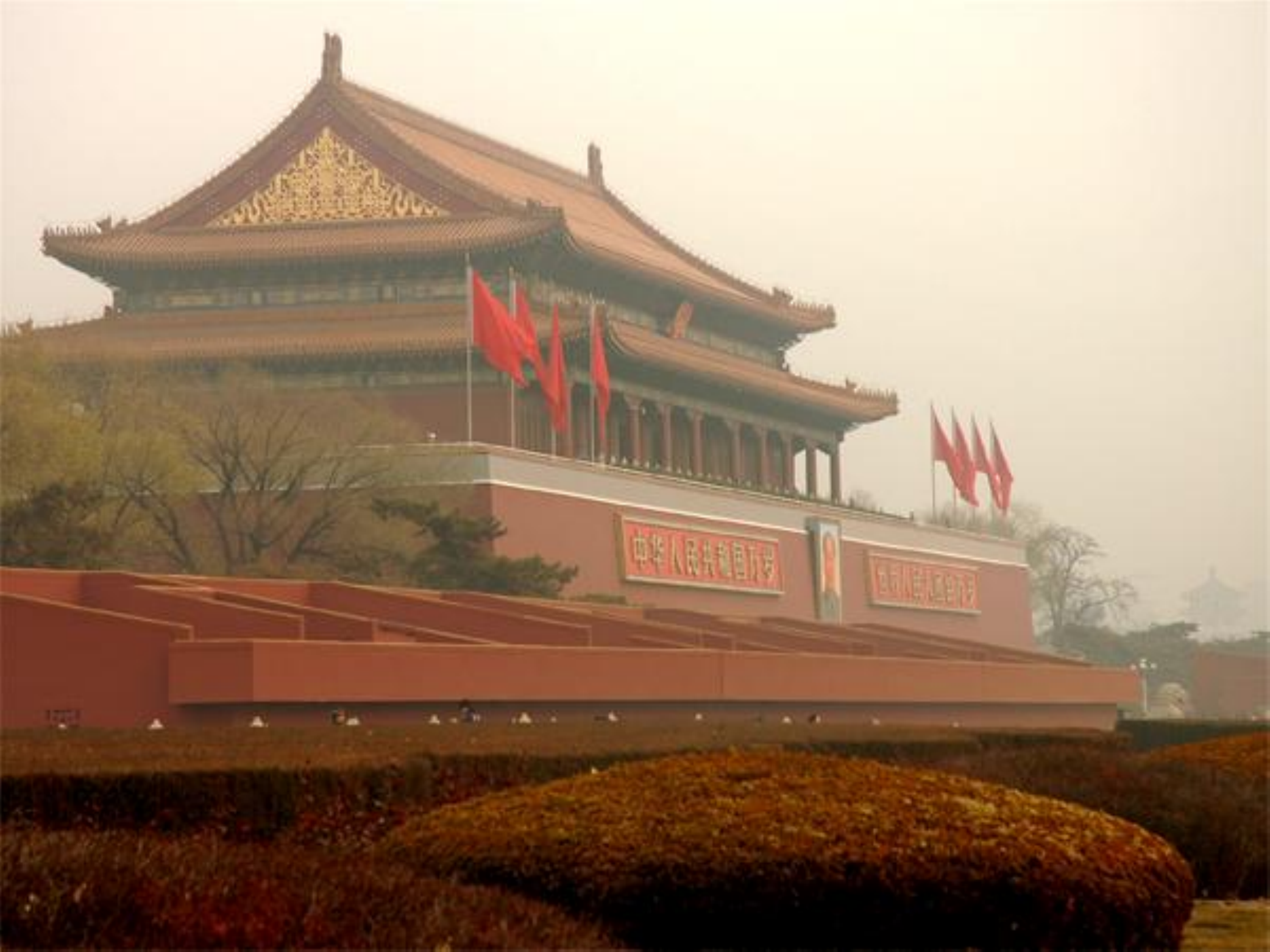}&
		\includegraphics[width=0.195\textwidth]{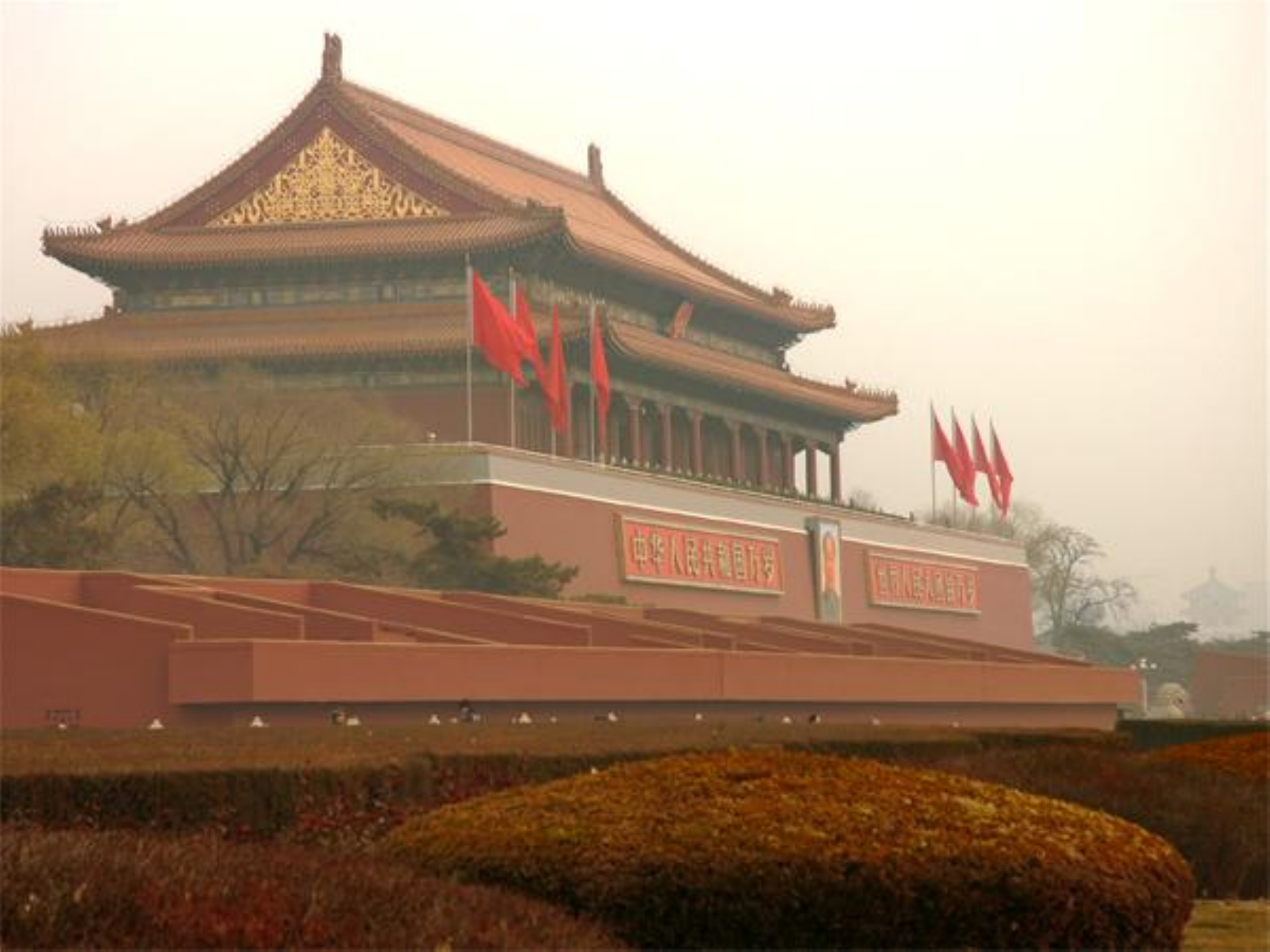}&
		\includegraphics[width=0.195\textwidth]{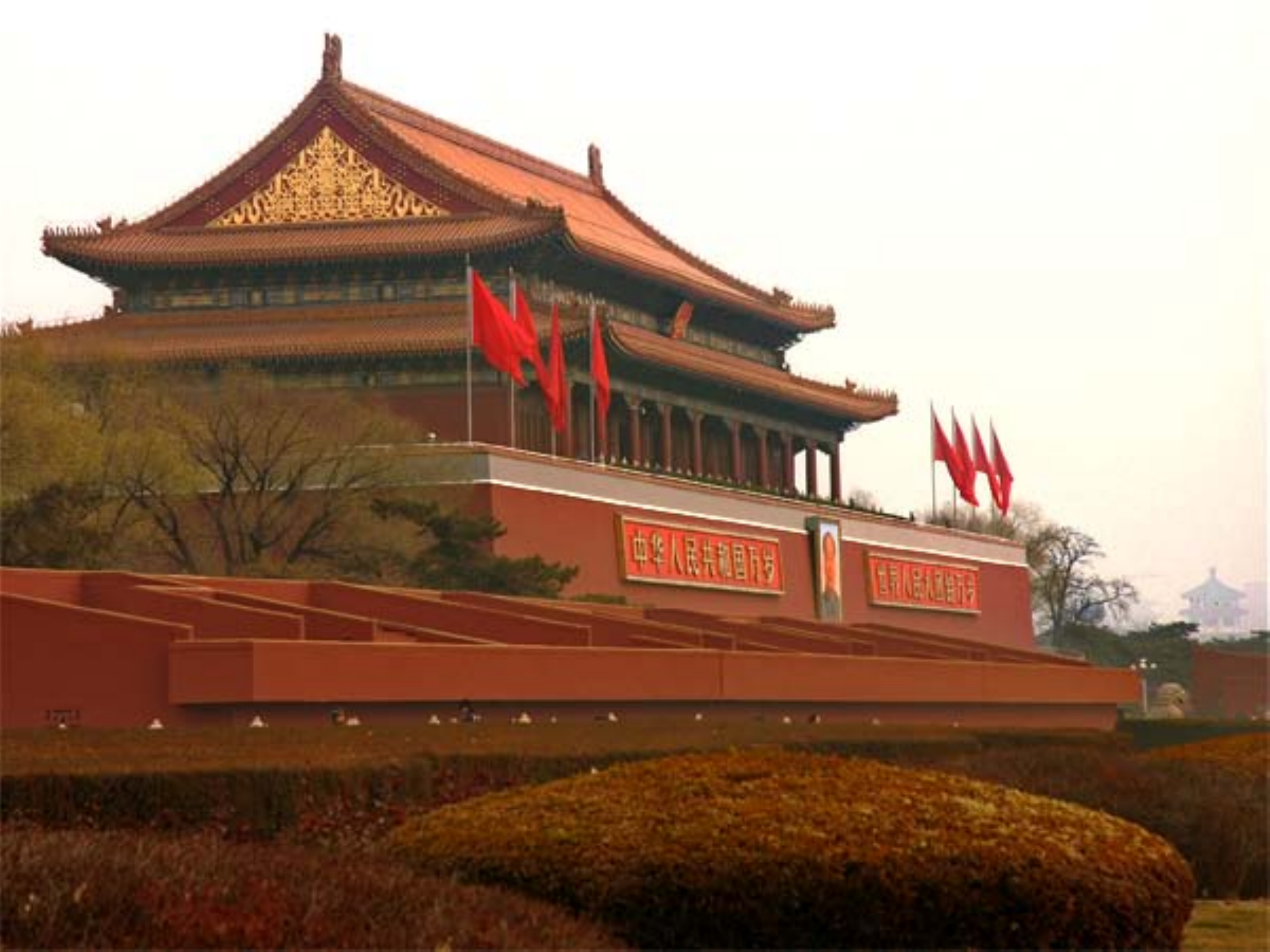}\\
		\includegraphics[width=0.195\textwidth]{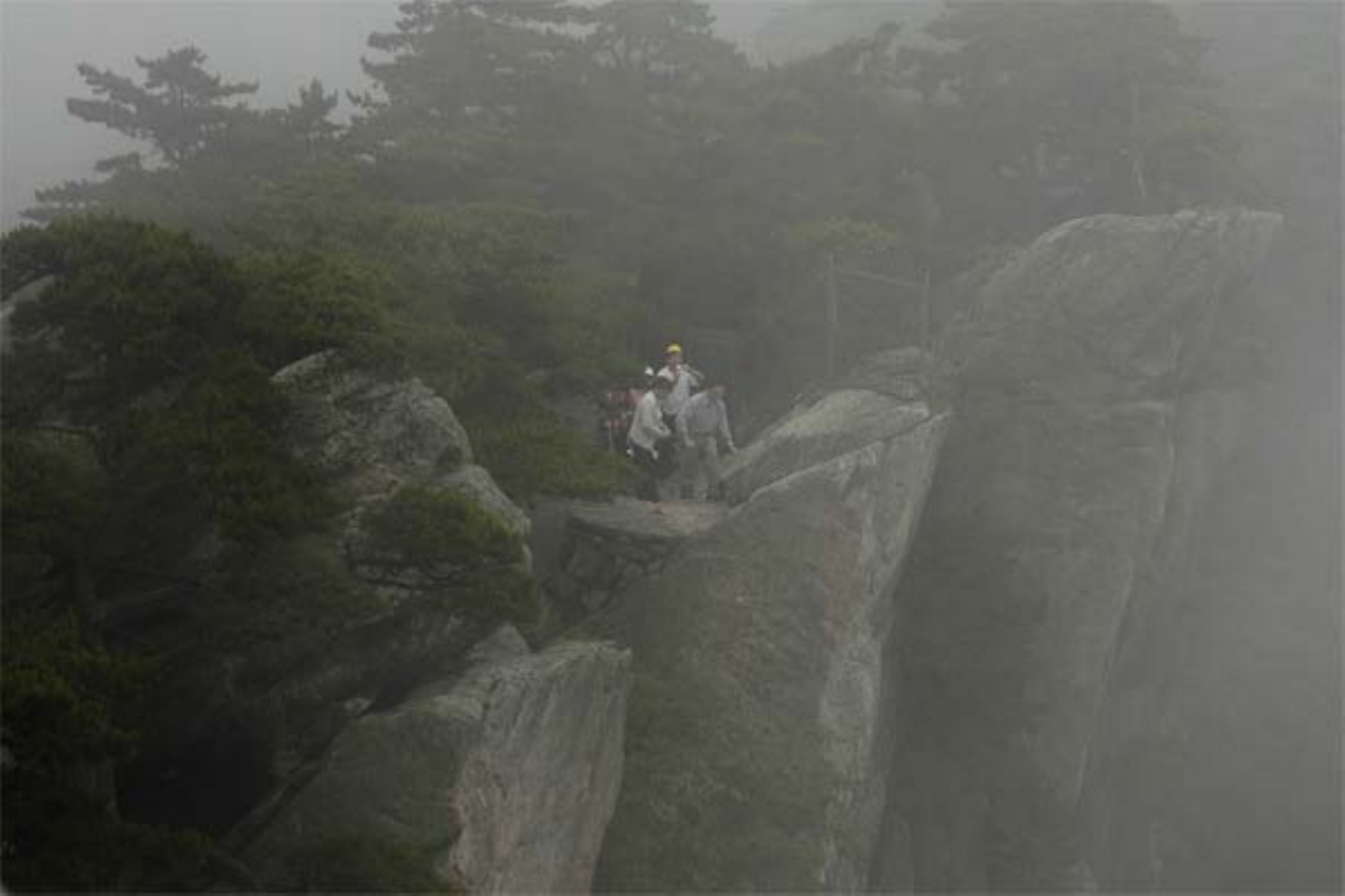}&
		\includegraphics[width=0.195\textwidth]{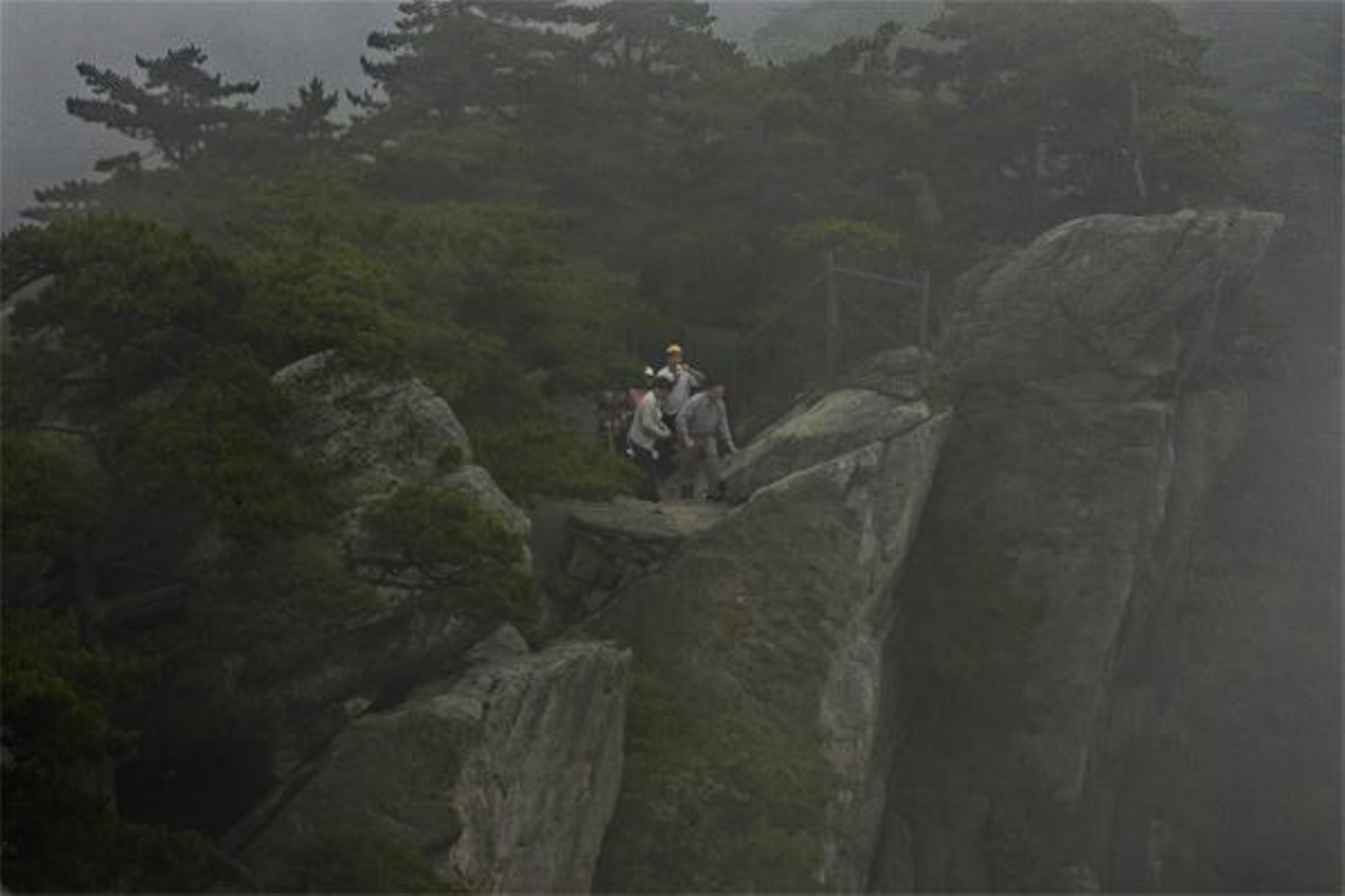}&
		\includegraphics[width=0.195\textwidth]{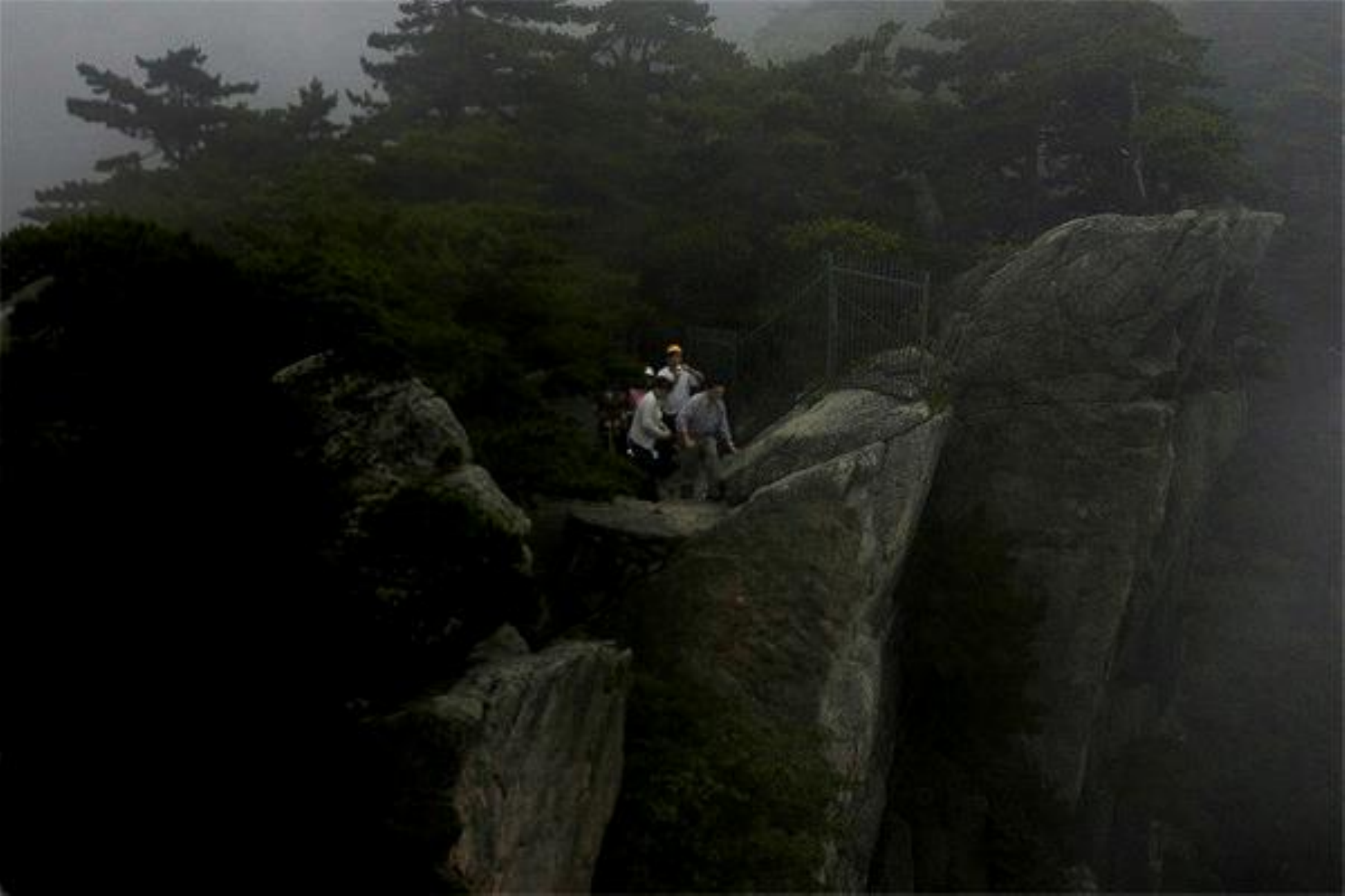}&
		\includegraphics[width=0.195\textwidth]{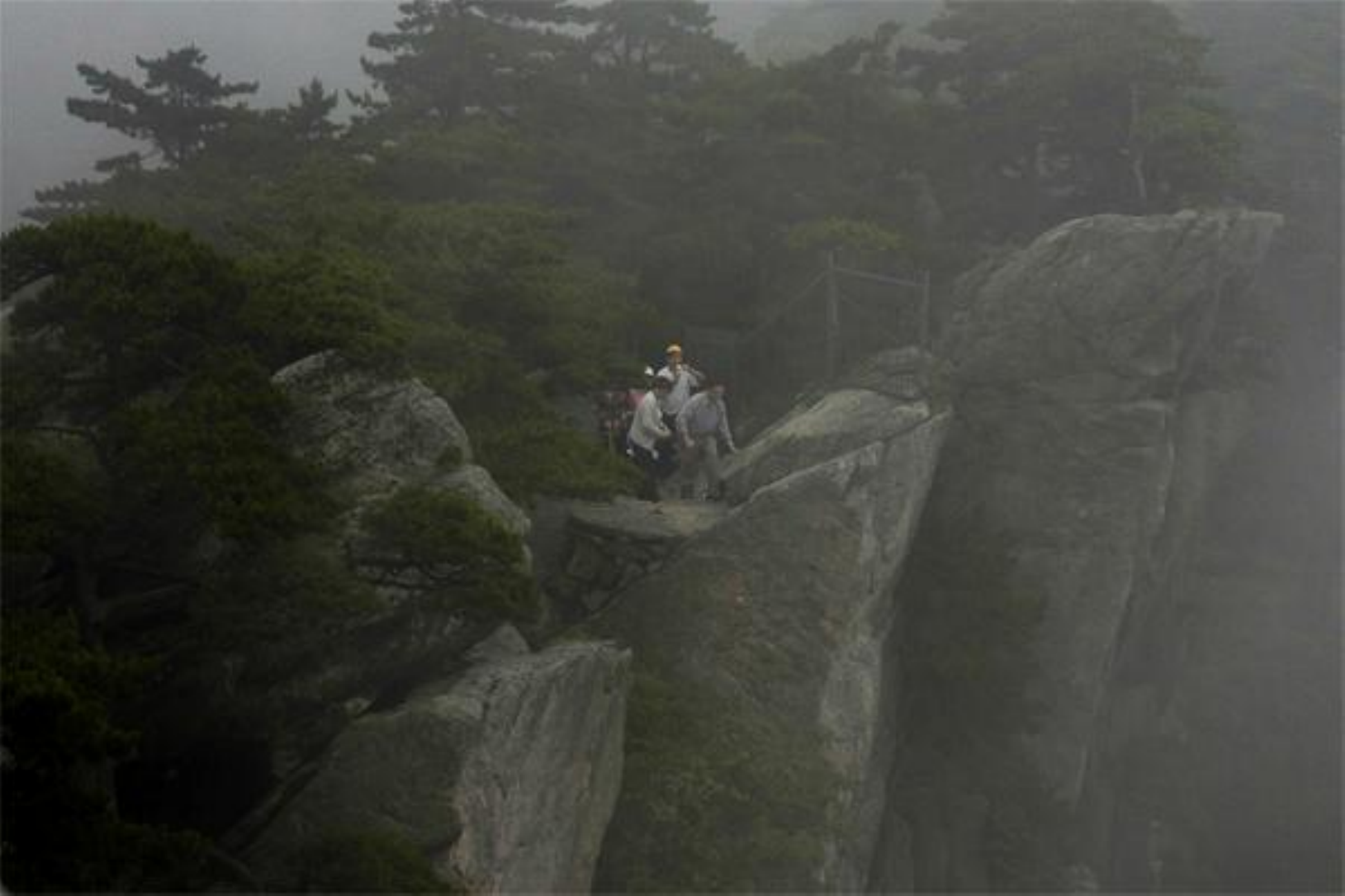}&
		\includegraphics[width=0.195\textwidth]{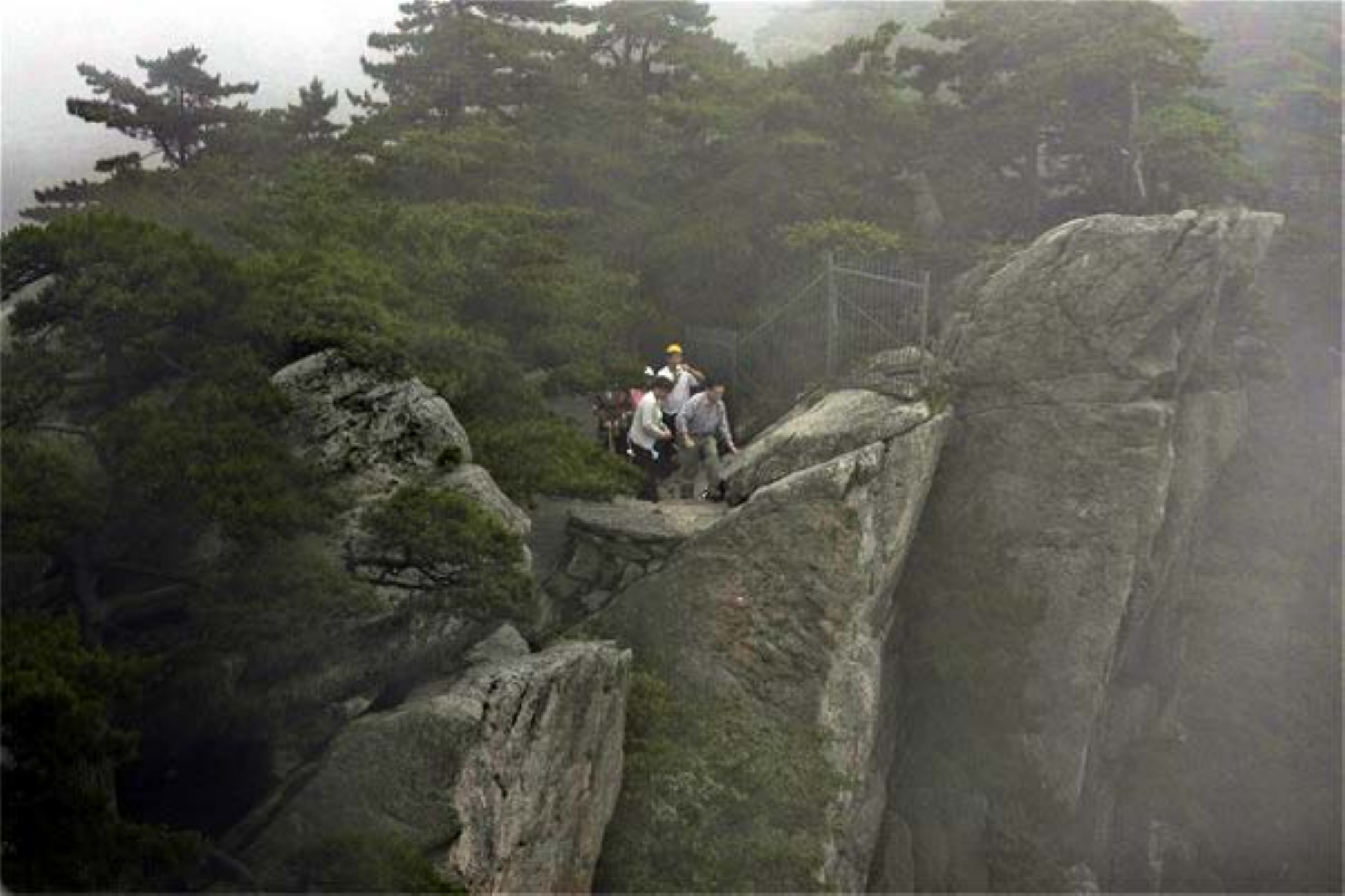}\\
		\includegraphics[width=0.195\textwidth]{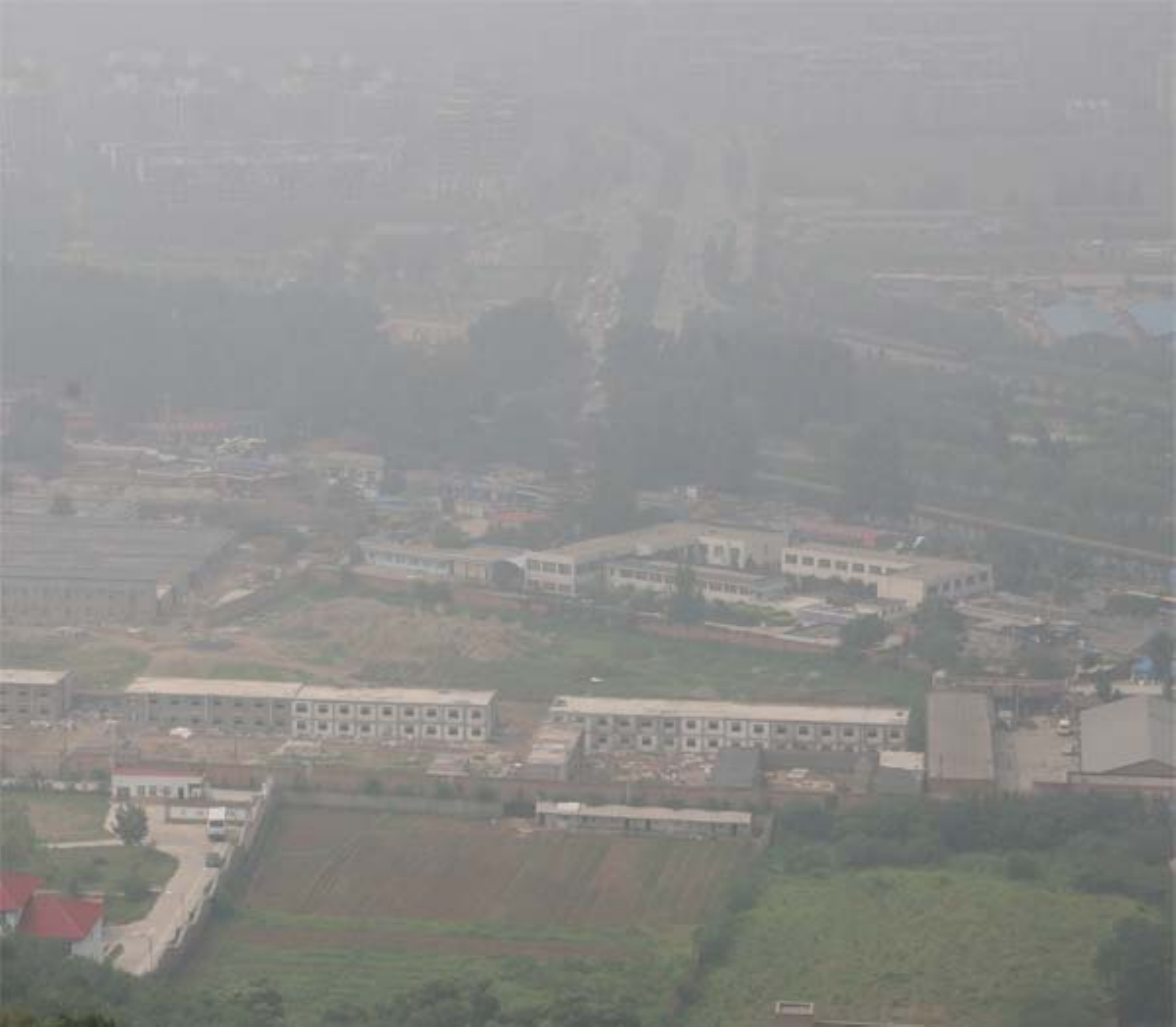}&
		\includegraphics[width=0.195\textwidth]{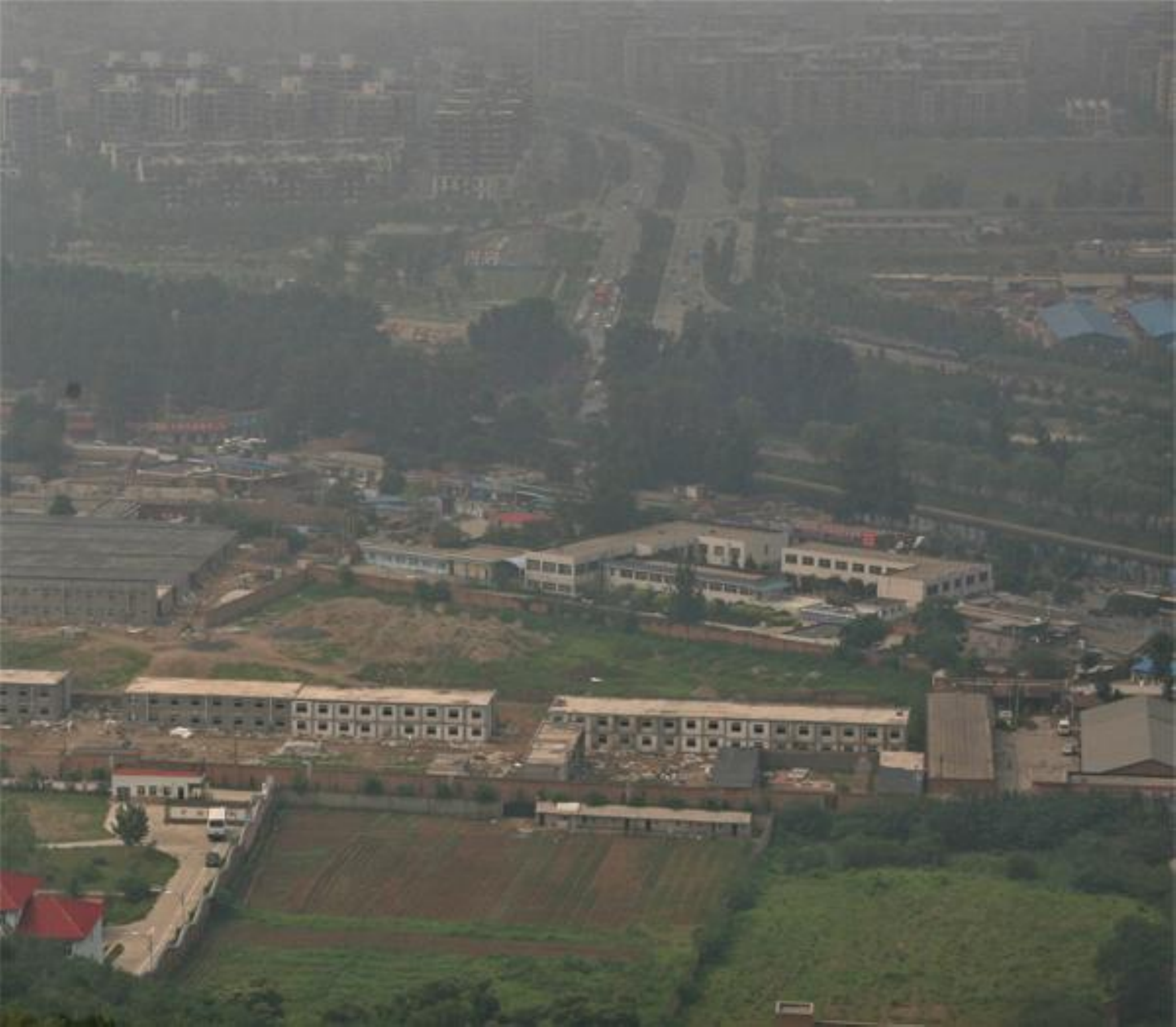}&
		\includegraphics[width=0.195\textwidth]{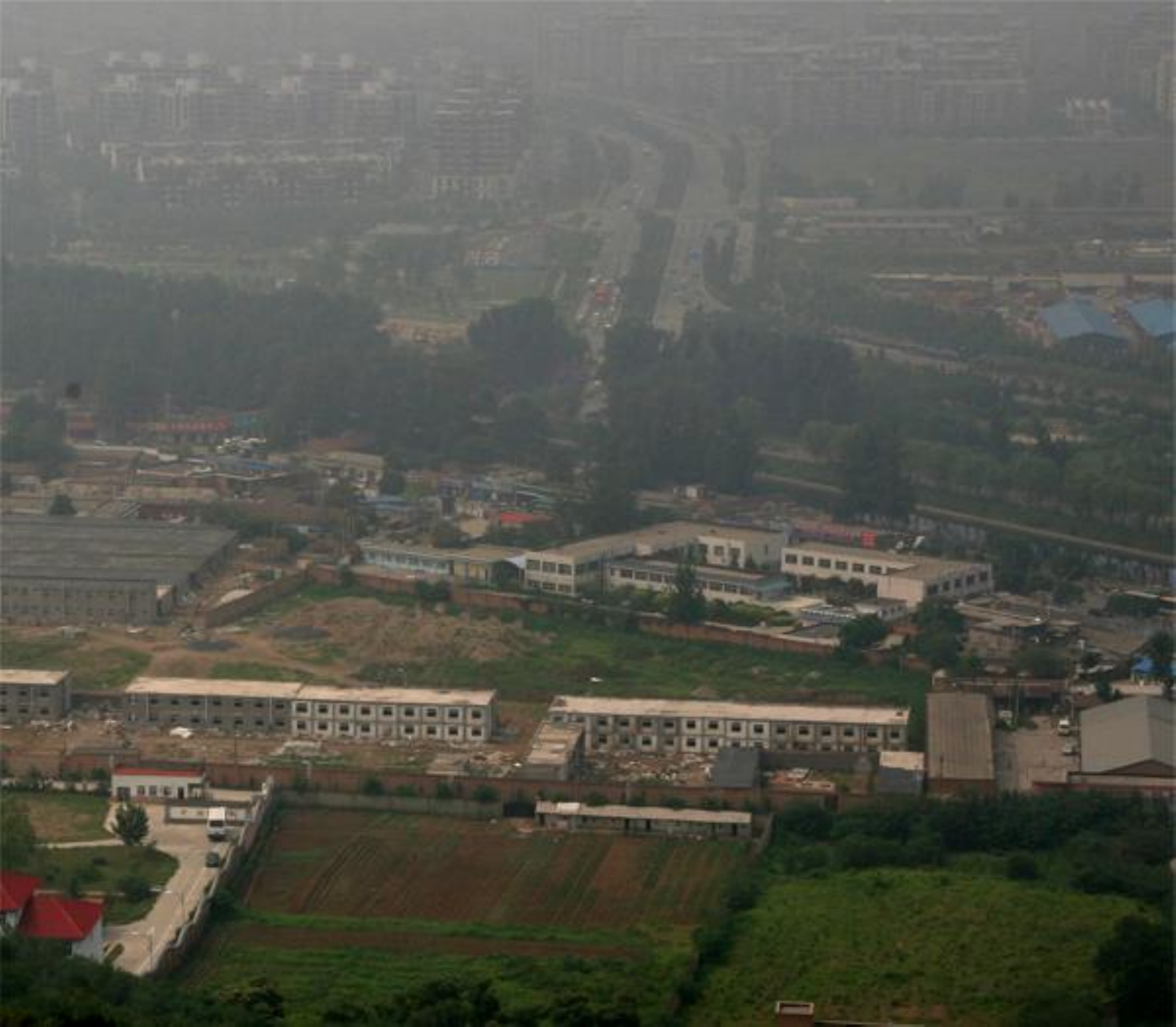}&
		\includegraphics[width=0.195\textwidth]{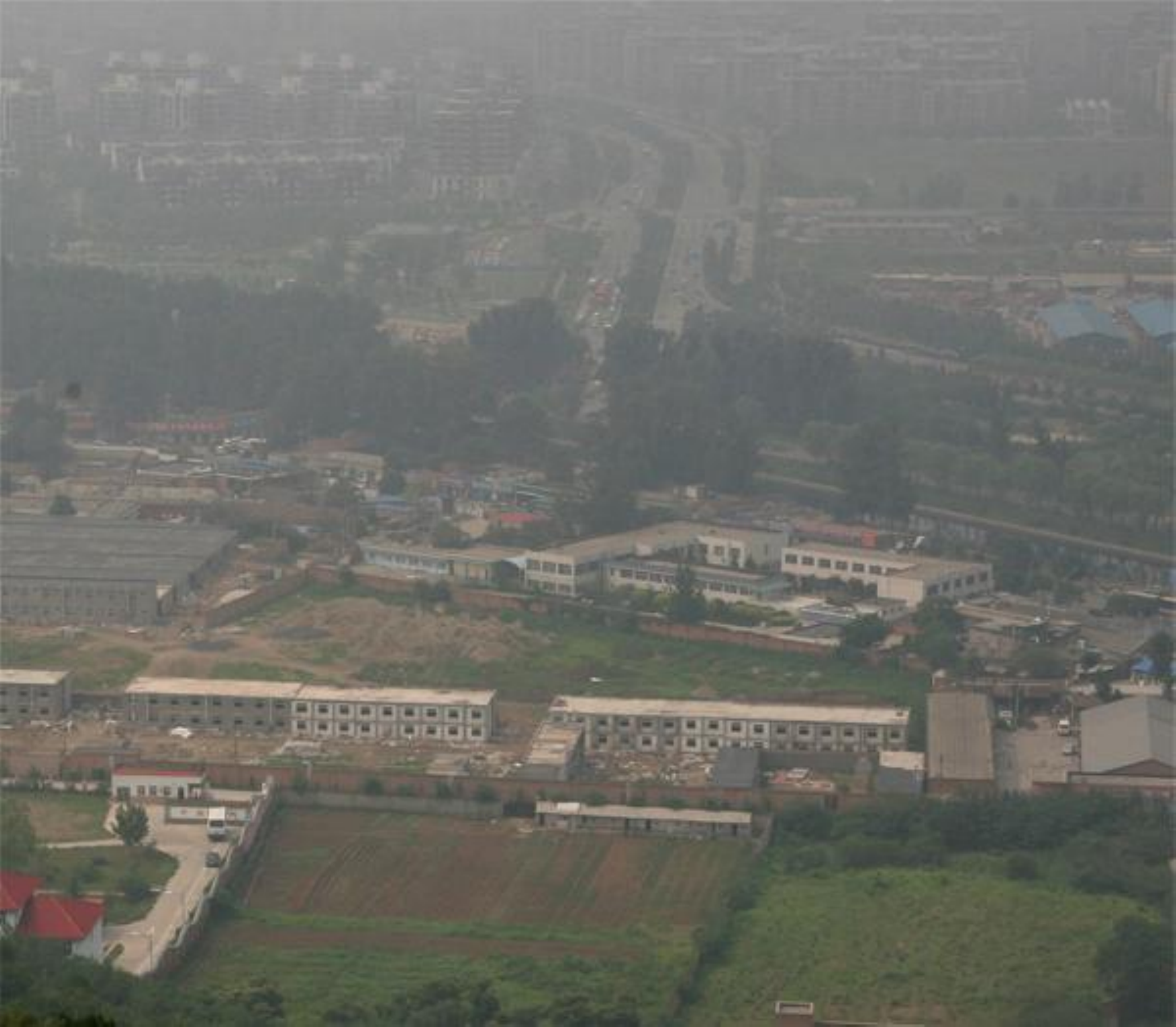}&
		\includegraphics[width=0.195\textwidth]{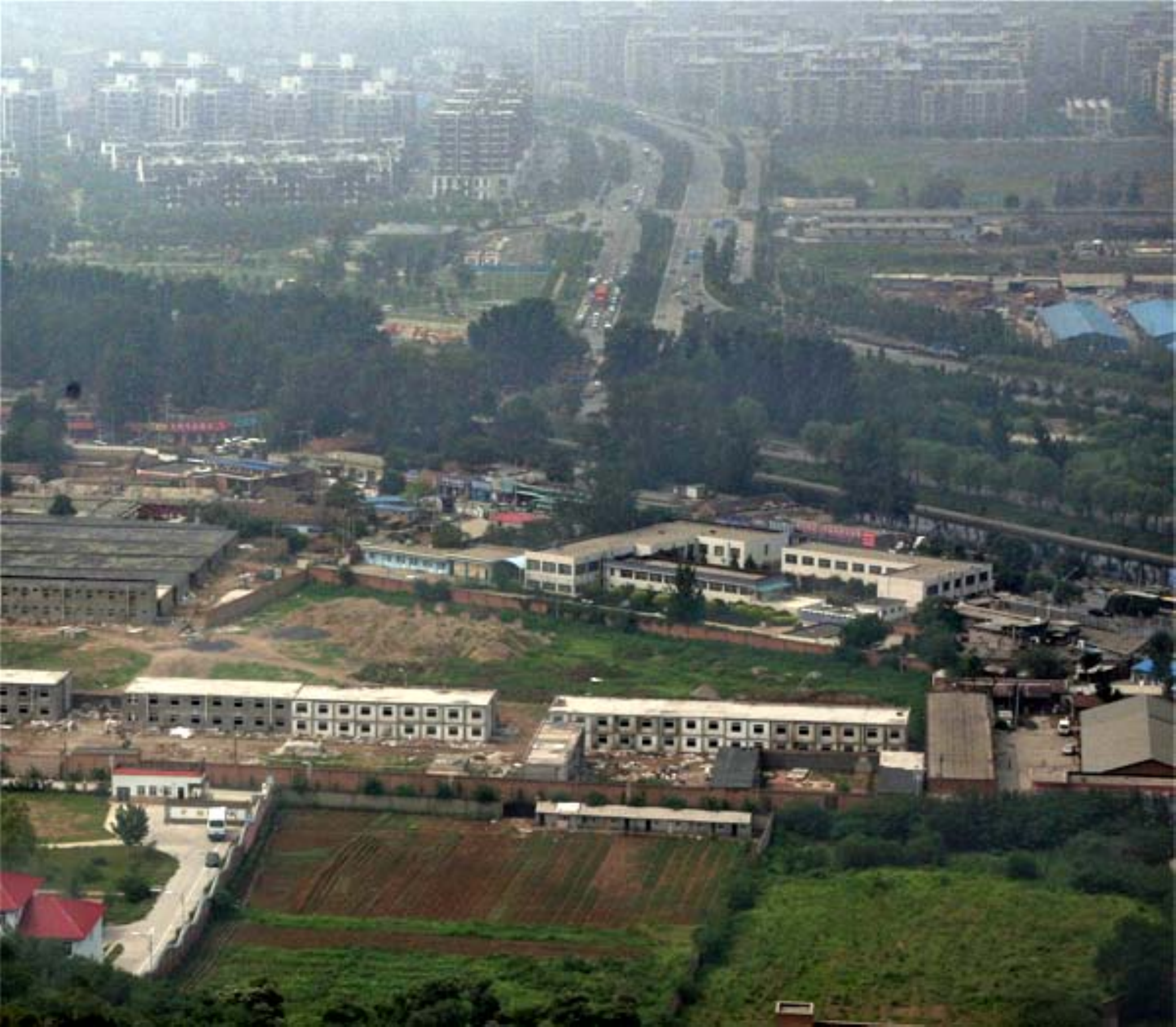}\\
		\footnotesize Hazy Input&\footnotesize~\cite{Meng2014Efficient} & \footnotesize~\cite{Cai2016DehazeNet} & \footnotesize~\cite{Ren2016Single} & \footnotesize Ours
	\end{tabular}
	\caption{The qualitative enhancement performance on challenging real-world hazy images collected by Fattal \cite{Fattal2014}.}
	\label{fig:readehazvisres}
\end{figure*}

\subsubsection{Single Image Haze Removal}
We compare our proposed framework with state-of-the-art approaches, including He~\cite{He2011Single}, Meng~\cite{Meng2014Efficient}, Cai~\cite{Cai2016DehazeNet}, Berman~\cite{Berman2016Non} and Ren~\cite{Ren2016Single}, for single image haze removal. In this task, we initialize the transmission based on existing prior (i.e., haze line~\cite{Berman2016Non}) and perform the DPE propagation to obtain the optimal transmission map. Then we estimate the latent clear image using the atmospheric scattering model as stated in Table~\ref{tab:summary}. We first report the quantitative performances ( i.e., the average PSNR, SSIM, and $L_1$ error~\cite{Li2017Haze}) of all the compared methods on two representative dehazing benchmarks (i.e., Fattal's~\cite{Fattal2014} and D-Hazy~\cite{AncutiDHazy}) in Table~\ref{tab:dehazquanres}. It is easy to observe that DPE achieves the best results among all the compared methods on all the test benchmarks. We then compare the estimated transmissions and recovered results on example images from Fattal's dataset in Figs.~\ref{fig:sydehazvisres} and \ref{fig:transmission-sydehazvisres}. Additional visual comparisons on example images from D-Hazy dataset is plotted in Fig.~\ref{fig:sydehazvisres2}. We also evaluate DPE on real-world hazy images and plot the enhancement results in Fig.~\ref{fig:readehazvisres}. From these quantitative and qualitative analyses, we observe that our method consistently out-performs all the compared dehazing methods.

\subsubsection{Underwater Image Enhancement}

Finally, we evaluate DPE on the task of underwater image enhancement. In this application, three different categories of algorithms, including layer decomposition~(i.e., \cite{li2014a}), fusion principle~(i.e., \cite{Ancuti2012Enhancing}) and transmission estimation (i.e., \cite{Berman2016Non}, \cite{UnderwaterHazeLines} and ours). Notice that here we follow \cite{UnderwaterHazeLines} to perform a histogram-based color correction as the post-process for three transmission-based methods. 
In Fig. \ref{fig:underes0}, we first compare the performance of transmission estimation for the work in \cite{Berman2016Non} and our DPE on an example underwater image. It can be seen that DPE obtains more accurate transmission, thus leads to the better enhanced image. Furthermore, we conduct experiments on example images collected by Berman et al.~\cite{UnderwaterHazeLines} (top two rows) and ourself (bottom two rows)\footnote{Based on the underwater robot picking contest: \url{http://www.cnurpc.org/}.}. We can see in Fig.~\ref{fig:underes} that DPE is able to obtain results with more details and better visual quality compared with other methods.

\begin{figure}[t]
	\begin{tabular}{c@{\extracolsep{0.1em}}c@{\extracolsep{0.1em}}c}
		\includegraphics[width=0.155\textwidth]{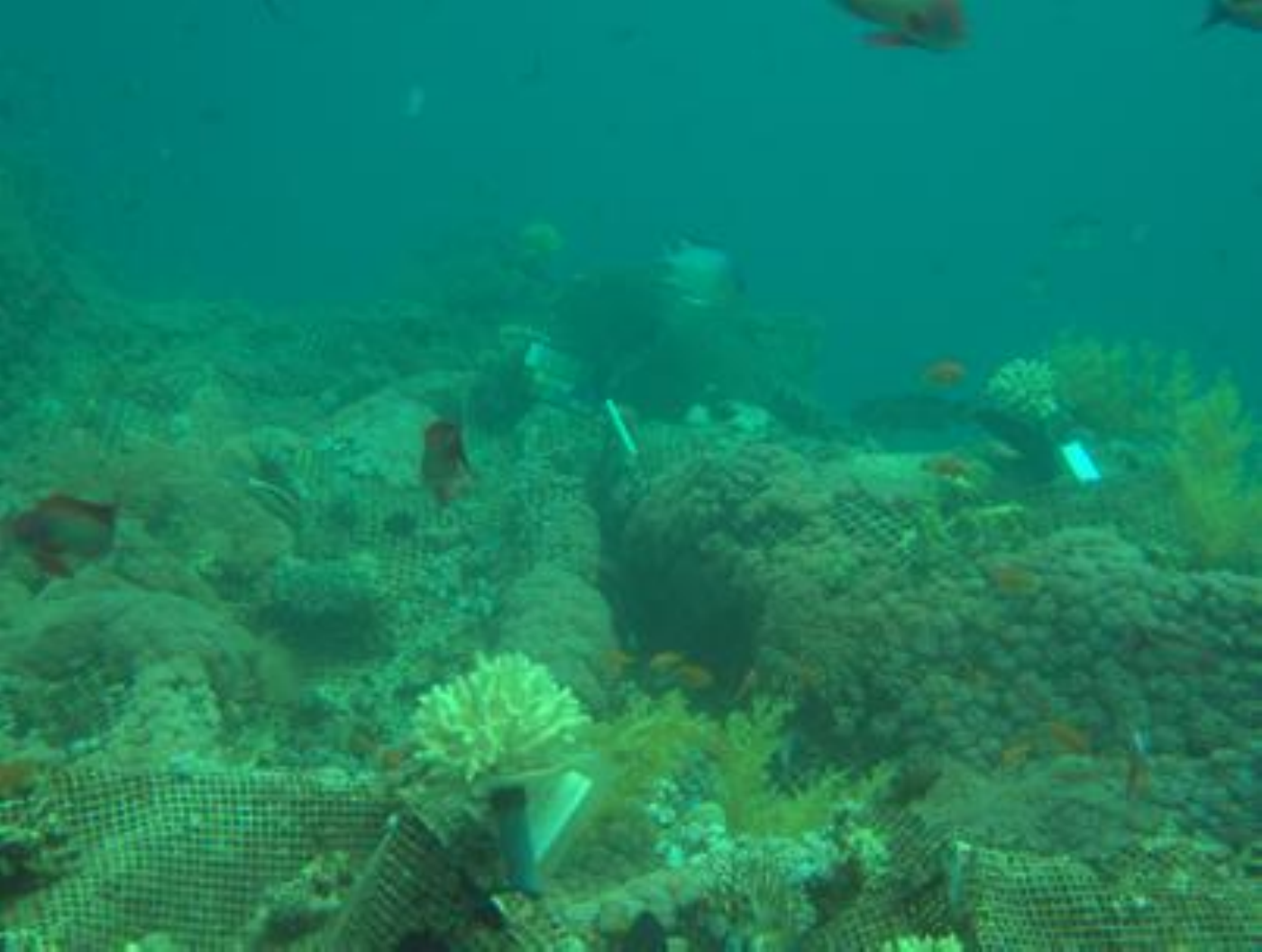}&
		\includegraphics[width=0.155\textwidth]{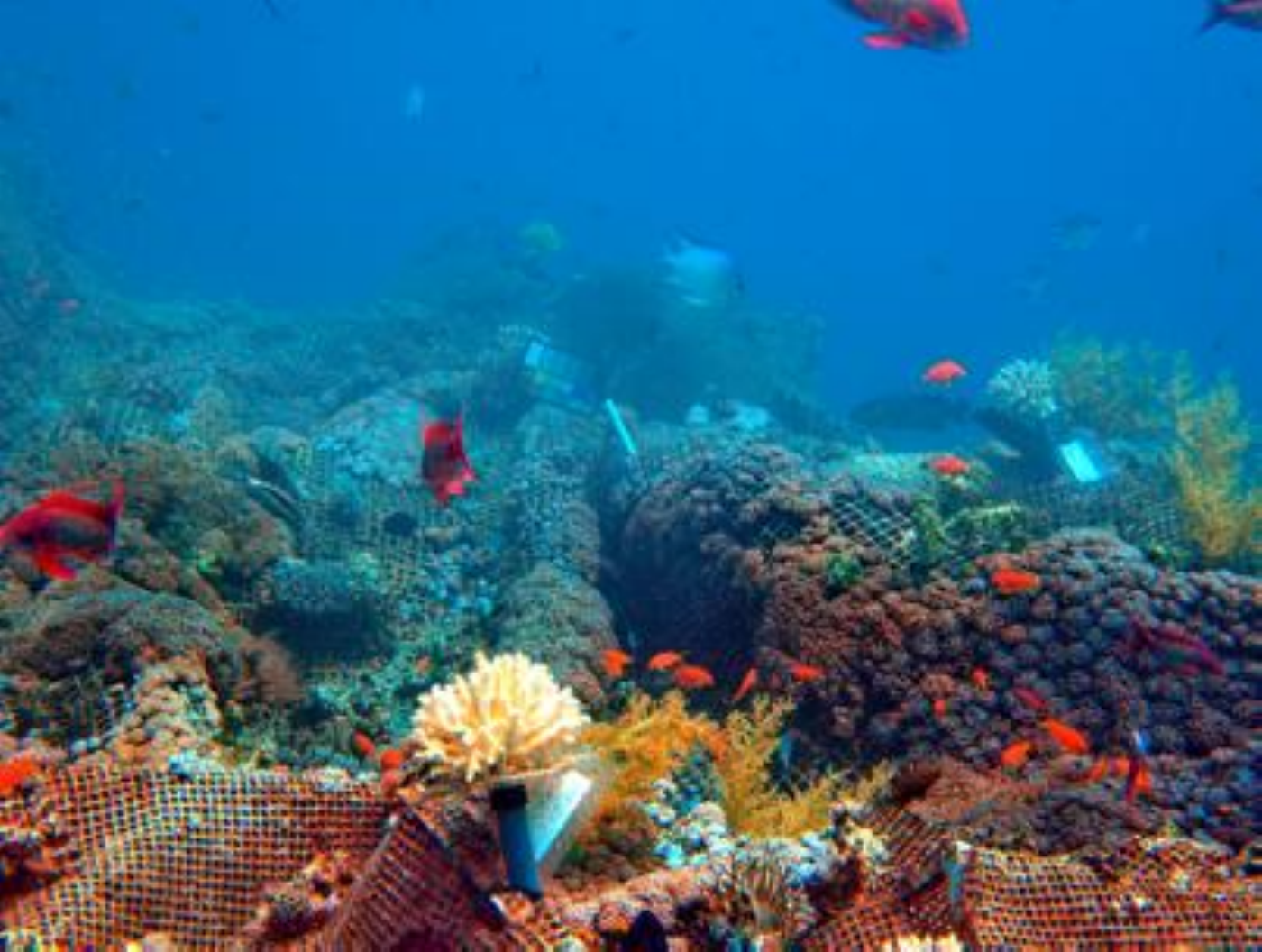}&
		\includegraphics[width=0.155\textwidth]{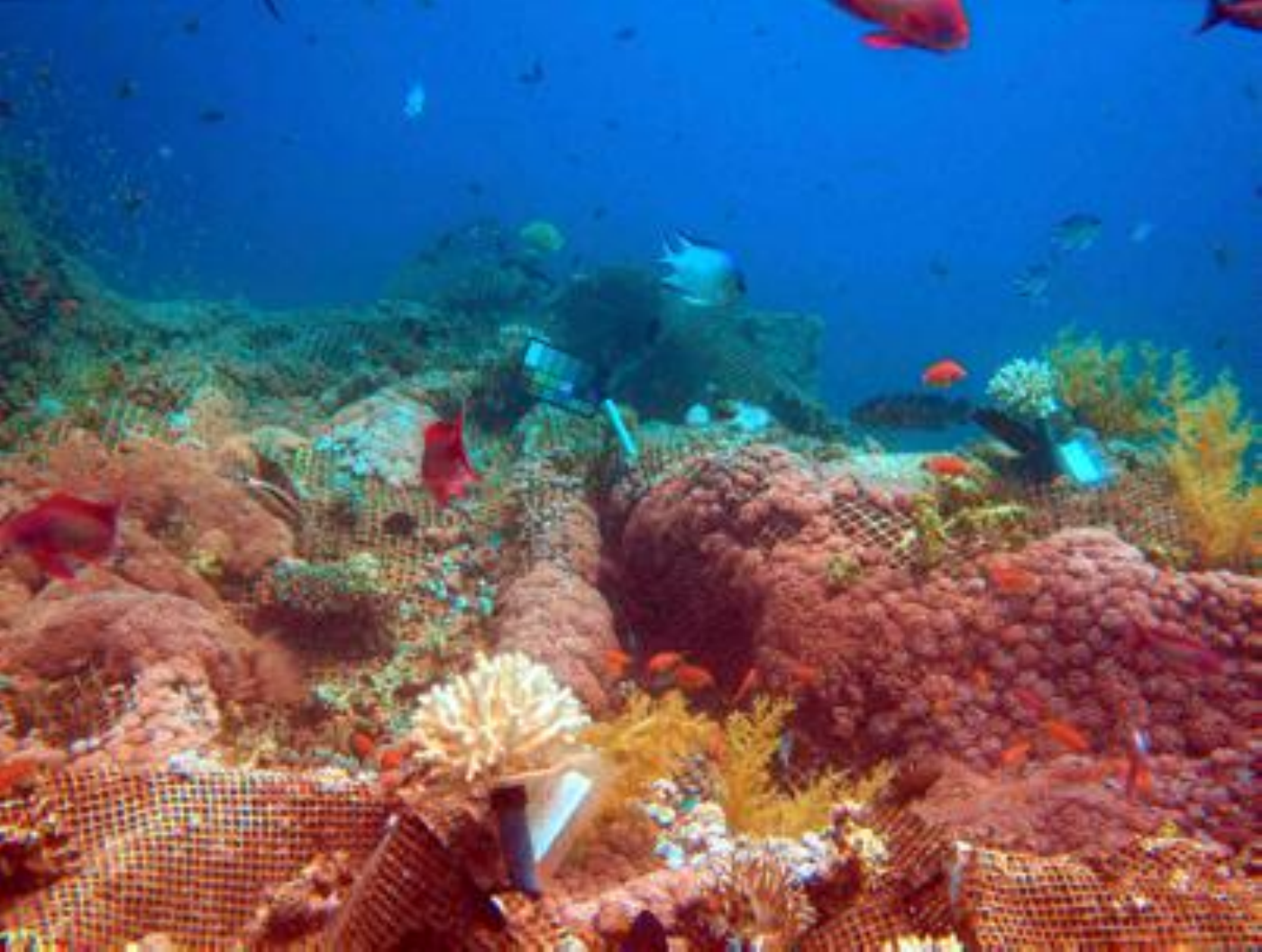}\\
		&
		\includegraphics[width=0.155\textwidth]{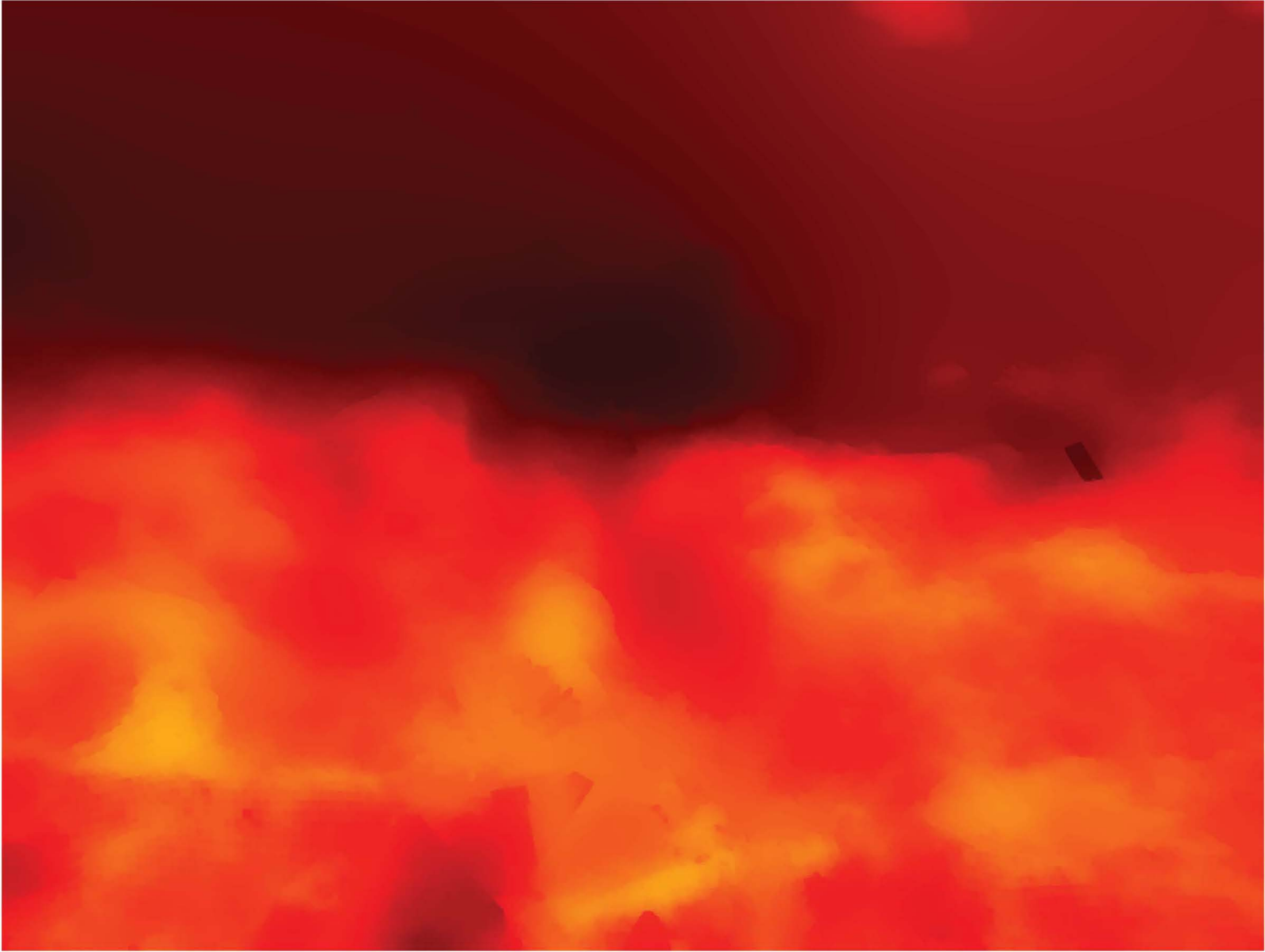}&
		\includegraphics[width=0.155\textwidth]{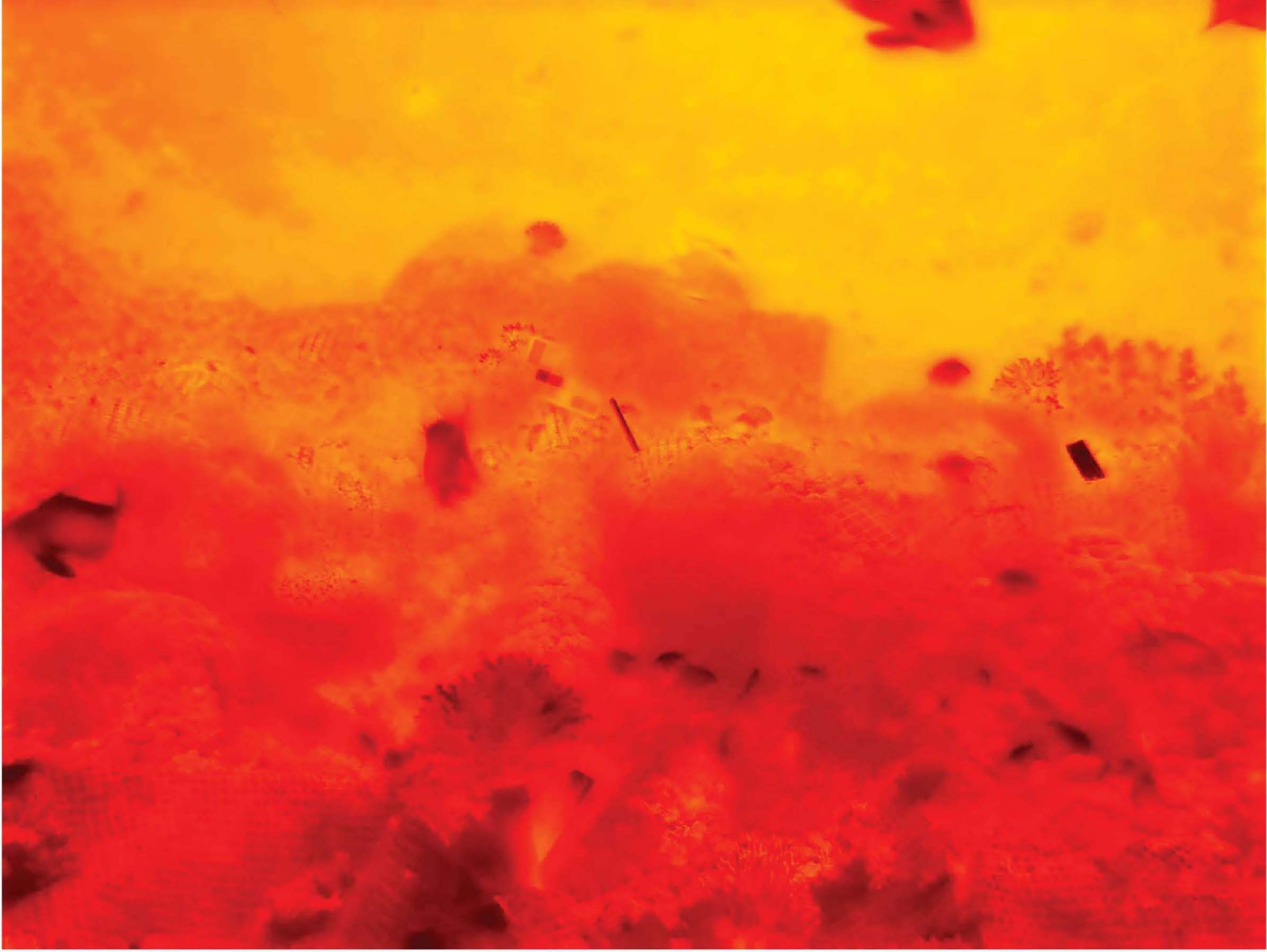}\\
		\footnotesize Underwater Input&\footnotesize~\cite{Berman2016Non}&\footnotesize Ours
	\end{tabular}
	\caption{Comparing the underwater enhancement (top row) and transmission estimation (bottom row) results on an example underwater image collected by Berman et al.~\cite{UnderwaterHazeLines}.}
	\label{fig:underes0}
\end{figure}

\begin{figure*}[t]
	\begin{tabular}{c@{\extracolsep{0.2em}}c@{\extracolsep{0.2em}}c@{\extracolsep{0.2em}}c@{\extracolsep{0.2em}}c@{\extracolsep{0.2em}}c}
		\includegraphics[width=0.16\textwidth]{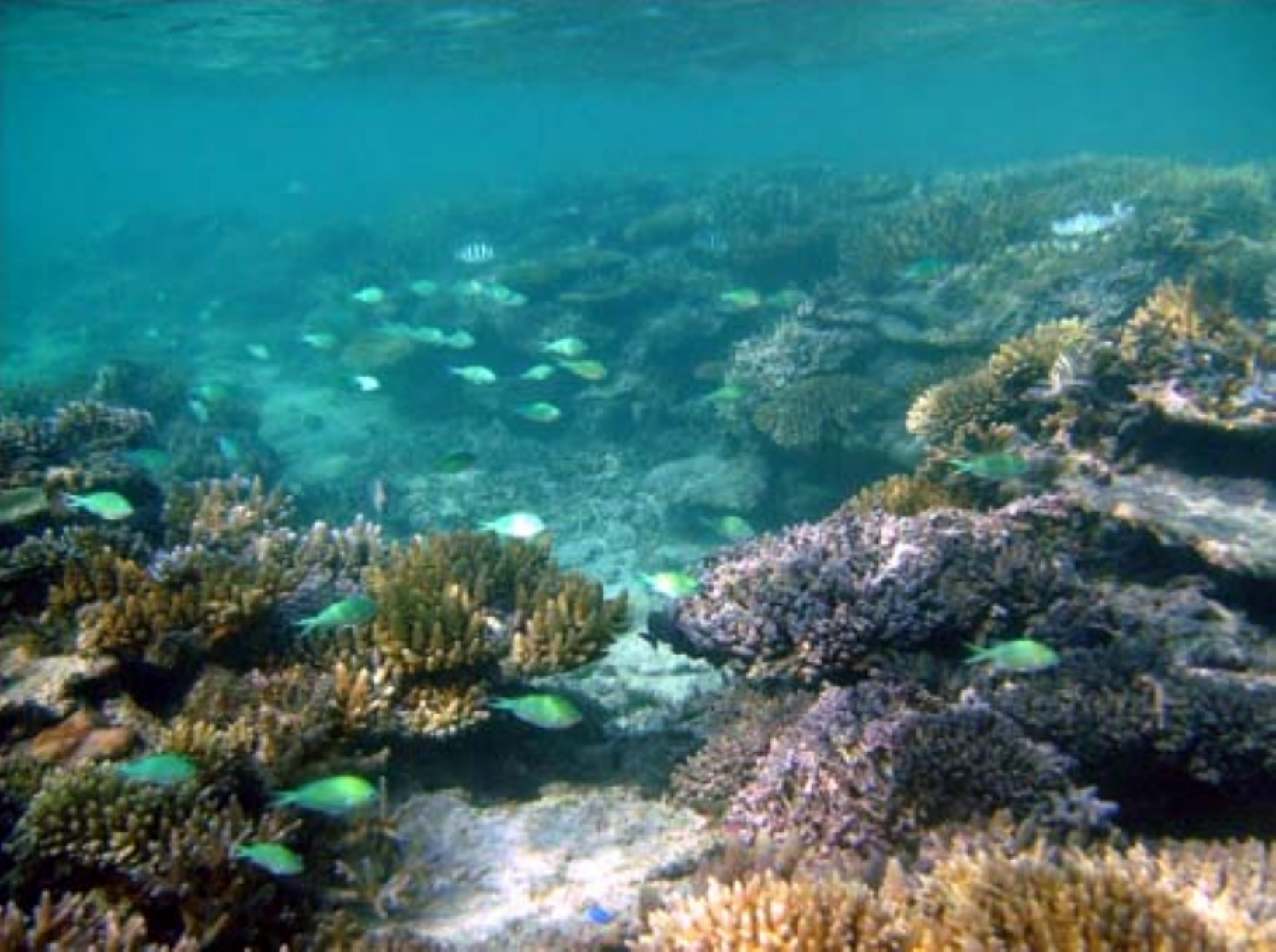}&
		\includegraphics[width=0.16\textwidth]{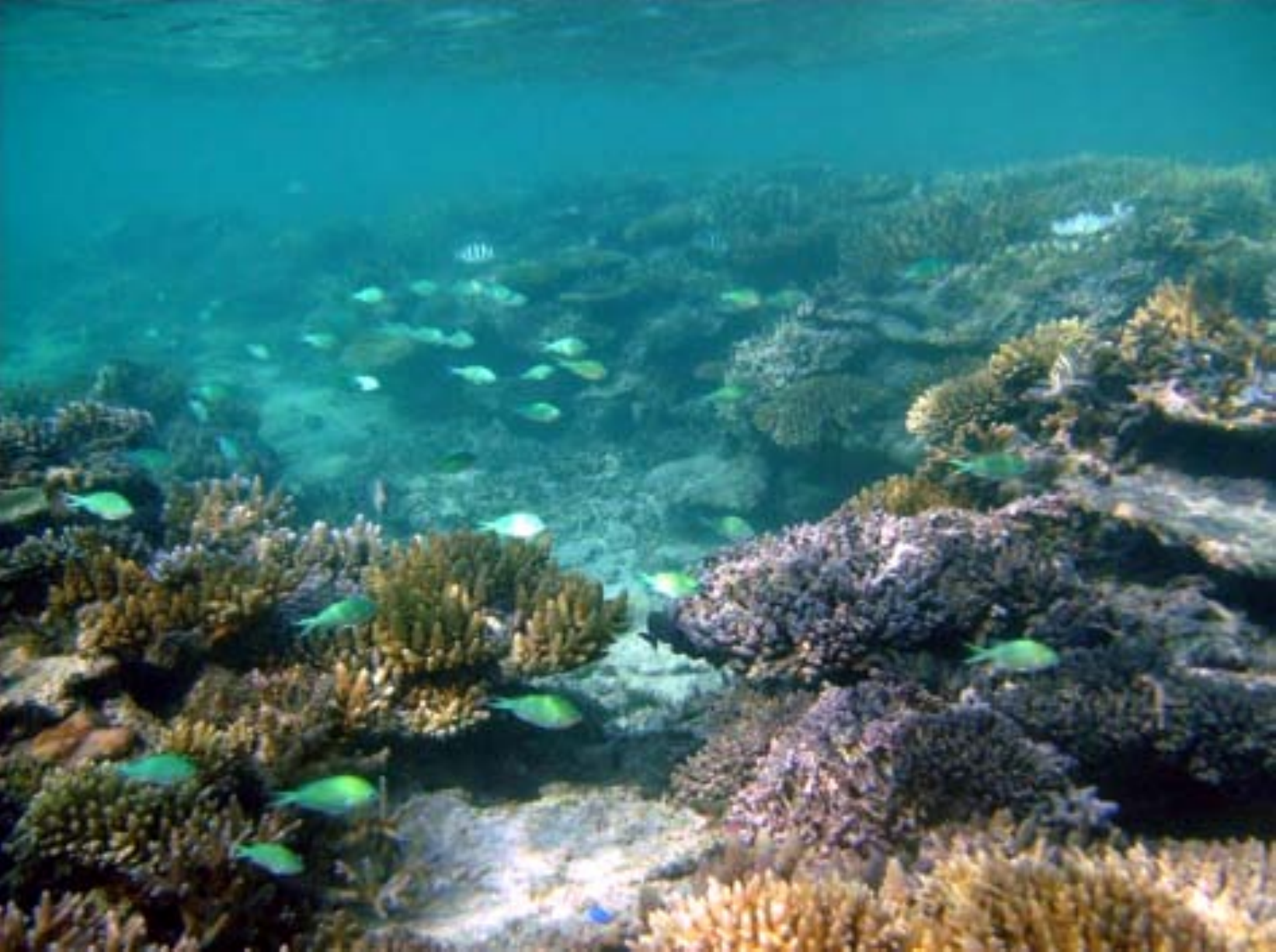}&
		\includegraphics[width=0.16\textwidth]{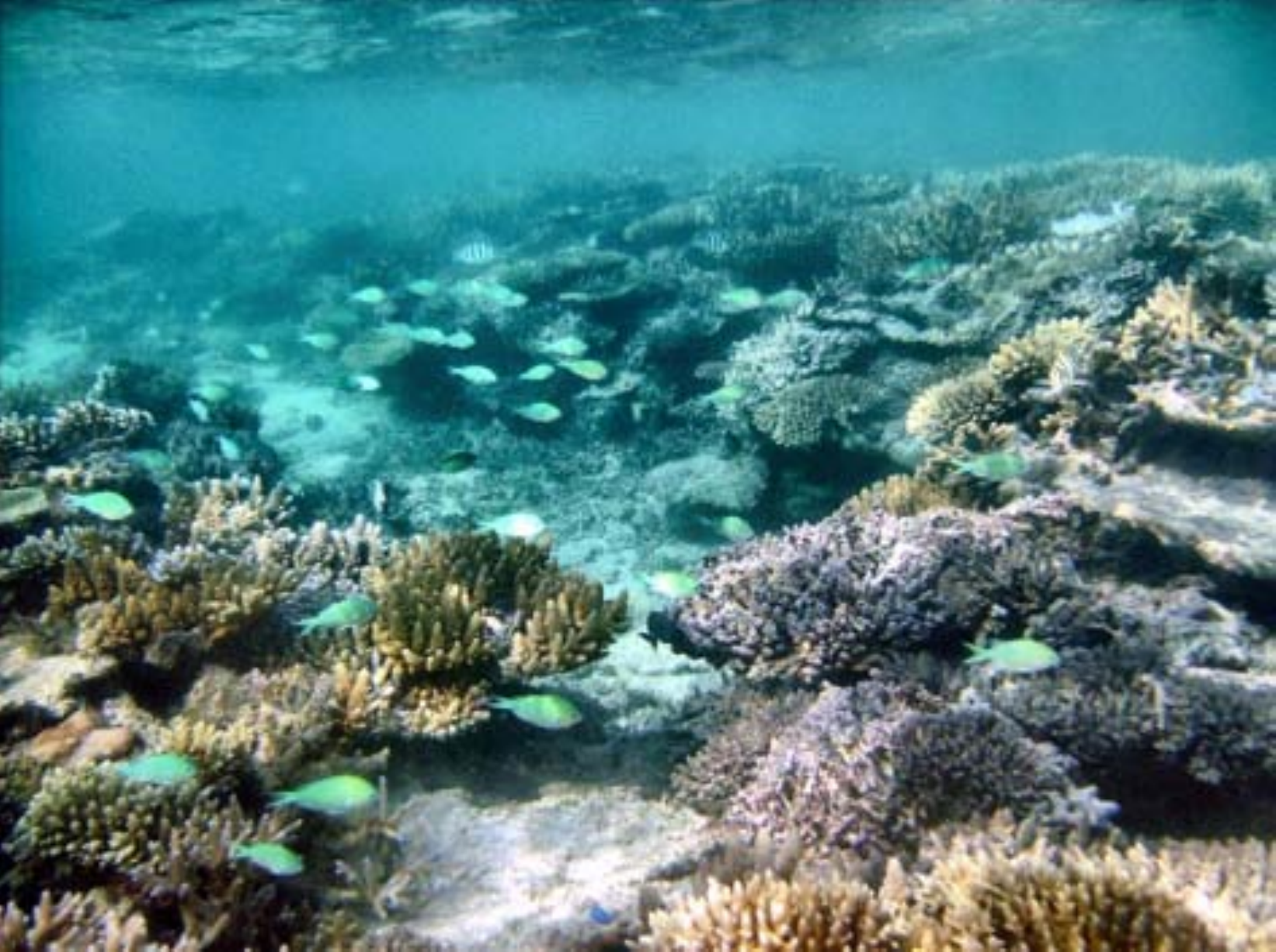}&
		\includegraphics[width=0.16\textwidth]{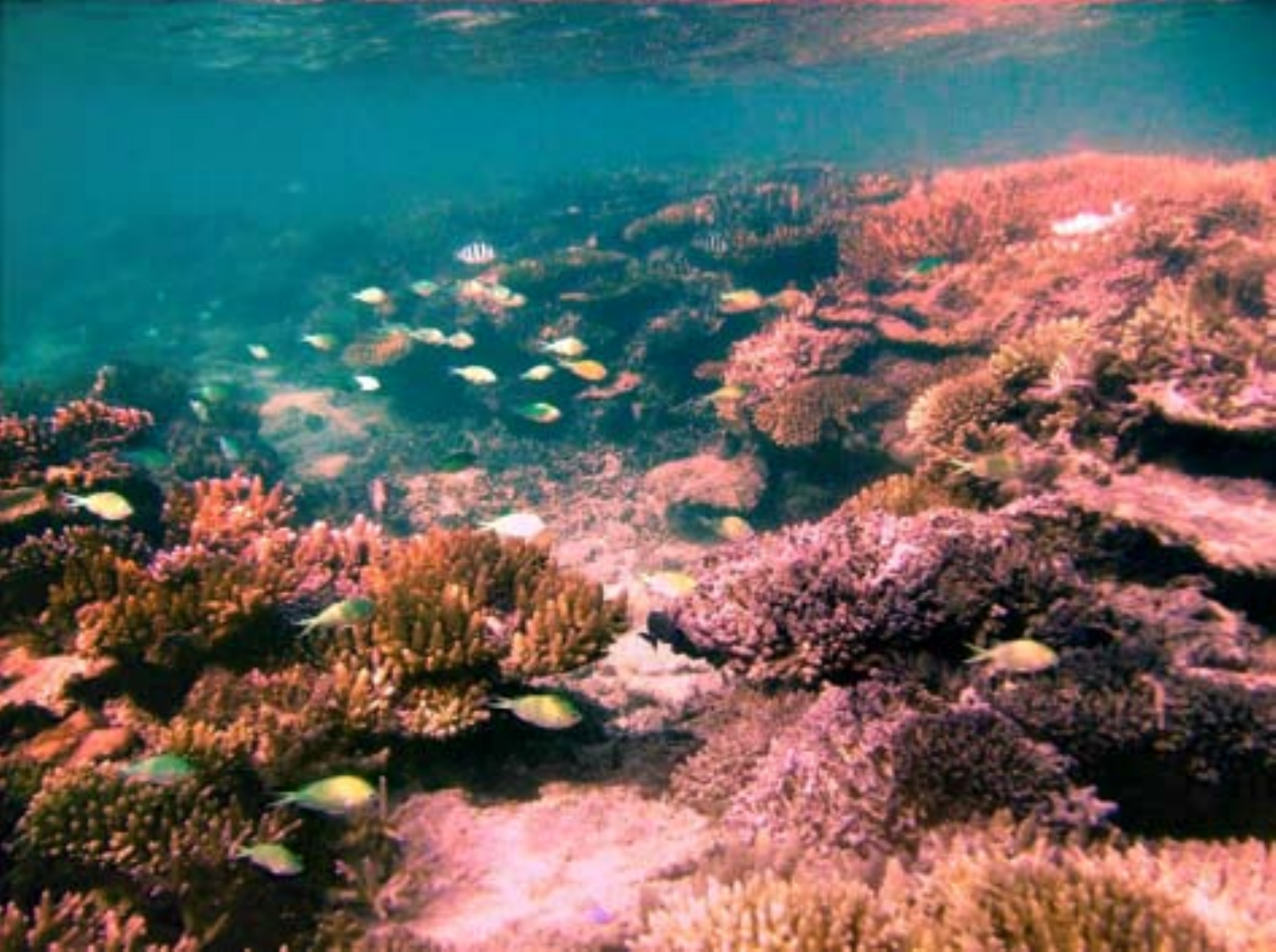}&
		\includegraphics[width=0.16\textwidth]{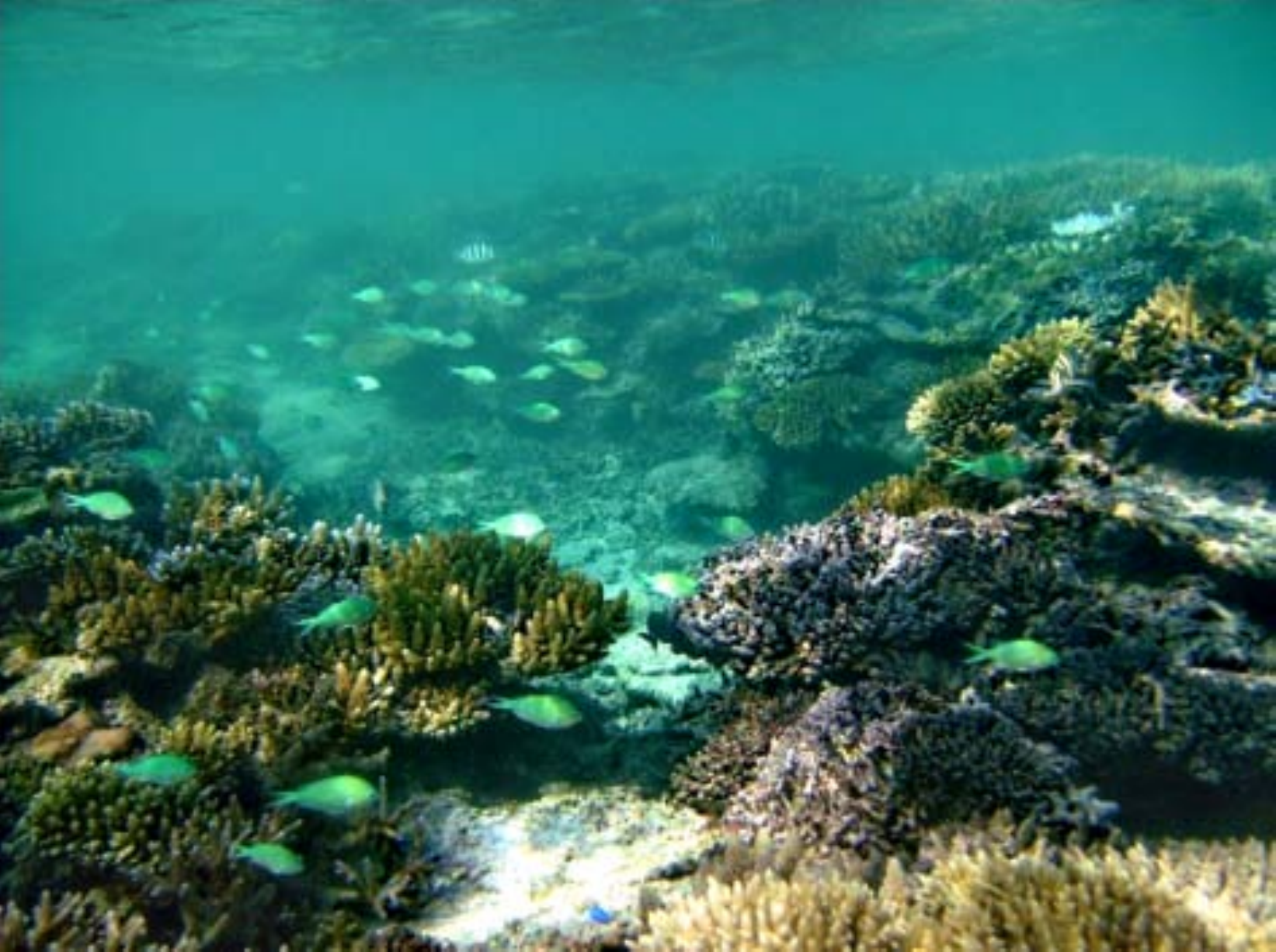}&
		\includegraphics[width=0.16\textwidth]{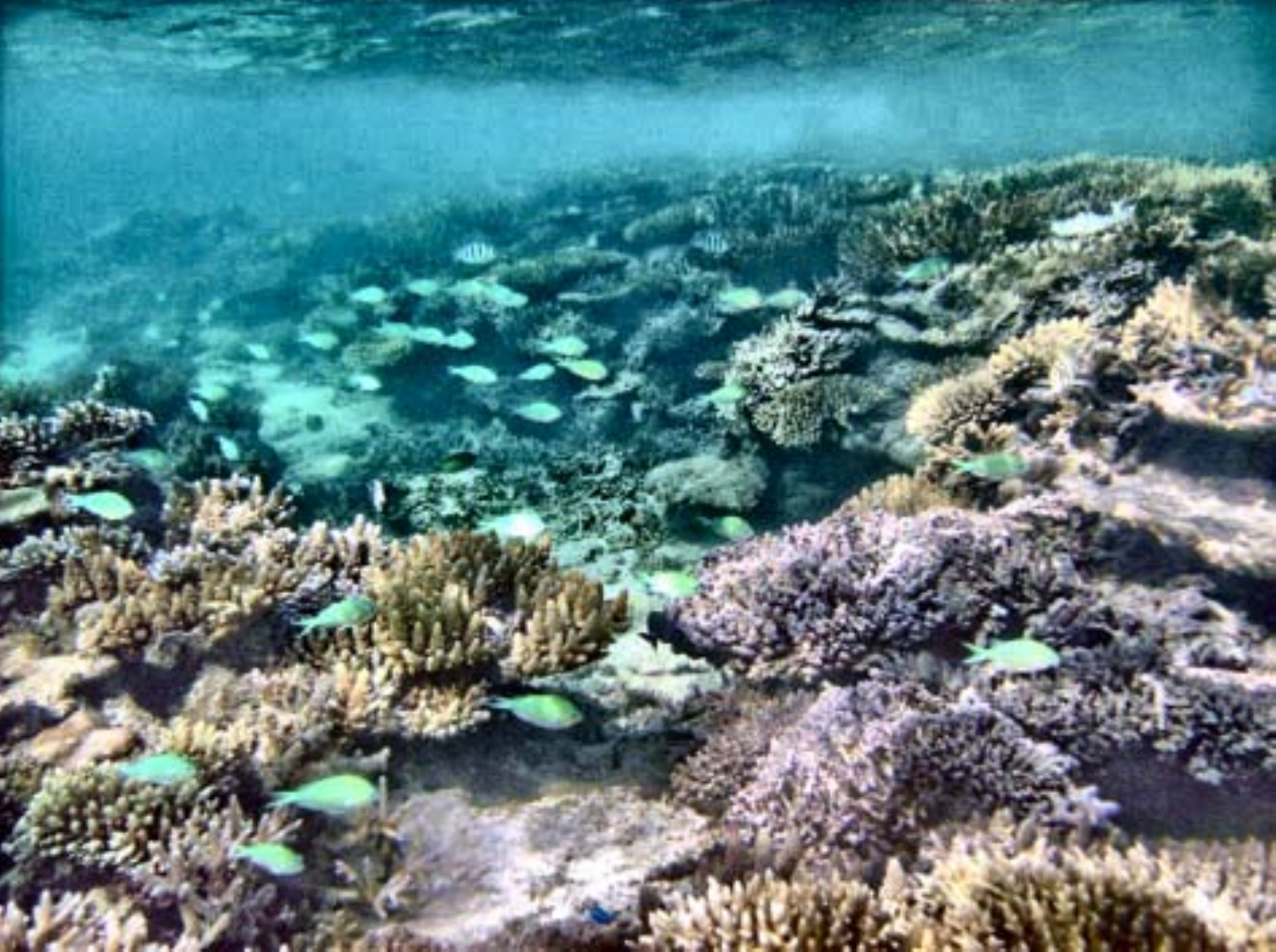}\\
		\includegraphics[width=0.16\textwidth]{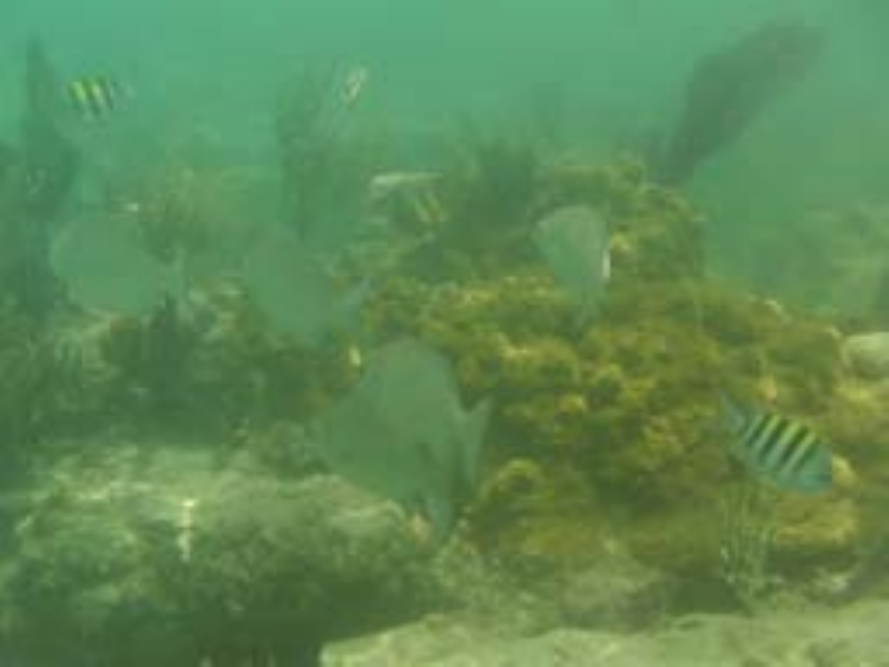}&
		\includegraphics[width=0.16\textwidth]{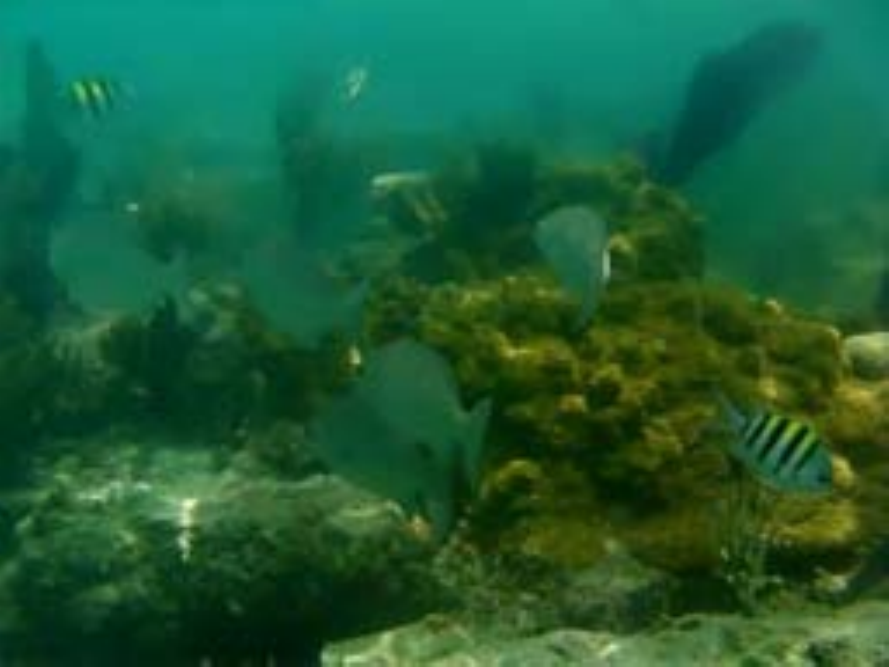}&
		\includegraphics[width=0.16\textwidth]{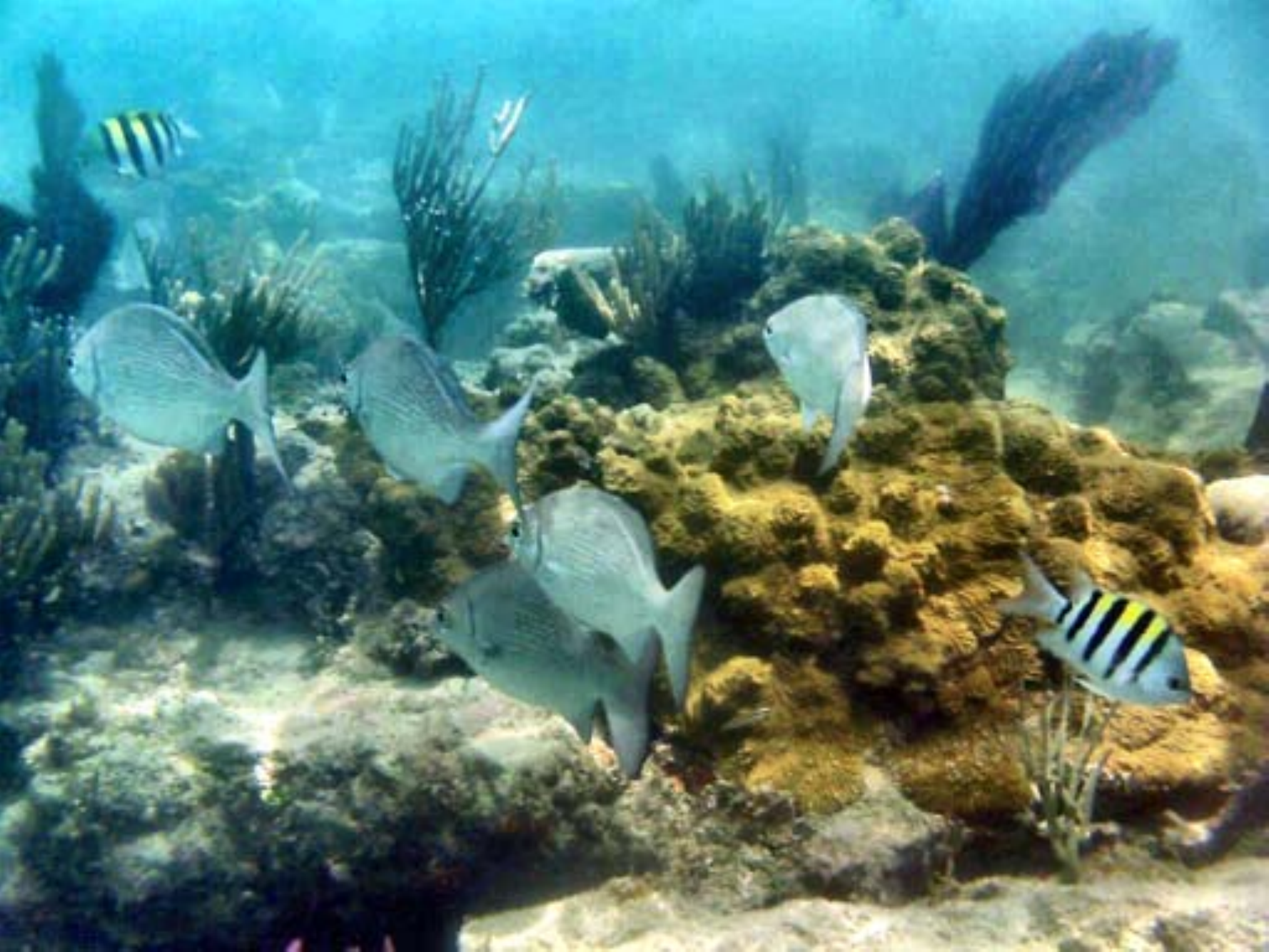}&
		\includegraphics[width=0.16\textwidth]{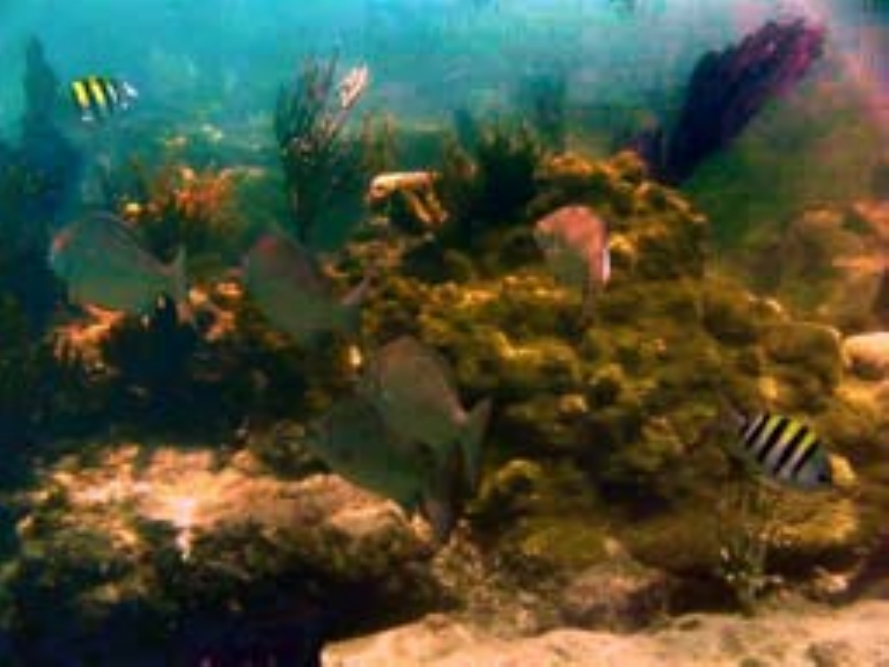}&
		\includegraphics[width=0.16\textwidth]{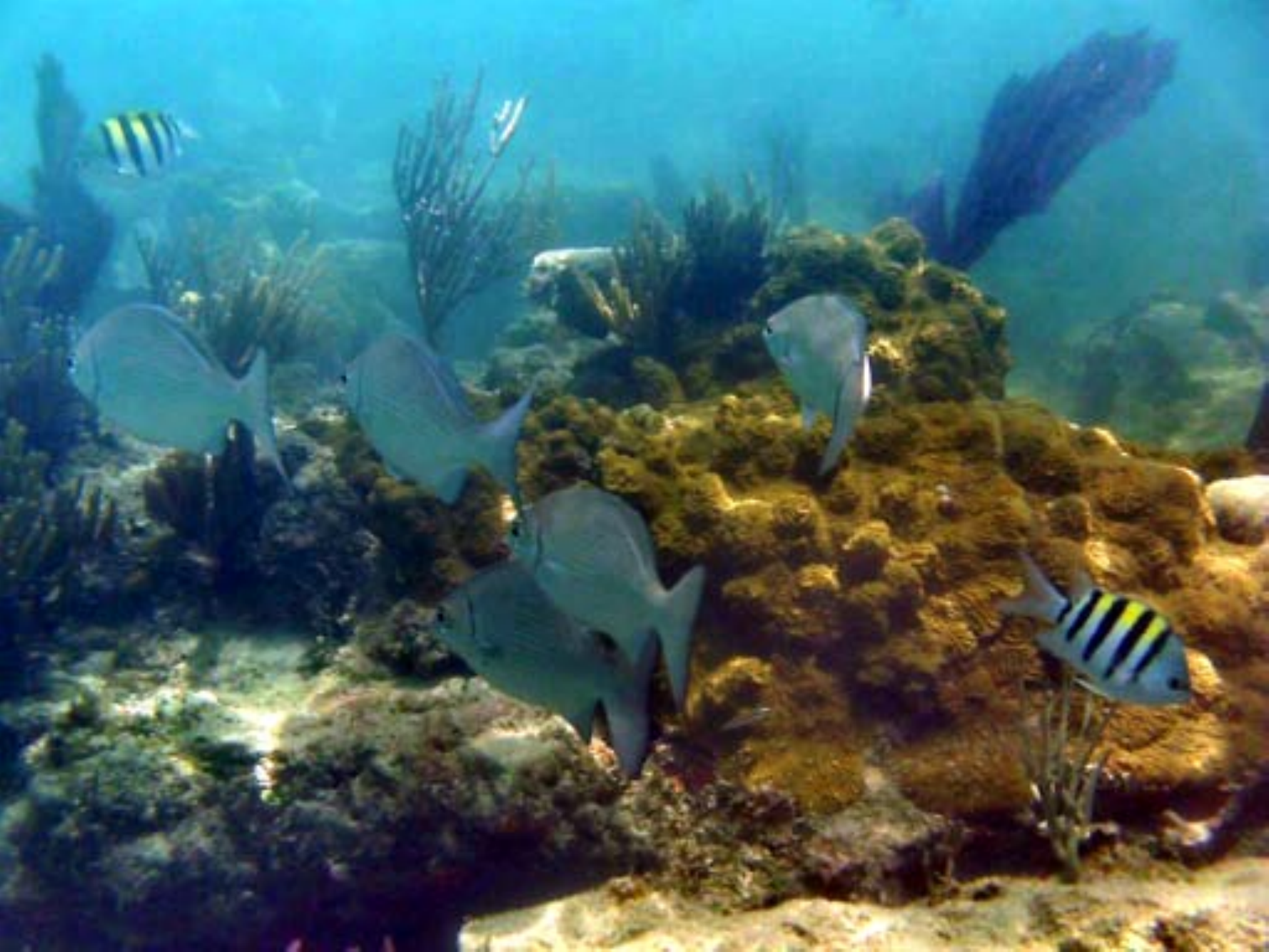}&
		\includegraphics[width=0.16\textwidth]{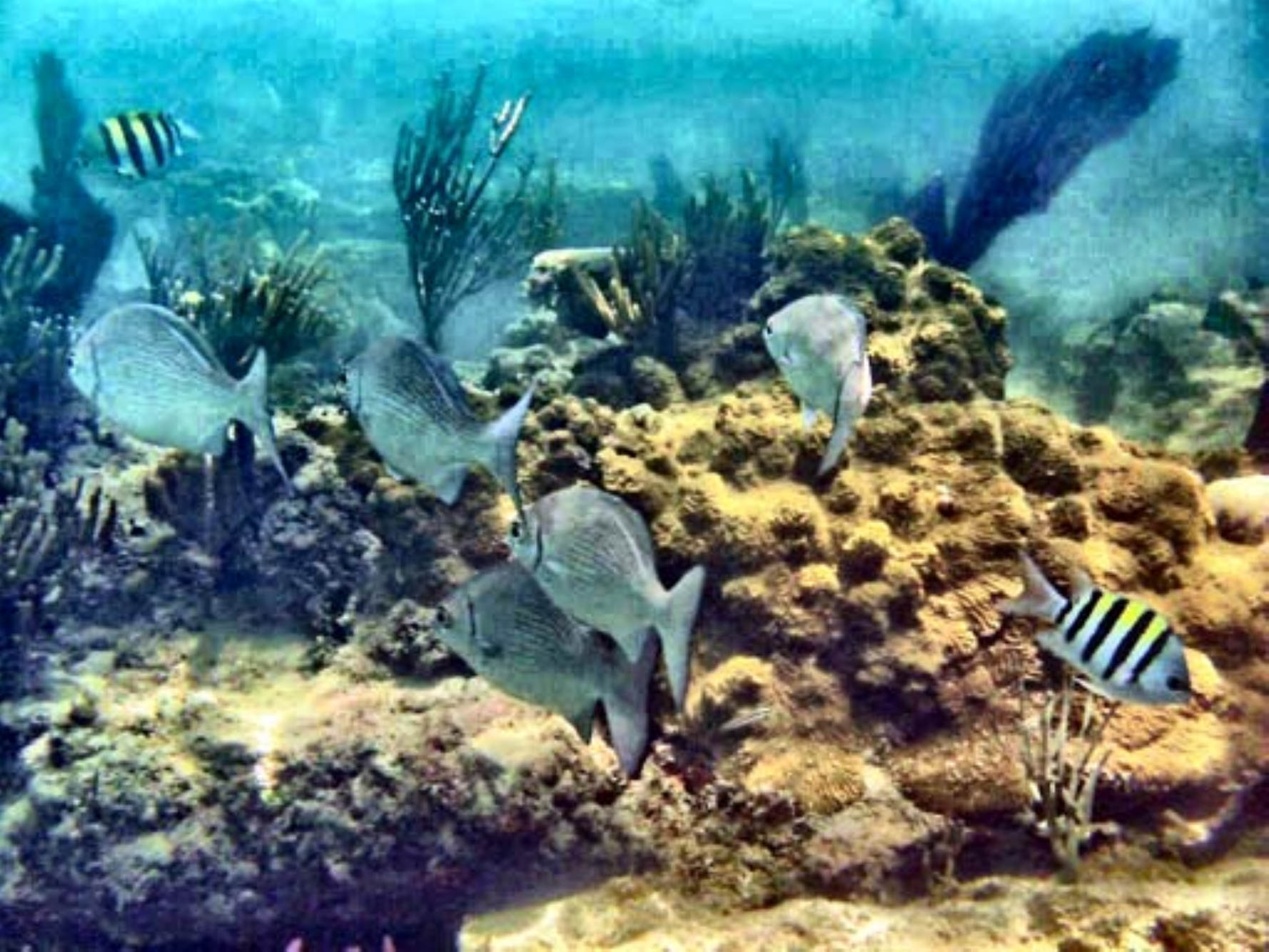}\\
		\includegraphics[width=0.16\textwidth]{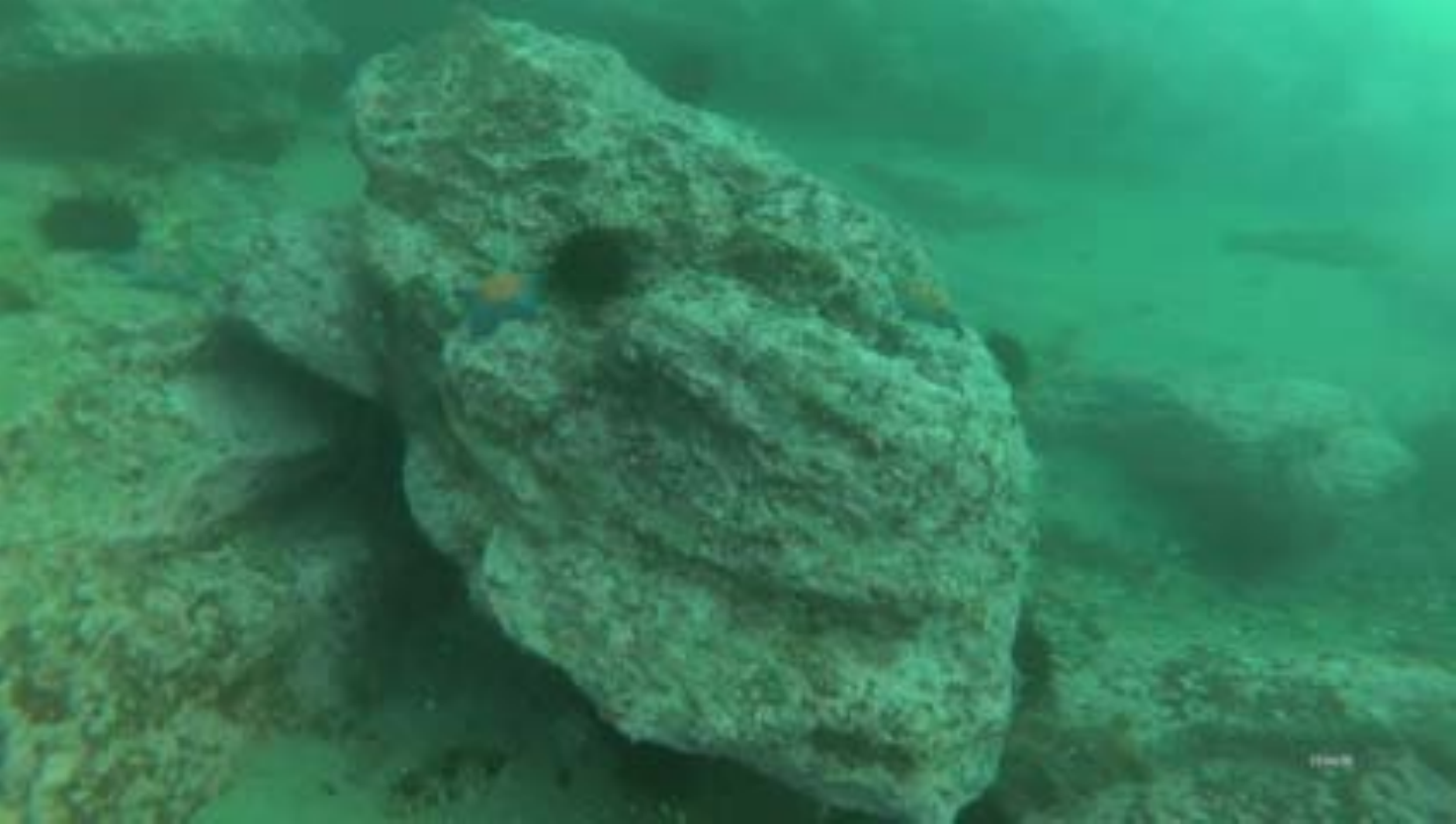}&
		\includegraphics[width=0.16\textwidth]{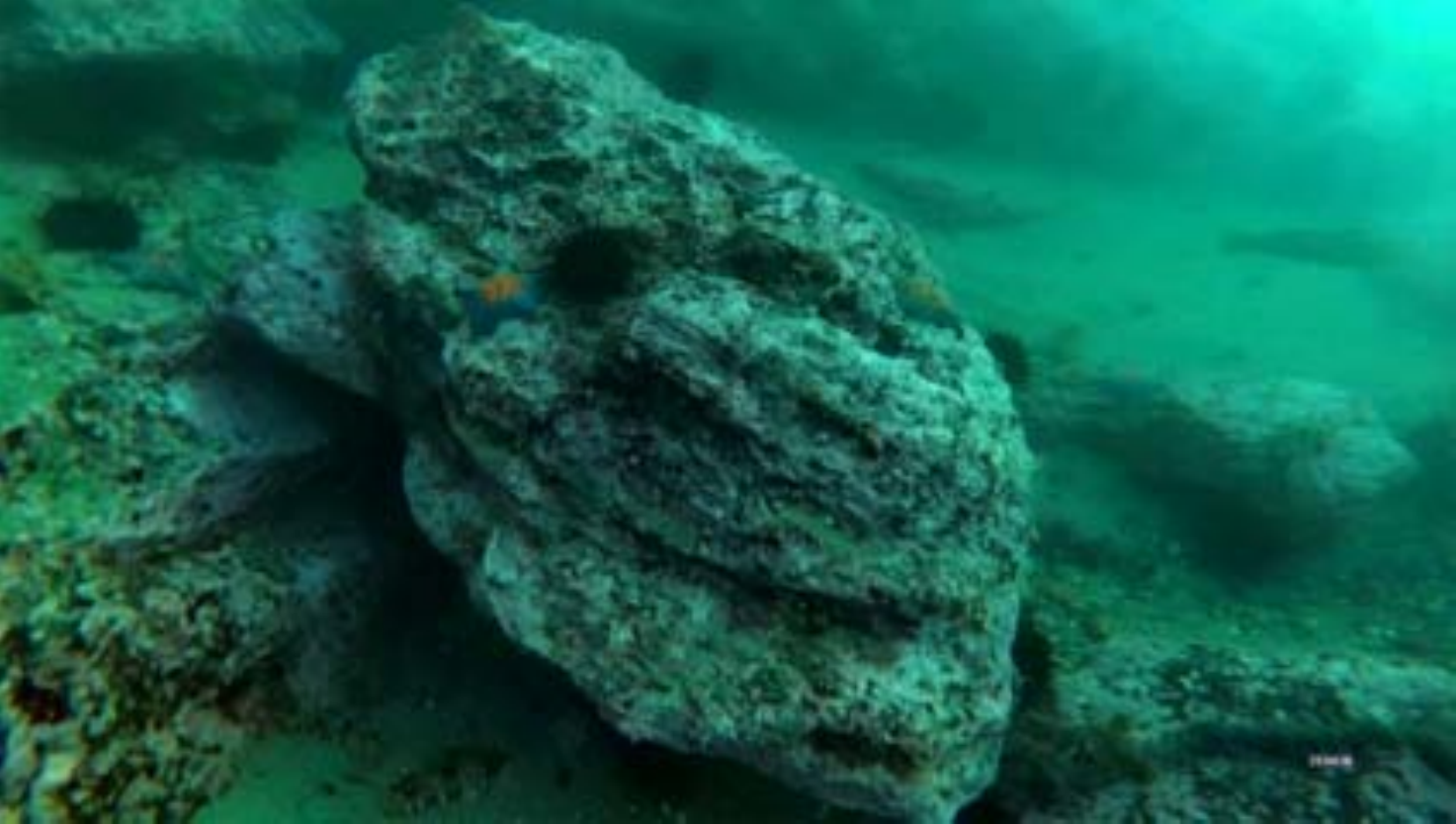}&
		\includegraphics[width=0.16\textwidth]{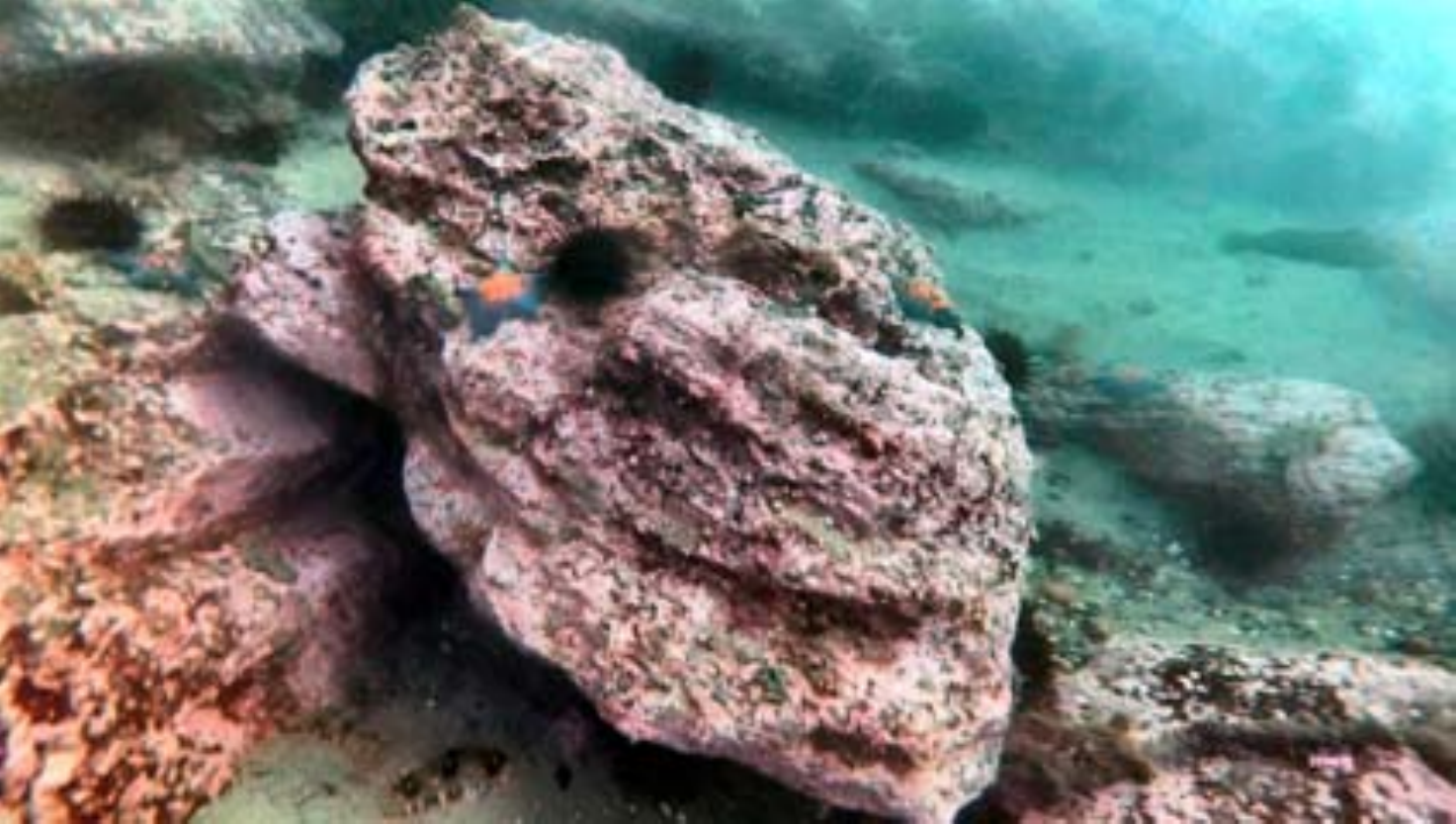}&
		\includegraphics[width=0.16\textwidth]{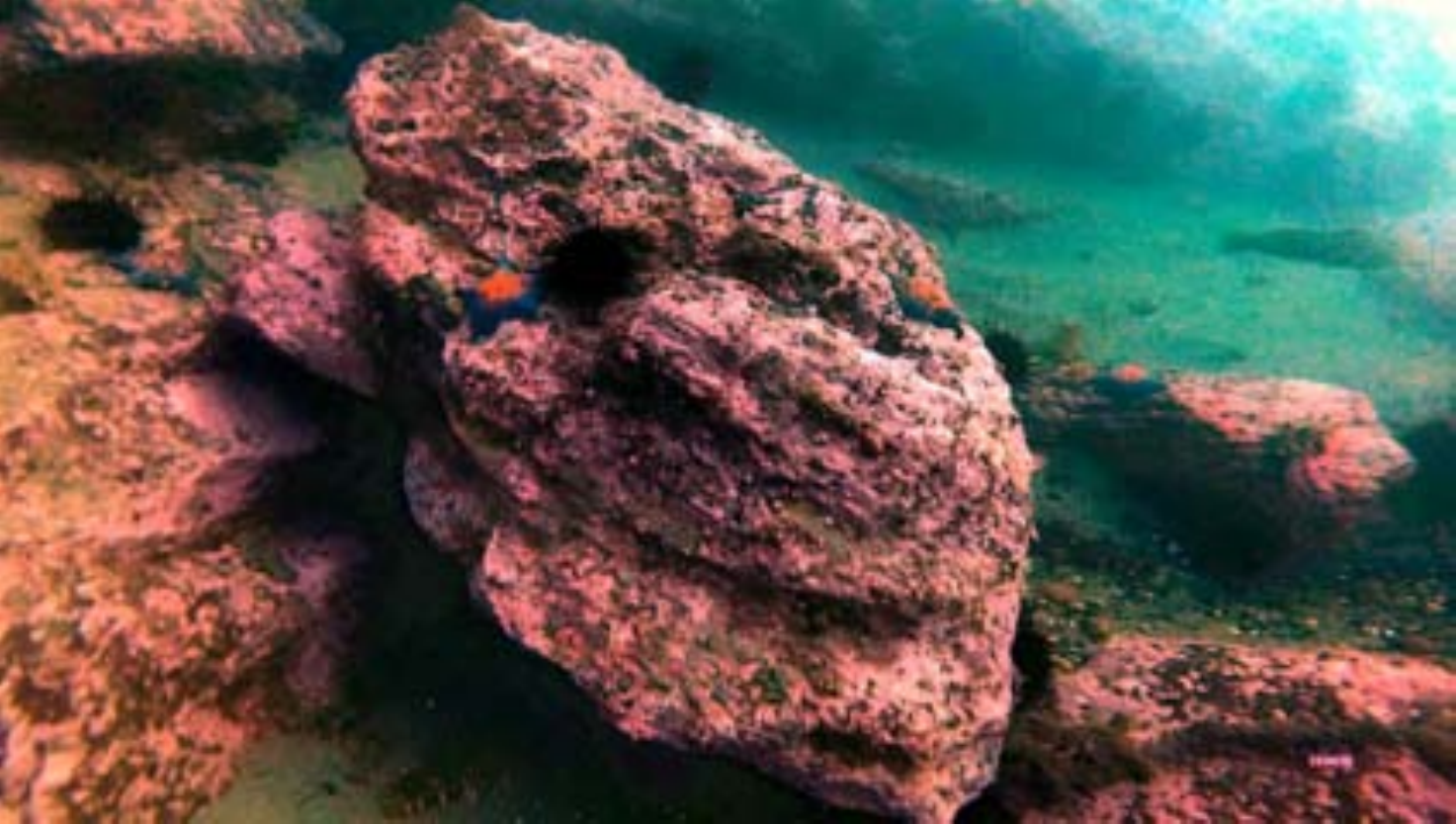}&
		\includegraphics[width=0.16\textwidth]{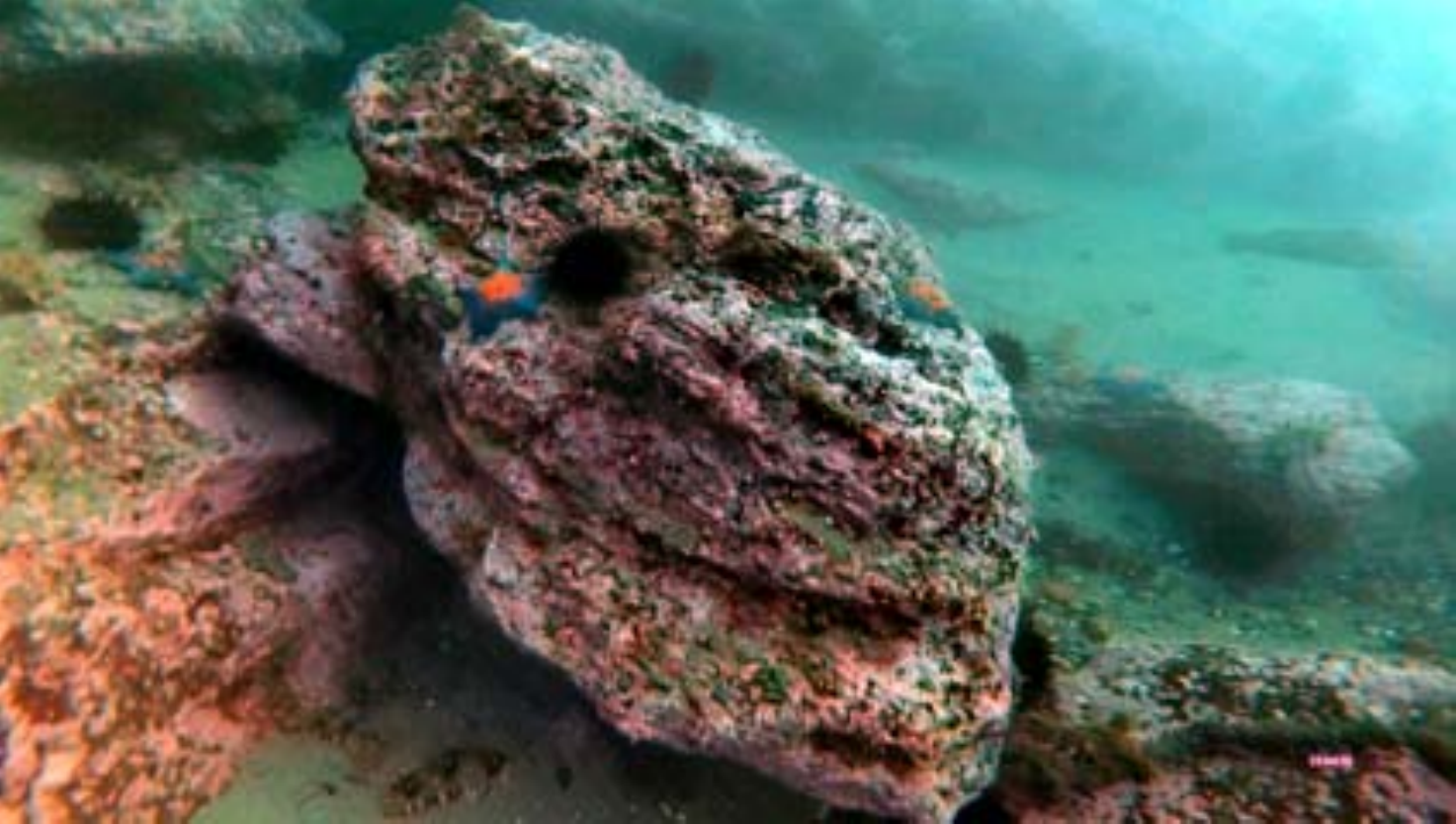}&
		\includegraphics[width=0.16\textwidth]{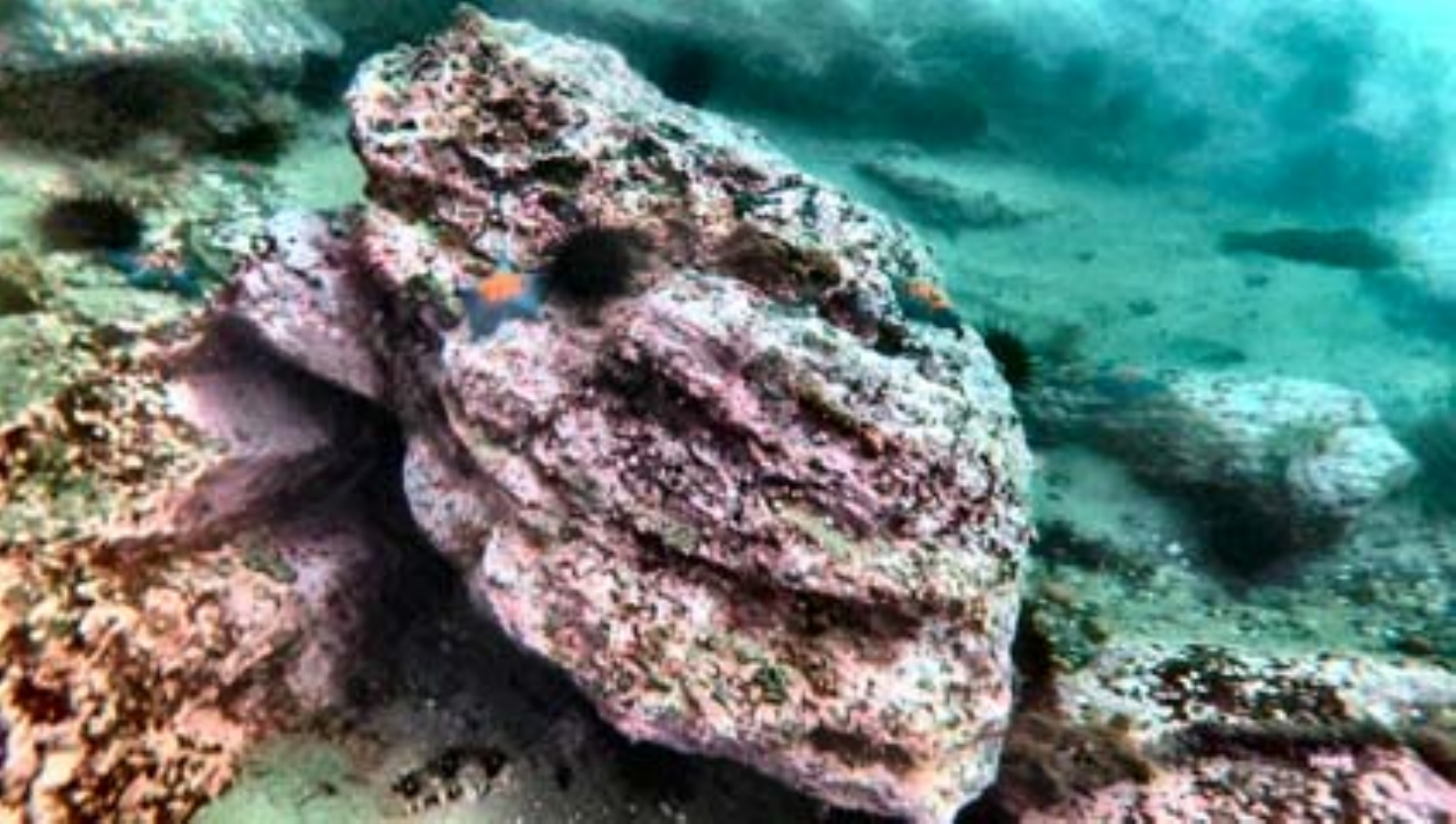}\\
		\includegraphics[width=0.16\textwidth]{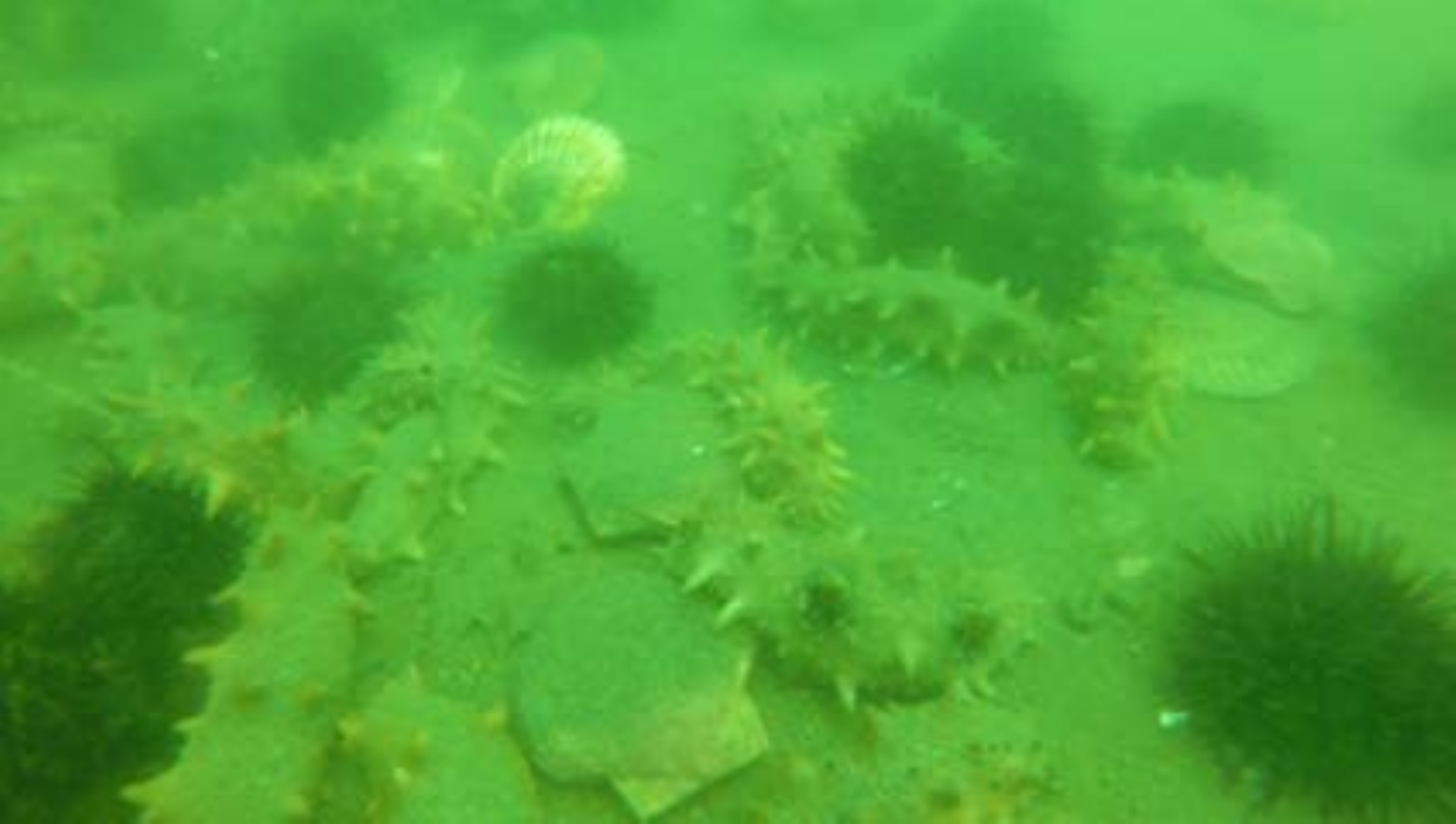}&
		\includegraphics[width=0.16\textwidth]{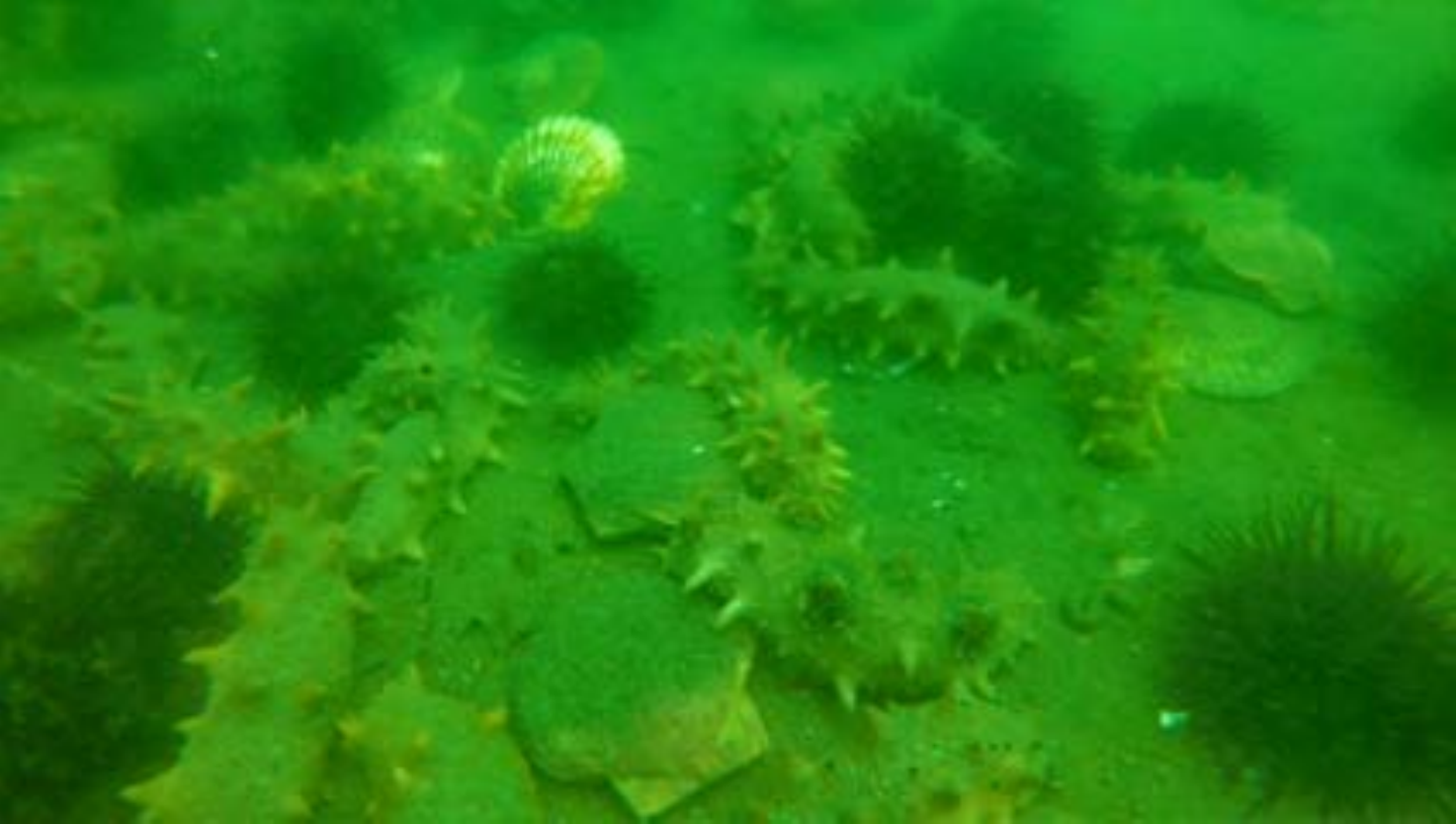}&
		\includegraphics[width=0.16\textwidth]{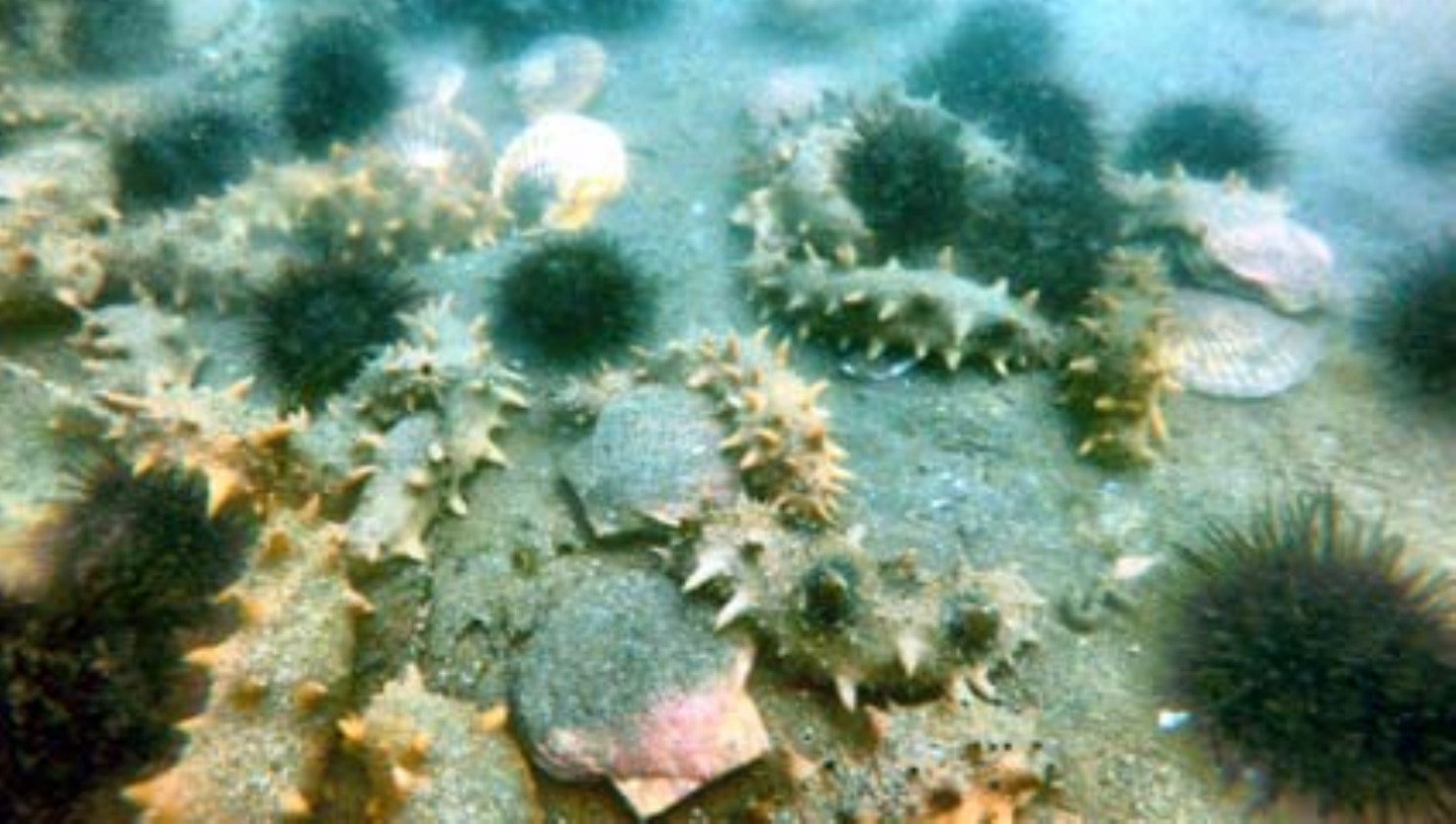}&
		\includegraphics[width=0.16\textwidth]{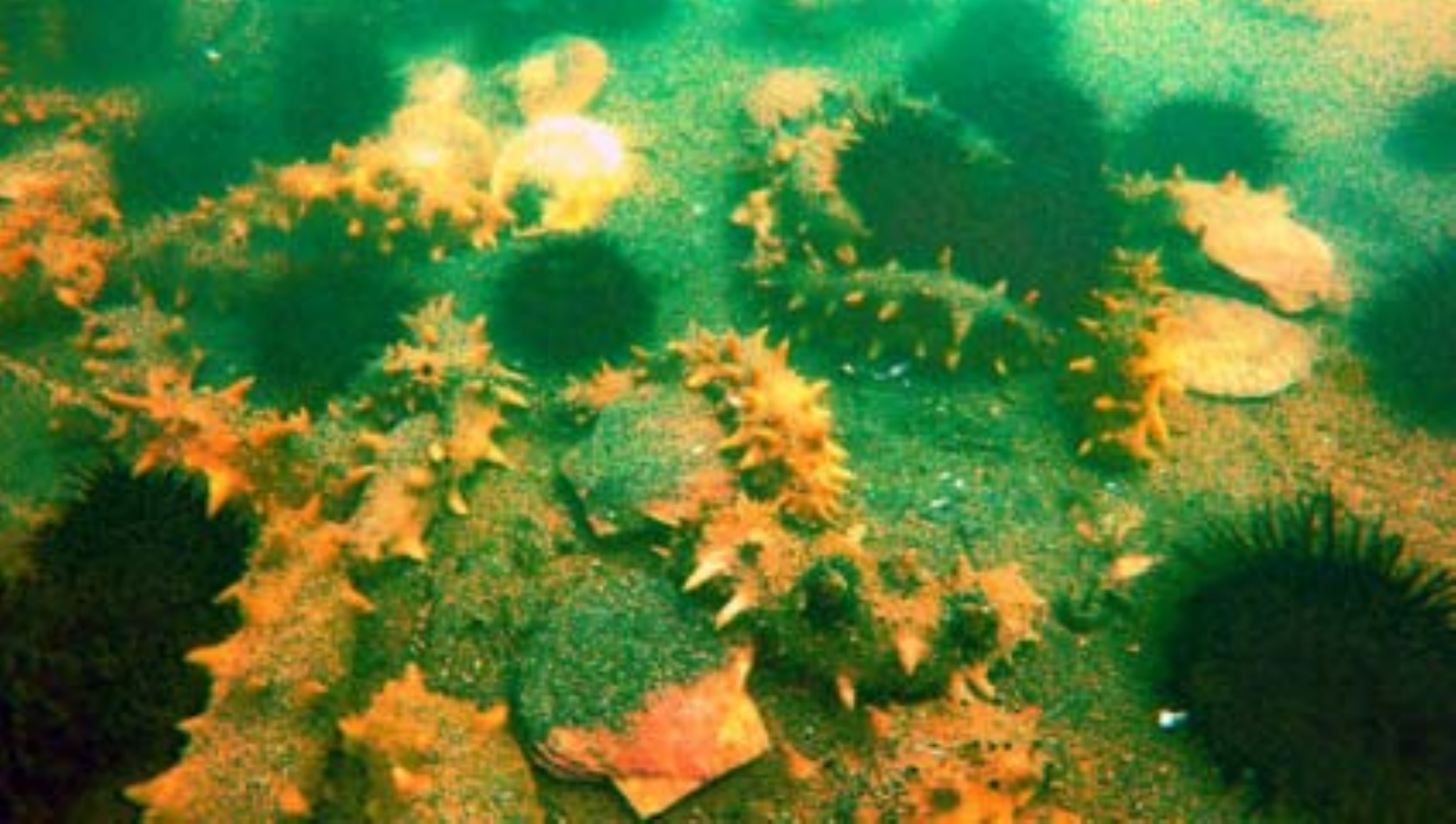}&
		\includegraphics[width=0.16\textwidth]{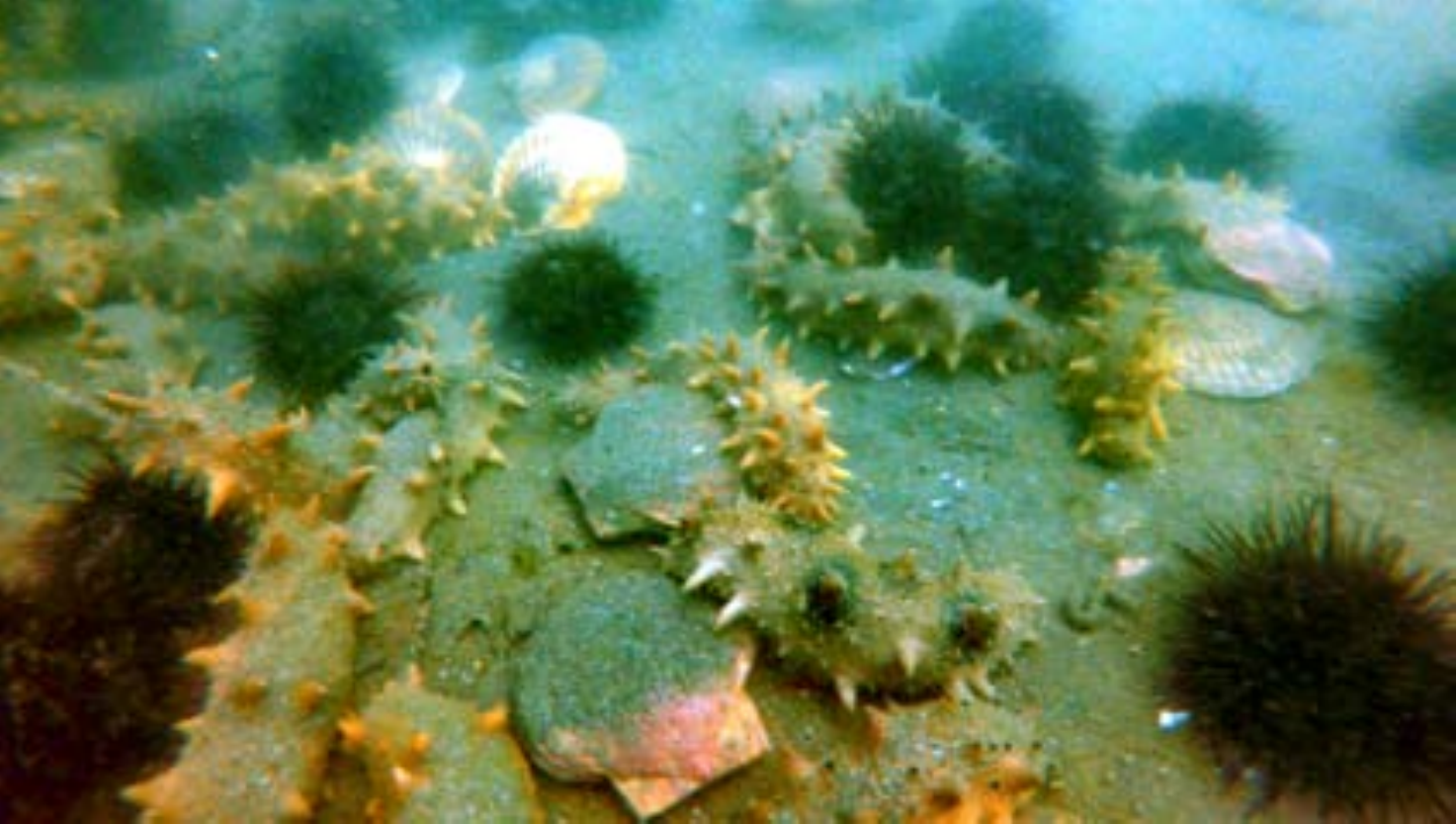}&
		\includegraphics[width=0.16\textwidth]{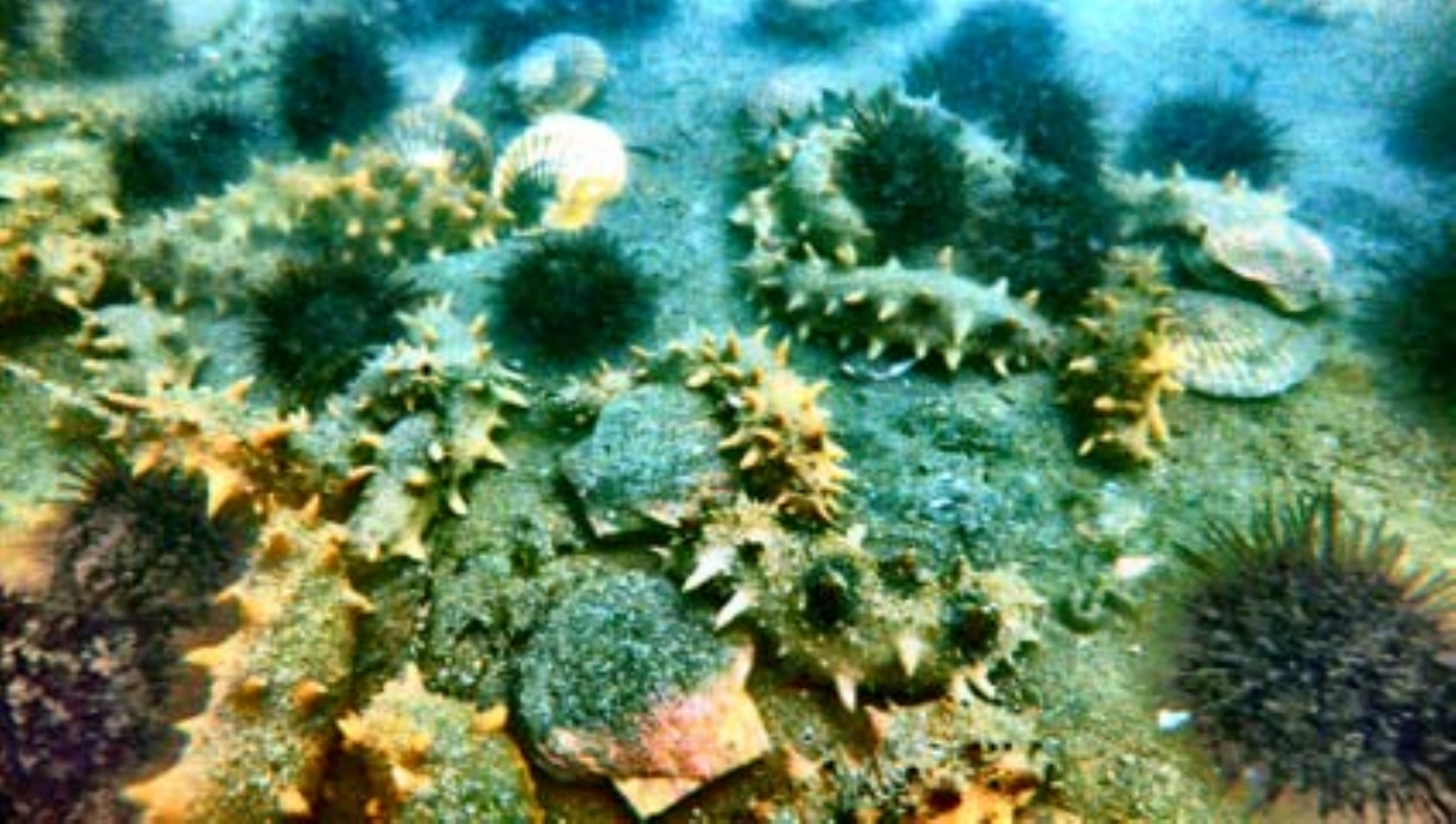}\\
		\footnotesize Underwater Input&\footnotesize~\cite{li2014a}&\footnotesize~\cite{Ancuti2012Enhancing}&\footnotesize~\cite{UnderwaterHazeLines}&\footnotesize~\cite{Berman2016Non}&\footnotesize Ours\\
	\end{tabular}
	\caption{The qualitative enhancement performance on challenging real-world underwater images collected by Berman et al. \cite{UnderwaterHazeLines} (top two rows) and ourself (bottom two rows).}
	\label{fig:underes}
\end{figure*}

\section{Conclusions}\label{sec:con}
In this paper, we developed a deep prior ensemble (DPE) framework to integrate domain-knowledge and information from training data to address image enhancement. By cascading three newly designed basic propagative building-blocks with a feedback control strategy, we actually establish
a theoretically convergent image propagation framework. The main advantage of DPE against conventional optimization-based approaches is that our iterations can successfully avoid unwanted local minimums by network-based descent directions. Meanwhile, we also improve the experience-based network structures by task-aware warm start and prior projection feedback control strategy. Extensive experimental results on various image enhancement tasks demonstrated that the proposed method can successfully provide favorable enhancement performance quantitatively and qualitatively.

\appendices
\section{Proofs of Our Theoretical Results}

In this appendix, we provide proofs for our main theoretical results in Proposition~\ref{prop:error}, Theorem~\ref{thm:converge} and Corollary~\ref{cor:converge}.

\subsection{Preliminaries}
It is necessary to first review and summarize some fundamental mathematical concepts (e.g., Kurdyka-{\L}ojasiewicz and semi-algebraic properties) in the following definition.
More details can also be found in~\cite{Attouch2010Proximal,Attouch2013Convergence} and the reference therein.

\begin{defi}\cite{Attouch2010Proximal,Attouch2013Convergence}
	The two necessary mathematical concepts are presented as follows.
	\begin{itemize}
		\item Kurdyka-{\L}ojasiewicz Property:
		Let $\varphi:\mathbb{R}^d\to(-\infty,\infty]$ be a proper lower semi-continuous function.
		Then function $\varphi$ is said to have Kurdyka-{\L}ojasiewicz (K{\L}) property at $\bar{x}\in\mathtt{dom}(\partial \varphi)$ if there exists $\eta\in(0,\infty]$, a neighborhood $U$ of $\mathbf{u}$ and a concave and continuous function $\phi:[0,\eta)\to\mathbb{R}_+$, such that for all
		$\mathbf{x}\in U\cap\{\mathbf{x}:\varphi(\mathbf{u})<\varphi(\mathbf{x})<\varphi(\mathbf{u})+\eta\}$,
		the following inequality holds
		\begin{equation}
		\phi'(\varphi(\mathbf{x})-\varphi(\mathbf{u}))\mathtt{dist}(0, \partial \varphi(\mathbf{x}))\geq 1.\label{eq:kl-property}
		\end{equation}
		If $\varphi$ satisfies the K{\L} property at each point of $\mathtt{dom}(\partial \varphi)$ then $\varphi$ is called a K{\L} function.
		\item Semi-algebraic Set and Function:
		A subset $\Omega$ of $\mathbb{R}^d$ is a real semi-algebraic set if there exits a finite number of real polynormial function
		$r_{ij}, h_{ij}:\mathbb{R}^d\to\mathbb{R}$ such that
		$\Omega=\cup_{j=1}^M\cap_{i=1}^N\left\{\mathbf{x}\in\mathbb{R}^d: r_{ij}(x)=0, h_{ij}(\mathbf{x})<0\right\}$.
		A function $\varphi:\mathbb{R}^d\to(-\infty,\infty]$ is called semi-algebraic if its graph
		$\mathtt{graph}(\varphi)=\{(\mathbf{x},a)\in\mathbb{R}^{d+1}: \varphi(\mathbf{x})=a\}$,
		is a semi-algebraic subset of $\mathbb{R}^{d+1}$.
	\end{itemize}
\end{defi}

We also recall the following property of the limiting sub-differential function~\cite{Attouch2013Convergence,Bolte2014Proximal} for our proof.

\begin{lemma}\label{lema:close_subgrad}
	Let $\varphi(\mathbf{x})$ be a proper and lower semi-continuous function.
	Suppose sequence $\{\mathbf{x}^k\}_{k\in\mathbb{N}}$ and its (limiting) sub-gradient $\mathbf{d}^k$ of $\varphi(\mathbf{x})$, i.e., $\mathbf{d}^k\in\partial\varphi(\mathbf{x}^k)$, have $\mathbf{x}^k\to\mathbf{x}^*$ and $\mathbf{d}^k\to\mathbf{d}^*$.
	If in addition $\varphi(\mathbf{x}^k)\to\varphi(\mathbf{x}^*)$
	as $k\to\infty$, then $\mathbf{d}^*\in\partial \varphi(\mathbf{x}^*)$.
\end{lemma}

\subsection{Proof of Proposition~\ref{prop:error}}\label{sec:prop-proof}

\begin{proof}
	From our propagation scheme and the expression of $\mathbf{m}^{k+1}$, we have
	\begin{equation}
	\begin{aligned}
	\mathbf{x}^{k+1}=&\mathtt{prox}_{\mathcal{X}_{\Omega}}(\ddot{\mathbf{x}}^{k+1}-\nabla\psi_{\eta^k}^k(\ddot{\mathbf{x}}^{k+1};\mathbf{y}))\\
	=&\mathtt{prox}_{\mathcal{X}_{\Omega}}(\mathbf{x}^{k+1}-\nabla\psi_{\eta^k}^k(\mathbf{x}^{k+1};\mathbf{y})+\mathbf{m}^{k+1}).
	\end{aligned}
	\end{equation}
	Thus from the definition of the proximal map $\mathtt{prox}$, the above equality is equal to
	\begin{equation}
	\mathbf{g}^{k+1} + \nabla\psi_{\eta^{k}}^{k}(\mathbf{x}^{k+1};\mathbf{y}) - \mathbf{m}^{k+1} = \mathbf{0},
	\end{equation}
	which is exactly the same form in the Condition~\ref{condition}.
	Thus, we conclude the assertion that the $\mathbf{m}^{k+1} = \ddot{\mathbf{x}}^{k+1}-\mathbf{x}^{k+1} + \nabla \psi_{\eta^k}^k(\mathbf{x}^{k+1};\mathbf{y}) - \nabla \psi_{\eta^k}^k(\ddot{\mathbf{x}}^{k+1};\mathbf{y})$ is an equivalent form to $\mathbf{m}^{k+1} = \mathbf{g}^{k+1} + \nabla\psi_{\eta^{k}}^{k}(\mathbf{x}^{k+1};\mathbf{y})$.
\end{proof}

\subsection{Proof of Proposition~\ref{prop:equiv}}

\begin{proof}
From the proof of Proposition \ref{prop:error} we can tell that, the highly nonlinear propagation result of DPE at $k$-th stage, i.e., the $\mathbf{x}^{k+1}$ calculated by Eq. \eqref{eq:iter}, satisfies
	\begin{equation}
	\mathbf{x}^{k+1}=\mathtt{prox}_{\mathcal{X}_{\Omega}}(\mathbf{x}^{k+1}-\nabla\psi_{\eta^k}^k(\mathbf{x}^{k+1};\mathbf{y})+\mathbf{m}^{k+1}).
	\end{equation} 
	Then with the definition of proximal map $\mathtt{prox}$, the equivalent formulation of the above equality:
	\begin{equation}\label{eq:eq}
	\mathbf{g}^{k+1} + \nabla\psi_{\eta^{k}}^{k}(\mathbf{x}^{k+1};\mathbf{y}) - \mathbf{m}^{k+1} = \mathbf{0},
	\end{equation}
	indicates that $\mathbf{x}^{k+1}$ can be regarded as an approximate solution of problem: $\min_{\mathbf{x}} \psi_{\eta^k}^k(\mathbf{x};\mathbf{y}) + \mathcal{X}_{\Omega}(\mathbf{x})$, by regarding $\mathbf{m}^{k+1}$ as the error to its first-order optimality condition,
	which also means that $\mathbf{x}^{k+1}$ can be regarded as a result of the following optimization problem
\begin{equation}
\min_{\mathbf{x}}  \psi_{\eta^k}^k(\mathbf{x};\mathbf{y}) + \mathcal{X}_{\Omega}(\mathbf{x}) - (\mathbf{m}^{k+1})^{\top}\mathbf{x}.
\end{equation}
\end{proof}

\subsection{Proof of Theorem~\ref{thm:converge}}\label{sec:thm-proof}
\begin{proof}	
	$1)$ First of all, we prove the sufficient descent property.
	From the equivalent reformulation (i.e., Eq.~\eqref{eq:equiv_form}) of our propagation scheme, we have that
	\begin{equation}
	\begin{aligned}
	&\psi_{\eta^k}^k(\mathbf{x}^{k+1};\mathbf{y}) + \mathcal{X}_{\Omega}(\mathbf{x}^{k+1}) - (\mathbf{m}^{k+1})^{\top}\mathbf{x}^{k+1}\\
	\leq&\psi_{\eta^k}^k(\mathbf{x}^{k};\mathbf{y}) + \mathcal{X}_{\Omega}(\mathbf{x}^{k}) - (\mathbf{m}^{k+1})^{\top}\mathbf{x}^{k}.
	\end{aligned}
	\end{equation}
	The above inequality can be clarified with the definitions of $\psi_{\eta^k}^k(\mathbf{x};\mathbf{y})$ and $\Psi(\mathbf{x};\mathbf{y})$, as
	\begin{equation}
	\begin{aligned}
	&\Psi(\mathbf{x}^{k};\mathbf{y}) - \Psi(\mathbf{x}^{k+1};\mathbf{y})\\
	\geq &\frac{\eta^k}{2}\|\mathbf{x}^{k+1}-\mathbf{x}^k\|^2 + (\mathbf{m}^{k+1})^{\top}(\mathbf{x}^{k}-\mathbf{x}^{k+1})\\
	\geq &\frac{\eta^k}{2}\|\mathbf{x}^{k+1}-\mathbf{x}^k\|^2 - (\frac{1}{\eta^k}\|\mathbf{m}^{k+1}\|^2+\frac{\eta^k}{4}\|\mathbf{x}^{k+1}-\mathbf{x}^k\|^2)\\
	\geq & (\frac{\eta^k}{4}-\frac{(c^k)^2}{\eta^k})\|\mathbf{x}^{k+1}-\mathbf{x}^k\|^2,
	\end{aligned}
	\end{equation}
	where the second inequality is established with Young's inequality and the last one comes from the Condition~\ref{condition}.
	Thus we have proved the assertion, i.e., Eq.~\eqref{eq:suff_des} with the definition of constant $\alpha_1$.
	
	On the other hand, since $\Psi(\mathbf{x};\mathbf{y})$ is a coercive function, that is, $\Psi(\mathbf{x};\mathbf{y})\rightarrow \infty$ as $\|\mathbf{x}\|\rightarrow \infty$, thus it surely brings the boundedness of sequence $\{\mathbf{x}^k\}_{k\in\mathbb{N}}$ with the sufficient descent property of $\Psi(\mathbf{x};\mathbf{y})$.

	$2)$ The second assertion in the Theorem~\ref{thm:converge} can be directly deduced from the formation of $\partial \Psi(\mathbf{x}^k;\mathbf{y})$.
	Since $\mathcal{X}_{\Omega}(\mathbf{x})$ is proper, lower semi-continuous and $\psi(\mathbf{x};\mathbf{y}):=f(\mathbf{x};\mathbf{y}) + g(\mathbf{x})$ is continuous differential, then we have
	\begin{equation}
	\begin{aligned}
	&\|\partial \Psi(\mathbf{x}^k;\mathbf{y})\|-\eta^{k-1}\|\mathbf{x}^k-\mathbf{x}^{k-1}\| \\
	\leq &\|\mathbf{g}^{k} + \nabla\psi_{\eta^{k-1}}^{k-1}(\mathbf{x}^{k};\mathbf{y})\| \leq c^{k-1}\|\mathbf{x}^{k}-\mathbf{x}^{k-1}\|,
	\end{aligned}
	\end{equation}
	which is directly deduced from the Condition~\ref{condition}.
	Thus, with the definition of $\alpha_2$, we have proved the second assertion by rewriting the above inequality.
	
	$3)$ From the sufficient descent property of $\Psi(\mathbf{x};\mathbf{y})$, we have
	\begin{equation}
	\begin{aligned}
	\sum_{k=0}^{N-1}\|\mathbf{x}^{k+1}-\mathbf{x}^k\|^2 \leq \alpha_1(\Psi(\mathbf{x}^0;\mathbf{y})-\Psi(\mathbf{x}^{N};\mathbf{y})),
	\end{aligned}
	\end{equation}
	for a positive integer $N$.
	Since $\Psi(\mathbf{x};\mathbf{y})$ is bounded from below, thus we have $\lim_{k\rightarrow \infty}\|\mathbf{x}^{k+1}-\mathbf{x}^k\|=0$ by taking the limit as $N\rightarrow \infty$.
	On the other hand, from $\|\mathbf{m}^{k+1}\|\leq c^{k}\|\mathbf{x}^{k+1}-\mathbf{x}^{k}\|$ we have $\mathbf{m}^{k+1} \rightarrow \mathbf{0}$ as $k\rightarrow\infty$.
	Furthermore, denoting $\mathbf{P}^k:=\eta^{k}(\mathbf{x}^{k-1}-\mathbf{x}^{k})+\mathbf{m}^k$, there obviously has $\mathbf{P}^k \in \partial\Psi(\mathbf{x}^k;\mathbf{y})$ and $\mathbf{P}^k\rightarrow \mathbf{0}$ as $k\rightarrow\infty$.
	
	Since $\{\mathbf{x}^k\}_{k\in\mathbb{N}}$ is bounded, then there exists a subsequence $\{\mathbf{x}^{k_l}\}_{l\in\mathbb{N}}$ such that $\mathbf{x}^{k_l}\rightarrow \mathbf{x}^{\ast}$ as $l\rightarrow\infty$.
	By letting step $k+1$ as $k_l-1$, then we have
	\begin{equation}
	\begin{aligned}
	&\psi_{\eta^{k_l}}^{k_l}(\mathbf{x}^{k_l};\mathbf{y}) + \mathcal{X}_{\Omega}(\mathbf{x}^{k_l}) - (\mathbf{m}^{k_l})^{\top}\mathbf{x}^{k_l}\\
	\leq&\psi_{\eta^{k_l}}^{k_l}(\mathbf{x}^{\ast};\mathbf{y}) + \mathcal{X}_{\Omega}(\mathbf{x}^{\ast}) - (\mathbf{m}^{k_l})^{\top}\mathbf{x}^{\ast}.
	\end{aligned}
	\end{equation}
	By taking $l\rightarrow\infty$, we have the following inequality with the first condition in the Theorem~\ref{thm:converge}
	\begin{equation}
	\limsup_{l\rightarrow\infty} \mathcal{X}_{\Omega}(\mathbf{x}^{k_l}) \leq \mathcal{X}_{\Omega}(\mathbf{x}^{\ast}).
	\end{equation}
	Then with the lower semi-continuous property of the function $\mathcal{X}_{\Omega}(\mathbf{x})$, we have $\lim_{l\rightarrow\infty}\mathcal{X}_{\Omega}(\mathbf{x}^{k_l})= \mathcal{X}_{\Omega}(\mathbf{x}^{\ast})$, which further indicates $\Psi(\mathbf{x}^{k_l};\mathbf{y}) \rightarrow \Psi(\mathbf{x}^{\ast};\mathbf{y})$ as $l\rightarrow\infty$.
	
	Together with the assertion of Lemma~\ref{lema:close_subgrad} we have $\mathbf{0} \in \Psi(\mathbf{x}^{\ast};\mathbf{y})$, which indicates that $\mathbf{x}^{\ast}$ is a critical point of $\Psi(\mathbf{x};\mathbf{y})$.
	Moreover, since $\Psi(\mathbf{x};\mathbf{y})$ is bounded from below and sufficient descent, $\Psi(\mathbf{x}^k;\mathbf{y})$ has limit value as $k\rightarrow \infty$.
	Together with $\Psi(\mathbf{x}^{k_l};\mathbf{y}) \rightarrow \Psi(\mathbf{x}^{\ast};\mathbf{y})$, we have concluded the proof.
\end{proof}
\subsection{Proof of Corollary~\ref{cor:converge}}\label{sec:cor-proof}
\begin{proof}
	Since $\Psi(\mathbf{x};\mathbf{y})$ is a semi-algebraic function, thus it satisfies K{\L} inequality at every point of $\mathtt{dom}(\partial \Psi(\mathbf{x};\mathbf{y}))$.
	From the Condition~\ref{condition} that $\{\mathbf{x}^k\}_{k\in\mathbb{N}}$ is a bounded sequence, then there exists a subsequence that converges to $\mathbf{x}^{\ast}$.
	With the Condition~\ref{condition}, then $\Psi$ has uniformized K{\L} property~\cite{Bolte2014Proximal} at the set of all limit points of $\{\mathbf{x}^k\}_{k\in\mathbb{N}}$.
	Since $\Psi(\mathbf{x};\mathbf{y})$ is sufficiently descent, then there exists $k_1$ such that for $k>k_1$,
	\begin{equation}
	\phi'(\Psi(\mathbf{x}^k;\mathbf{y})-\Psi(\mathbf{x}^{\ast};\mathbf{y}))\mathtt{dist}(0, \partial\Psi(\mathbf{x}^k;\mathbf{y}))\geq 1
	\end{equation}
	From the concavity of $\phi$, we get
	\begin{equation}
	\begin{aligned}
	&\Lambda_{k,k+1}\\
	&:=\phi(\Psi(\mathbf{x}^k;\mathbf{y})-\Psi(\mathbf{x}^{\ast};\mathbf{y})) - \phi(\Psi(\mathbf{x}^{k+1};\mathbf{y})-\Psi(\mathbf{x}^{\ast};\mathbf{y}))\\
	&\geq\phi'(\Psi(\mathbf{x}^k;\mathbf{y})-\Psi(\mathbf{x}^{\ast};\mathbf{y}))(\Psi(\mathbf{x}^k;\mathbf{y})-\Psi(\mathbf{x}^{k+1};\mathbf{y}))\\
	&\geq\frac{\Psi(\mathbf{x}^k;\mathbf{y})-\Psi(\mathbf{x}^{k+1};\mathbf{y})}{\mathtt{dist}(0, \partial\Psi(\mathbf{x}^k;\mathbf{y}))}\geq \frac{\alpha_1\|\mathbf{x}^{k+1}-\mathbf{x}^k\|^2}{\alpha_2\|\mathbf{x}^{k}-\mathbf{x}^{k-1}\|},
	\end{aligned}
	\end{equation}
	where the last inequality comes from the Condition~\ref{condition} and the second assertion in the Theorem~\ref{thm:converge}.
	Then we have
	\begin{equation}
	\begin{aligned}
	2\|\mathbf{x}^{k+1}-\mathbf{x}^k\| \leq& 2(\frac{\alpha_2}{\alpha_1}\Lambda_{k,k+1}\|\mathbf{x}^k-\mathbf{x}^{k-1}\|)^{\frac{1}{2}}\\
	\leq& \frac{\alpha_2}{\alpha_1}\Lambda_{k,k+1} + \|\mathbf{x}^k-\mathbf{x}^{k-1}\|.
	\end{aligned}
	\end{equation}
	Summing up the above inequality from $k_1+1$ to $k$ yields
	\begin{equation}
	\begin{aligned}
	&2\sum_{i=k_1+1}^k\|\mathbf{x}^{i+1}-\mathbf{x}^i\|\leq \sum_{i=k_1+1}^k\|\mathbf{x}^i-\mathbf{x}^{i-1}\| + \frac{\alpha_2}{\alpha_1}\Lambda_{k_1+1,k+1}\\
	\leq& \sum_{i=k_1+1}^k\|\mathbf{x}^{i+1}-\mathbf{x}^i\| + \|\mathbf{x}^{k_1+1}-\mathbf{x}^{k_1}\| + \frac{\alpha_2}{\alpha_1}\Lambda_{k_1+1,k+1},
	\end{aligned}
	\end{equation}
	where the first inequality comes from the definition of $\Lambda_{k,k+1}$.
	Thus we have the following inequality for any $k >k_1$, with the fact that $\psi\geq 0$
	\begin{equation}
	\begin{aligned}
	&\sum_{i=k_1+1}^k\|\mathbf{x}^{i+1}-\mathbf{x}^i\|\\
	\leq &\|\mathbf{x}^{k_1+1}-\mathbf{x}^{k_1}\| + \phi(\Psi(\mathbf{x}^{k_1+1};\mathbf{y})-\Psi(\mathbf{x}^{\ast};\mathbf{y})),
	\end{aligned}
	\end{equation}
	which indicates that $\{\mathbf{x}^k\}_{k\in\mathbb{N}}$ has finite length, i.e.,
	\begin{equation}
	\sum_{k=1}^{\infty}\|\mathbf{x}^{k+1}-\mathbf{x}^k\|\leq \infty.
	\end{equation}
	Further on, the sequence $\{\mathbf{x}^k\}_{k\in\mathbb{N}}$ is a Cauchy sequence which converges to a critical point of $\Psi(\mathbf{x};\mathbf{y})$.
\end{proof}




\ifCLASSOPTIONcaptionsoff
  \newpage
\fi

\bibliographystyle{IEEEtran}
\bibliography{reference}

\begin{IEEEbiography}[{\includegraphics[width=1in,height=1.25in,clip,keepaspectratio]{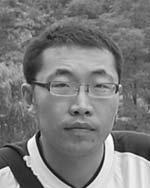}}]{Risheng Liu} received the BSc and PhD degrees both in mathematics from the Dalian University of Technology in 2007 and 2012, respectively. He was a visiting scholar in the Robotic Institute of Carnegie Mellon University from 2010 to 2012. He served as Hong Kong Scholar Research Fellow at
	the Hong Kong Polytechnic University from 2016 to 2017. He is currently an associate professor with the Key Laboratory for Ubiquitous Network and Service Software of Liaoning Province, Internal School of Information and Software Technology, Dalian University of Technology. His research interests include machine learning, optimization, computer vision and multimedia. He was a co-recipient of the IEEE ICME Best Student Paper Award in both 2014 and 2015. Two papers were also selected as Finalist of the Best Paper Award in ICME 2017. He is a member of the IEEE and ACM.
\end{IEEEbiography}

\begin{IEEEbiography}[{\includegraphics[width=1in,height=1.25in,clip,keepaspectratio]{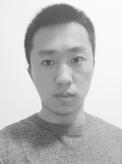}}]{Long Ma} received the B.E. degree in Information and Computing Science from Northeast Agricultural University, Harbin, China, in 2016. He is currently pursuing the master’s degree in software engineering at Dalian University of Technology, Dalian, China. His research interests include computer vision, image enhancement and machine learning. 
\end{IEEEbiography}

\begin{IEEEbiography}[{\includegraphics[width=1in,height=1.25in,clip,keepaspectratio]{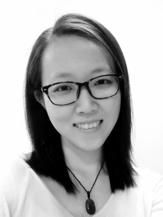}}]{Yiyang Wang}
received the B.Sc and Ph.D degree both in Mathematics from Dalian University of Technology, in 2012 and 2017. She is now a postdoc in the Institute of Atmospheric Sciences from Fudan University and Shanghai Institute of Meteorological Science. Her main research interests include machine learning, optimization and image processing.

\end{IEEEbiography}

\begin{IEEEbiography}[{\includegraphics[width=1in,height=1.25in,clip,keepaspectratio]{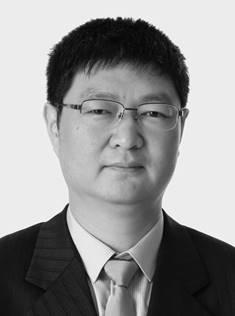}}]{Lei Zhang}
(M’04-SM’14-F’18) received his B.Sc. degree in 1995 from Shenyang Institute of Aeronautical Engineering, Shenyang, P.R. China, and M.Sc. and Ph.D degrees in Control Theory and Engineering from Northwestern Polytechnical University,
Xi’an, P.R. China, respectively in 1998 and 2001, respectively. From 2001 to 2002, he was a research associate in the Department of Computing, The Hong Kong Polytechnic University. From January 2003 to January 2006 he worked as a Postdoctoral Fellow in the Department of Electrical and Computer Engineering, McMaster University, Canada. In 2006, he joined as an Assistant Professor with the Department of Computing, The Hong Kong Polytechnic University, where he has been a Chair Professor, Since 2017. He has published over 200 papers in those areas. His research interests include computer vision, pattern recognition, image and video analysis, and biometrics. As of 2018, his
publications have been cited over 33,000 times in the literature. Prof. Zhang is an Associate Editor of IEEE Trans. on Image Processing, SIAM Journal of Imaging Sciences and Image and Vision Computing, etc. He is a “Web of Science Highly Cited Researcher” from 2015 to 2017. More information can be found in his homepage \url{http://www4.comp.polyu.edu.hk/∼cslzhang/}.
\end{IEEEbiography}

\end{document}